\documentclass[a4paper,10pt]{article}
\usepackage[utf8]{inputenc}
\usepackage[T1]{fontenc}
\usepackage{amsmath}
\usepackage{amssymb}

\usepackage{empheq}
\usepackage{lmodern}
\usepackage{float}

\usepackage[svgnames]{xcolor}

\usepackage{colortbl}
\newcommand{\fillred}{\cellcolor{red!25}}
\newcommand{\fillgreen}{\cellcolor{green!25}}
\newcommand{\fillorange}{\cellcolor{orange!25}}

\usepackage{hyperref}
\hypersetup{
  colorlinks=true,
  citecolor=LimeGreen,
  linkcolor=Magenta,
  urlcolor=Blue
}

\usepackage{geometry}
\newtheorem{Theorem}{Theorem}[section]
\newtheorem{assumption}[Theorem]{Assumption}
\newtheorem{proposition}[Theorem]{Proposition}

\newtheorem{remark}[Theorem]{Remark}

\usepackage{chngpage}
\usepackage{array}
\newcolumntype{R}[1]{>{\raggedleft\arraybackslash }b{#1}}
\newcolumntype{L}[1]{>{\raggedright\arraybackslash }b{#1}}
\newcolumntype{C}[1]{>{\centering\arraybackslash }b{#1}}
\newcolumntype{P}[1]{>{\centering\arraybackslash}p{#1}}

\newcommand{\norm}[1]{\left\|#1\right\|}
\newcommand{\abs}[1]{\left|#1\right|}

\DeclareMathOperator{\Tr}{{\text{Tr}}}

\DeclareMathOperator{\dif}{{\text{d}\!}}
\DeclareMathOperator{\R}{\mathbb{R}}

\usepackage{listings}

\def\Lc{{\cal L}}
\def \E{\mathbb{E}}
\def \M{\mathbb{M}}

\def\Lc{{\cal L}}
\def\x{\times}
\def \Fb{\overline F}
\def\1{{\bf 1}}

\newcommand{\DPar}[2]{\frac{\partial\,\! #1}{\partial\,\! #2}}

\usepackage{bbm}

\usepackage[backend=biber, style=alphabetic,
sorting=ynt]{biblatex}
\begin{document}

\title{Machine Learning for semi linear PDEs}
\author{Quentin Chan-Wai-Nam  \footnote{EDF R\&D} \thanks{quentin.chan-wai-nam@edf.fr}
\and
Joseph Mikael \footnote{EDF R\&D} \thanks{joseph.mikael@edf.fr}
\and 
Xavier Warin \footnote{EDF R\&D \& FiME, Laboratoire de Finance des March\'es de l'Energie} \thanks{xavier.warin@edf.fr}
}

\maketitle

\abstract{
Recent machine learning algorithms dedicated to solving semi-linear PDEs are improved by using different neural network architectures and different parameterizations. These algorithms are compared to a new one that solves a fixed point problem by using deep learning techniques. This new algorithm appears to be competitive in terms of accuracy with the best existing algorithms.
\vspace{5mm}

\\ }

{\bf Acknowledgments.} 
{The authors would like to thank Simon F\'ecamp for useful discussions and technical advices.}

\vspace{5mm}

\section{Introduction}
This paper is devoted to the resolution in high dimension of equations of the form
\begin{align}
-\partial_t u-\Lc u =f(t,x,u,Du);\qquad & u_T=g,& t<T,~x\in\R^d,
\label{eq:PDEInit}
\end{align}
with a  non-linearity $f(t,x,y,z)$ in the solution and its gradient, a  bounded terminal condition $g$ and a diffusion generator $\Lc$ satisfying
\begin{align}
\label{eq:PDE}
\mathcal{L}u := \frac{1}{2} \Tr \left(\sigma \sigma^\top(t, x) D^2 u(t, x)\right) + \mu(t, x).Du(t, x).
\end{align}
where $\mu$ is a function defined on $\R \times \R^d$ with values in $\R^d$, $\sigma$ is a function defined on $\R \times \R^d$ with values in $\M^d$ the set of  $d \times d$ matrix and  $\mathcal{L}$ is the generator associated to the forward process:
\begin{flalign}
X_t^{0,x} = x + \int_0^t \mu(s,X_s) \dif s+  \int_0^t \sigma(s,X_s) \dif W_s,
\label{eq:SDE}
\end{flalign}
with  $W_t$ a $d$-dimensional Brownian motion.

Traditional deterministic methods (e.g. finite elements method) dedicated to solving numerically non linear Partial Differential Equations (PDE) suffer from the curse of dimensionality and one cannot hope to solve equations of dimension greater than 4 or 5 with this kind of methods.

Based on the resolution of the BSDE associated to the PDE first exhibited in \cite{pardoux1990adapted} and using the time discretization scheme proposed in \cite{bouchard2004discrete}, some effective algorithms based on regressions manage to solve non linear PDEs in dimension above 4 (see \cite{gobet2005regression,lemor2006rate}). As shown in \cite{gobet2016linear} this technique is the source of a lot of research. Among others, we may refer to  \cite{fahim2011probabilistic} which generalizes this technique to full non linear equations by using the Second Order Backward Equation framework proposed in \cite{cheridito2007second}. This regression technique uses some basis functions that can be either some global polynomials as in \cite{longstaff2001valuing} or some local polynomials as proposed in \cite{bouchard2012monte}: therefore this methodology  still faces the curse of dimensionality and can only solve some problems in dimension below 7 or 8.

Recently, \cite{henry2016branching,bouchard2017numerical,bouchard2017numerical2,warin2017variations} proposed to solve high dimensional PDE by using a branching method and a time step randomization applied to the Feyman-Kac representation  of the PDE.
In the case of semi-linear PDE's, a differentiation technique using some Malliavin weights as proposed in \cite{fournie1999applications} allows to estimate the gradient $Du$ of the solution. Unfortunately, branching techniques are only limited to small maturities, some small non-linearities and mainly to non-linearities that are polynomial in $u$ and $Du$.

Most recently, three other methods try to solve this difficult problem:
\begin{itemize}
\item \cite{warin2018nestingsMC,warin2018monte} propose a very simple technique based only on nesting Monte Carlo applied on the Feynman-Kac representation of the PDE. The convergence of the algorithm is demonstrated. As shown by the estimators in \cite{warin2018monte}, this technique is far more effective if the Lipschitz coefficients associated to the non linearity and the maturity of the problem are not too high.
\item In \cite{weinan2017linear,weinan2016multilevel,hutzenthaler2017multi}, the authors develop an algorithm based on Picard iterations, multi-level techniques and automatic differentiation once again applied to the Feynman-Kac representation of the PDE. The time integration appearing in this representation is achieved by quadrature and the authors are able to solve some PDEs in very high dimension. However, tuning this algorithm can be difficult due to the number of methodologies involved in the resolution. Recently \cite{hutzenthaler2018overcoming} combined the ideas of \cite{weinan2017linear,weinan2016multilevel,hutzenthaler2017multi} and \cite{henry2016branching,warin2018monte} to show that some modified Picard iteration algorithm for non-linearities in $u$ applied to the heat equation can be solved with a polynomial complexity with both the dimension and the reciprocal of the required accuracy. However no numerical results are given to confirm the result.
\item At last \cite{han2017overcoming,weinan2017deep} propose a deep learning based technique called \emph{Deep BSDE} (DBSDE) to solve semi-linear PDEs. \cite{beck2017machine} extends the latter methodology to full non linear equations. This approach is based on an Euler discretization of the forward underlying SDE with solution $X_t$  and of the BSDE associated to the problem. The algorithm tries to learn the values $u$ and  $z= \sigma^{\top} Du$ at each time step of the Euler scheme so that a forward simulation of $u$ till maturity $T$  matches the target $g(X_T)$.
\cite{raissi2018forward} introduces a version of the DBSDE algorithm in which a neural network tries to learn $u$ by calculating $Du$ by automatic differentiation and incorporates the constraints associated to the Euler discretization of the BSDE in the loss function. 
These deep learning-based techniques seem to be very effective but no current result justifies their convergence. It is then difficult to know their limitations.
\end{itemize}
The previously described methods are all interesting but in the present paper, we focus on machine learning-based algorithms.

The objectives of this paper are:
\begin{itemize}
\item to give an improved version of the DBSDE algorithm using different networks architecture and parameterizations,
\item to develop a new deep learning-based algorithm, mixing some features coming from \cite{warin2018monte} and \cite{hutzenthaler2017multi},
\item to compare numerically how these algorithms compare each other. Particularly, we will see that that the new algorithm is competitive with the improved \emph{Deep BSDE} algorithm found in term of accuracy.
\item to give a demonstration showing that, under some simplifying assumptions, the loss of the new algorithm can go to zero and that, when the driver is independent of the gradient, a loss going to zero implies a convergence of the scheme to the true solution of the problem.
\end{itemize}

All algorithms used can be found at
\url{https://gitlab.com/14chanwa/ml_for_semilinear_pdes.git}.

\section{Existing Deep BSDE algorithms}
\label{sec:biblio_dbsde}

The DBSDE algorithm proposed in \cite{han2017overcoming,weinan2017deep} starts from the BSDE representation of \eqref{eq:PDEInit} first proposed in \cite{pardoux1990adapted}: 
\begin{align}
\label{eq:bsde}
u(t,X_t)= u(0, X_0)- \int_0^t f(s,X_s,u(s,X_s),Du(s,X_s)) ds + \int_0^t Du(s,X_s)^{\top} \sigma(s,X_s) dW_s.
\end{align}

For a set of time steps $t_0=0<t_1< \ldots < t_N = T$, we use an Euler scheme to approximate   $(X_{t_i})_{i =1...N}$ from equation \eqref{eq:SDE}   by:
\begin{align*}
X_{t_{i+1}} \approx X_{t_{i}} + \mu(t_i, X_{t_i}) (t_{i+1} - t_{i}) + \sigma(t_i, X_{t_i}) (W_{t_{i+1}} - W_{t_{i}}). 
\end{align*}
In the same way, an approximation of equation \eqref{eq:bsde}
is obtained by the Euler scheme:
\begin{align*}
u(t_{i+1}, X_{t_{i+1}}) &= u(t_{i}, X_{t_{i}}) - f(t_i, X_{t_i},u(t_{i}, X_{t_{i}}), D u(t_{i}, X_{t_{i}}))(t_{i+1} - t_{i}) +  \\
&  D u(t_{i}, X_{t_{i}})^{\top} \sigma(t_{i}, X_{t_{i}}) (W_{t_{i+1}} - W_{t_i}).
\end{align*}

In the initial DBSDE algorithm, neural networks are supposed to output an approximate of $$\kappa_{t_i}:= \sigma(t_{i}, X_{t_{i}})^{\top} D u(t_{i}, X_{t_{i}})$$ from the vector of \textit{features} $X_{t_i}$. In the machine learning language, the realizations of $(X_{t_i})_{i =1...N}$ represent the data. The parameters  $\theta$ of the neural networks are estimated with a stochastic gradient descent which objective is to minimize the loss $\ell(\theta):= \E\big[(u(T, X_T) - g(X_T))^2\big],$ as $g(X_T)$ corresponds to the target of  $u(T,X_T)$ due to the  terminal condition $u(T, X_T) = g(X_T)$.

\begin{figure}[ht]
\centering
\includegraphics[height=4.5cm]{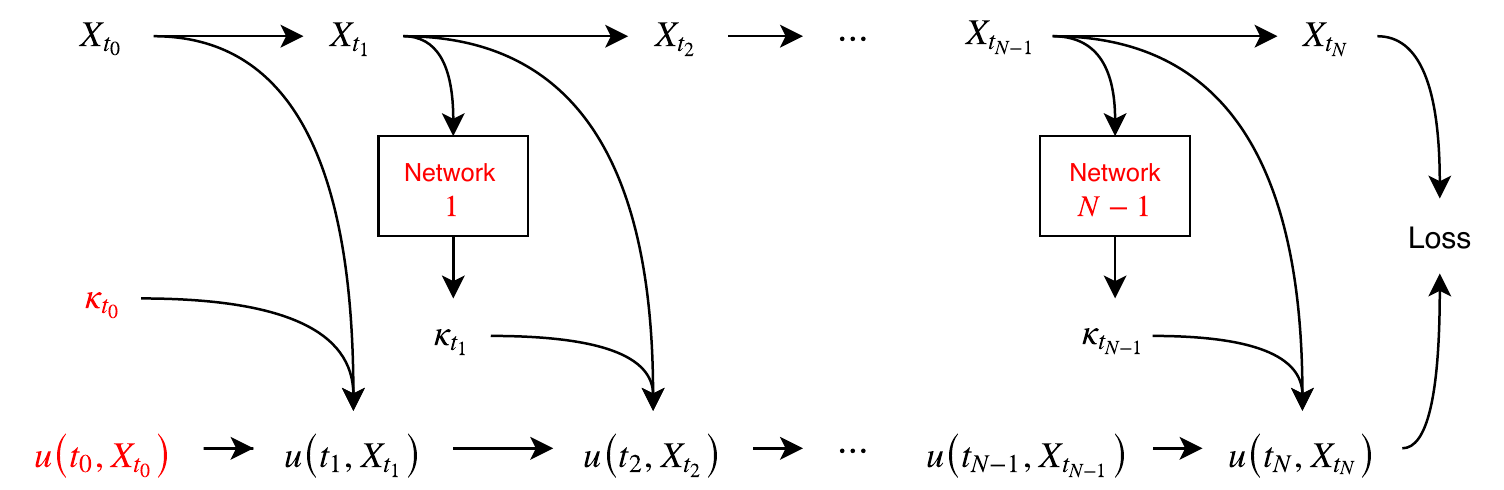}
\caption{\cite{han2017overcoming} original graph. The parameters of the graph are represented in \textcolor{red}{red}.}
\label{fig:jentzen_graph}
\end{figure}
The architecture described in \cite{han2017overcoming,weinan2017deep} and in Figure (\ref{fig:jentzen_graph}) consists in building $N-1$ feed forward neural networks to estimate $(\kappa_{t_i})_{i=1, \ldots N-1}$. The number of weights to be estimated is roughly $N\times \hbox{nb layers} \times\hbox{layer size}$. By construction, there is no link between the gradients of two successive and possibly close time steps. We will see in Section \ref{sec:other_architectures} that we can add a global structure to the architecture ensuring a consistency between two gradients of two close time steps.

To help the neural network converge, \cite{fujii2017asymptotic} consists in learning the residue to a prior on the gradient to be learned. The prior is derived from an asymptotic expansion of first order.

\cite{raissi2018forward} proposes to approximate directly the function $u$ with a neural network, enforcing the Euler discretization scheme softly in the loss function:
\begin{align}
\ell(\theta) &:= \mathbb{E} \left[ \sum_{i=0}^{N-1} \Phi(t_i, t_{i+1}, X_{t_i}, Y_{t_{i}}, Y_{t_{i+1}}, W_{t_{i+1}} - W_{t_i}) + (g(X_T) - Y_T)^2 \right] \label{eq:raissiloss}
\end{align}
\begin{align*}
\Phi(t_i, t_{i+1}, X_{t_i}, Y_{t_{i}}, Y_{t_{i+1}}, W_{t_{i+1}} - W_{t_i}) &= \left(Y_{t_{i+1}} - Y_{t_{i}} + f(t_i, X_{t_i}, Y_{t_i}, \sigma^\top (t_i, X_{t_i}) \widehat{Z}_{t_i}) (t_{i+1} - t_i) \right.\\
&\left. \qquad - \widehat{Z}_{t_i}^\top \sigma (t_i, X_{t_i})  (W_{t_{i+1}} - W_{t_i})\right)^2
\end{align*}
where $Y_{t_i}$ should approximate $u(t_i, X_{t_i})$, with $Y_{t_i} := \mathcal{N}(t_i, X_{t_i})$ where $\mathcal{N}$ is a neural network,  and $\widehat{Z}_{t_i} := \widehat{D} Y_{t_i}$ where $\widehat{D}$ is the automatic differentiation operator relative to $X_{t_i}$ in TensorFlow, so that $\widehat{Z}_{t_i}$ should approximate $Du(t_i, X_{t_i})$. We point out that: 

\begin{enumerate}
\item Deep learning-based algorithms have the advantage of returning values of $Y_{t_i}$ and $Z_{t_i}$ along the chosen time discretization, given trajectories $W_{t_0}, \cdots, W_{t_{N-1}}$.
\item The algorithms described in \cite{han2017overcoming,weinan2017deep} do not return $u(t, x)$, $Du(t, x)$ for any $t, x$ outside of a given trajectory. The algorithm in \cite{raissi2018forward} could provide $u(t, x)$, $Du(t, x)$ for any $(t, x)$, but we have little guarantee that the value would be exact for instance for a $t$ that does not correspond to any of the $t_i$'s, or a combination $(t, x)$ never seen by the algorithm along trajectories.
\item Deep learning-based algorithms use the Euler discretization scheme to approximate $Y_{t_i}$, either as hard constraints as in \cite{han2017overcoming, weinan2017deep} or soft constraints as in \cite{raissi2018forward} using the loss \eqref{eq:raissiloss}.
\item An Euler scheme might not be necessary for computing $X_{t_i}$ from $W$, when the SDE \eqref{eq:SDE} can be solved exactly.
\end{enumerate}
The algorithms \cite{han2017overcoming,weinan2017deep} and \cite{raissi2018forward} can solve semilinear PDEs along trajectories with a fixed initial condition $X_0$. In \cite{han2017overcoming}, the authors point out that the algorithm can be adapted to a varying initial $X_0$ by adding a supplementary neural network $X_0 \mapsto (Y_0, \kappa_{t_0})$ at the left of the network described in Figure \ref{fig:jentzen_graph}.

\cite{han2017overcoming,weinan2017deep} and \cite{raissi2018forward} use the same discretized integration scheme, but with different formulations that may lead to different behaviors. However, by enforcing integration constraints, the formulation presented in \cite{han2017overcoming,weinan2017deep} helps the algorithm convergence by reducing the complexity of the function estimated by the neural networks. From our tests, we find that the convergence of the formulation proposed by \cite{han2017overcoming,weinan2017deep} is overall faster. Moreover, the latter formulation allows us to use recurrent schemes, as described in the Section \ref{sec:sharingparameters}. In the present work we adopt the latter formulation.

\section{Different neural networks architectures}
\label{sec:other_architectures}

To our knowledge, the various DBSDE solver only uses standard fully-connected (FC) feed forward neural networks. The use of different network architectures might improve the results: it is the case for instance in other areas like computer vision with convolutional neural networks, or natural language processing with recurrent neural networks. In our case, the use of specific structures could 1) limit the growth of the number of networks weights to be estimated when, for example the time discretization increases and 2) improve the convergence of the optimization algorithm, for instance by reducing numerical instabilities known in machine learning as ``vanishing gradients'' or ``exploding gradients''. In the following sections, we propose several neural network architectures. Our neural networks approximate $\kappa_{t_i} \simeq Du(t_i, X_{t_i})$ instead of $\sigma^\top (t_i, X_{t_i}) Du(t_i, X_{t_i})$ as in \cite{han2017overcoming,weinan2017deep}, as we find that doing this does not change significantly the behaviour of the algorithms while enabling some equations to be formulated more conveniently. In the following, $Y_{t_i}$ denotes the approximation of $u(t_i, X_{t_i})$ by the different algorithms. Apart from the neural networks architecture, we investigate three different improvements:

\begin{itemize}
    \item First, using \emph{batch normalization} \cite{ioffe2015batchnormalization} may help networks to learn faster and better. This technique is difficult to adapt to recurrent neural networks (or if so, only handles stationary signals by not sharing the moving statistics through time, as in \cite{cooijmans2016recurrent} for instance). Batch normalization is used in \cite{han2017overcoming,weinan2017deep}. 
    \item Second, using the ELU (\emph{exponential linear unit}) activation function may accelerate the learning process \cite{clevert2015fast}. It appears that it also reduces the need for regularization or batch normalization.
    \item Third, using residual learning, which consists in making identity shortcut connections between several hidden layers, may help accelerate the learning process by reducing numerical instabilities such as vanishing gradients in very deep neural networks \cite{he2016residuallearning}.
\end{itemize}
These improvements are described in Section \ref{sec:improvments}.

\subsection{Architectures building a different neural networks for each time step}
\label{sec:improvments}

\begin{figure}[ht]
\begin{adjustwidth}{-1in}{-1in}
\centering
\begin{tabular}{|P{4cm}|P{4cm}|P{4cm}|}
\hline
\includegraphics[scale=0.6]{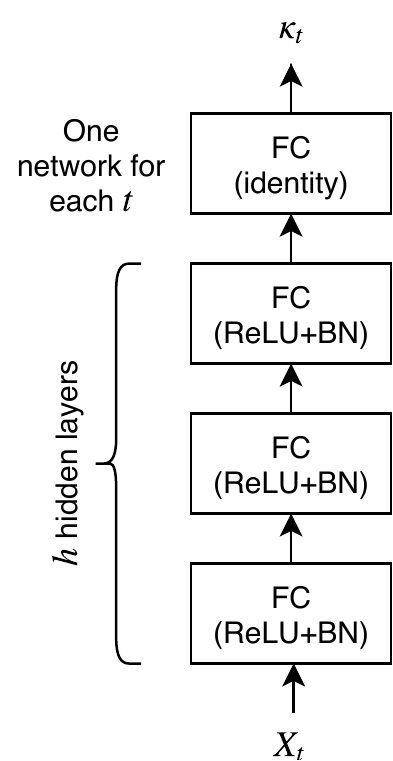} &
\includegraphics[scale=0.6]{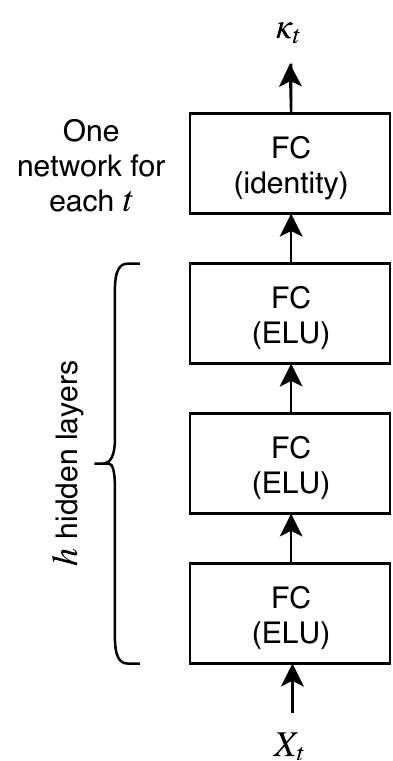} &
\includegraphics[scale=0.6]{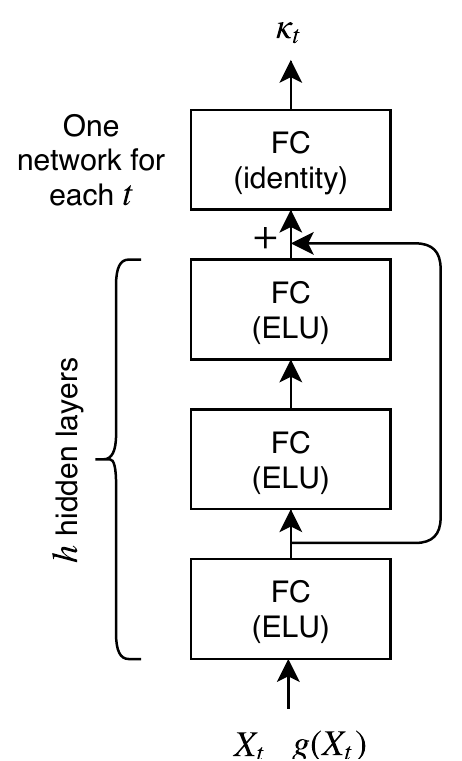} \\
\hline
\textbf{a.} FC DBSDE & 
\textbf{b.} FC ELU & 
\textbf{c.} FC Residual \\
\hline
\end{tabular}
\end{adjustwidth}
\caption{Using one different network for every timestep. Each network is composed of $h$ hidden layers and an output layer with identity activation. The network \textbf{a.} is the one used in \cite{han2017overcoming, weinan2017deep}, with \emph{fully-connected} (FC) hidden layers with ReLU activation function and \emph{batch normalization} (BN), but we find that not using BN on the output layer yields better results. Compared to network \textbf{a.}, network \textbf{b.} uses ELU activation functions instead of ReLU and do not implement batch normalization. Finally, network \textbf{c.} also takes as input $g(X_t)$ and adds residual connections every $2$ hidden layers starting from the output of the first hidden layer (if the number of hidden layers $h$ is even, then the last residual connection only skips one hidden layer). Adding $Y_t$ as an input to these networks makes the optimization algorithm unstable.} \label{networks:fc}
\end{figure}

The DBSDE solvers in \cite{han2017overcoming,weinan2017deep,beck2017machinela} use a different fully-connected feed forward neural network at each time step $t_i$ to estimate the gradients at time $t_i$, as described in Figure \ref{fig:jentzen_graph}. Each of these networks take as input $X_{t_i}$. Formally, each $\kappa_{t_i}$ is estimated with a specific network $\mathcal{N}_i : x \in \R^d \mapsto \kappa \in \R^d$ with parameters $\theta_i$:
\begin{align}
\kappa_{t_i} &:= \mathcal{N}_{i}^{\theta_i} (X_{t_i}). 
\end{align}
We modify the original DBSDE network by:
\begin{itemize}
    \item not using batch normalization in the final layer (see Figure \ref{networks:fc}). We find that this improves the final results compared to the original network. This network is referred to as the \textit{FC DBSDE} network. 
    \item not using batch normalization and changing the activation function from a ReLU to an ELU. This corresponds to \textit{FC ELU} network in the following.
    \item adding residual connections, i.e. adding identity shortcut connections between several hidden layers. We call this network \textit{FC Residual} network. 
\end{itemize}
The corresponding networks \textbf{a.}, \textbf{b.} and \textbf{c.} in our comparative study are presented in Figure \ref{networks:fc}.

\subsection{Architectures building one single neural network for all the time steps}

In this Section we propose two architectures which share the neural networks parameters through time. Our objectives are twofold: on the one hand, we want to reduce the number of weights to be estimated by the gradient descent, and on the other hand, we want to add some regularity in the estimated gradient. On the latter point, even if the gradient is not stationary in time one can expect that for sufficiently regular solutions, the gradient between two close time steps should be close for a given $x$.

Note that in order to build a single network shared through all the time steps, we have to add a dimension (namely the time dimension) to the problem to handle non-stationarities. We thus address whether it is faster and/or more accurate to estimate the weights of $N$ neural networks having $d$ input features as in Section \ref{sec:improvments} or to estimate the weights of one single neural network having $d + 1$ input features.

\subsubsection{Sharing parameters through time}\label{sec:sharingparameters}

\begin{figure}[ht]
\centering
\includegraphics[height=4.5cm]{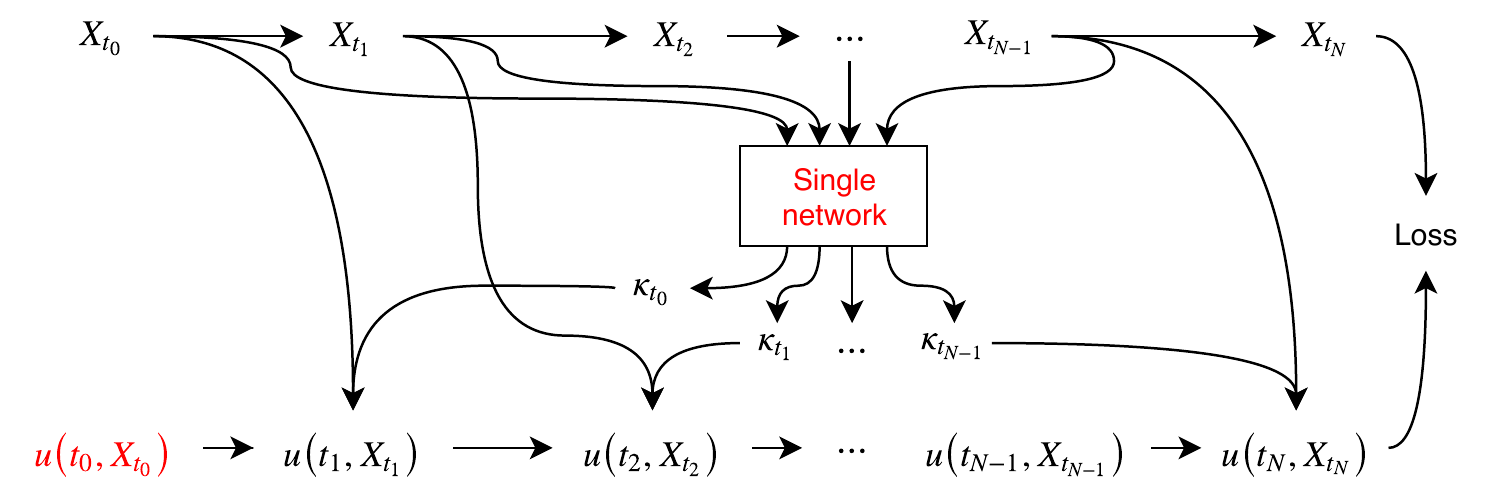}
\caption{Graph with common parameters. The parameters of the graph are represented in \textcolor{red}{red}.}
\label{fig:merge}
\end{figure}

Similarly as \cite{raissi2018forward} (but on the gradient), we propose to share the parameters of the networks for each time step, i.e. we use a single network $\mathcal{N}^\theta : (t, x) \in \R^{d+1} \mapsto \kappa \in \R^d$ instead of $N$ networks defined on $\R^{d}$ (see Figure \ref{fig:merge}):
\begin{align}
\kappa_{t_i} &:= \mathcal{N}^\theta (t_i, X_{t_i}).
\end{align}
This architecture should be easier to optimize, since the parameters are linked more closely to the loss function. Note that we cannot use \emph{batch normalization} with this formalism as the distribution of $X$ is likely to be non-stationary. $\kappa_{t_0}$ is also obtained as an output of the network. In what follows, this architecture is referred to as the \textit{Merged Deep BSDE}. 

Moreover, we find it helpful to feed the neural networks not only with $X_t$, but also with other variables known at instant $t$, such as $Y_t$ and $g(X_t)$: thus, we write 
\begin{align}
\kappa_{t_i} &:= \mathcal{N}^\theta (t_i, X_{t_i}, Y_{t_i}, g(X_{t_i})).
\end{align}
If we consider $Y_{t_i}$ to be the output of the neural structure at time $t_i$, then the previous formulation is a \emph{recurrent neural network}: $Y_{t_i}$ depends directly on the output of a previous call to the neural network and is fed as input to the following call of the neural network.

\begin{figure}[ht]
\begin{adjustwidth}{-1in}{-1in}
\centering
\begin{tabular}{|P{4cm}|P{4cm}|P{4cm}|}
\hline
\includegraphics[scale=0.6]{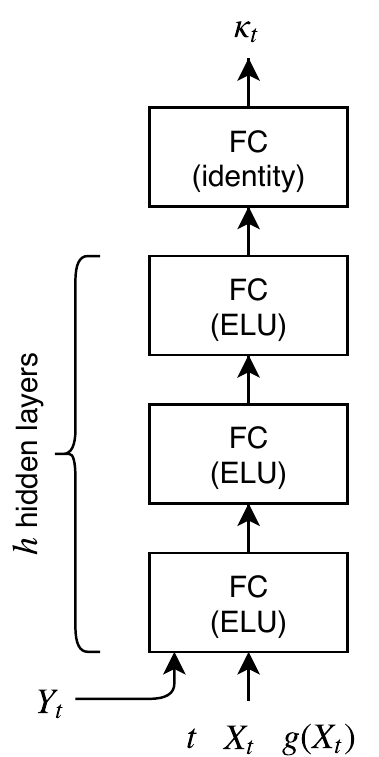} &
\includegraphics[scale=0.6]{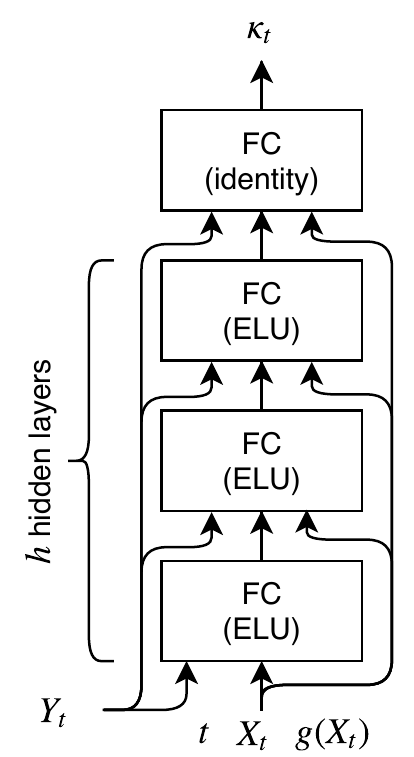} &
\includegraphics[scale=0.6]{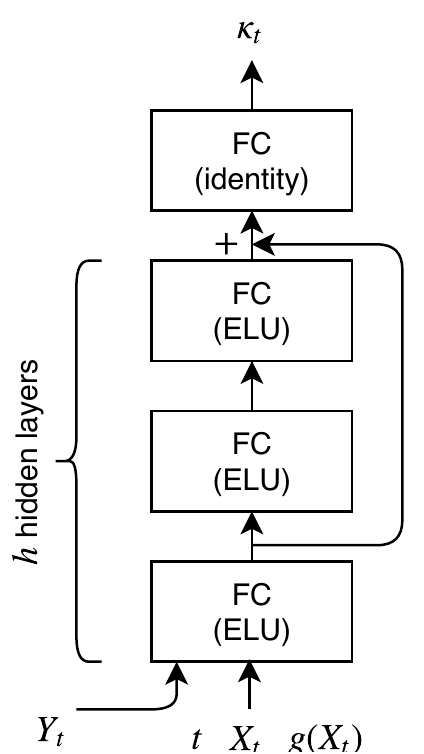} \\
\hline
\textbf{d.} FC Merged & 
\textbf{e.} FC Merged Shortcut &
\textbf{f.} FC Merged Residual \\
\hline
\end{tabular}
\end{adjustwidth}
\caption{Sharing the same network parameters for every timestep. Each network is composed of $h$ hidden layers and an output layer with identity activation. Network \textbf{d.} takes as input $t$, $X_t$, $g(X_t)$ and $Y_t$, and uses ELU activation functions. Network \textbf{e.} is the same but injects the input layer as a supplementary input to each hidden layer -- we call these \emph{shortcut connections}. Finally, network \textbf{f.} is the same as \textbf{d.} but adding \emph{residual connections} every several hidden layers, like in network \textbf{c.}.}
\label{networks:fcmerged}
\end{figure}

The corresponding networks in our comparative study are presented in Figure \ref{networks:fcmerged}.

\subsection{Adding a temporal structure with LSTM networks}

Due to Euler discretization error, the target  $g(X_T)$ cannot be reached exactly and the loss function cannot be perfectly zeroed. By allowing $\kappa_t$ to depend not only on variables realization at date $t$ but also on long and short term dependencies, we might counteract discretization errors and find some strategies with a smaller loss than with simple fully-connected feed forward network.

Recurrent neural networks with memory, LSTM networks for instance \cite{article:lstm, blog:colah2015understanding}, are networks which use an internal state to build short and long-term dependencies when applied to a sequence. These networks proved very efficient in performing tasks on sequences \cite{blog:karpathy2015unreasonablernn}. Formally, if $m_{t} \in \R^p$ is the state at time $t$, the equation would write

\begin{align}
(\kappa_{t_i}, m_{t_i}) := \mathcal{N}^\theta(t_i, X_{t_i}, m_{t_{i-1}})
\end{align}
where $m_{t_{i-1}}$ is a parameter and $\kappa_{t_0}$ is an output of the network. Using these types of networks might enable the network to build its own input feature through $m_t$, as well as compensating for long term effects such as the discretization error.

\begin{figure}[ht]
\begin{adjustwidth}{-1in}{-1in}
\centering
\begin{tabular}{|P{4cm}|P{4cm}|P{4cm}|P{4cm}|}
\hline
\includegraphics[scale=0.6]{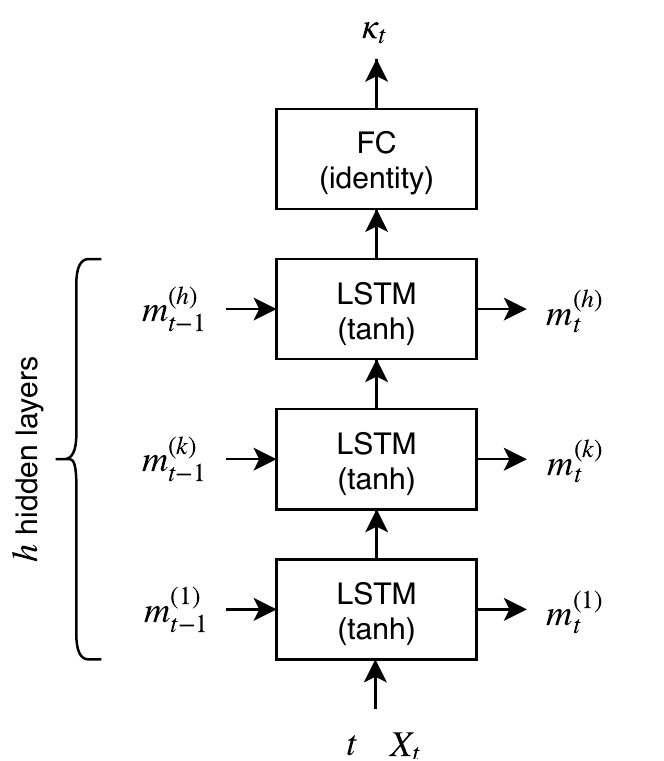} &
\includegraphics[scale=0.6]{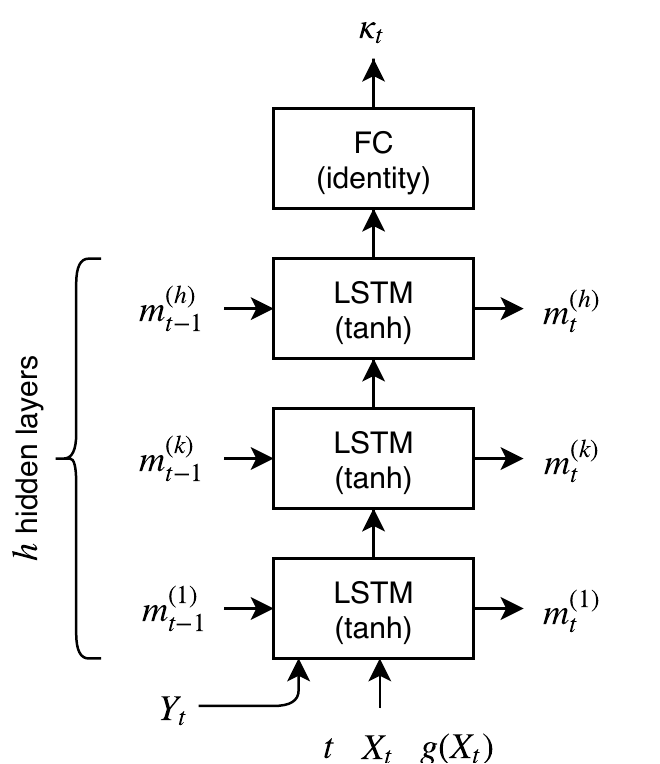} &
\includegraphics[scale=0.6]{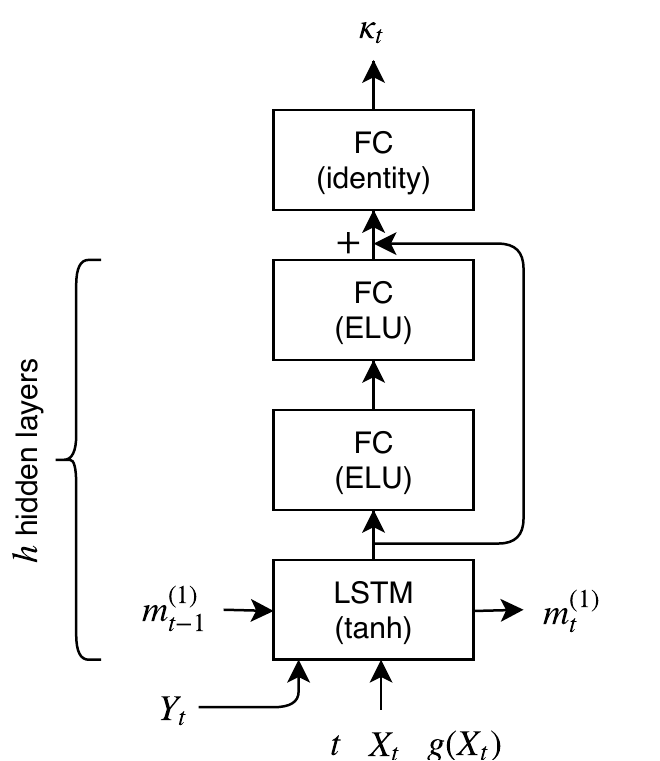} &
\includegraphics[scale=0.6]{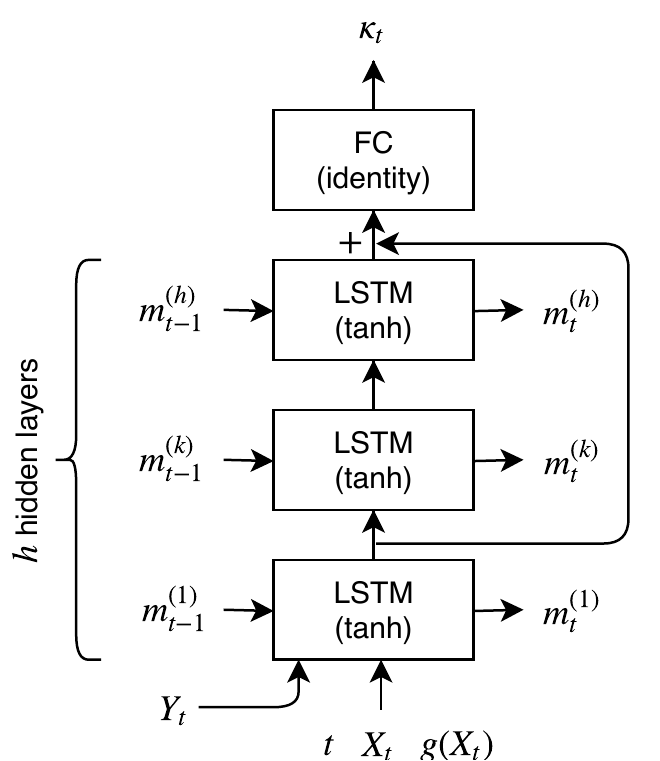} \\
\hline
\textbf{g.} LSTM & 
\textbf{h.} Augmented LSTM & 
\textbf{i.} Hybrid LSTM & 
\textbf{j.} Residual LSTM \\
\hline
\end{tabular}
\end{adjustwidth}
\caption{LSTM-based algorithms. Network \textbf{g.} is composed of $h$ stacked LSTM, which takes $t$ and $X_t$ as input, with a state $m_{t} = (m_{t}^{(1)}, \cdots, m_{t}^{(h)})$, and an output layer with identity activation function. Network \textbf{h.} is the same network but taking $Y_t$ and $g(X_t)$ as supplementary inputs. Network \textbf{i.} is a combination of the Merged network \textbf{f.}, but replacing the first hidden layer by a LSTM -- thus the network is composed of $1$ hidden LSTM layer and $h-1$ hidden FC (ELU) layers. Finally, network \textbf{j.} is the same as \textbf{h.} but adding residual connections every few hidden layers, as described for network \textbf{f.}.} 
\label{networks:lstm}
\end{figure}

Similarly, we can feed the neural network with other variables at instant $t_i$ :

\begin{align}
(\kappa_{t_i}, m_{t_i}) := \mathcal{N}^\theta(t_i, X_{t_i}, Y_{t_i}, g(X_{t_i}), m_{t_{i-1}}).
\end{align}
The corresponding networks in our comparative study are presented in Figure \ref{networks:lstm}.

\clearpage
\section{Two new machine learning algorithms}
\label{sec:new_algorithms}

The previously described algorithms are based on an Euler scheme with a time step discretization and try to estimate the function value and its derivative at these discrete values.

We propose two algorithms not relying on an Euler scheme for the BSDE but which try to estimate the global function $u$ as a functional of $t \in [0,T]$ and $x \in \R^d$. Note that the Euler scheme is still necessary to calculate the forward process \eqref{eq:SDE} when it cannot be exactly simulated.

In this Section  $ \rho(x)=\lambda^u x^{u-1} \frac{e^{-\lambda x}}{\Gamma(u)}$  is the density of a random variable with a Gamma law. The associated  cumulative distribution function is   
\begin{equation}
F(x) =\frac{\gamma(u,\lambda x)}{\Gamma(u)},
\label{eq:CCF}
\end{equation}
where $\gamma(s, x) =  \int_0^x t^{s-1} e^{-t} dt$ is the incomplete gamma function. Let us denote  $$ \Fb(t):= 1-F(t).$$

Note that $u=1$ corresponds to the case of the exponential law that we will use in practice in all our experiments.

Denoting by $\E_{t,x}$ the expectation operator conditional on  $X_t=x$ at time $t\leq T$, the representation of the solution $u$ from the Feynman-Kac formula (valid under regularity assumptions on the terminal function and the coefficients of equation \eqref{eq:SDE}) is given by:
\begin{flalign}
u(t,x) 
= & 
\E_{t,x} \Big[ \Fb(T-t)\frac{g(X_T)}{ \Fb(T-t)}+\int_t^T \frac{f(s,X_{s},u(s,X_{s}), Du(s, X_s))}{\rho(s-t)}\rho(s-t)ds\Big]
\label{eq:U}
\end{flalign}

Introducing a random variable $\tau$ of density $\rho$ we get 
\begin{flalign}
u(t,x)  =&
\E_{t,x} \big[\phi\big(t, t+\tau ,X_{t+\tau}, u(t+\tau ,X_{t+\tau}), Du(t+\tau ,X_{t+\tau}) \big)\big],
\label{eq:feynman}
\end{flalign}
with
\begin{flalign*}
\phi(s, t,x,y,z) &:= \frac{\1_{\{t\ge T\}}}{ \Fb(T-s)}g(x)\!+\! \frac{\1_{\{t<T\}}}{\rho(t -s)} f(t,x,y,z).
\end{flalign*}

\begin{figure}[ht]
\centering
\includegraphics[scale=0.8]{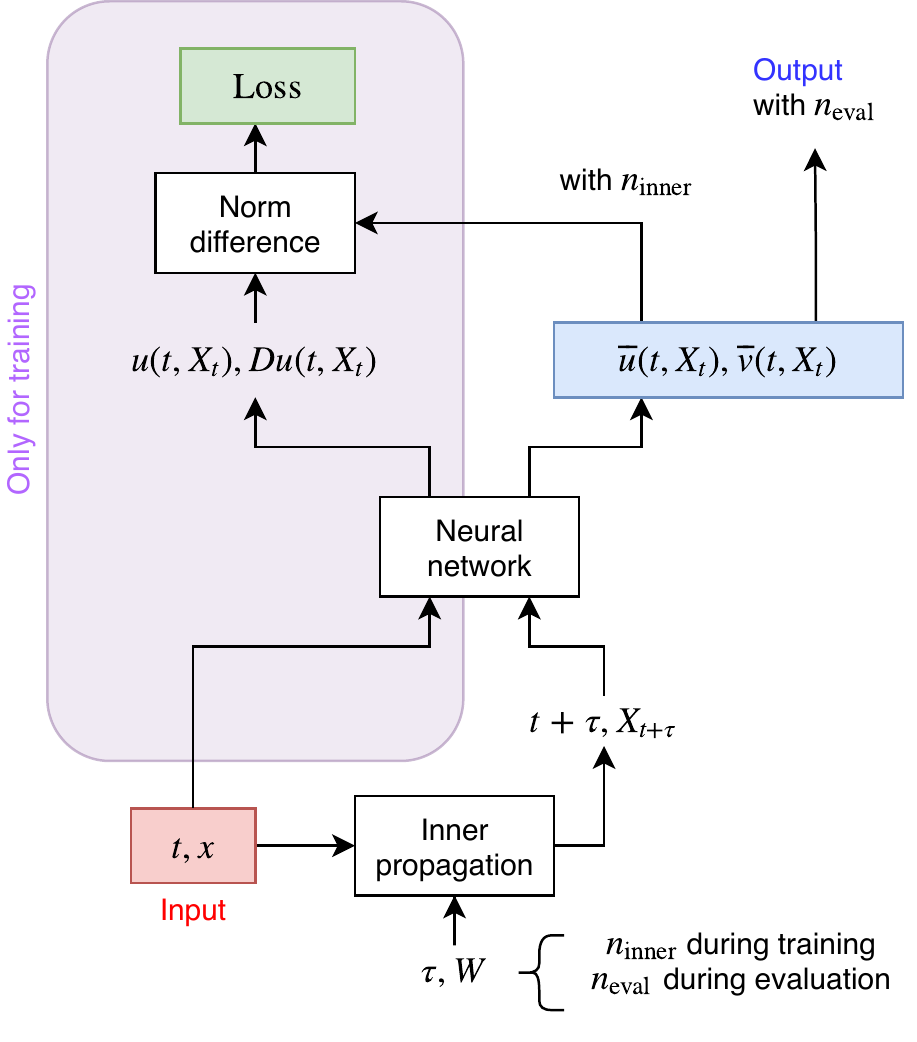} 
\caption{Our new machine learning algorithm. First, some $(t, x)$ are given as inputs. During training, the discretized operator $T$ is computed using a number $n_\text{inner}$ of realizations of $\tau, W$. The loss is the norm difference between $(\bar{u}, \bar{v})$, and $(u, Du)$. During evaluation, the discretized operator $T$ is computed using $n_\text{eval}$ realizations of $\tau, W$ and the corresponding $(\bar{u}, \bar{v})$ are outputted.}
\label{algo:deepnesting}
\end{figure}

We then propose two schemes to solve the problem. An informative diagram is presented in Figure \ref{algo:deepnesting}.

 \subsection{A first scheme}
 \label{sec:firstSchme}
 Following the idea of \cite{henry2016branching,warin2018nestingsMC}
 we explain how to calculate  $Du(t,x)$ using Malliavin weights. We isolate two cases:
 
 \begin{itemize}
 \item If the coefficients $\mu$ and $\sigma$ are constant, then the process $X_{t+\tau}^{t,x}$  solution of equation \eqref{eq:SDE} is given by
 \begin{align}
 \label{eq:SDE_EUL1}
 X_{t+\tau}^{t,x} = x + \mu \tau + \sigma (W_{t+\tau} -W_{t})
 \end{align}
 and we define the antithetic variable :
 \begin{align}
 \label{eq:SDE_EUL1ANT}
 \widehat{X}_{t+\tau}^{t,x} = x + \mu \tau - \sigma (W_{t+\tau} -W_{t}).
 \end{align}
 \item 
 When the coefficients are not constant, an Euler Scheme with a step $\Delta t$  is still necessary. In this case, we denote $J = \lfloor \frac{\tau}{\Delta t} \rfloor$. If $\tau < \Delta t$ then equations \eqref{eq:SDE_EUL1} and \eqref{eq:SDE_EUL1ANT} are used, otherwise
 \begin{align}
 X_{t+ (i+1) \Delta t}^{t,x} = & X_{t +i \Delta t}^{t,x} + \mu(t+ i \Delta t, X_{t +i \Delta t}^{t,x}) \Delta t + \sigma(t+ i \Delta t, X_{t +i \Delta t}^{t,x}) (W_{t+ (i+1) \Delta t}-W_{t+ i \Delta t}) , \quad  i=0, J-1 \nonumber \\
 X_{t+ \tau }^{t,x} = & X_{t+J \Delta t}^{t,x} + \mu(t+ J \Delta t, X_{t +J \Delta t}^{t,x}) (\tau -J \Delta t) + \sigma(t+ J \Delta t, X_{t +J \Delta t}^{t,x}) (W_{t+ \tau}-W_{t+ J \Delta t})
 \label{eq:EL}
 \end{align}
 and $\widehat{X}$ is defined by
 \begin{align}
 \widehat{X}_{t+ \Delta t}^{t,x} = & x + \mu(t, x) \Delta t - \sigma(t, x) (W_{t+  \Delta t}-W_{t}), \\
 \widehat{X}_{t+ (i+1) \Delta t}^{t,x} = & \widehat{X}_{t +i \Delta t}^{t,x} + \mu(t+ i \Delta t, \widehat{X}_{t +i \Delta t}^{t,x}) \Delta t + \sigma(t+ i \Delta t, \widehat{X}_{t +i \Delta t}^{t,x}) (W_{t+ (i+1) \Delta t}-W_{t+ i \Delta t}) , \quad  i=1, J-1 \nonumber \\
 \widehat{X}_{t+ \tau }^{t,x} = & \widehat{X}_{t+J \Delta t}^{t,x} + \mu(t+ J \Delta t, \widehat{X}_{t +J \Delta t}^{t,x}) (\tau -J \Delta t) + \sigma(t+ J \Delta t, \widehat{X}_{t +J \Delta t}^{t,x}) (W_{t+ \tau}-W_{t+ J \Delta t}).
 \label{eq:ELANT}
 \end{align}
 \end{itemize}
 As defined in \cite{warin2018monte}, an estimator of the gradient of $u$ is given by:
 \begin{align}
 Du(t,x) = & \E_{t,x} \Big[ \sigma^{-\top} \frac{W_{(t+ (\tau \ \wedge \Delta t) ) \wedge T}-W_{t}}{\tau \wedge (T-t) \wedge \Delta t} \frac{1}{2} \big( \phi(t,t+\tau, X_{t+\tau}^{t,x}, u(t+\tau, X_{t+\tau}^{t,x}), Du(t+\tau, X_{t+\tau}^{t,x})) -  \nonumber \\
 &  \phi(t,t+\tau, \widehat{X}_{t+\tau}^{t,x}, u(t+\tau, \widehat{X}_{t+\tau}^{t,x}), Du(t+\tau, \widehat{X}_{t+\tau}^{t,x})) \big) \big].
 \label{eq:DU}
 \end{align}
 \begin{remark}
 When the process can be exactly simulated, the Euler scheme can be avoided: the Malliavin weights have to be modified accordingly as shown for example in the numerical example in \cite{warin2018monte} for an Ornstein-Uhlenbeck process. 
In this case the previous equation \eqref{eq:DU} is adapted by taking $\Delta t = \tau$.
\end{remark}

For two metric spaces $E$, $F$, we introduce $Lip(E,F)$ the set of Lipschitz continuous functions defined on $E$ with values in $F$.

Following the ideas of  \cite{hutzenthaler2017multi}, we  introduce  the operator $T :Lip([0,T] \times \R^d, \R^{d+1}) \longrightarrow Lip([0,T] \times \R^d, \R^{d+1})$, such that to $( u, v) \in ( Lip([0,T] \times \R^d, \R) \times Lip([0,T] \times \R^d, \R^d))$ is  associated $(\bar u, \bar v)$ solution of
\begin{align}
\bar u & = \frac{1}{2}\E_{t,x} \big[ \phi\big(t, t+\tau ,X_{t+\tau}^{t,x}, u(t+\tau ,X_{t+\tau}^{t,x}), v(t+\tau ,X_{t+\tau}^{t,x}) \big) + \phi\big(t, t+\tau ,\widehat{X}_{t+\tau}^{t,x}, u(t+\tau ,\widehat{X}_{t+\tau}^{t,x}), v(t+\tau ,\widehat{X}_{t+\tau}^{t,x}) \big) \big] \nonumber \\
\bar v & = \E_{t,x} \Big[ \sigma^{-\top} \frac{W_{(t+ (\tau  \wedge \Delta t )) \wedge T}-W_{t}}{\tau \wedge (T-t) \wedge \Delta t} \frac{1}{2} \big( \phi(t,t+\tau, X_{t+\tau}^{t,x}, u(t+\tau, X_{t+\tau}^{t,x}), v(t+\tau, X_{t+\tau}^{t,x})) -  \nonumber \\
& \quad  \phi(t,t+\tau, \widehat{X}_{t+\tau}^{t,x}, u(t+\tau, \widehat{X}_{t+\tau}^{t,x}), v(t+\tau, \widehat{X}_{t+\tau}^{t,x})) \big) \big].
\label{eq:fixedPoint}
\end{align}

Instead of trying to solve the problem with some fixed point iteration as in \cite{hutzenthaler2017multi}, we propose to solve the problem with a machine learning technique
by defining the loss function $\ell$ for
$U \in Lip([0,T] \times \R^d, \R^{d+1})$ by:
\begin{align}
\ell(U) =  \E[ || U -T U||^2  ]
\label{eq:lossNewAlgo}
\end{align}
Note that the operator $T$ necessitates to calculate an expectation involved in the loss function.
This expectation may be calculated with only a few thousand samples $n_\text{inner}$ by a Monte Carlo approximation.\\
To be more explicit, we sample once for all
the $\tau$ and $W$ Brownian increments that appear in   \eqref{eq:NewAlgoDis}, and we suppose that $(u(\theta, t,x)  ,v(\theta,t,x) )  := \mathcal{N}^\theta(t, X_{t})$ so that $u$ is a  $\R$ valued  function and $v$ a $\R^d$ valued function parameterized by a single neural network with parameters $\theta$.
We introduce the discrete version of equation \eqref{eq:fixedPoint}:
\begin{align}
\bar u(\theta,t,x) = & \frac{1}{2}\frac{1}{n_\text{inner}} \sum_{i=1}^{n_\text{inner}} \big[\phi\big(t, t+\tau^i ,X_{t+\tau^i}^{t,x,i}, u(\theta,t+\tau^i ,X_{t+\tau^i}^{t,x,i}), v(\theta,t+\tau^i  ,X_{t+\tau^i}^{t,x,i}) \big) +  \nonumber \\
& \quad \phi\big(t, t+\tau^i  ,\widehat{X}_{t+\tau^i }^{t,x,i}, u(\theta,t+\tau^i  ,\widehat{X}_{t+\tau^i }^{t,x,i}), v(\theta+t+\tau ,\widehat{X}_{t+\tau^i }^{t,x,i}) \big)\big] \nonumber \\
\bar v(\theta,t,x) = & \frac{1}{2}\frac{1}{n_\text{inner}} \sum_{i=1}^{n_\text{inner}} \Big[ \sigma^{-\top} \frac{W_{(t+(\tau^i   \wedge \Delta t) ) \wedge T}^i -W_{t}^i }{\tau^i  \wedge (T-t) \wedge \Delta t} \big( \phi(t,t+\tau^i , X_{t+\tau^i}^{t,x,i},  u(\theta,t+\tau^i, X_{t+\tau^i}^{t,x,i}), v(\theta,t+\tau^i, X_{t+\tau^i}^{t,x,i})) -  \nonumber \\
& \quad  \phi(t,t+\tau^i, \widehat{X}_{t+\tau^i}^{t,x,i}, u(\theta,t+\tau^i, \widehat{X}_{t+\tau^i}^{t,x,i}), v(\theta,t+\tau, \widehat{X}_{t+\tau^i}^{t,x,i})) \big) \big] 
\label{eq:NewAlgoDis}
\end{align}

We introduce $\zeta$ an uniform random variable on $[0,T]$ and
$X^{0,x}_{\zeta}$ a random variable obtained as the solution of equation \eqref{eq:SDE} potentially with an Euler scheme. Then the loss function is defined by :
\begin{align}
\ell(\theta) &:= \mathbb{E} \left[  \left(\bar u(\theta, \zeta ,X^{0,x}_{\zeta}) - u(\theta,\zeta ,X^{0,x}_{\zeta}) \right)^2 +
\sum_{j=1}^d \left(\bar v_j(\theta, \zeta ,X^{0,x}_{\zeta}) - v_j(\theta,\zeta ,X^{0,x}_{\zeta}) \right)^2 \right] \label{eq:LossFirstScheme}
\end{align}

Ideally, the number of samples  $n_\text{inner}$ used in equation \eqref{eq:NewAlgoDis} should be very high to limit the bias in calculating $\tilde u$, $\tilde v$.
However, the number of terms in the loss function grows up linearly with $n_\text{inner}$ and this leads to an increase in computing time and memory usage in TensorFlow: the automatic differentiation computing cost to calculate the gradient grows up at least  linearly with the number of terms.

 However this representation of the solution allows to get a solution $u, Du$ on $[0,T] \times \R^d$. 
 Once the convergence with the machine learning algorithm is achieved, we get a representation $u$ and $Du$ on $[0,T] \times \R^d$ but
 with a limited number of inner samples, so that a bias is present.

One way to reduce the bias consists in repeating the calculation with different values drawn for the inner samples. A more effective way to get a better estimation of $u$ and $Du$ at point $x$ and date $0$ (or any pointwise estimation) consists in achieving a post-processing: from a very high number of particles $n_\text{eval} \gg n_\text{inner}$ we evaluate $u(0,x)$ and its derivative $Du(0,x)$ by replacing $n_\text{inner}$ by $n_\text{eval}$ in equation \eqref{eq:NewAlgoDis}. The idea is that equation \eqref{eq:fixedPoint} gives the solution as a sum of a function of $f$ involving $u$ and $v$  and a function depending on the terminal value so independent on the error made on  $u$ and $v$ using a limited number of samples in \eqref{eq:NewAlgoDis}. It seems natural to use a processing with a very high number of trajectories that will permit at least to kill the bias on the second term independent on $u$ and $v$.

\subsection{A second scheme}
\label{sec:secondScheme}

In the second scheme, the gradient is not parameterized independently by the neural network, but rather obtained directly by differentiating the $u$ function using TensorFlow. Noting $\widehat{D} u$ the TensorFlow automatic differentiation operator applied to the function $u$, the operator $T:Lip([0,T] \times \R^d, \R) \longrightarrow Lip([0,T] \times \R^d, \R)$  associates $u \in Lip([0,T] \times \R^d, \R)$ to $\bar u,$ solution of
\begin{align}
\bar u(t,x) = & \frac{1}{2}\E_{t,x} \big[\phi\big(t, t+\tau ,X_{t+\tau}^{t,x}, u(t+\tau ,X_{t+\tau}^{t,x}), \widehat{D} u(t+\tau ,X_{t+\tau}^{t,x}) \big) + \nonumber \\
& \phi\big(t, t+\tau ,\widehat{X}_{t+\tau}^{t,x}, u(t+\tau ,\widehat{X}_{t+\tau}^{t,x}), \widehat{D} u(t+\tau ,\widehat{X}_{t+\tau}^{t,x}) \big)\big] 
\label{eq:fixedPointBis}
\end{align}

As in the previous algorithm, the equation \eqref{eq:fixedPointBis} is discretized with a given number of samples $\tau$, $W_\tau$ chosen once for all. The function  $u$ is approximated by a neural network: $u(\theta, t,x)  := \mathcal{N}^\theta(t, X_{t})$  and the previous operator is approximated by:
\begin{align}
\bar u(\theta,t,x) = & \frac{1}{2}\frac{1}{n_\text{inner}} \sum_{i=1}^{n_\text{inner}} \big[\phi\big(t, t+\tau^i ,X_{t+\tau^i}^{t,x,i}, u(\theta,t+\tau^i ,X_{t+\tau^i}^{t,x,i}), \widehat{D}u(\theta,t+\tau^i  ,X_{t+\tau^i}^{t,x,i}) \big) +  \nonumber \\
& \quad \phi\big(t, t+\tau^i  ,\widehat{X}_{t+\tau^i }^{t,x,i}, u(\theta,t+\tau^i  ,\widehat{X}_{t+\tau^i }^{t,x,i}), \widehat{D}u(\theta,t+\tau ,\widehat{X}_{t+\tau^i }^{t,x,i}) \big)\big]
\label{eq:secondScheme}
\end{align}
Then we could use a loss function only involving $u$ :
\begin{align}
\ell(\theta) &:= \mathbb{E} \left[  \left(\bar u(\theta, \zeta ,X^{0,x}_{\zeta}) - u(\theta,\zeta ,X^{0,x}_{\zeta}) \right)^2 \right] \label{eq:LossSecondScheme1}
\end{align}

As we will show in Section \ref{sec:comp_new_arch}, it turns out that we can achieve better results by also computing an estimator for $Du$:
\begin{align}
\bar v(\theta,t,x) & = \E_{t,x} \Big[ \sigma^{-\top} \frac{W_{(t+ (\tau  \wedge \Delta t )) \wedge T}-W_{t}}{\tau \wedge (T-t) \wedge \Delta t} \frac{1}{2} \big( \phi(t,t+\tau, X_{t+\tau}^{t,x}, u(\theta,t+\tau, X_{t+\tau}^{t,x}), \widehat{D}u(\theta,t+\tau, X_{t+\tau}^{t,x})) -  \nonumber \\
& \quad  \phi(t,t+\tau, \widehat{X}_{t+\tau}^{t,x}, u(\theta,t+\tau, \widehat{X}_{t+\tau}^{t,x}), \widehat{D}u(\theta,t+\tau, \widehat{X}_{t+\tau}^{t,x})) \big) \big]
\label{eq:SecondSchemevbar}
\end{align}
and including a term in $\widehat{D}u$ in the loss function:
\begin{align}
\ell(\theta) &:= \mathbb{E} \left[  \left(\bar u(\theta, \zeta ,X^{0,x}_{\zeta}) - u(\theta,\zeta ,X^{0,x}_{\zeta}) \right)^2 +
\sum_{j=1}^d \left( \bar{v}_j(\theta, \zeta ,X^{0,x}_{\zeta}) - \widehat{D} u_j(\theta,\zeta ,X^{0,x}_{\zeta}) \right)^2 \right] \label{eq:LossSecondScheme2}
\end{align}

Once the loss function is minimized and an estimation of $u$ is achieved, a more accurate estimation of $\bar u$ at date $0$ is achieved by solving equation \eqref{eq:fixedPointBis} with a very high number of simulations as in the first algorithm.

 \subsection{Presentation of the networks used}
\begin{figure}[H]
\begin{adjustwidth}{-1in}{-1in}
\centering
\begin{tabular}{|P{4.5cm}|P{4.5cm}|P{4.5cm}|}
\hline
\includegraphics[scale=0.6]{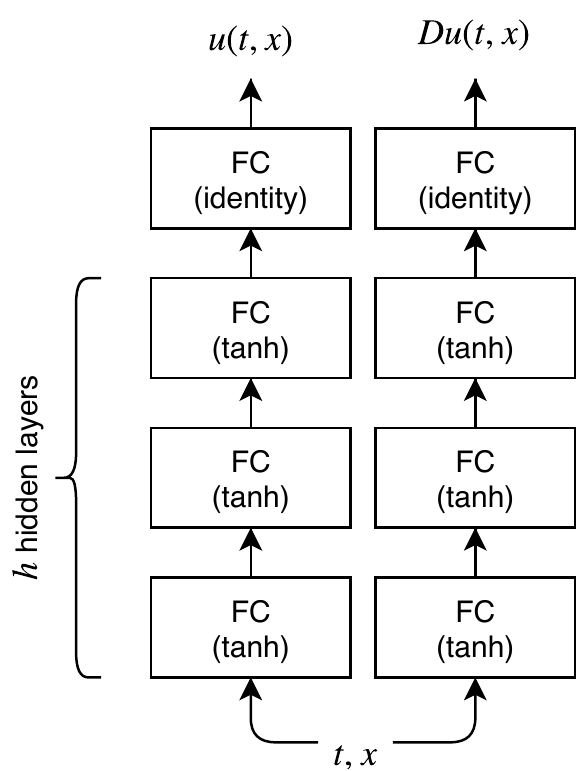} &
\includegraphics[scale=0.6]{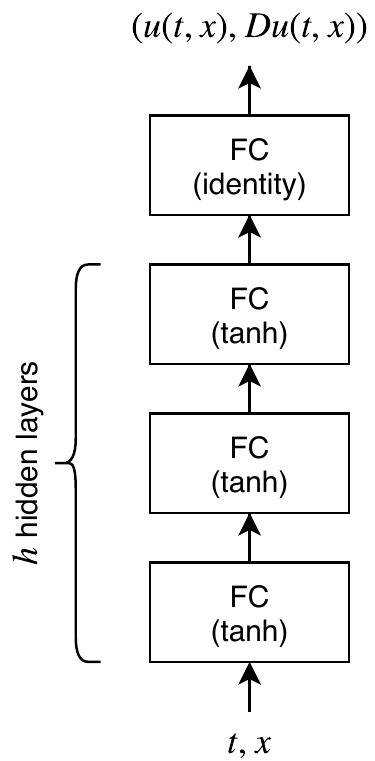} &
\includegraphics[scale=0.6]{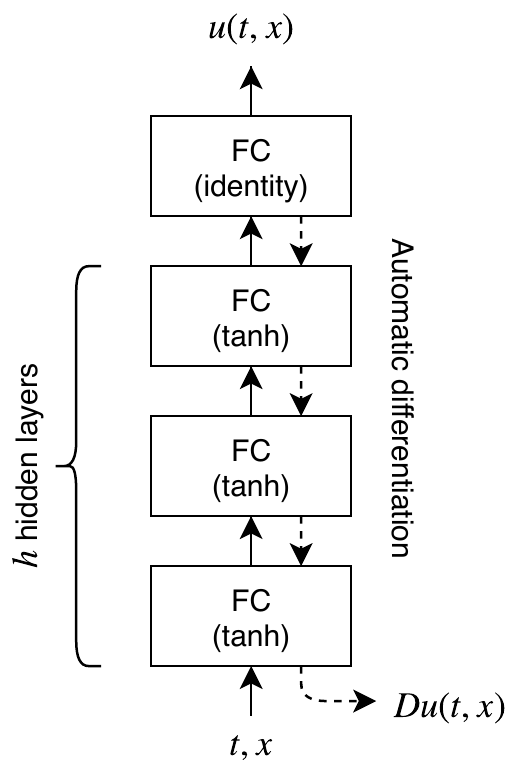} \\
\hline
\textbf{A} Separated & 
\textbf{B.} Shared &
\textbf{C.} Automatic differentiation \\
\hline
\end{tabular}
\end{adjustwidth}
\caption{Networks used for our new algorithm. \textbf{A.} Separated and \textbf{B.} Shared correspond to the first scheme and \textbf{C.} AutoDiff to the second scheme. Network \textbf{C bis.} is the same as \textbf{C.} but using the loss function \eqref{eq:LossSecondScheme1}.}
\label{networks:deepnesting}
\end{figure}

 The networks we chose to compare are represented in Figure \ref{networks:deepnesting}.

 \paragraph{First scheme} The network \textbf{A.} consists in two separate networks to approximate $u$ and $Du$ respectively. The network \textbf{B.} consists in a single network to approximate $u$ and $Du$ (the output is the concatenation of these two quantities). \textbf{A.} and \textbf{B.} use the loss function \eqref{eq:LossFirstScheme}.

 \paragraph{Second scheme} The network \textbf{C.} computes $Du$ using automatic differentiation and the loss function \eqref{eq:LossSecondScheme2}. We also compared the network \textbf{C bis.} which computes $Du$ using automatic differentiation but rather uses the loss \eqref{eq:LossSecondScheme1} with no cost on $Du$.

\subsection{Numerical and convergence consideration}

In all the examples, we use a feed forward network and used an exponential law for $\tau$ taking $u=1$ in equation \eqref{eq:CCF}.

From a numerical point of view, because of the number of terms involved in the loss function, the computation time cannot compete with the ones obtained by \cite{han2017solving}, but we hope it may be more accurate. Besides on numerical example we encounter cases in which the method \cite{han2017solving} fails to converge if the initial values are not initialized close to the solution. In our test, the two new algorithms  do not seem to have this problem.

Because we use a gradient method on a non convex problem we cannot prove that the method converges to the solution: the method might converge to a local minimum, but it seems less prove to this flaw than the method \cite{han2017solving}. However, it is possible to give some converging heuristic as done in \cite{sirignano2018dgm} neglecting the bias due to the use of a finite number of inner particles, i.e. supposing that the inner expectation is calculated exactly and that the dates $\tau$ follow a gamma law.

 We make the following assumptions:
 \begin{assumption}
 The SDE \eqref{eq:SDE} has constant coefficients
 \label{ass:SDE}
 \end{assumption}
 \begin{assumption}
 \label{ass::lipF}
 $f$ is  bounded, uniformly Lipschitz in $u$, $v$ with constant $K$ :
 \begin{flalign}
 | f(t, x, y, z) - f(t, x, \hat y,  \hat z) | \le K (|y-\hat y| + || z - \hat z||)  \quad  \forall t \in \R^{+}, x \in \R^d, (y, \hat y) \in \R^2, (z, \hat z) \in \R^d \times \R^d  
 \end{flalign}
 \end{assumption}
 \begin{assumption}
 \label{ass::gBound}
 $g$ is bounded.
 \end{assumption}
  \begin{assumption}
 \label{ass::uReg}
 Equation  \eqref{eq:PDEInit} has an unique solution $u \in C^{1,2}([0,T] \x \R^d)$, such that
 \begin{itemize}
 \item $u$ is $\theta$-H\"older with $\theta \in (0,1]$ in time with constant $\hat K$ :
 \begin{align*}
 | u(t,x) - u(\tilde t, x) | \le \hat K | t- \tilde{t}|^\theta \quad \quad \quad \forall (t,\tilde{t},x) \in [0,T] \times [0,T] \times \R^d,
 \end{align*}
 \item $u(t,x)$ has a quadratic growth in $x$ uniformly in $t$,
 \end{itemize}
 \end{assumption}

A small parameter $\epsilon$ being chosen, under assumption \ref{ass:SDE}, we can define a compact $\Omega^{t,x}_{\epsilon}$ such that the probability that the solution of the SDE \eqref{eq:SDE} on $[0,T]$ leaves the compact is  smaller than $\epsilon$. Then $P(  X_{\zeta}^{t,x} \in  \Omega^{t,x}_{\epsilon} \quad  \forall \zeta \in [t, T]) \ge 1 - \epsilon.$

We note $\psi$ the sigmoid function and we introduce the set of functions:
\begin{align*}
\kappa_n = \left \{ \xi \; : \; (t,x) \in \R^{1+d} \mapsto \R \ni \xi(t,x) = \sum_{i=1}^n \beta_i \psi\left( \alpha_{1,i} t + \sum_{j=2}^{d+1} \alpha_{j,i} x_j +c_i\right) \right \}
\end{align*}
where $\theta = (\beta_1, \dots, \beta_n, \alpha_{1,1}, \dots, \alpha_{d,n}, c_1, \dots, c_n) \in \R^{3n+ nd}$. We set $\kappa = \cup_{i}^\infty \kappa_i$.

 We suppose that $(u(\theta, t,x)  ,v(\theta,t,x) )  \in \kappa^{d+1}$  is  a parametrized  estimation of  the solution $(u, Du)$  and that equation \eqref{eq:fixedPoint} is slightly  modified: for $(\zeta, y) \in  [0, T] \times \Omega^{0,x}_\epsilon$ we localize the previous $(\bar u, \bar v)$ by
 \begin{align}
 \bar u^\epsilon(\theta,\zeta,y) = & \frac{1}{2}\E_{\zeta,y} \big[ \big\{ \phi\big(\zeta, \zeta+\tau ,X_{\zeta+\tau}^{\zeta,y}, u(\theta,\zeta+\tau ,X_{\zeta+\tau}^{\zeta,y}), v(\theta, \zeta+\tau ,X_{\zeta+\tau}^{\zeta,y}) \big) + \nonumber \\
 & \phi\big(\zeta, \zeta+\tau ,\widehat{X}_{\zeta+\tau}^{\zeta,y}, u(\theta,\zeta+\tau ,\widehat{X}_{\zeta+\tau}^{\zeta,y}), v( \theta, \zeta+\tau ,\widehat{X}_{\zeta+\tau}^{\zeta,y}) \big) \big\} 1_{X^{\zeta,y}_{\zeta+\tau} \in \Omega_\epsilon^{0,x}} \big] \nonumber \\
 \bar v^\epsilon(\theta,\zeta,y) = &\E_{\zeta,y} \Big[ 1_{X^{\zeta,y}_{\zeta + \tau} \in \Omega_\epsilon^{0,x} } \sigma^{-\top} \frac{W_{(\zeta+ \tau ) \wedge T}-W_{\zeta}}{\tau \wedge (T-\zeta)} \frac{1}{2} \big( \phi(\zeta,\zeta+\tau, X_{\zeta+\tau}^{\zeta,y}, u( \theta,\zeta+\tau, X_{\zeta+\tau}^{\zeta,y}), v( \theta, \zeta+\tau, X_{\zeta+\tau}^{\zeta,y})) -  \nonumber \\
 & \quad  \phi(\zeta,t+\tau, \widehat{X}_{\zeta+\tau}^{\zeta,y}, u(\theta, \zeta+\tau, \widehat{X}_{\zeta+\tau}^{\zeta,y}), v(\theta,\zeta+\tau, \widehat{X}_{\zeta+\tau}^{\zeta,y})) \big) \big],
 \label{eq:fixedPointdemo}
 \end{align}
 where we have suppose that $\tau$ is sampled  by a gamma law with $u<1$.

At last, we also localize the loss function \eqref{eq:LossFirstScheme}:
\begin{align}
\ell^\epsilon(\theta) &:= \mathbb{E} \left[ 1_{X^{t,x}_{\zeta} \in \Omega^{0,x}_\epsilon}  \left( \left(\bar u^\epsilon(\theta, \zeta ,X^{0,x}_{\zeta}) - u(\theta,\zeta ,X^{0,x}_{\zeta}) \right)^2 +
\sum_{j=1}^d \left(\bar v^\epsilon_j(\theta, \zeta ,X^{0,x}_{\zeta}) - v_j(\theta,\zeta ,X^{0,x}_{\zeta}) \right)^2  \right) \right] \label{eq:LossFirstSchemeLoc}
\end{align}

We first show in Proposition \ref{prop:minloss} demonstrated in Section \ref{sec:demo1} of the Appendix that it is possible to find a sequence of elements of $\kappa^{d+1}$ such that the loss function \eqref{eq:LossFirstSchemeLoc} is as small as desired.
\begin{proposition}
\label{prop:minloss}
Under assumptions \ref{ass:SDE}, \ref{ass::lipF}, \ref{ass::gBound}, \ref{ass::uReg}, for every $\epsilon$, there exists a constant $\hat K$ depending on $u$ such that there exists $(u(\theta, t,x)  ,v(\theta,t,x) )  \in \kappa^{d+1}$ satisfying:
\begin{flalign*}
\ell^\epsilon(\theta) \le \hat K \epsilon
\end{flalign*}
where $\ell^\epsilon(\theta)$ is given by equation \eqref{eq:LossFirstSchemeLoc}.
\end{proposition}

At last we suppose that the non linearity only depend on $u$, such that  only the first equation in \eqref{eq:fixedPointdemo} is used:
\begin{align}
\bar u^\epsilon(\theta,\zeta,y) = & \frac{1}{2}\E_{\zeta,y} \big[ \big(\phi\big(\zeta, \zeta+\tau ,X_{\zeta+\tau}^{\zeta,y}, u(\theta,\zeta+\tau ,X_{\zeta+\tau}^{\zeta,y}) \big) + \nonumber \\
& \phi\big(\zeta, \zeta+\tau ,\widehat{X}_{\zeta+\tau}^{\zeta,y}, u(\theta,\zeta+\tau ,\widehat{X}_{\zeta+\tau}^{\zeta,y}), \big) \big) 1_{X^{\zeta,y}_{\zeta +\tau} \in \Omega_\epsilon^{0,x}} \big],
\label{eq:simpliFixedPint}
\end{align}
where $\phi$ is independent of $Du$.

Thus we can use the loss function depending only on $u$:
\begin{align}
\ell^\epsilon(\theta) &:= \mathbb{E} \left[ 1_{X^{t,x}_{\zeta} \in \Omega^{0,x}_\epsilon}  \left( \bar u^\epsilon\left(\theta, \zeta ,X^{0,x}_{\zeta}\right) - u\left(\theta,\zeta ,X^{0,x}_{\zeta} \right)  \right)^2 \right] \label{eq:spLossFirstSchemeLoc}
\end{align}
Then we can state the next proposition proved in the appendix and showing that if the Lipschitz constant associated to $f$ is small enough then  a loss function  going to zero assure a convergence of the numerical solution to the solution of the continuous problem.

\begin{proposition}
Suppose that  $f$ is independent of $Du$, that assumptions \ref{ass::gBound}, \ref{ass::uReg}, \ref{ass:SDE} are satisfied, that $f$ is uniformly Lipschitz in $u$, and that the loss function is given by \eqref{eq:spLossFirstSchemeLoc}. \\
If we have a sequence $(\epsilon_n, \theta_n)$ where $\epsilon_n$ goes to $0$ and $\theta_n \in \R^{3n+nd}$  such that $\ell^{\epsilon_n}(\theta_n)$ goes to zero then 
for all compact  $\hat K \in \R^d$ 
\begin{flalign}
\E\left[   \int_0^T 1_{X^{0,x}_{s} \in \hat K}  (u( s,X^{0,x}_s) - u^{\epsilon_n}( \theta_n, s,X^{0,x}_s))^2  ds \right] \longrightarrow 0
\end{flalign}
as $n$ goes to infinity.
\label{prop:convergence}
\end{proposition}

\section{Experiments \& results on DBSDE algorithms}

We choose to test our algorithms on the following PDEs, which details can be found in the Appendix:
\begin{itemize}
\item A Black-Scholes Barenblatt equation \ref{pde:bsbarenblatt} from \cite{raissi2018forward}.
\item The Hamilton-Jacobi-Bellman equation \ref{pde:hjb} corresponding to a control problem, presented in \cite{weinan2017deep,han2017overcoming}, with a non-bounded terminal condition $g(x) = 0.5 \log\left(1+\norm{x}^2 \right)$.
\item A toy example \ref{pde:warin2} with an oscillating solution $u(t, x) = \exp(a (T-t))\cos(\sum_{i=1}^d x_i)$ and a non-linearity in $\left(y \sum_{i=1}^d z_i\right)^2$.
\item An equation \ref{pde:richou} from \cite{richou2010phd} close to HJB, but with a non-Lipschitz terminal condition $g(x) = \sum_{i=1}^d \left(\max\left\{0, \min\left[ 1, x_i\right] \right\}\right)^\alpha$.
\item A toy example \ref{pde:cir} with an oscillating solution and a CIR model for $X$.
\item A toy example \ref{pde:warin3} with an oscillating solution and a non-linearity in $y / (\sum_{i=1}^d z_i)$.
\end{itemize}
In the whole section $\delta_t$ stands for the size of the time step.



\subsection{Influence of the hyperparameters and methodology}

\subsubsection{Learning parameters}

\paragraph{Batch size} As \cite{han2017overcoming,weinan2017deep,raissi2018forward}, we use the Adam optimizer \cite{ruder2016overview}. The batch size is a key parameter of this algorithm. We find that in our case, using large batches speeds the algorithm up without giving up much learning efficiency. In the following, we choose a batch size of $M=300$.

\paragraph{Learning rate} As described in \cite{bengio2012practicalrecommendations}, the learning rate is arguably the training hyperparameter that has the biggest influence on training. Theoretical work indicate that the optimal learning rate for a given problem is close to ``the biggest value before the algorithm diverges'' up to a factor $2$. One also advocates the use of \emph{learning rate schedules}, i.e. changing the learning rate during training, to achieve lower losses when a loss floor is reached. Moreover, as shown in \cite{han2016dlapproxforscp}, lower learning rates enable reaching lower losses and stabilizing the learning process in the final stages. We choose not to tune the learning rate for each of our networks and hyperparameter choice, but rather use an adaptive strategy. We initially choose a learning rate of $\eta = 0.01$, which we find to be a reasonable starting value, albeit $10$ times larger than the one proposed in the original article on Adam \cite{kingma2014adam} -- some of our algorithms would diverge during training for $\eta = 0.02$.  Working with periods of $1000$ iterations (gradient descent updates), we keep trace of the mean test loss over the period. Between two periods, we check if the mean of the losses has decreased less than $5\%$. If it is not the case, we consider that we reached a loss plateau, and we divide the learning rate by $2$ for the next period. This adaptive strategy enables to explore various learning rate scales during training, achieving lower losses and avoiding fluctuations at the end of training.

\paragraph{Initialization}In our neural networks, we use Xavier initialization \cite{glorot2010understanding} for the weights and a normal initialization for the biases. We find that a good initialization of $Y_0$ is also necessary: if the initial guess is far from the optimum, the algorithm would converge very slowly or get stuck in local optima, especially for our LSTMs. In order to make a reasonable guess, we initialize with $Y_0 := \mathbb{E}[g(X_T)]$, which is the solution corresponding to $f := 0$.

\paragraph{Regularization} In machine learning, regularizing the optimization process by adding a term in the loss penalizing high weights often helps the network. We find that in our case, adding $L^2$ regularization on the weights of the neurons (not on the biases) degrades the convergence and the precision of the algorithm: it is not surprising as our data is not noisy nor redundant, and thus the network do not experience overfitting. We thus do not regularize our network in the following.

\paragraph{Centering and rescaling} Neural networks tend to have convergence issues when their inputs are not scaled and centered, and perform best when the inputs follow a normal distribution, especially with our LSTMs and $\tanh$ activation functions. In our case, the inputs $t$ and $Y$ are not Gaussian, but $X$ is close to a Gaussian when $\mu$ is small. Since we will use the same network for each time step, we scale and center all the inputs $(t_i, X_{t_i}, Y_{t_i}, g(X_{t_i}))$ for all $t_i$'s with the same coefficients, so that they take values in $\sim [-1, 1]$, as described in Section \ref{subsubsection:standardtraining}.

\paragraph{Number of hidden layers and hidden layer sizes}

We investigate the influence of the number of hidden layers and the hidden layer sizes on the convergence of the algorithms, and the precision of the results. We denote by $h$ the number of hidden layers and $w$ the size of the hidden layers. In practice, the optimal $w$ and $h$'s depend on the equation under consideration and the network use. We find that for most networks, for a fixed $w$, setting $h$ over $2$ or $3$ increases greatly the difficulty of the problem and degrades the convergence speed without increasing significantly the precision of the results. Increasing the hidden layer size $w$ does not have a significant impact over a certain value -- we find that suitable values are $h=2$ and $w \simeq d$ or $2d$.

\subsubsection{Standard training procedure} \label{subsubsection:standardtraining}

If not precised otherwise, we use the following hyperparameters and training procedure:
\begin{itemize}
\item If applicable, initialize $Y_0$ to $\mathbb{E}[g(X_T)]$, weights with Xavier initialization and biases from a normal distribution.
\item Use centered and rescaled neural network inputs $\widetilde{t}, \widetilde{X}, \widetilde{Y}, \widetilde{g}(X)$, defined as
\[ 
\widetilde{t} = \frac{t - (T-\delta_t)/2}{(T-\delta_t)/2}, \qquad
\widetilde{X} = \frac{X - X_\text{mean}}{X_\text{std}}, \qquad
\widetilde{Y} = \frac{Y - Y_\text{mean}}{\abs{Y_\text{mean}}}, \qquad
\widetilde{g}(X) = \frac{g(X) - Y_\text{mean}}{\abs{Y_\text{mean}}},
\]
where, if $M$ is a number of samples (we took $M=10000$), we compute beforehand:
\[
X_\text{mean} =  \underset{\textrm{\shortstack{$i=0..N$\\ $m=1..M$}}}{\text{mean}} X_{t_i}^{(m)}, \qquad 
X_\text{std} =  \underset{\textrm{\shortstack{$i=0..N$\\ $m=1..M$}}}{\text{std}} X_{t_i}^{(m)}, \qquad
Y_\text{mean} =  \underset{\textrm{\shortstack{$i=0..N$\\ $m=1..M$}}}{\text{mean}} g(X_{t_i}^{(m)}).
\]
\item Evaluate the test loss each $100$ iterations during training with a separate test set of size $1000$.
\item Use the Adam optimizer \cite{kingma2014adam} with the recommended parameters, but with a decreasing learning rate strategy: set initially $\eta \leftarrow \eta_0 = 10^{-2}$, let $\ell_j$ be the test loss evaluate at iteration $j$ and
\[ \widehat{\ell}_k = \underset{j=1000k..1000(k+1)-1}{\text{mean}} \ell_{j}.\]
If at a step $j=1000(k+1)$, 
\[ \frac{\widehat{\ell}_{k} - \widehat{\ell}_{k+1}}{\widehat{\ell}_{k}} < 5\% \]
then $\eta \leftarrow \eta / 2$. In practice, since we only evaluate the test loss every $100$ iterations $j$, $\widehat{\ell}_{k}$ is a mean over $10$ values.
\item We recall we use a batch size of $300$ in the Adam optimizer.
\item For the new algorithm described in Section \ref{sec:new_algorithms}, we use $\lambda=0.5$ (or $\lambda = 1.0$ if precised) and $n_\text{inner} = 10000$ during training.
\end{itemize}

\paragraph{Error measures}After $16000$ iterations, we retain the set of parameters that generated the lowest test loss during training. Finally, we compute the deterministic quantities:
\begin{align}
\text{Relative error on $Y_0$} &= \frac{\abs{Y_0 - Y_{0, \text{ref}}}}{\abs{Y_{0, \text{ref}}}}, \label{eq:relerrY0} \\
\text{Relative error on $Z_0$} &= \frac{\norm{Z_0 - Z_{0, \text{ref}}}_2^2}{\norm{Z_{0, \text{ref}}}_2^2}, \label{eq:relerrZ0}
\end{align}
and we compute a final test loss and the following expectations using a final test set of size $1500$:
\begin{align}
\text{Integral error on $Y$} &= \mathbb{E} \left[ \delta_t \left(\frac{\abs{Y_0 - Y_{0, \text{ref}}} + \abs{Y_T - Y_{T, \text{ref}}}}{2}  + \sum_{i=1}^{N-1} \abs{Y_{t_{i}} - Y_{t_{i}, \text{ref}}} \right) \right], \label{eq:interrY}\\
\text{Integral error on $Z$} &= \mathbb{E} \left[ \delta_t \left(\frac{\norm{Z_0 - Z_{0, \text{ref}}}_2^2 + \norm{Z_T - Z_{T, \text{ref}}}_2^2}{2}  + \sum_{i=1}^{N-1} \norm{Z_{t_{i}} - Z_{t_{i}, \text{ref}}}_2^2 \right) \right]. \label{eq:interrZ}
\end{align}
We also compute the errors on $Y$ and on $Z$ along the trajectories. 
For our new algorithm, we use $n_\text{eval} = 100000$ during evaluation to compute the integral errors on $Y$ \eqref{eq:interrY} and $Z$ \eqref{eq:interrZ} and $n_\text{eval}=1000000$ to compute the (pointwise) relative errors on $Y_0$ \eqref{eq:relerrY0} and $Z_0$ \eqref{eq:relerrZ0}. Since the evaluation is computationally intensive, we use $10$ trajectories to compute \eqref{eq:interrY} and \eqref{eq:interrZ}.

We compute precise baselines for $Y_0$ and $Z_0$, or use closed formulas, as presented in the Appendix. For the integral errors, we use the closed formulas or $50000$ realizations in Monte-Carlo solutions.

In the following results, if not precised otherwise, we compute all means and quantiles using $5$ independent runs for each simulation.

\clearpage
\subsection{Numerical results: \emph{Deep BSDE}} \label{subs:numres:deepbsde}

The number of parameters for each network is represented in Table \ref{table:numparamdeepbsde}. The FC Merged networks \textbf{d.} and \textbf{f.} have significantly less parameters than the other networks. In the following, we fix $h=2$ and $w=2d$ for all networks.

\begin{table}[ht]
\begin{adjustwidth}{-1.5cm}{-1.5cm}
\centering
\begin{tabular}{|p{3.7cm}|p{8.7cm}|r|r|}
\hline
Network & Number of parameters & $d=10$ & $d=100$\\
\hline
\textbf{a.} FC DBSDE & $ 1 + d + (N-1)\left[2d^2 + 4d + (2d)^2 + 4d + 2d^2 + d \right]$ & 88 121 & 8 009 201\\
\hline
\textbf{b.} FC ELU & $1 + d + (N-1)\left[2d^2 + 2d + (2d)^2 + 2d + 2d^2 + d\right]$ & 84 161 & 7 969 601\\
\hline
\textbf{c.} FC Residual & $1 + d + (N-1)\left[(d+1)(2d) + 2d + (2d)^2 + 2d + 2d^2 + d\right]$ & 86 141 & 7 989 401\\
\hline
\textbf{d.} FC Merged & $1 + (d+3)(2d) + 2d + (2d)^2 + 2d + 2d^2 + d$ & 911 & 81 101\\
\hline
\textbf{e.} FC Merged Shortcut & $1 + (d+3)(2d) + 2d + (3d+3)(2d) + 2d + (3d+3)d + d$ & 1 301 & 112 001\\
\hline
\textbf{f.} FC Merged Residual & $1 + (d+3)(2d) + 2d + (2d)^2 + 2d + 2d^2 + d$ & 911 & 81 101\\
\hline
\textbf{g.} LSTM & $1 + 16d + 4[(d+1)(2d) + 2d] + 4[(2d)^2 + 2d] + 2d^2 + d$ & 3 011 & 264 101\\
\hline
\textbf{h.} Augmented LSTM & $1 + 16d + 4[(d+3)(2d) + 2d] + 4[(2d)^2 + 2d] + 2d^2 + d$ & 3 171 & 265 701\\
\hline
\textbf{i.} Hybrid LSTM & $1 + 8d + 4[(d+3)(2d) + 2d] + (2d)^2 + 2d + 2d^2 + d$ & 1 831 & 144 301 \\
\hline
\textbf{j.} Residual LSTM & $1 + 16d + 4[(d+3)(2d) + 2d] + 4[(2d)^2 + 2d] + 2d^2 + d$ & 3 171 & 265 701\\
\hline
\end{tabular}
\end{adjustwidth}
\caption{Number of parameters for each of or networks for \emph{Deep BSDE} for $2$ hidden layers of size $2d$ for an equation in dimension $d$, with $N=100$ time steps. We count, in this order: the contributions of the initial conditions, the first hidden layer, the other hidden layer, the output layer. We also show the number of parameters for $d=10$ and $d=100$.} \label{table:numparamdeepbsde}
\end{table}

\subsubsection{Influence of the number of time steps} \label{subs:inf:ntsteps}

\begin{figure}[p]
\begin{adjustwidth}{-1.5cm}{-1.5cm}
\centering
\textbf{1.}\includegraphics[height=4cm, trim=0cm 0cm 5cm 0cm, clip]{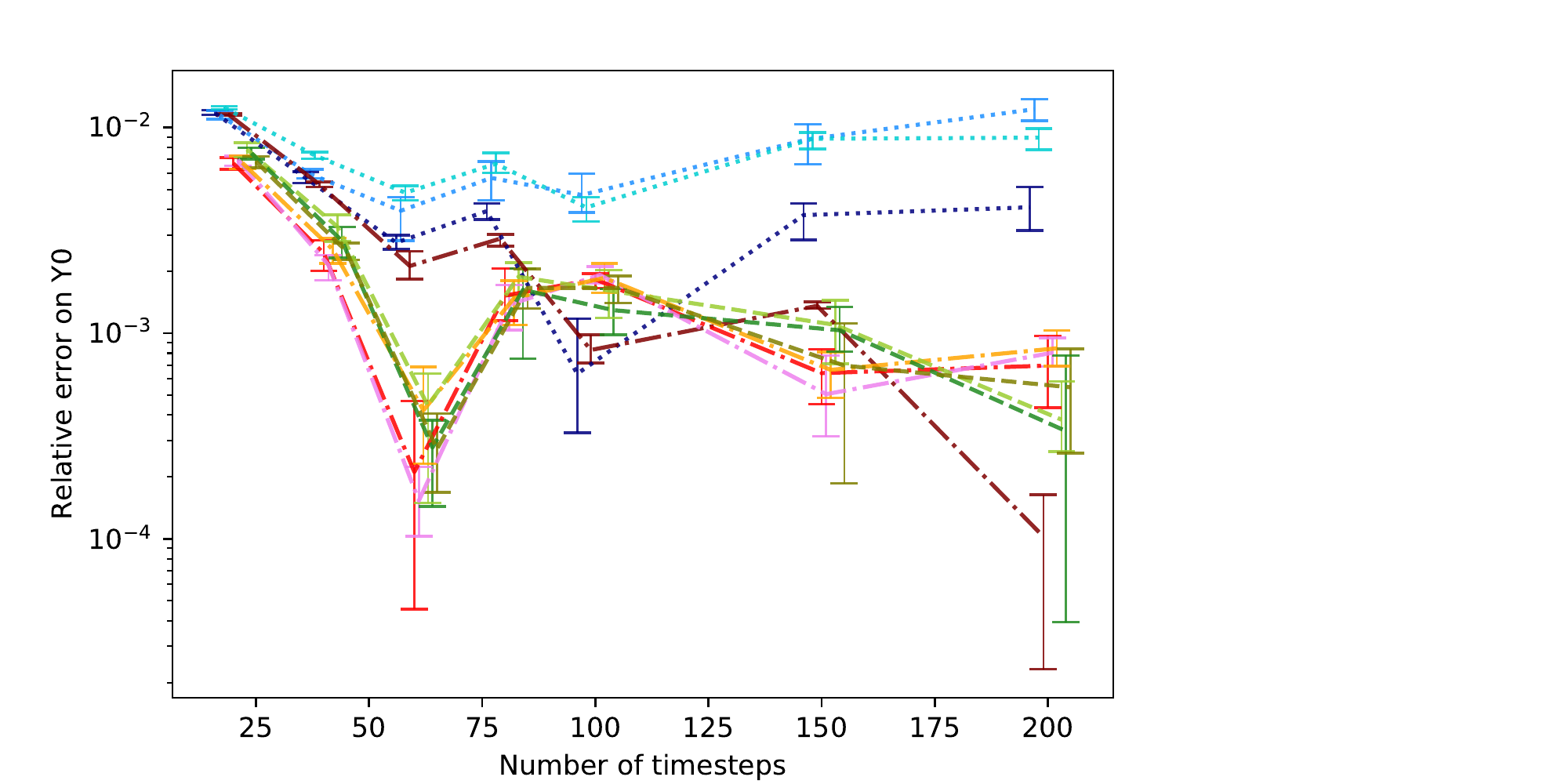}
\textbf{2.}\includegraphics[height=4cm]{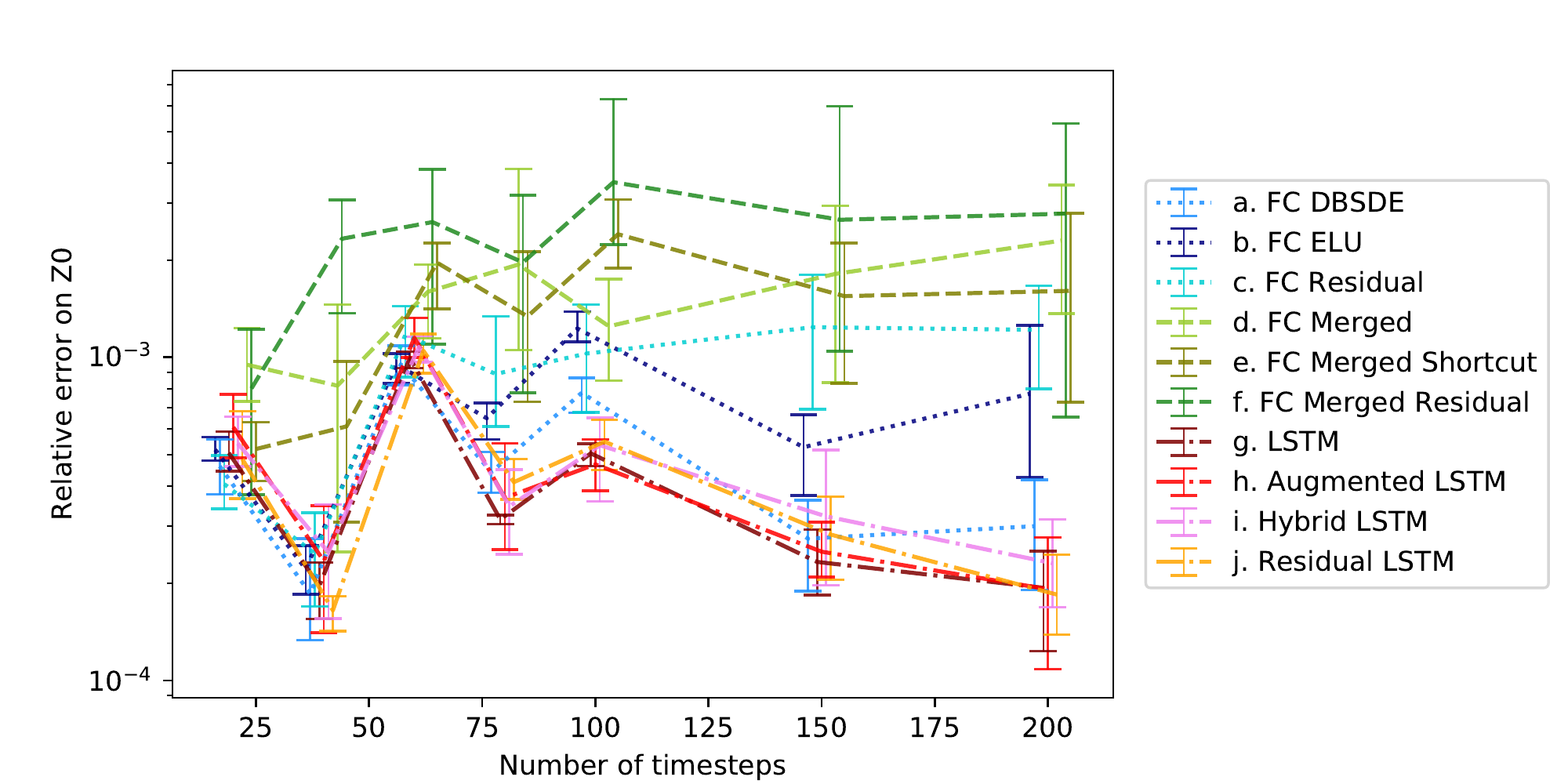} \\
\textbf{3.}\includegraphics[height=4cm, trim=0cm 0cm 5cm 0cm, clip]{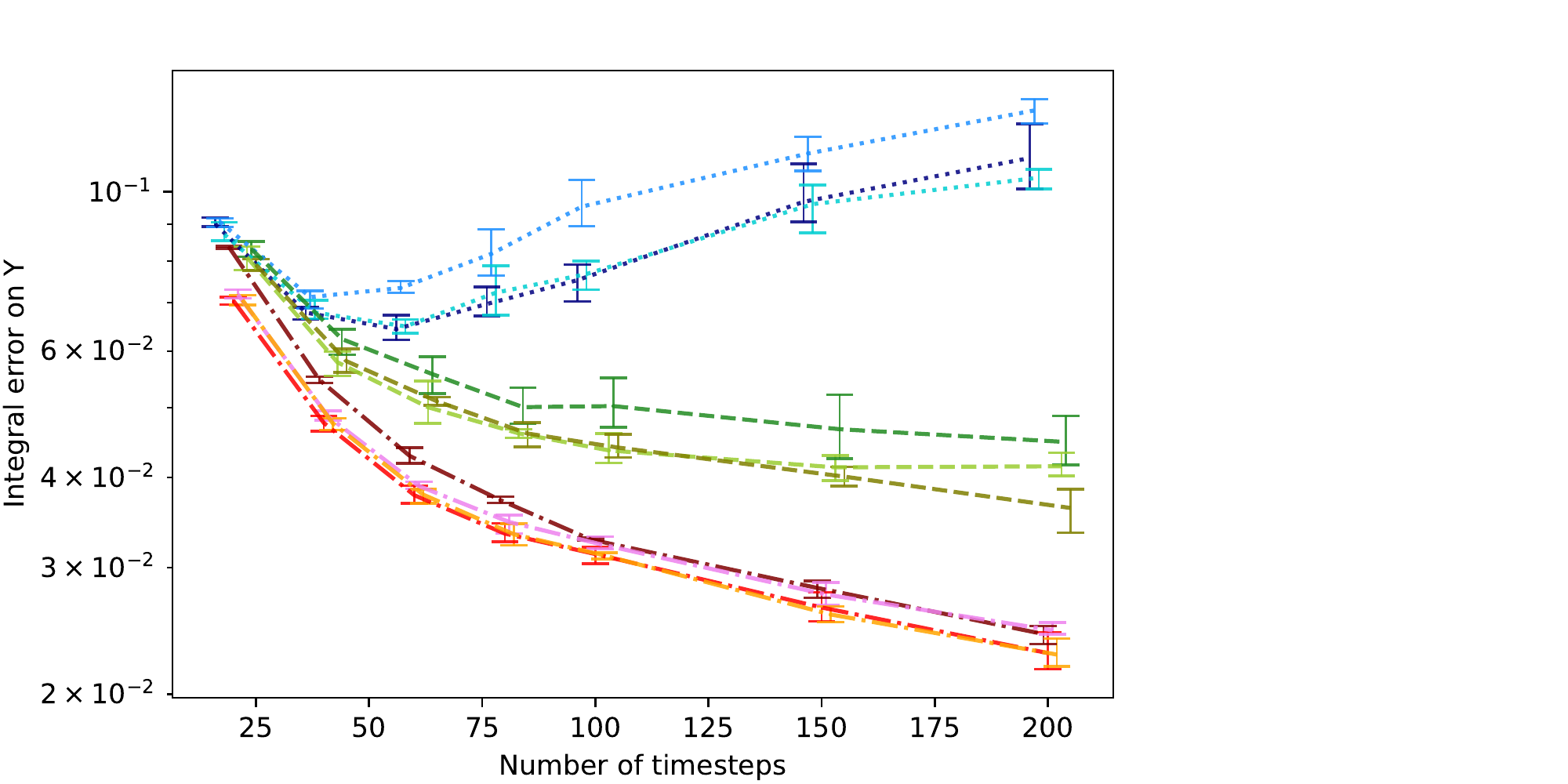}
\textbf{4.}\includegraphics[height=4cm]{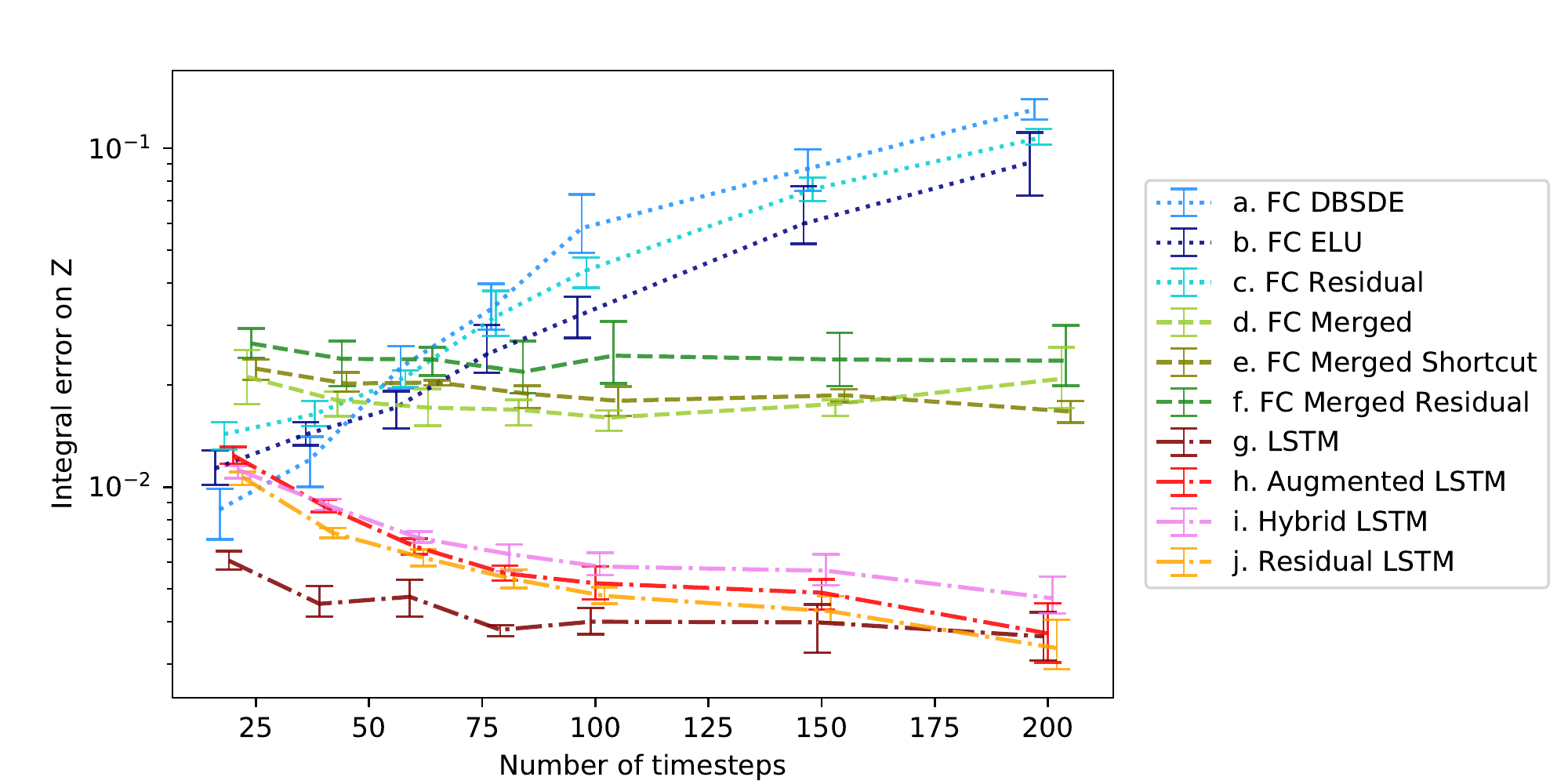} \\
\textbf{5.}\includegraphics[height=4cm]{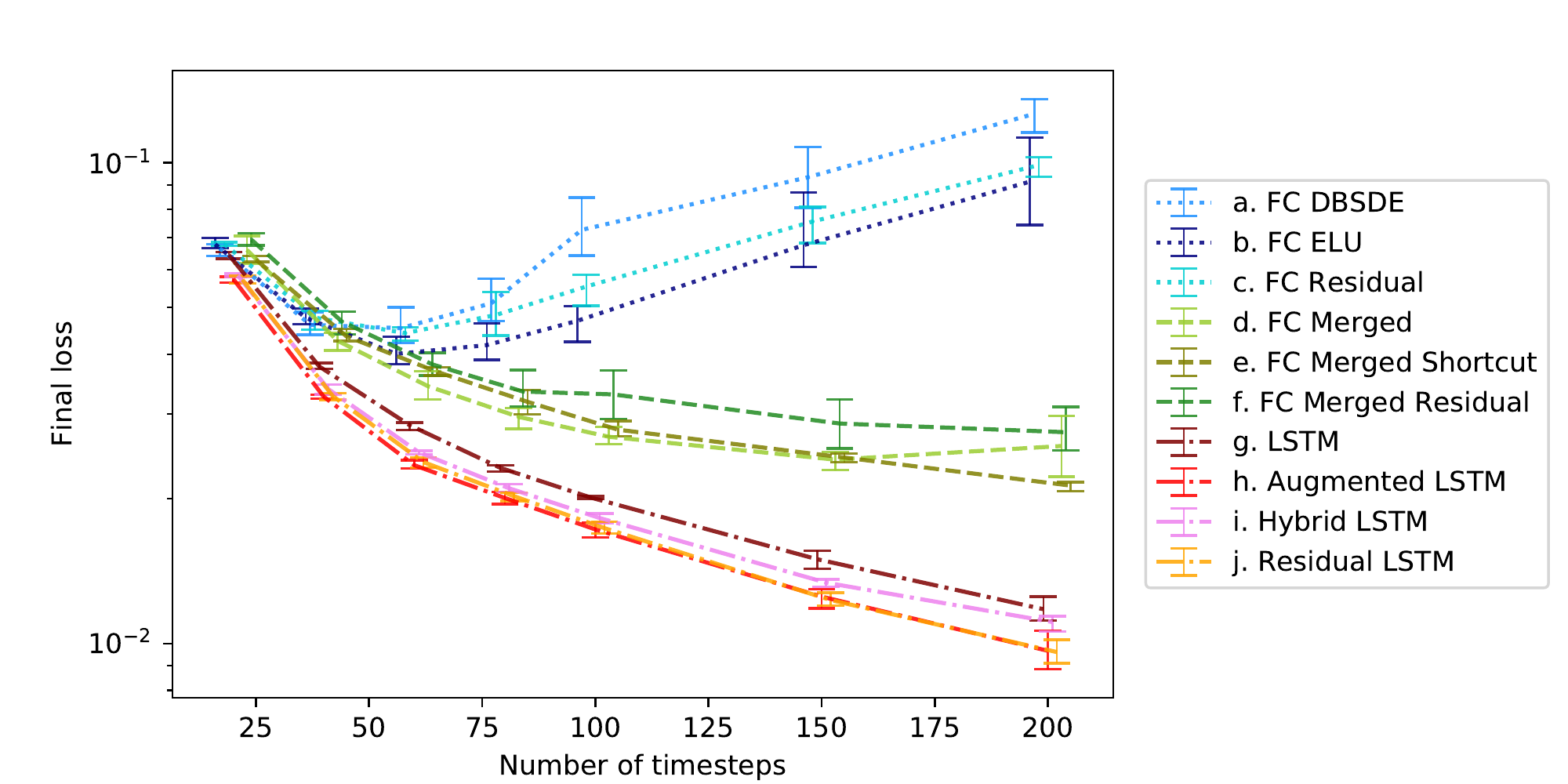} \\[0.5cm]
\textbf{6.}\includegraphics[height=4cm, trim=0cm 0cm 5cm 0cm, clip]{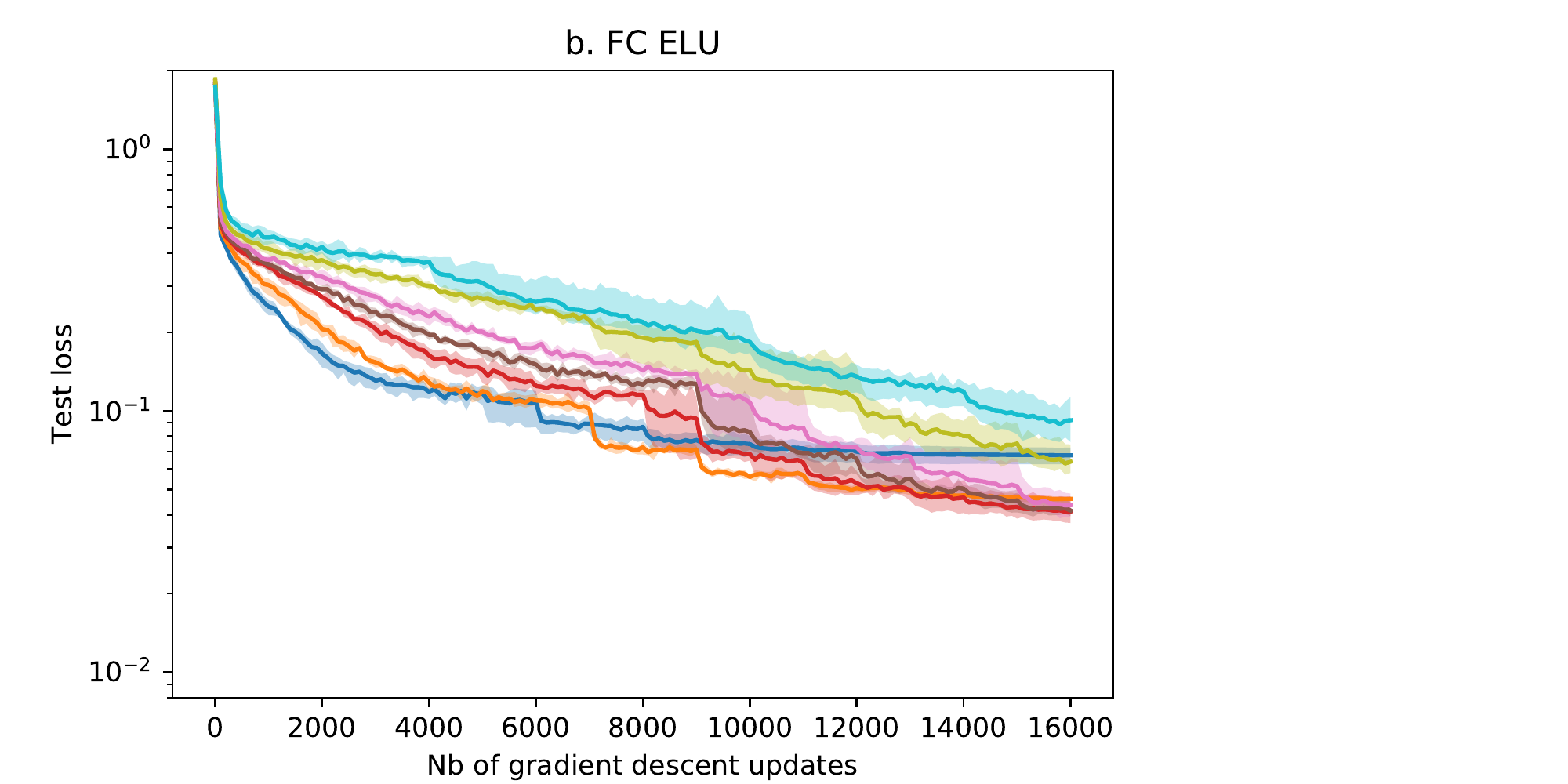}
\textbf{7.}\includegraphics[height=4cm]{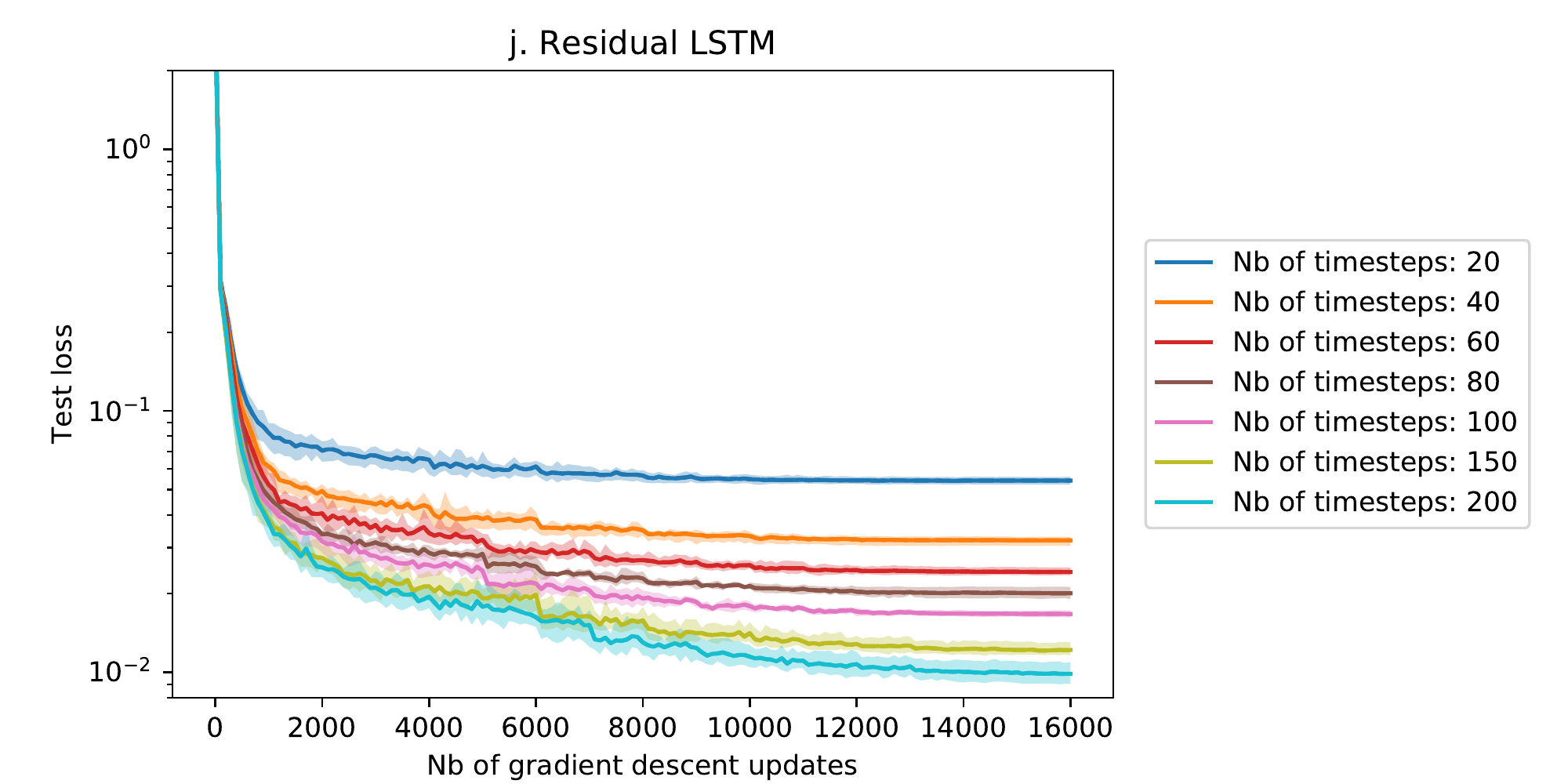}
\end{adjustwidth}

\caption{Influence of the \textbf{number of time steps} on equation \ref{pde:richou} ($d=10$) on $5$ different runs, with the standard learning hyperparameters (see \ref{subsubsection:standardtraining}). \textbf{1.}, \textbf{2.}: we represent the mean of the integral error on $Y$ \eqref{eq:interrY} and the integral error on $Z$ \eqref{eq:interrZ} and the $5-95\%$ confidence intervals. Our LSTM-based networks perform better than the other networks. Merged network perform better than FCs. \textbf{3.}, \textbf{4.}: we represent the final relative error on $Y_0$ \eqref{eq:relerrY0} and the relative error on $Z_0$ \eqref{eq:relerrZ0} (corresponding to the lowest test loss obtained during training). \textbf{5.} We represent the final test loss. Our LSTM-based networks and the merged network perform better when the number of time steps increase, whereas this hurts the convergence of not merged networks, that tend to perform worse: this is shown in \textbf{6.} and \textbf{7.} where we see the LSTM-based network achieves lower losses with a higher number of time steps, whereas the other network can not.}
\label{fig:influence_ntsteps_richoud10}
\end{figure}

We investigate the influence of the number of time steps on the convergence and on the precision of the results. On equation \ref{pde:richou} ($d=10$, Figure \ref{fig:influence_ntsteps_richoud10}), LSTM networks and Merged networks perform better overall than the other networks. The increase of the number of time steps benefits to networks with shared parameters, especially LSTMs, while it degrades strongly the convergence of networks \textbf{a.}, \textbf{b.} and \textbf{c.}.

\begin{figure}[p]
\begin{adjustwidth}{-1.5cm}{-1.5cm}
\centering
\textbf{1.}\includegraphics[height=4cm, trim=0cm 0cm 5cm 0cm, clip]{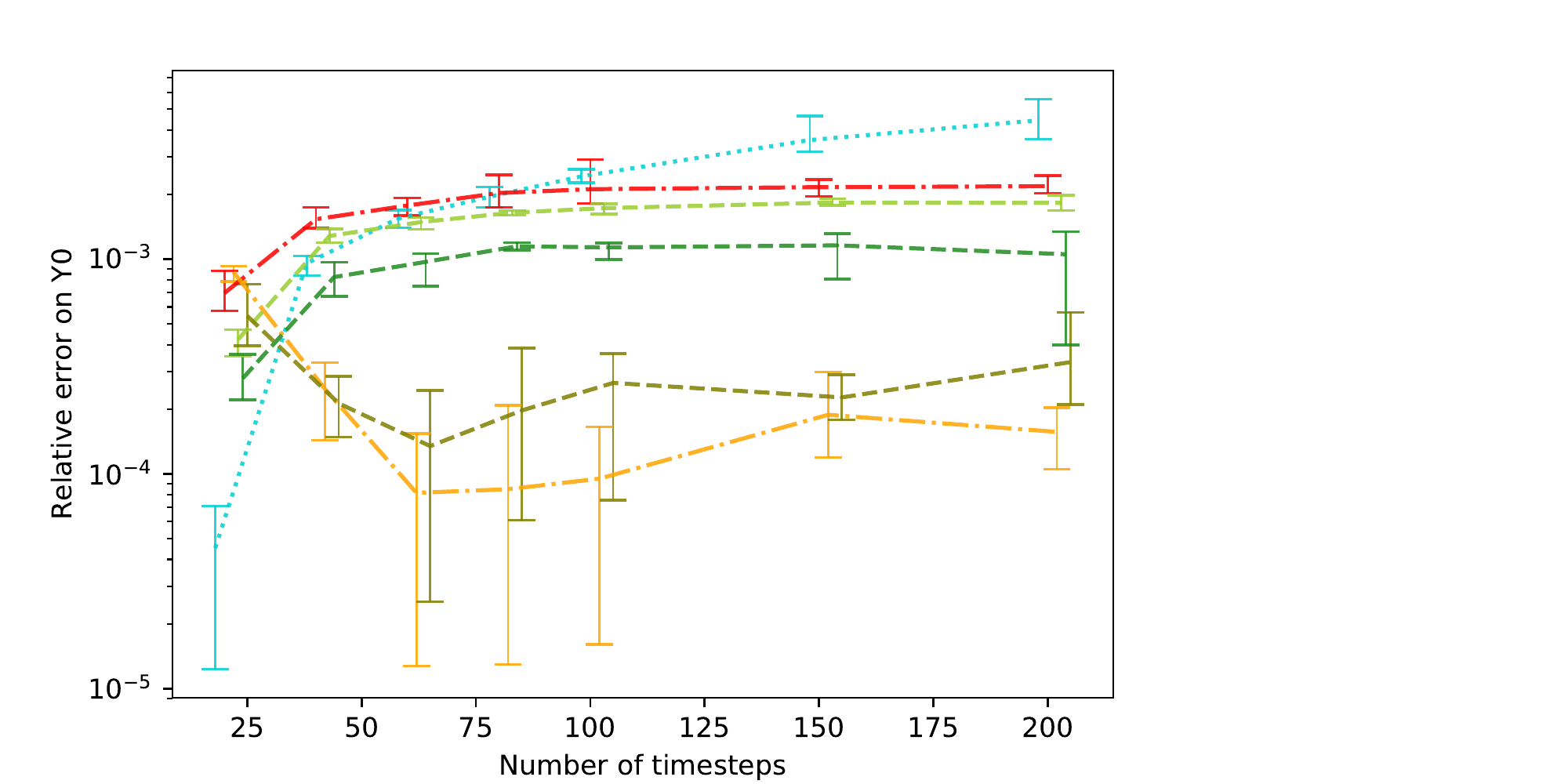}
\textbf{2.}\includegraphics[height=4cm]{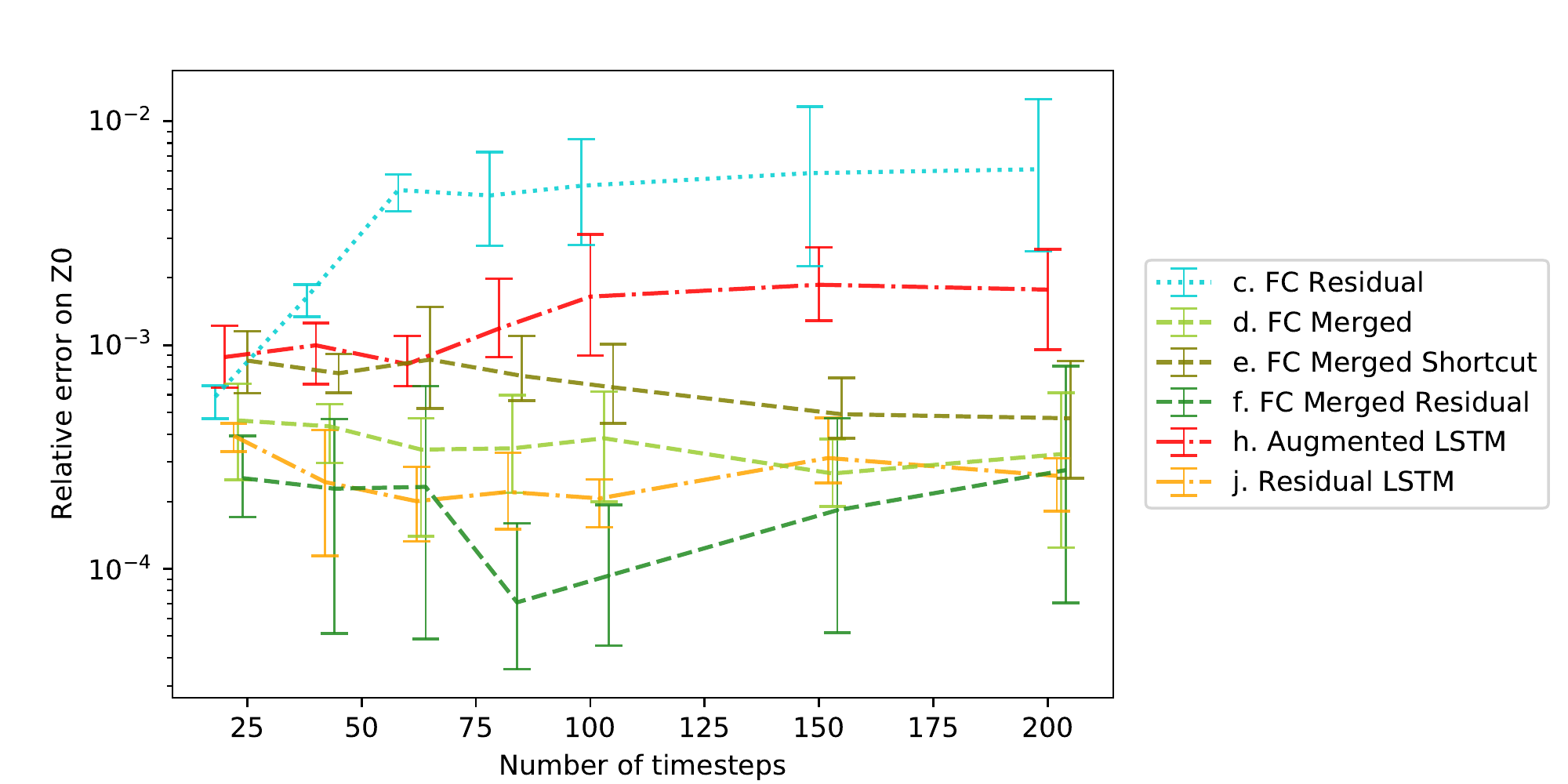} \\
\textbf{3.}\includegraphics[height=4cm, trim=0cm 0cm 5cm 0cm, clip]{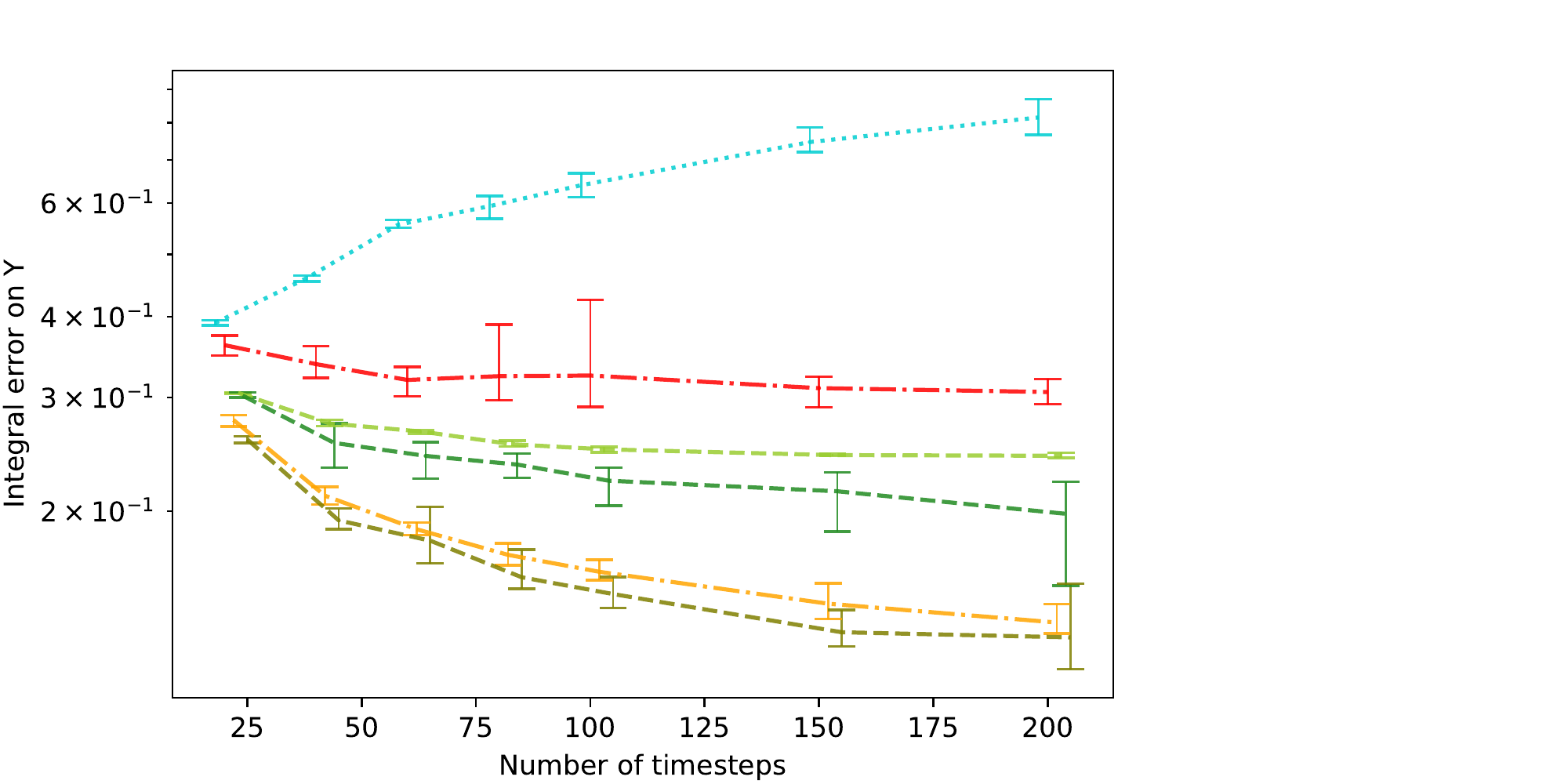}
\textbf{4.}\includegraphics[height=4cm]{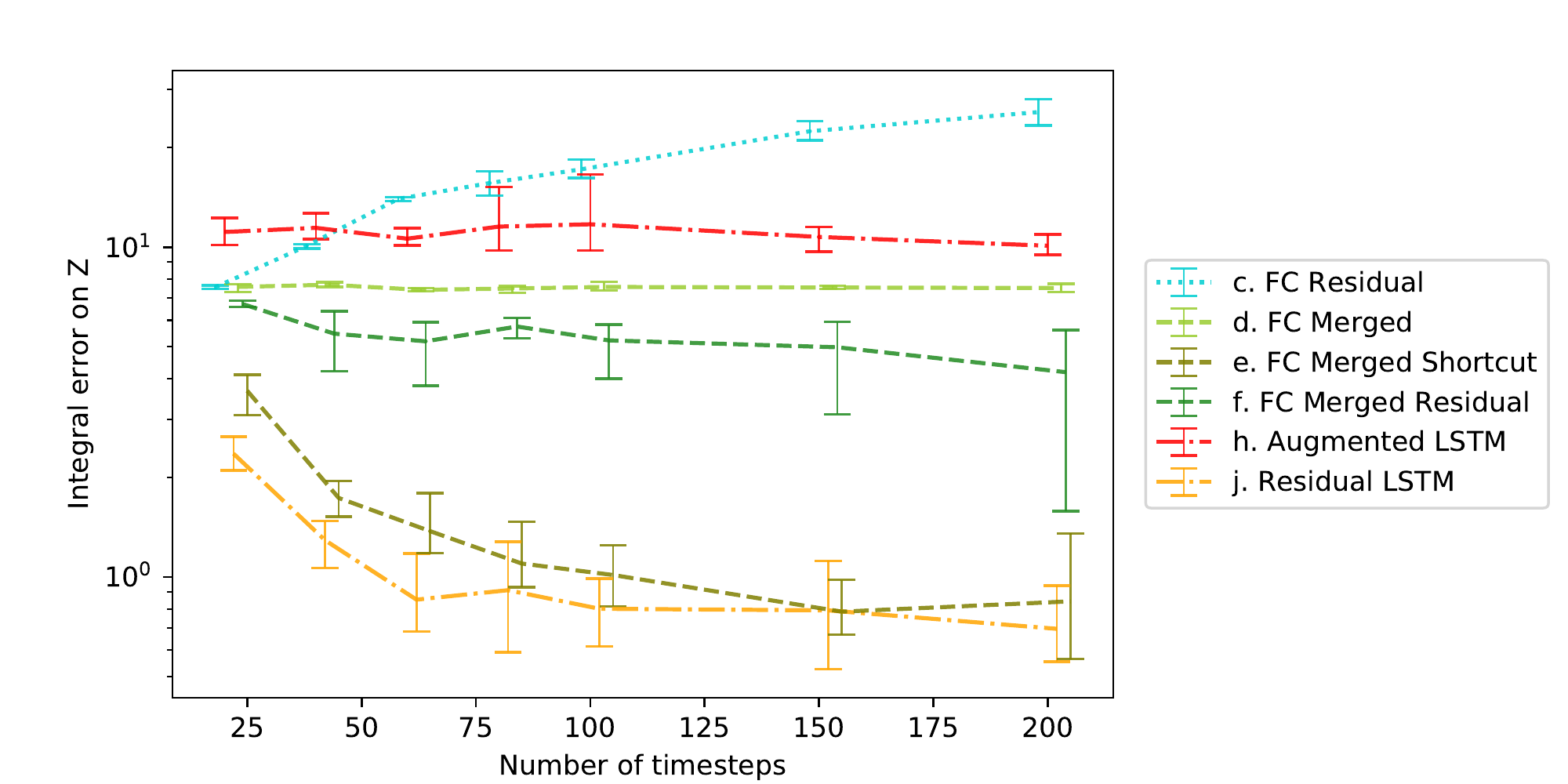} \\
\textbf{5.}\includegraphics[height=4cm]{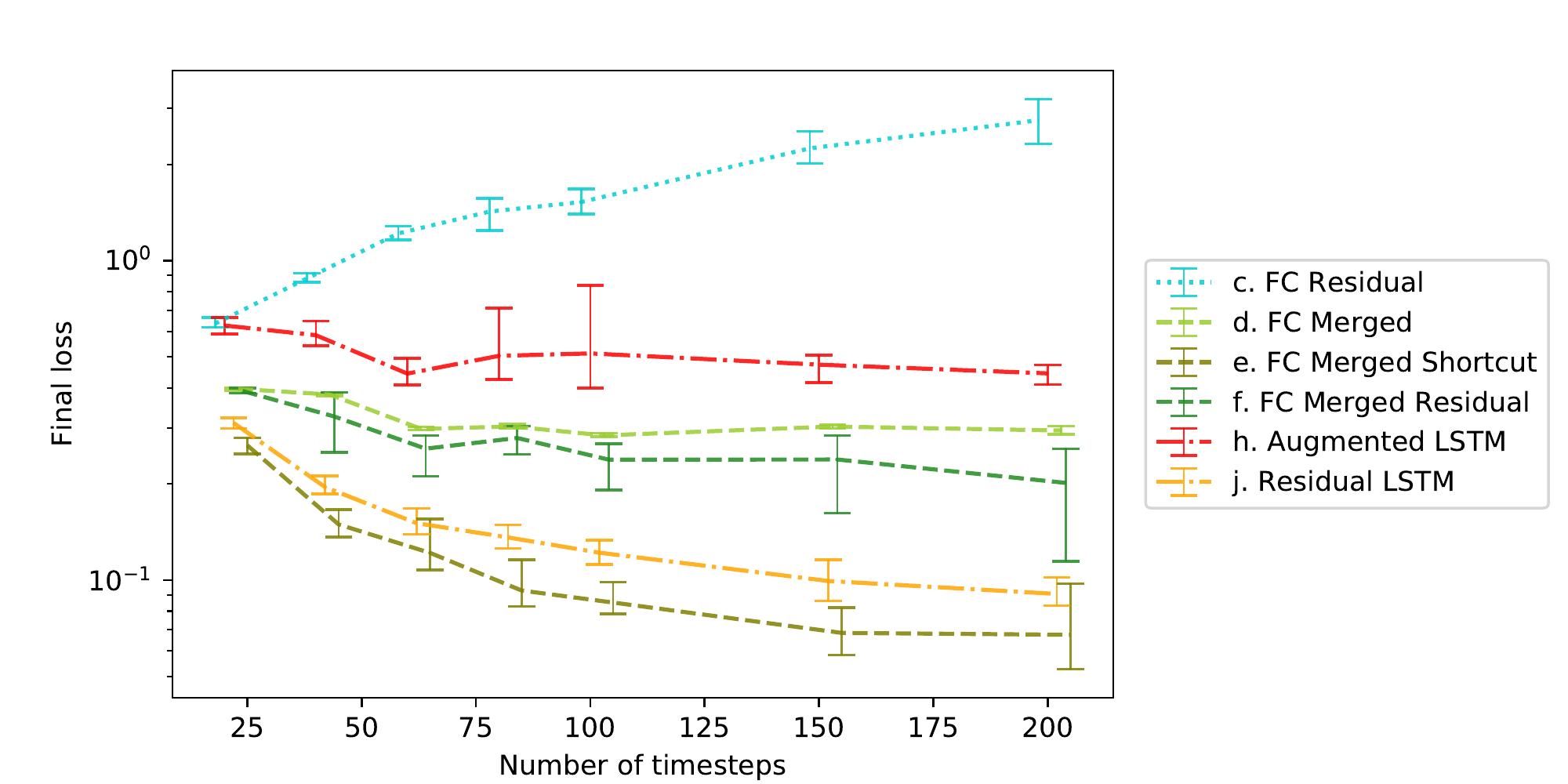} \\[0.5cm]
\textbf{6.}\includegraphics[height=4cm, trim=0cm 0cm 5cm 0cm, clip]{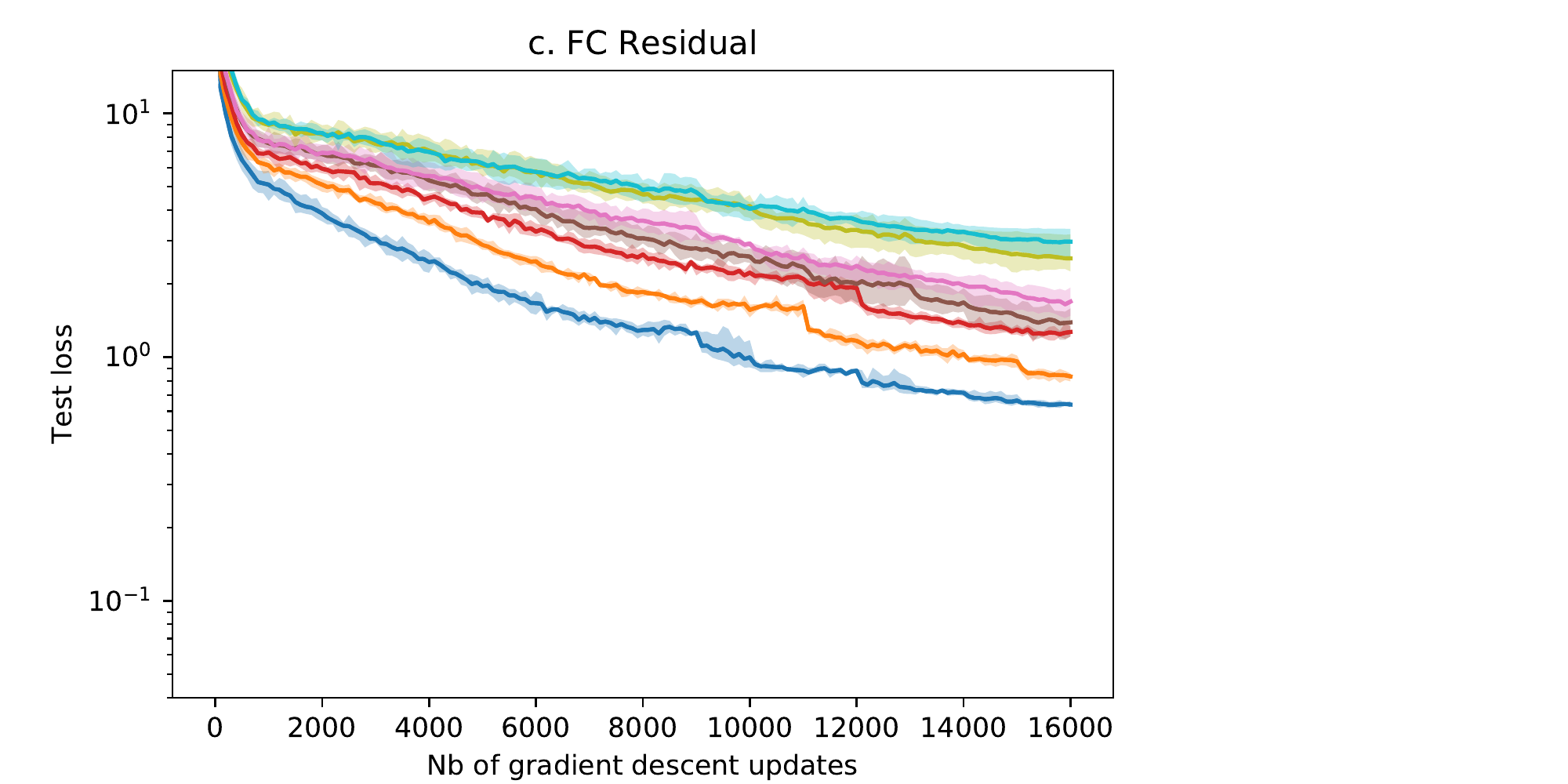}
\textbf{7.}\includegraphics[height=4cm]{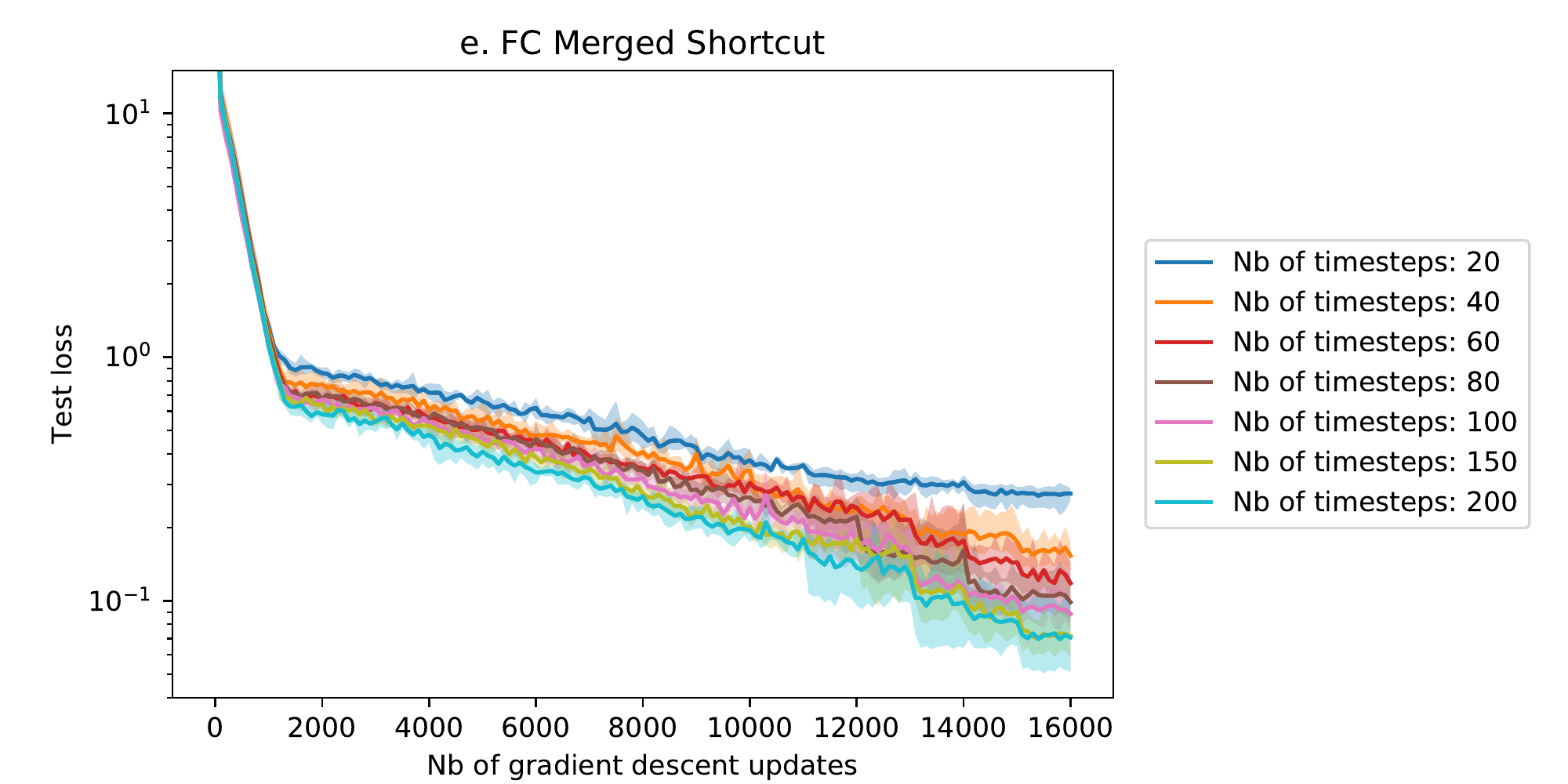}
\end{adjustwidth}

\caption{Influence of the \textbf{number of time steps} on equation \ref{pde:bsbarenblatt} ($d=100$) on $5$ different runs, with the standard learning hyperparameters (see \ref{subsubsection:standardtraining}). We do not represent networks \textbf{a.}, \textbf{b.}, \textbf{g.}, \textbf{i.}, which losses are much higher and did not decrease during training. \textbf{1.}, \textbf{2.}: we represent the final relative error on $Y_0$ \eqref{eq:relerrY0} and the relative error on $Z_0$ \eqref{eq:relerrZ0} (corresponding to the lowest test loss obtained during training). \textbf{3.}, \textbf{4.}: we represent the mean of the integral error on $Y$ \eqref{eq:interrY} and the integral error on $Z$ \eqref{eq:interrZ} and the $5-95\%$ confidence intervals. \textbf{5.}: we represent the final test losses. We represent the test losses during training for \textbf{c.} and \textbf{e.} in \textbf{6.} and \textbf{7.} and show the merged network see better convergence with a higher number of time steps, while the not merged network does not.}
\label{fig:influence_ntsteps_bsbarenblattd100}
\end{figure}

On equation \ref{pde:bsbarenblatt} ($d=100$, Figure \ref{fig:influence_ntsteps_bsbarenblattd100}), the networks \textbf{a.}, \textbf{b.}, \textbf{c.}, \textbf{g.} and \textbf{i.} fail to converge (the loss remains constant and  high). Overall, the LSTM with a residual structure \textbf{j.} and the Merged network with a shortcut structure \textbf{e.} perform noticeably better than the other networks, and their precision increase with the number of time steps.

\begin{figure}[p]
\begin{adjustwidth}{-1.5cm}{-1.5cm}
\centering
\textbf{1.}\includegraphics[height=4cm, trim=0cm 0cm 5cm 0cm, clip]{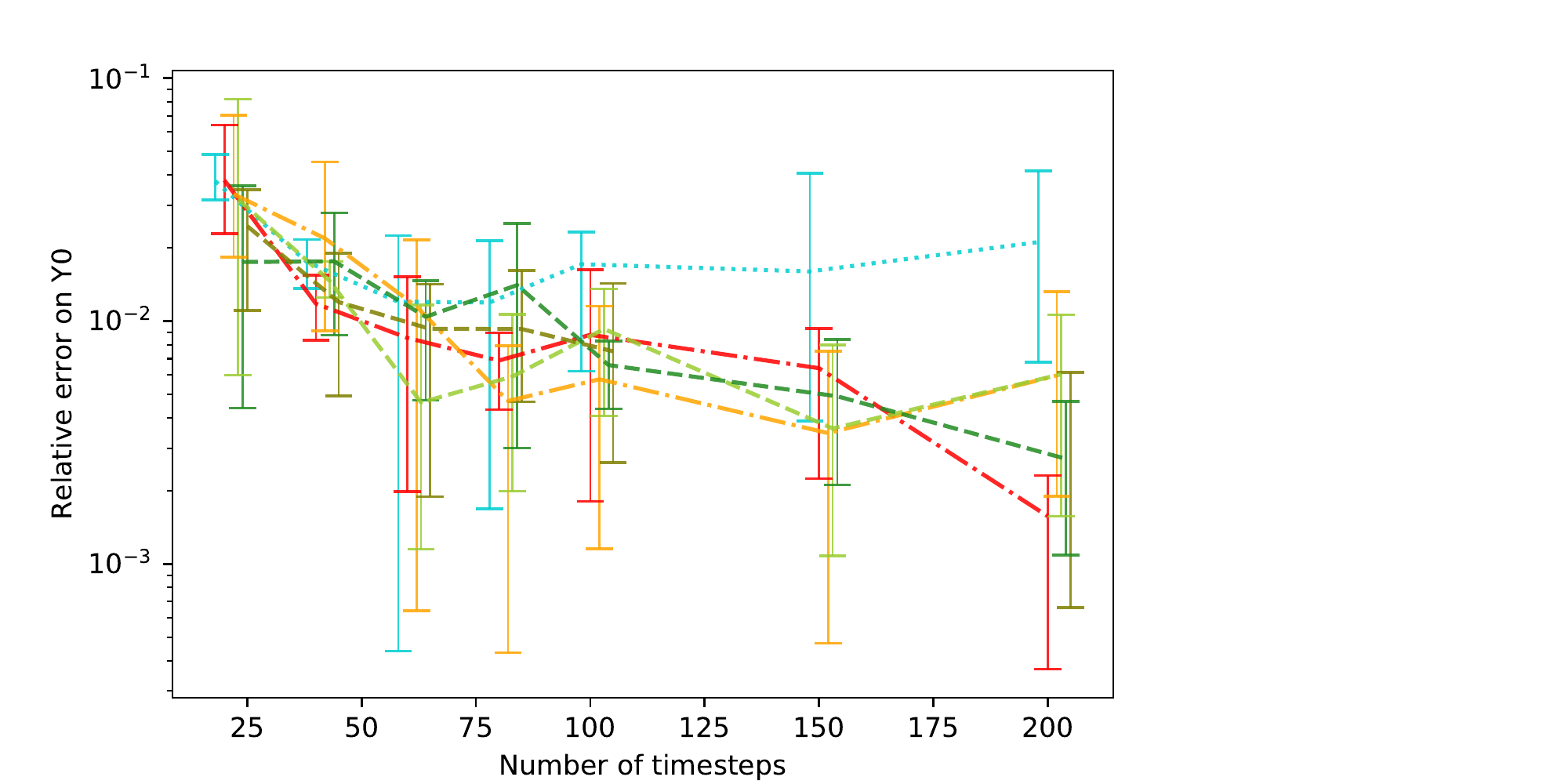}
\textbf{2.}\includegraphics[height=4cm]{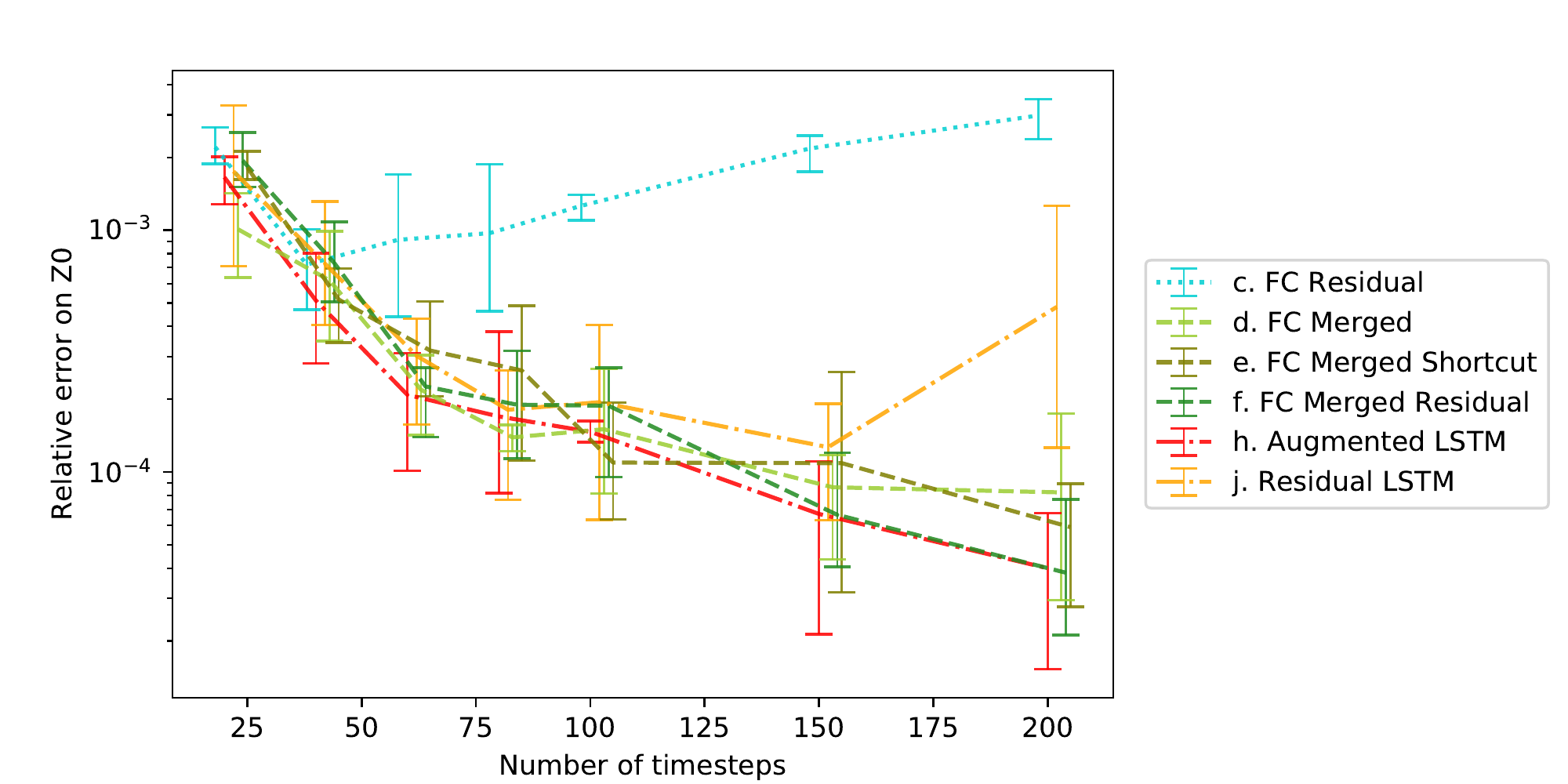} \\
\textbf{3.}\includegraphics[height=4cm, trim=0cm 0cm 5cm 0cm, clip]{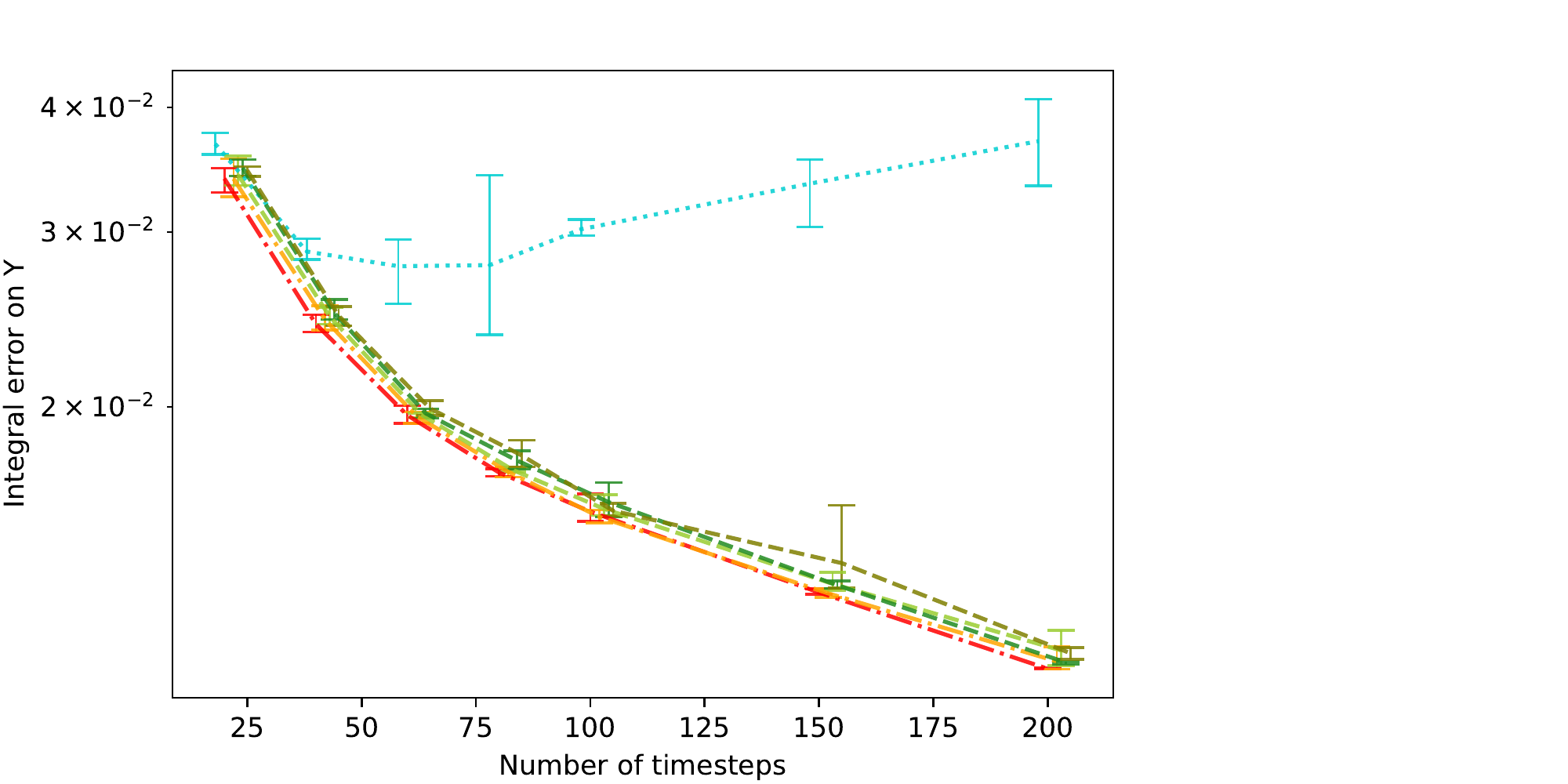}
\textbf{4.}\includegraphics[height=4cm]{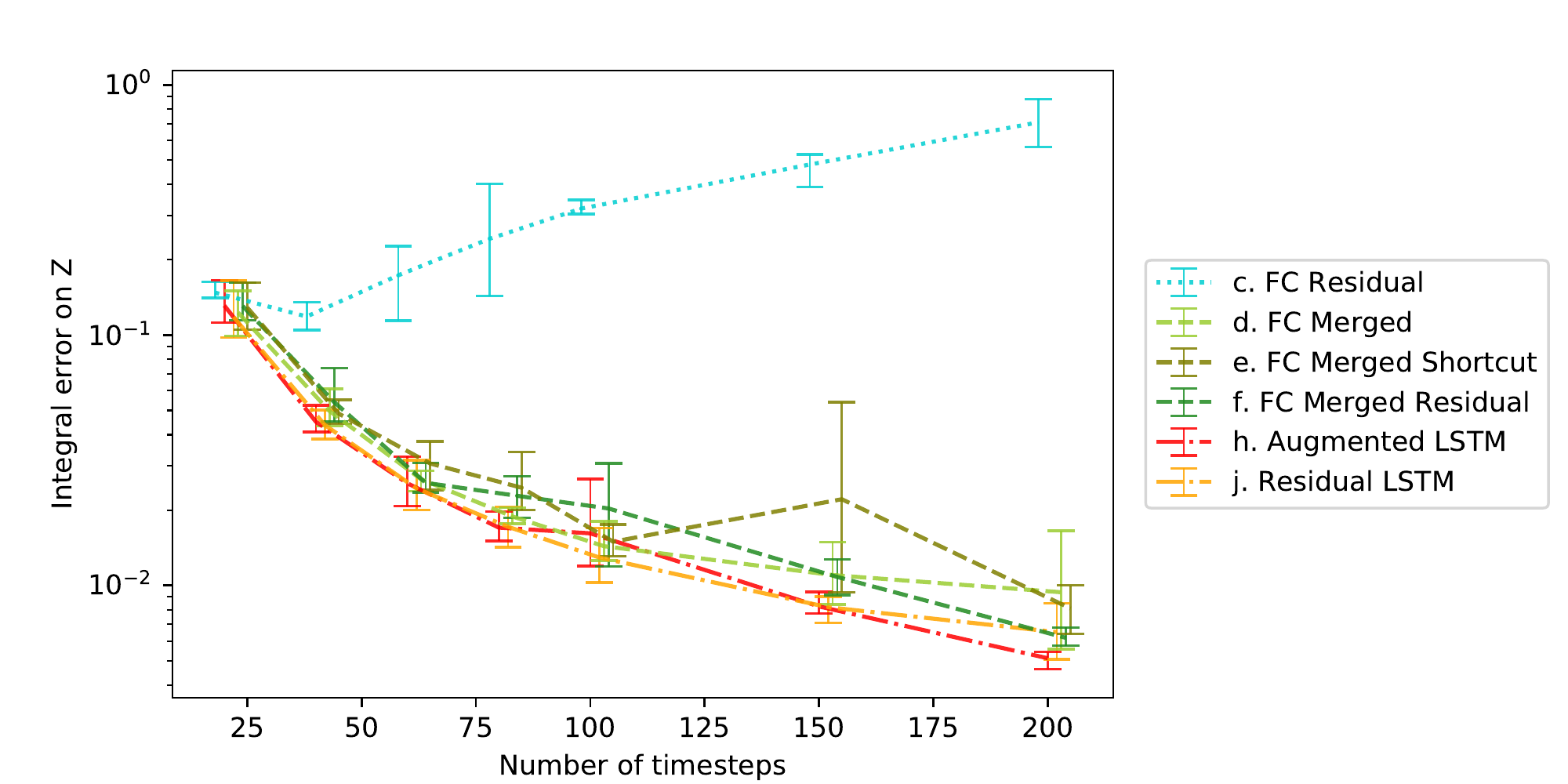} \\
\textbf{5.}\includegraphics[height=4cm]{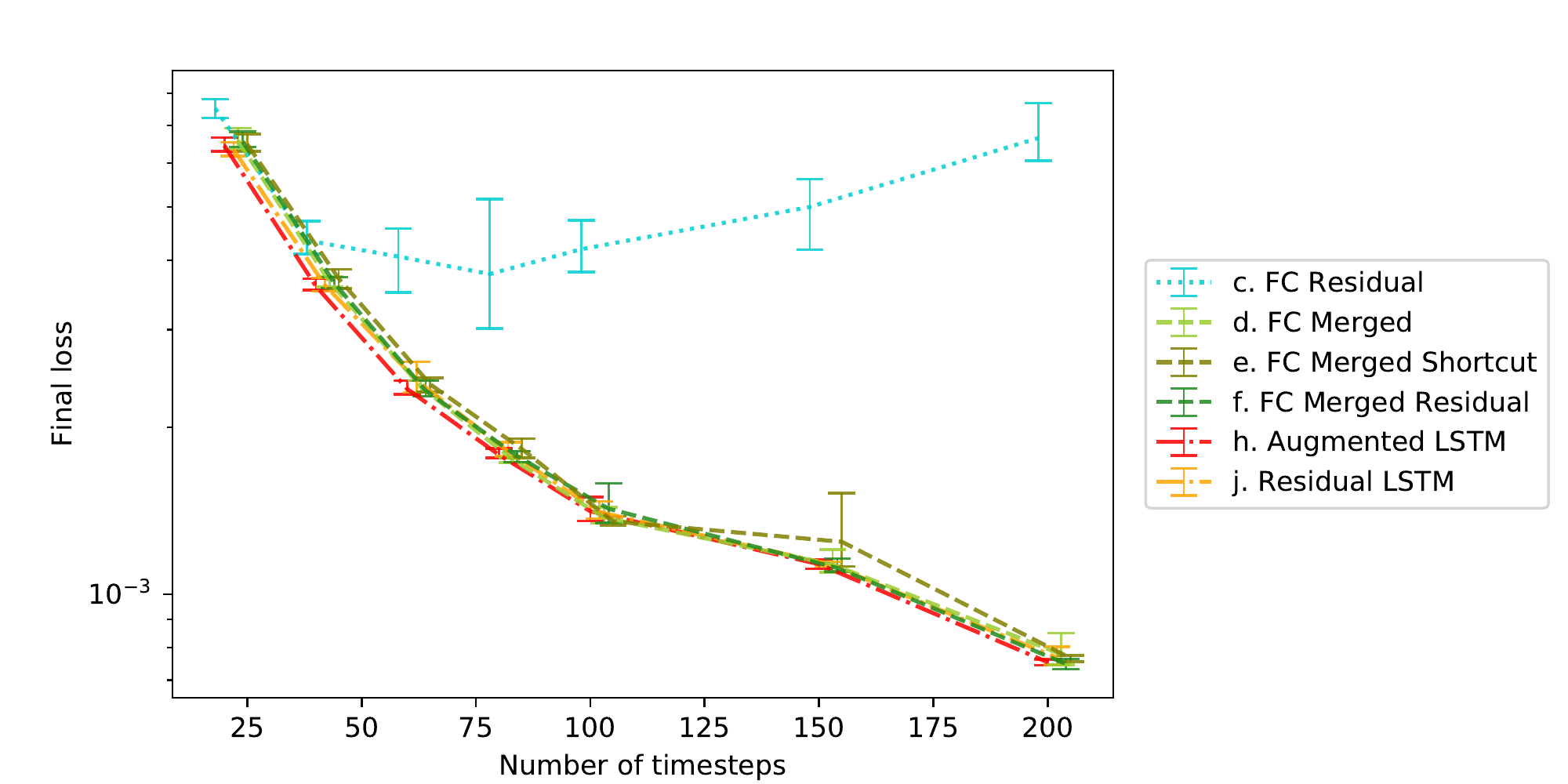} \\[0.5cm]
\textbf{6.}\includegraphics[height=4cm, trim=0cm 0cm 5cm 0cm, clip]{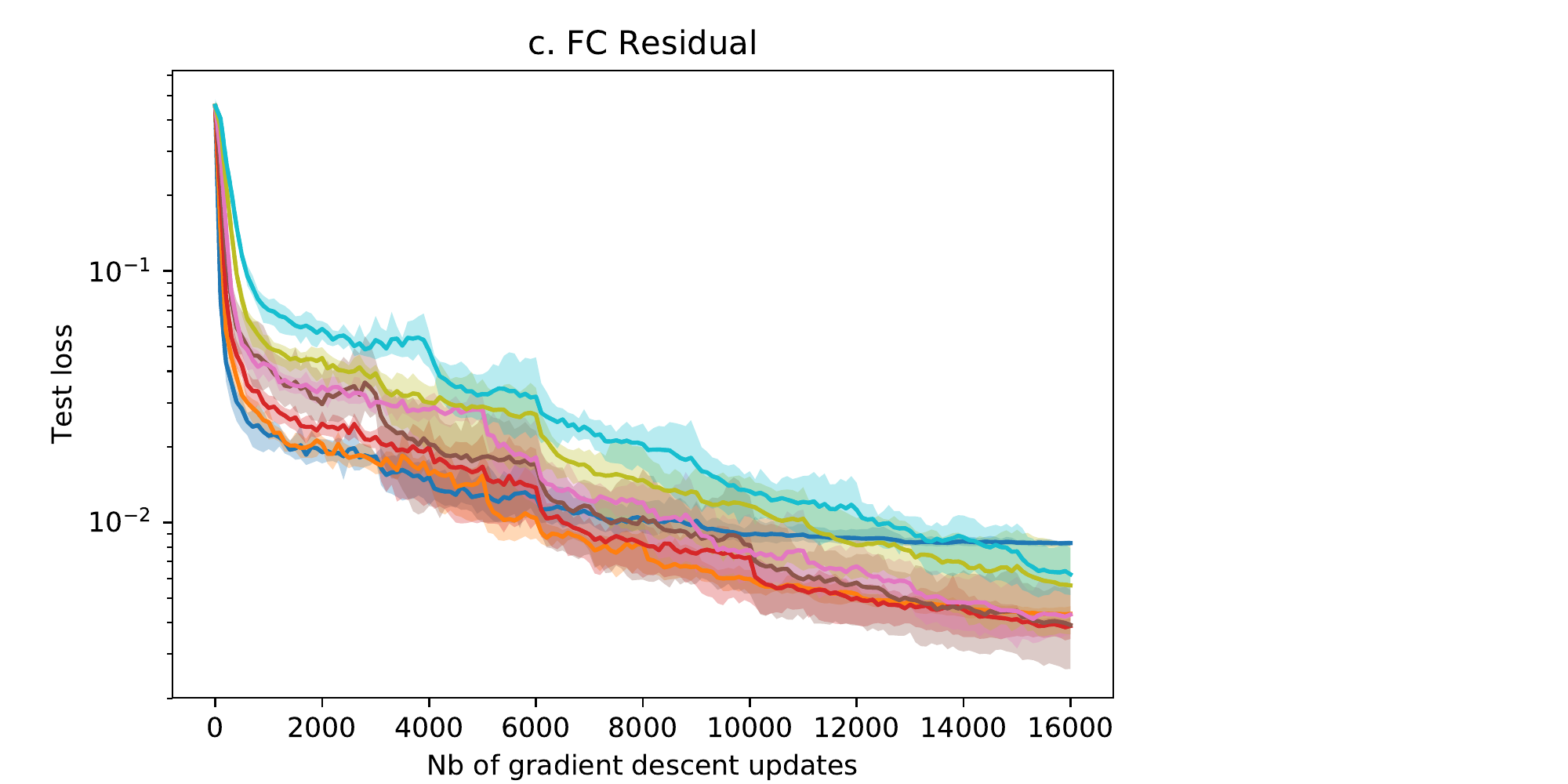}
\textbf{7.}\includegraphics[height=4cm]{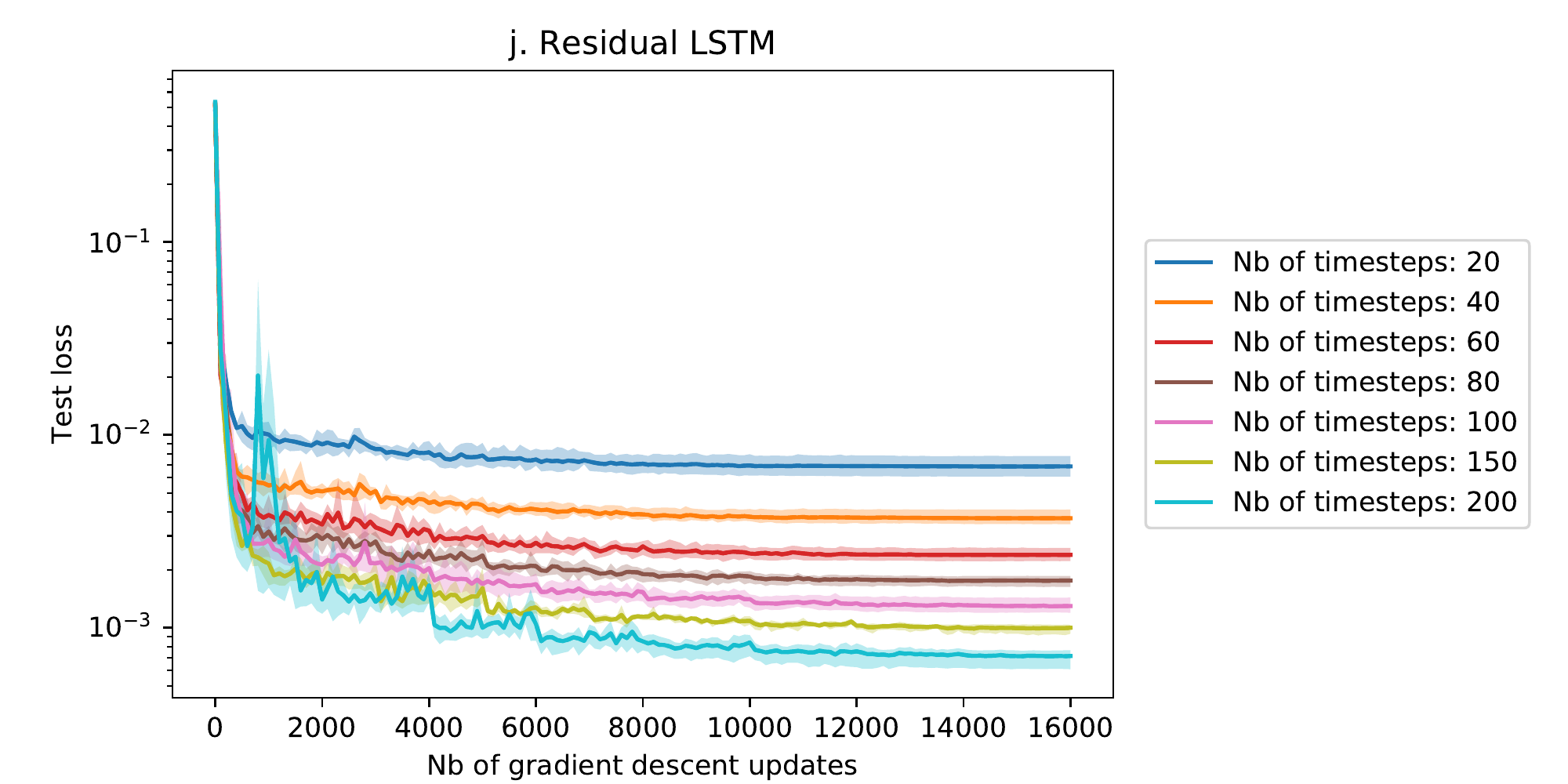}
\end{adjustwidth}

\caption{Influence of the \textbf{number of time steps} on equation \ref{pde:cir} ($d=100$) on $5$ different runs, with the standard learning hyperparameters (see \ref{subsubsection:standardtraining}). We do not represent network \textbf{g.}, which losses are much higher and did not decrease during training. Note that networks \textbf{f.} and \textbf{i.} proved unstable during training with a large number of time steps. \textbf{1.}, \textbf{2.}: we represent the final relative error on $Y_0$ \eqref{eq:relerrY0} and the relative error on $Z_0$ \eqref{eq:relerrZ0} (corresponding to the lowest test loss obtained during training). \textbf{3.}, \textbf{4.}: we represent the mean of the integral error on $Y$ \eqref{eq:interrY} and the integral error on $Z$ \eqref{eq:interrZ} (on all the trajectory) and the $5-95\%$ confidence intervals. \textbf{5.}: we represent the final test losses. We represent the test losses during training for \textbf{c.} and \textbf{j.} in \textbf{6.} and \textbf{7.} and show the merged network see better convergence with a higher number of time steps, while the not merged network does not.}
\label{fig:influence_ntsteps_cird100}
\end{figure}

On equation \ref{pde:cir} ($d=100$, Figure \ref{fig:influence_ntsteps_cird100}), the networks \textbf{a.}, \textbf{b.}, \textbf{c.} perform worse than the Merged and LSTM networks. The other networks show very similar performance, while the network \textbf{i.} shows some instabilities.

Overall, Merged and LSTM networks show significant improvements on the results of \emph{Deep BSDE} compared to the other architectures. Increasing the number of time steps generally improve the results of the Merged and LSTM networks, while degrading or having no effects on the standard FC network.

\begin{table}[ht]
\begin{adjustwidth}{-1.5cm}{-1.5cm}
\centering
\begin{tabular}{|p{4cm}|p{2.5cm}|p{2.5cm}|p{2.5cm}|}
\hline
Network & \ref{pde:richou} ($d=10$) & \ref{pde:bsbarenblatt} ($d=100$) & \ref{pde:cir} ($d=100$) \\
\hline
\textbf{a.} FC DBSDE & $60$ & \fillred & \fillred \\
\hline
\textbf{b.} FC ELU & $60$ & \fillred & \fillred \\
\hline
\textbf{c.} FC Residual & $60$ & $20$ & $60$ \\
\hline
\textbf{d.} FC Merged & $150$ & $100$ & \fillgreen $200+$ \\
\hline
\textbf{e.} FC Merged Shortcut & $200+$ & $200$ \fillgreen & \fillgreen $200+$ \\
\hline
\textbf{f.} FC Merged Residual & $200+$  & $200$ & \fillgreen $200+$ \\
\hline
\textbf{g.} LSTM & $200+$ & \fillred & \fillred \\
\hline
\textbf{h.} Augmented LSTM & $200+$ \fillgreen & $100$ & \fillgreen $200+$ \\
\hline
\textbf{i.} Hybrid LSTM & $200+$ \fillgreen & \fillred & \fillorange $200+$ \\
\hline
\textbf{j.} Residual LSTM & $200+$ \fillgreen & $200$ \fillgreen & \fillgreen $200+$ \\
\hline
\end{tabular}
\end{adjustwidth}
\caption{Number of time steps that yield the better results (among $20$, $40$, $60$, $80$, $100$, $150$, $200$ or $200+$ for larger values). The cell is filled in red if the algorithm did not converge (had a final loss much greater than the other algorithms), in green if it corresponds the best results, in orange if it is unstable.}
\end{table}

\clearpage
\subsubsection{Influence of the non-linearity in the driver}

In this section, we investigate the influence of the non-linearity in the driver on the results of our algorithms. We use the equations \ref{pde:warin2} and \ref{pde:warin3}, in which the parameter $r$ is the rescale factor in $f$: the non-linearity increases with $r$.

Generally speaking, the final test loss and the integral error on $Y$ tend to decrease with $r$, while the other errors increase with $r$.

\begin{figure}[p]
\begin{adjustwidth}{-1.5cm}{-1.5cm}
\centering
\textbf{1.}\includegraphics[height=4cm, trim=0cm 0cm 5cm 0cm, clip]{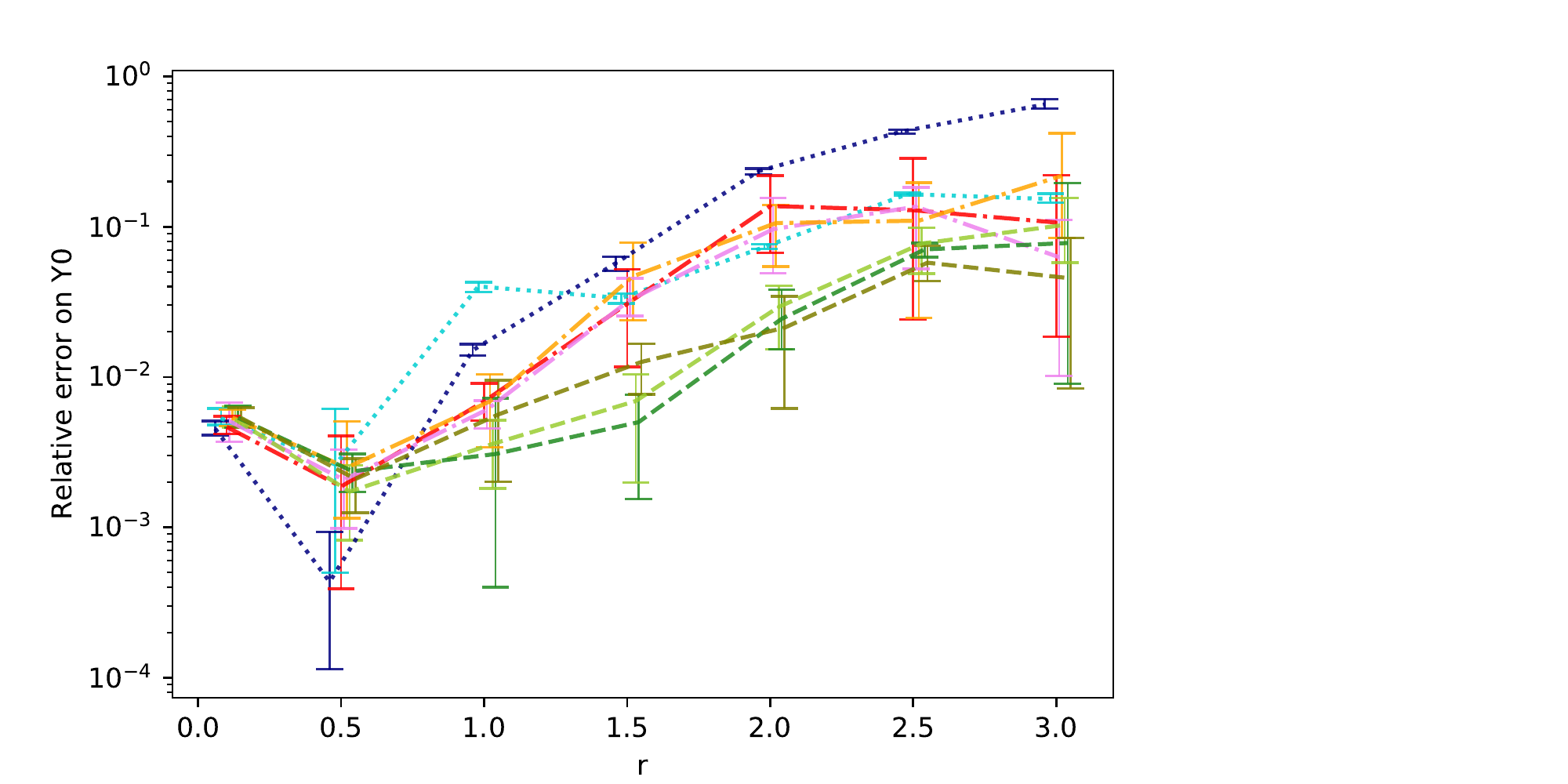}
\textbf{2.}\includegraphics[height=4cm]{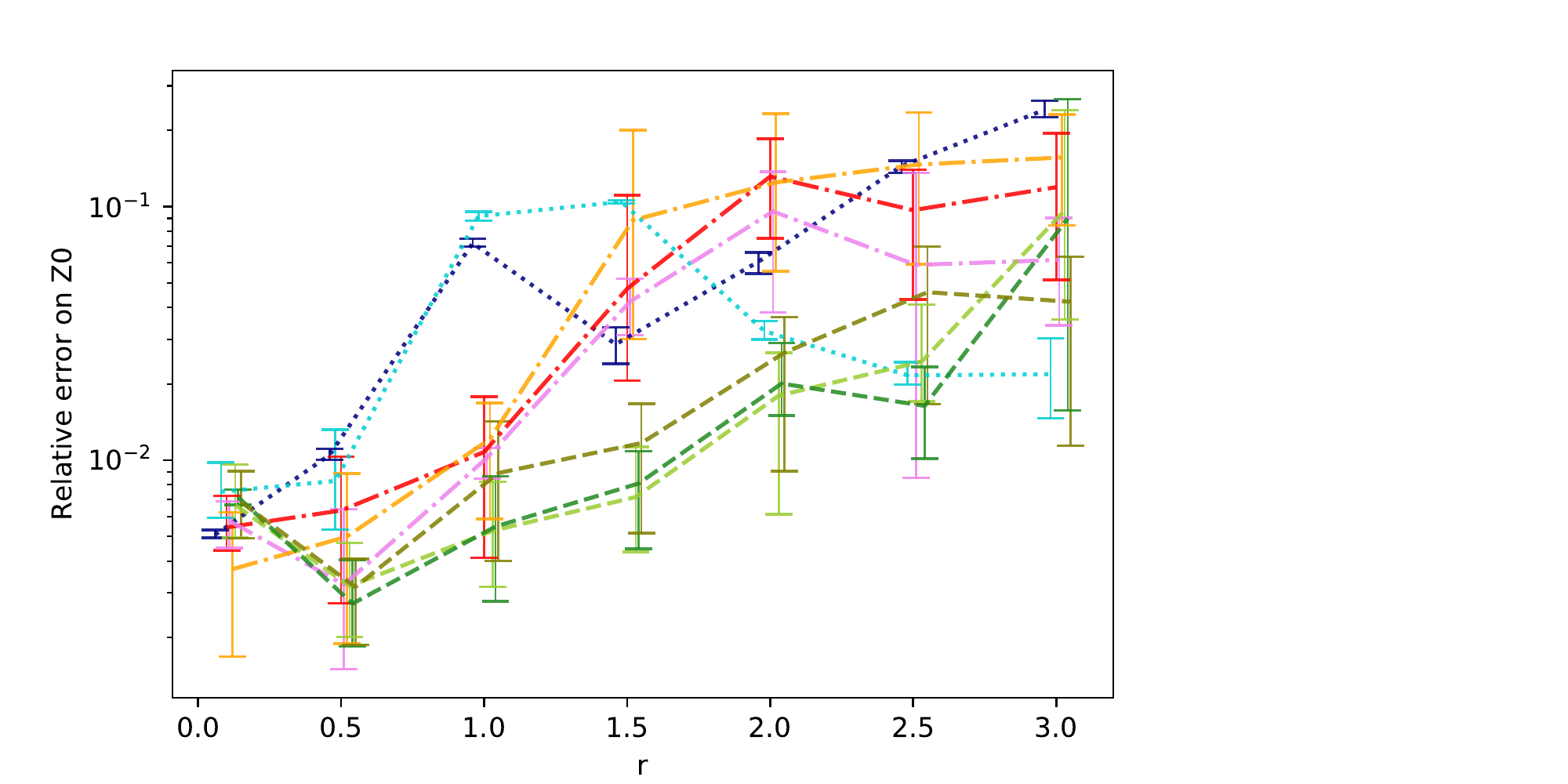} \\
\textbf{3.}\includegraphics[height=4cm, trim=0cm 0cm 5cm 0cm, clip]{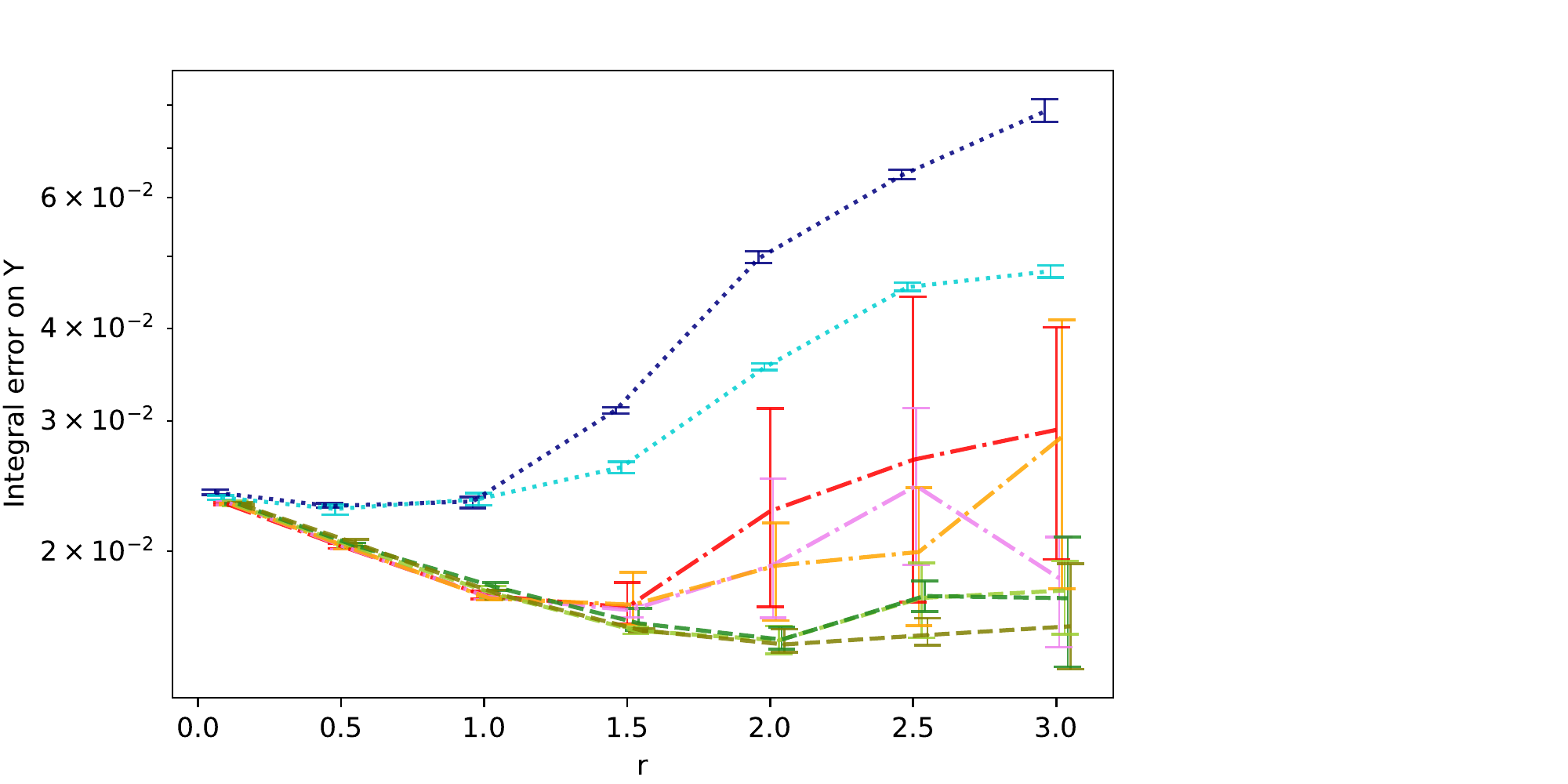}
\textbf{4.}\includegraphics[height=4cm]{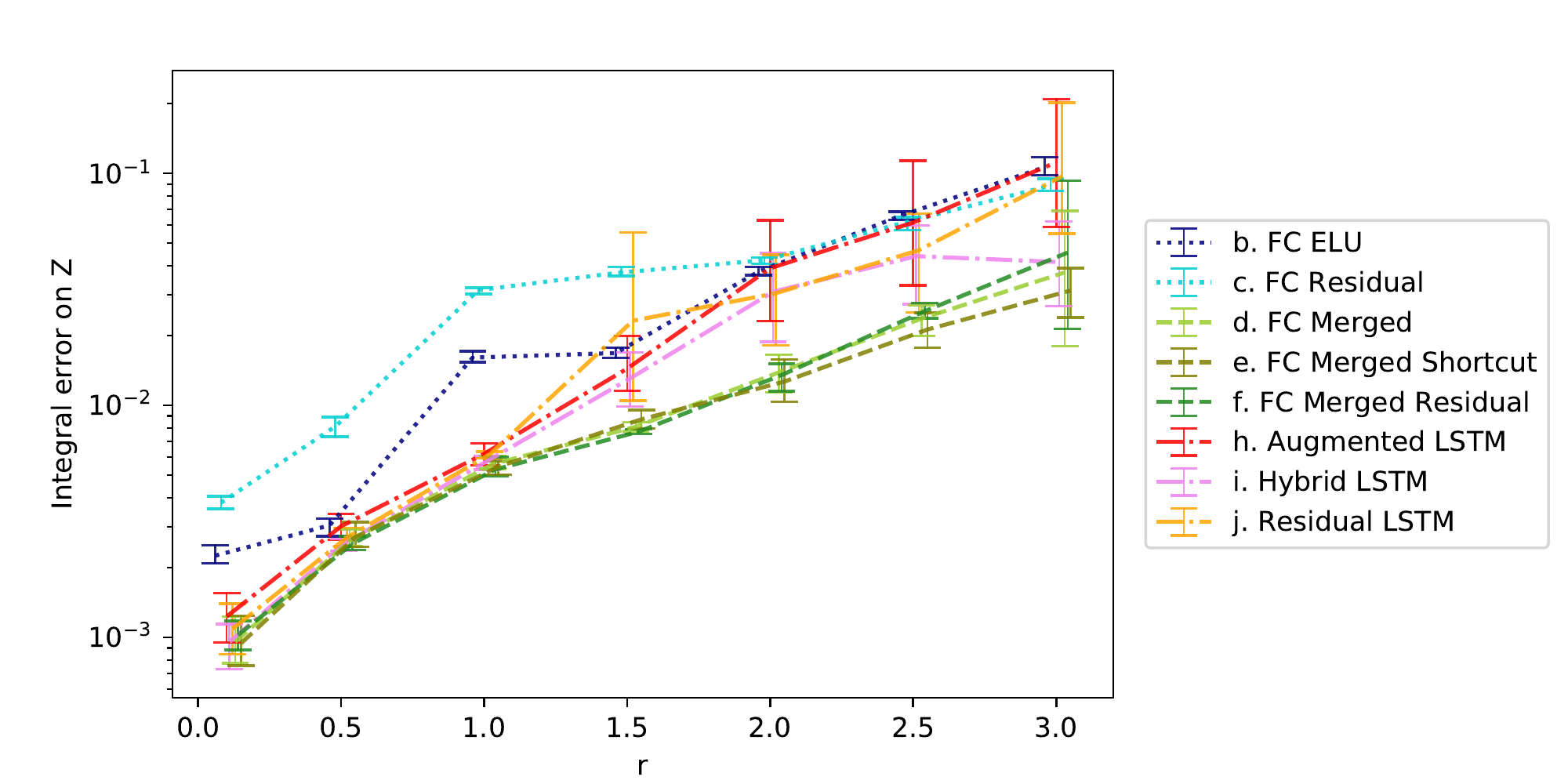} \\
\textbf{5.}\includegraphics[height=4cm]{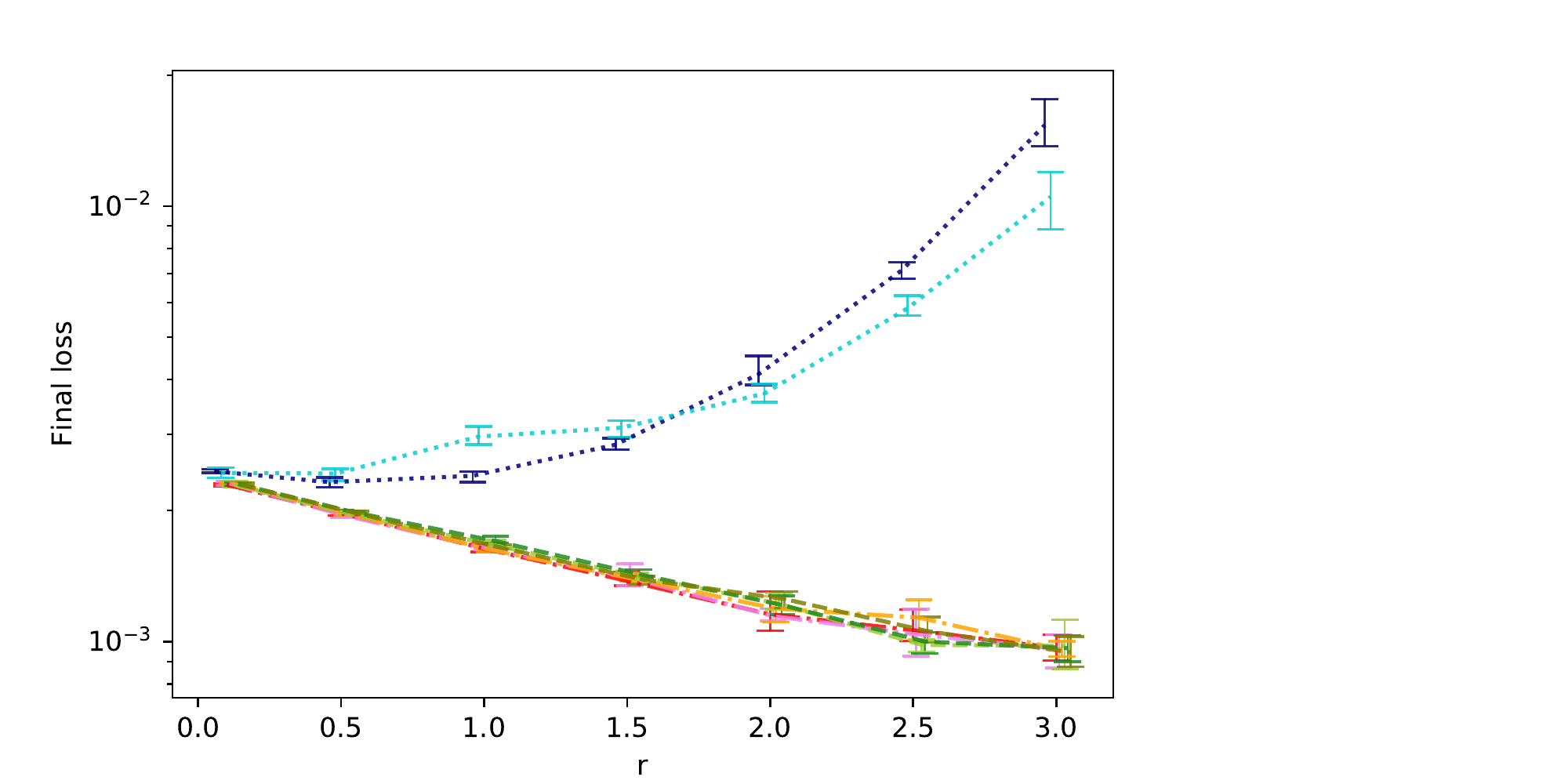} \\[0.5cm]
\textbf{6.}\includegraphics[height=4cm, trim=0cm 0cm 5cm 0cm, clip]{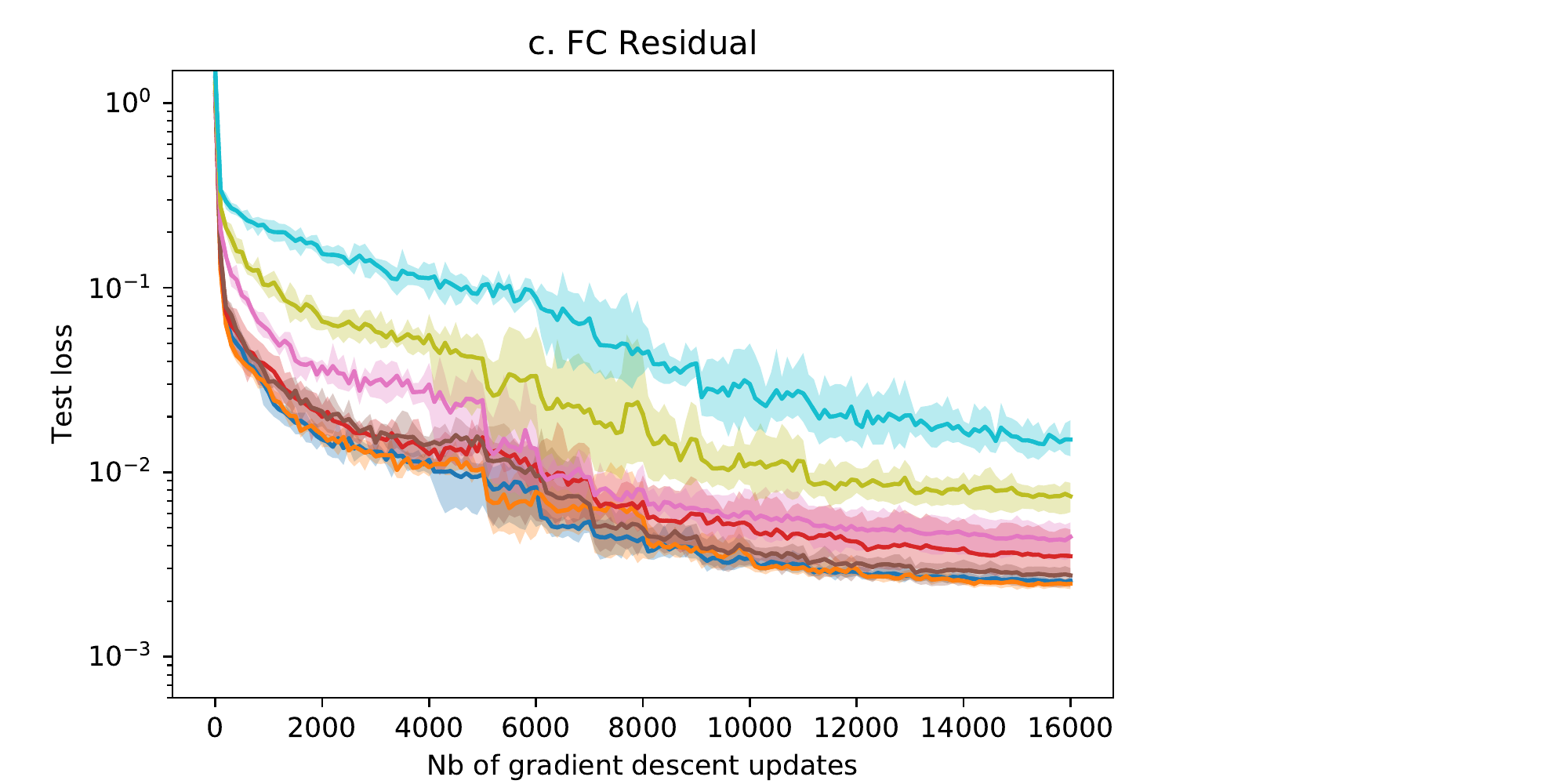}
\textbf{7.}\includegraphics[height=4cm]{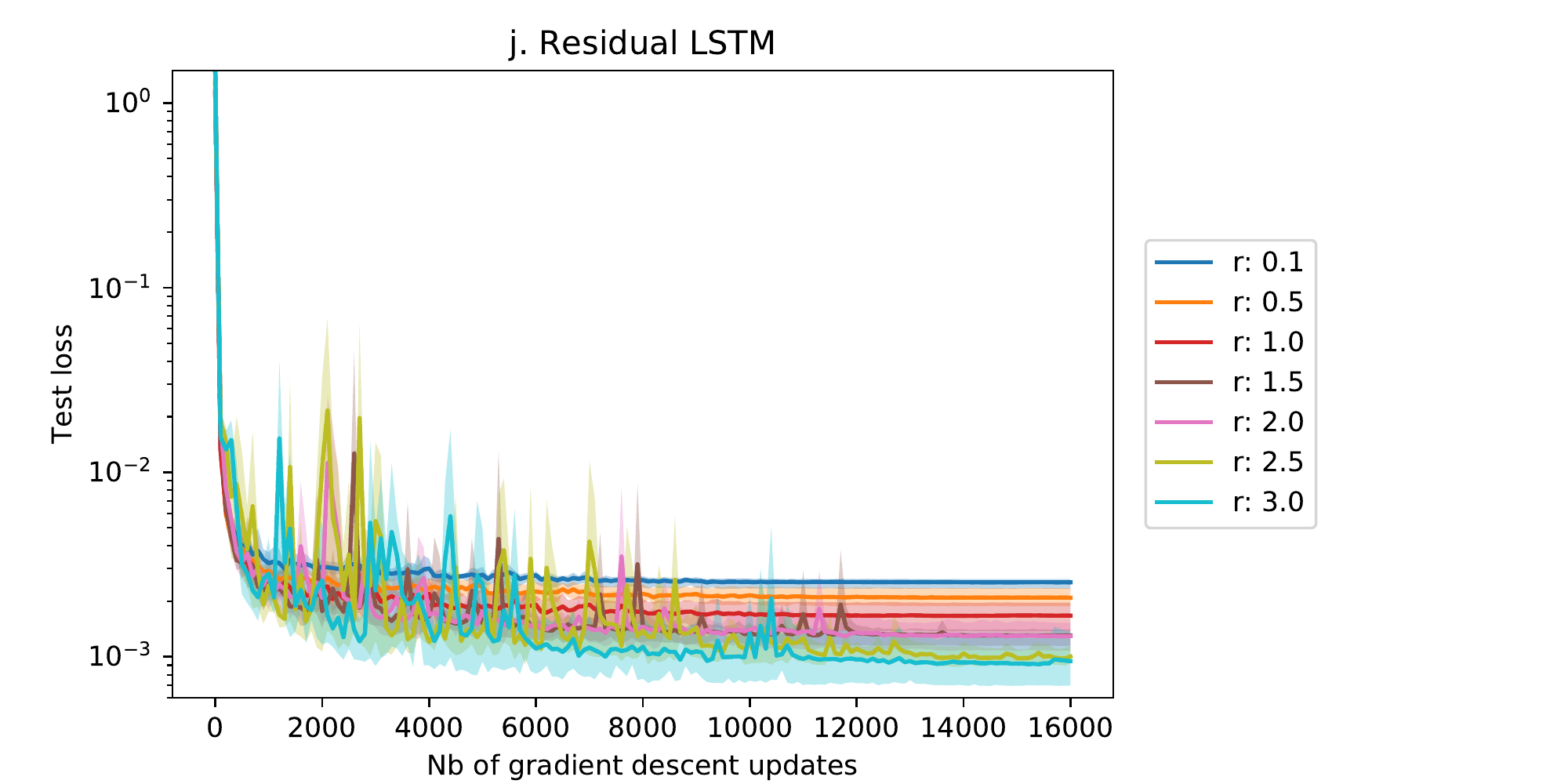}
\end{adjustwidth}
\caption{Influence of the \textbf{non-linearity} on $f$ (parameter $r$) on equation \ref{pde:warin2} ($d=10$, $N=100$). We represent \textbf{1.} the relative error on $Y_0$ \eqref{eq:relerrY0}, \textbf{2.} the relative error on $Z_0$ \eqref{eq:relerrZ0}, \textbf{3.} the integral error on $Y$ \eqref{eq:interrY}, \textbf{4.} the integral error on $Z$ \eqref{eq:interrZ}, \textbf{5.} the final test loss. The mean and the $5\%$ and $95\%$ quantiles are computed on $5$ independent runs and represented with the lines and error bars. Networks \textbf{a.} and \textbf{g.} do not converge on this example (their losses do not decrease during training and remain significantly higher than the other networks) and are not represented here. Interestingly, the final test loss decreases with $r$ while the other error measures tend to increase. In \textbf{6.} and \textbf{7.}, we represent the evolution of the test loss during training.}
\label{fig:infr:warin2}
\end{figure}

On equation \ref{pde:warin2} ($d=10$, Figure \ref{fig:infr:warin2}), the networks \textbf{a.} and \textbf{g.} do not converge (remain at a high loss during training). The Merged networks perform better than the other networks overall -- the LSTM networks yield similar performance but show slightly higher errors for high $r$'s. The final losses of the standard FC networks increase greatly with $r$, showing these networks have difficulties to converge.

\begin{figure}[p]
\begin{adjustwidth}{-1.5cm}{-1.5cm}
\centering
\textbf{1.}\includegraphics[height=4cm, trim=0cm 0cm 5cm 0cm, clip]{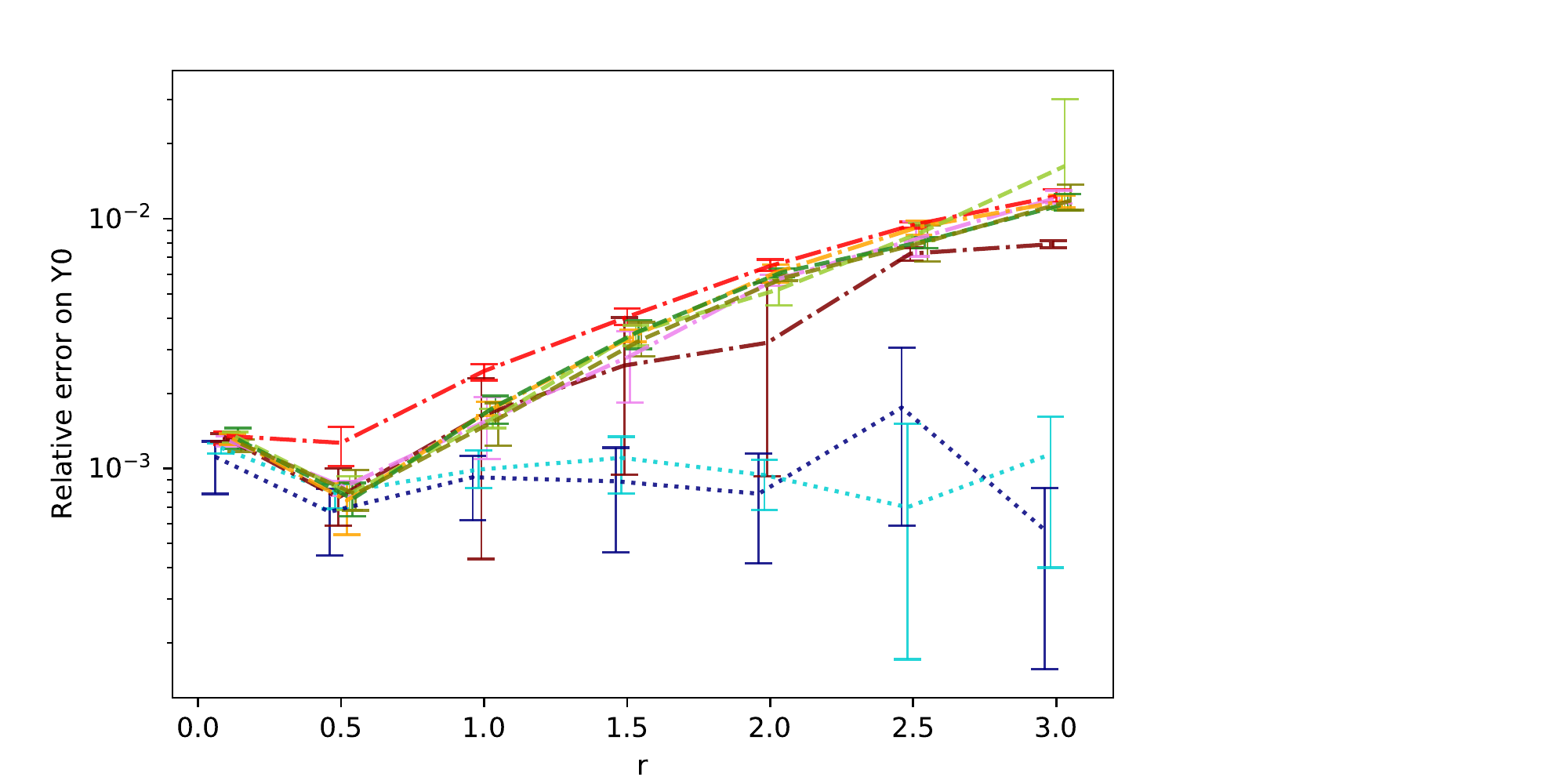}
\textbf{2.}\includegraphics[height=4cm]{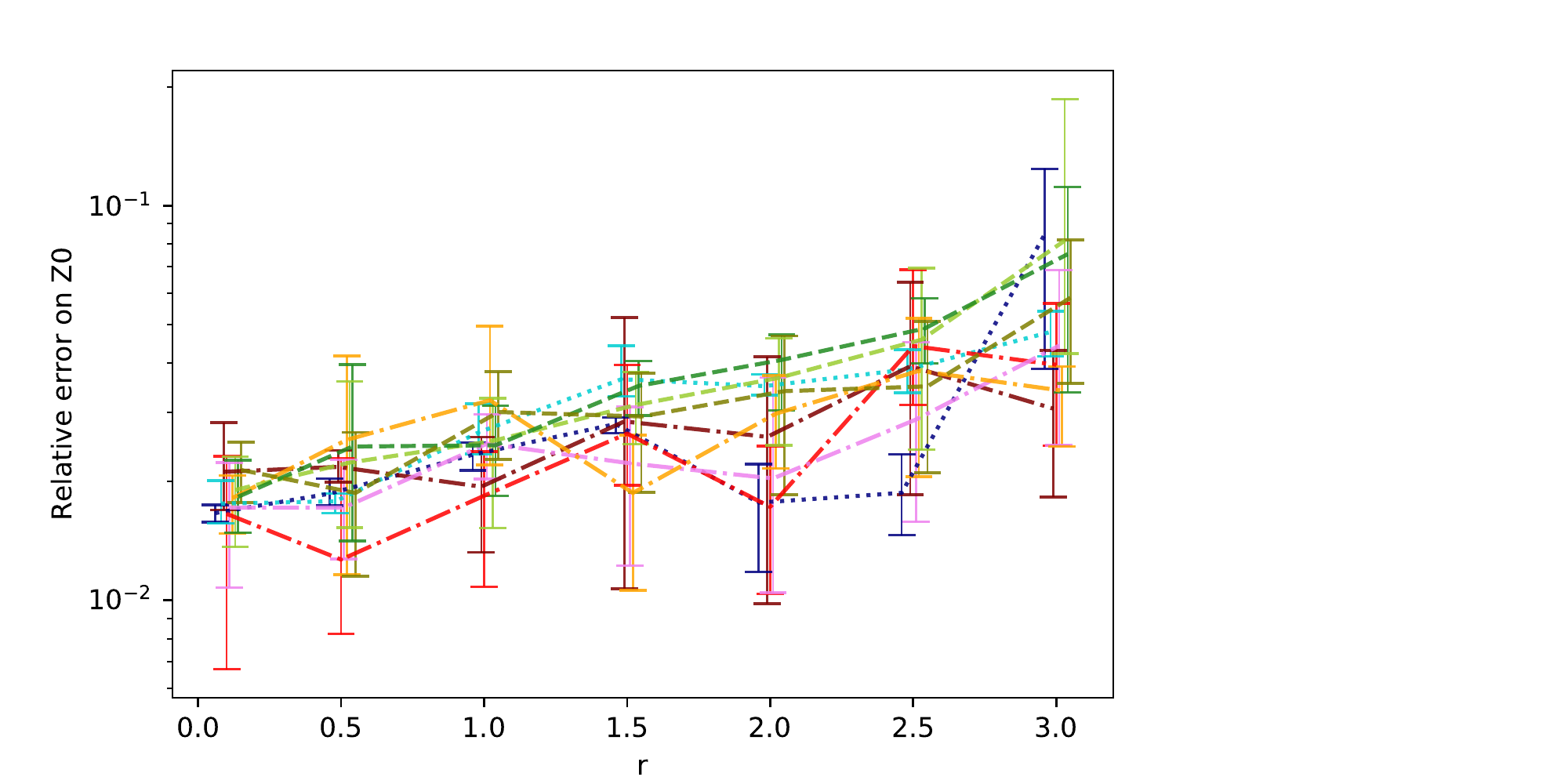} \\
\textbf{3.}\includegraphics[height=4cm, trim=0cm 0cm 5cm 0cm, clip]{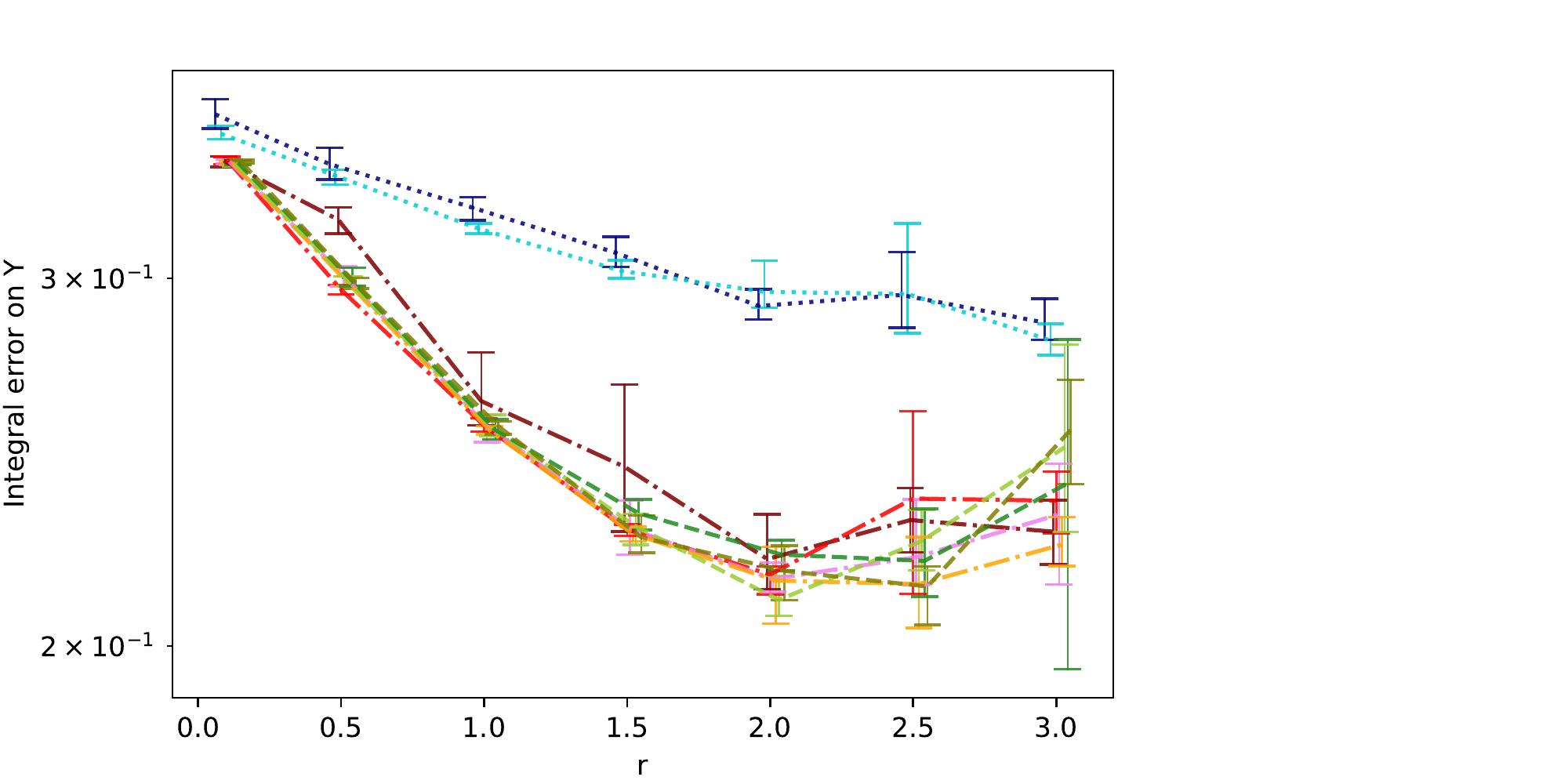}
\textbf{4.}\includegraphics[height=4cm]{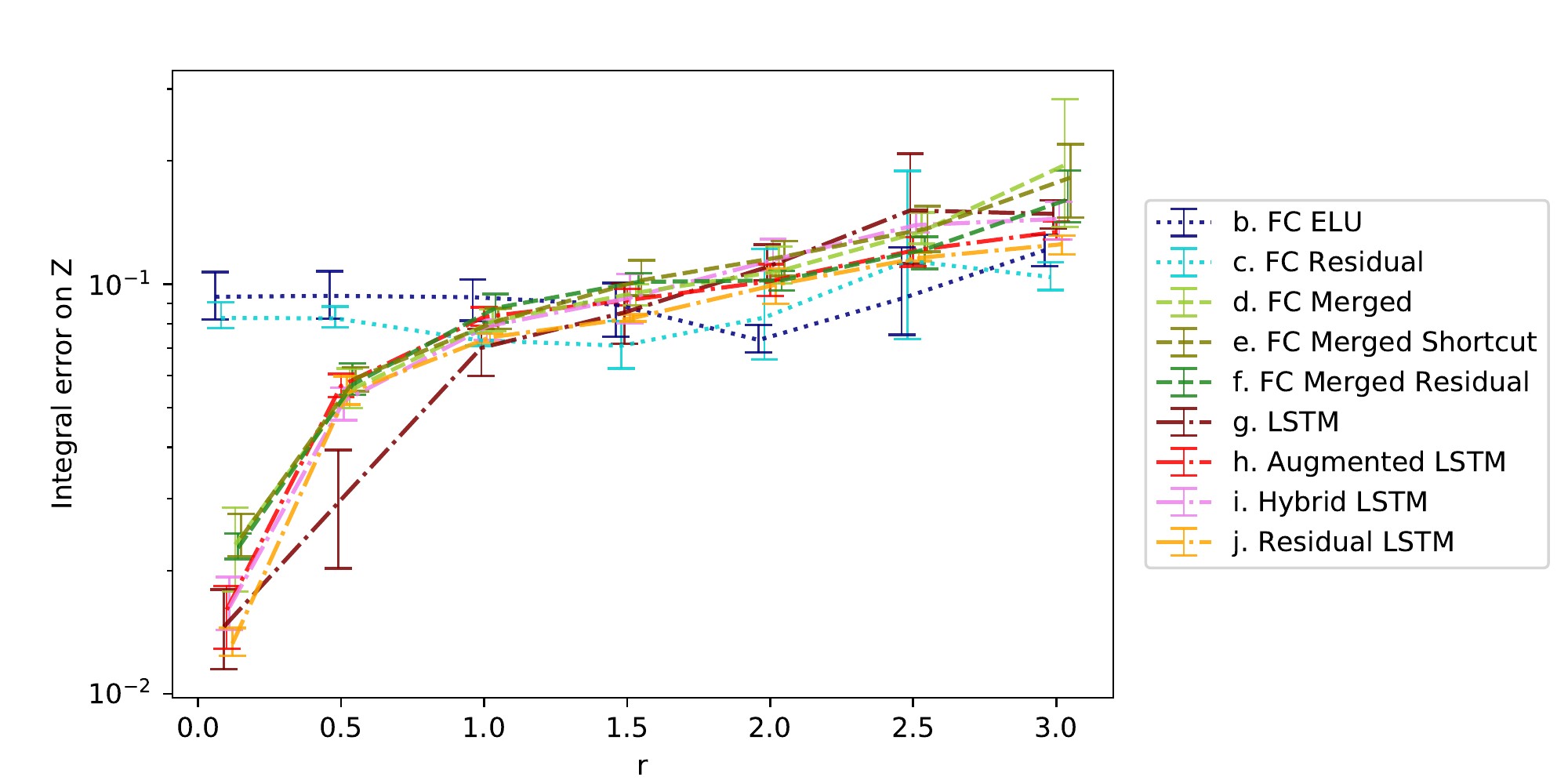} \\
\textbf{5.}\includegraphics[height=4cm]{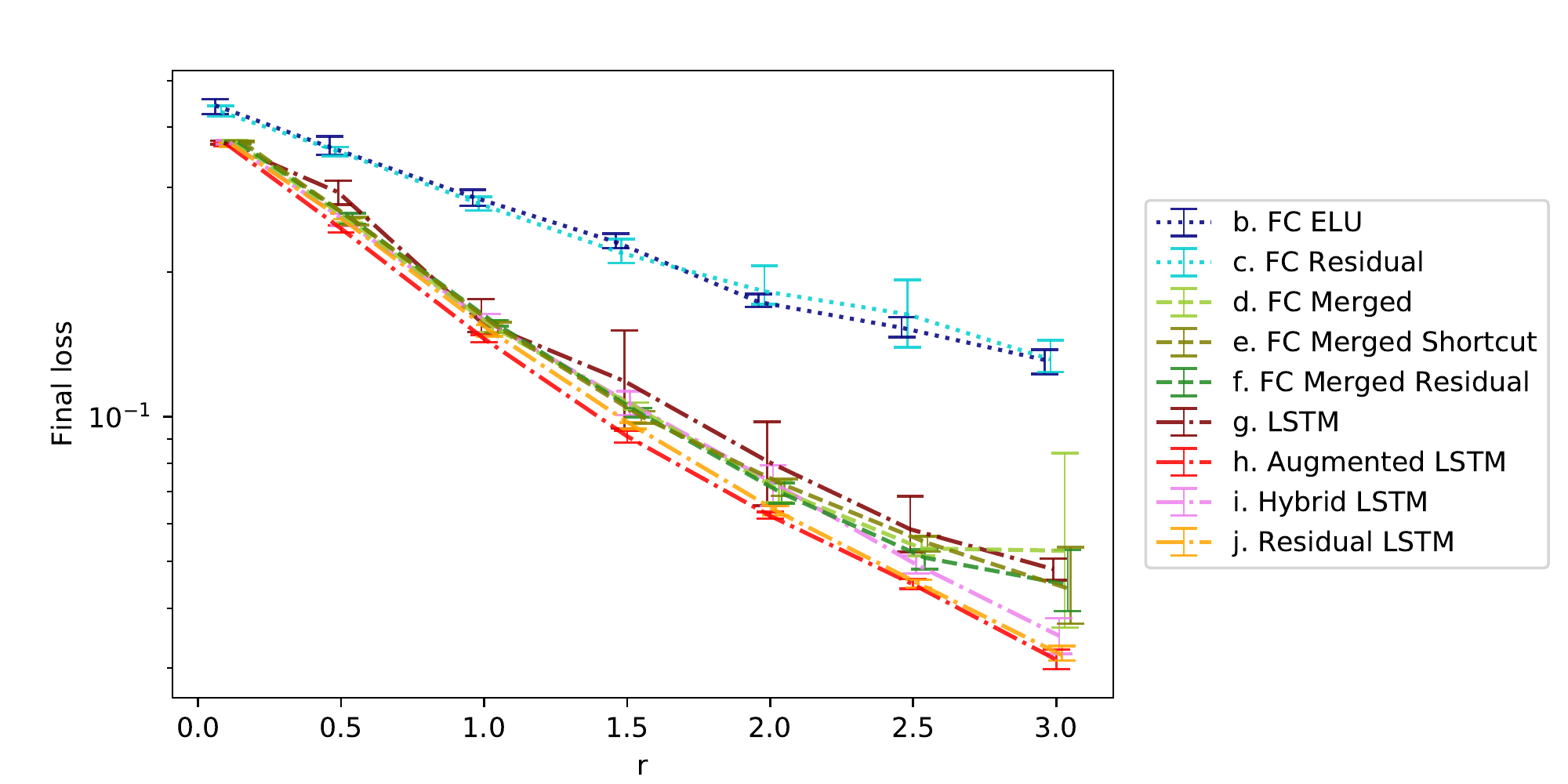} \\[0.5cm]
\textbf{6.}\includegraphics[height=4cm, trim=0cm 0cm 5cm 0cm, clip]{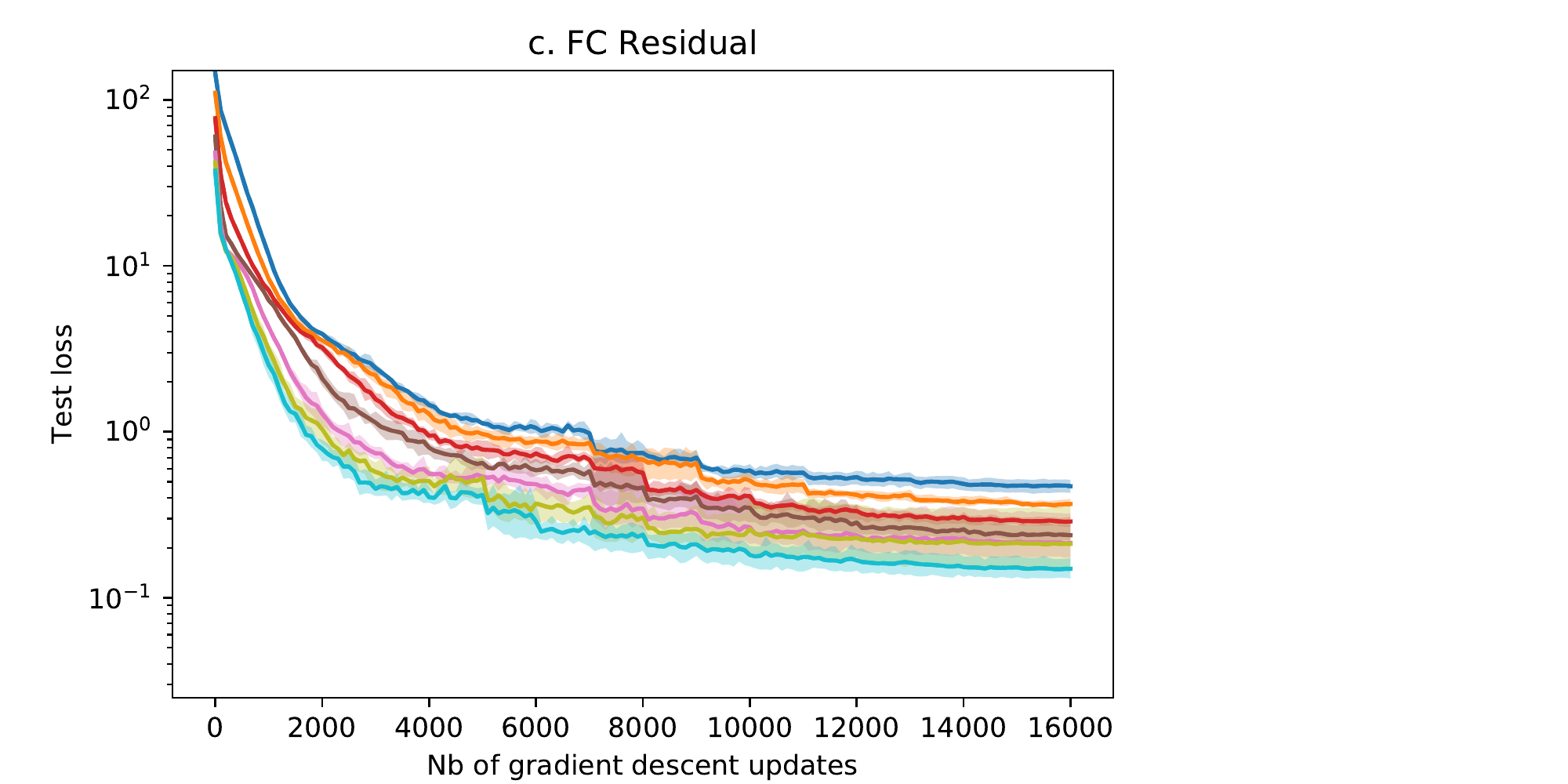}
\textbf{7.}\includegraphics[height=4cm]{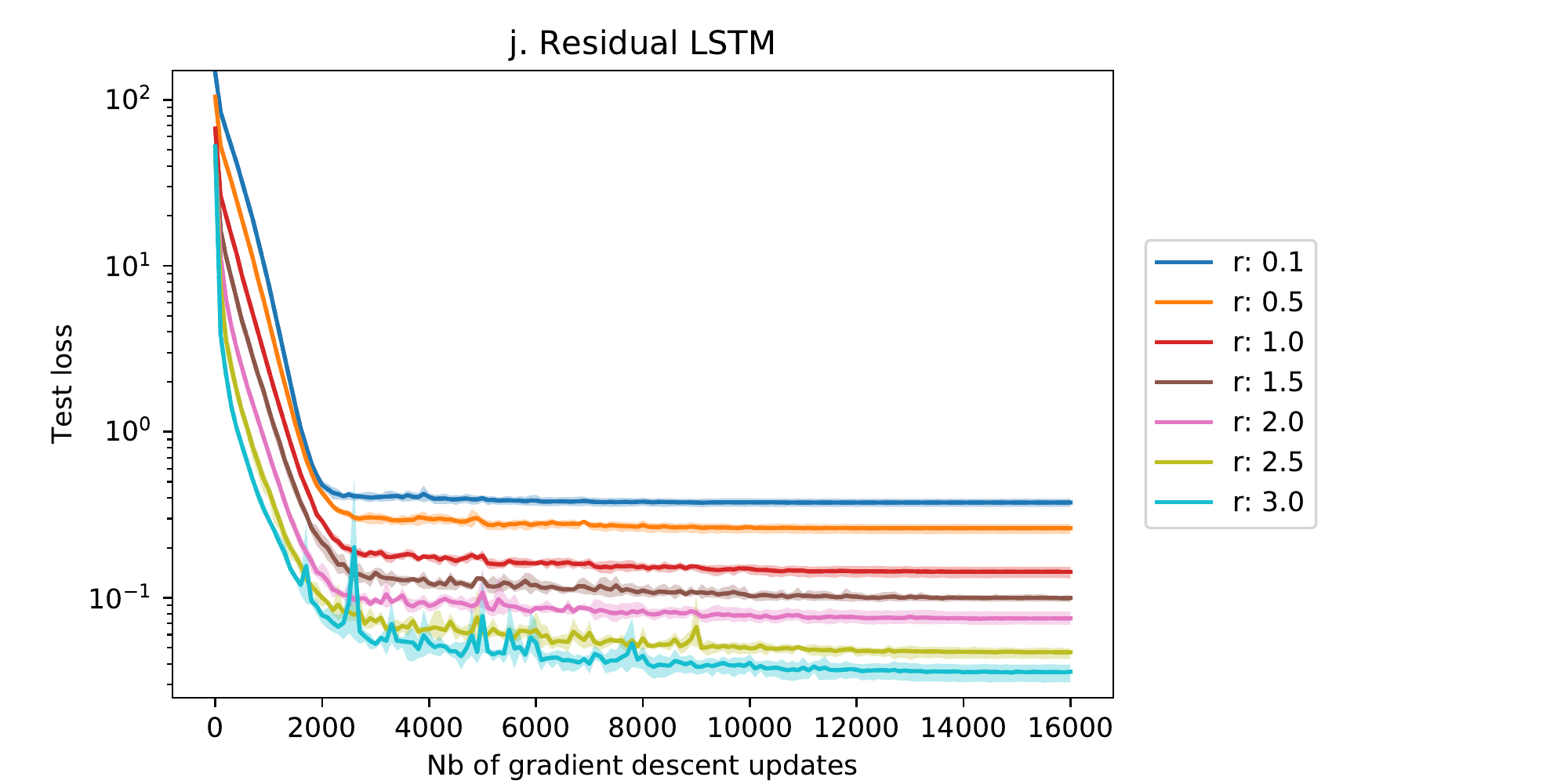}
\end{adjustwidth}
\caption{Influence of the \textbf{non-linearity} on $f$ (parameter $r$) on equation \ref{pde:warin3} ($d=10$, $N=100$). We represent \textbf{1.} the relative error on $Y_0$ \eqref{eq:relerrY0}, \textbf{2.} the relative error on $Z_0$ \eqref{eq:relerrZ0}, \textbf{3.} the integral error on $Y$ \eqref{eq:interrY}, \textbf{4.} the integral error on $Z$ \eqref{eq:interrZ}, \textbf{5.} the final test loss. The mean and the $5\%$ and $95\%$ quantiles are computed on $5$ independent runs and represented with the lines and error bars. The network \textbf{a.} do not converge on this example (its loss does not decrease during training and remain significantly higher than the other networks) and is not represented here. In \textbf{6.} and \textbf{7.}, we represent the evolution of the test loss during training.}
\label{fig:infr:warin3}
\end{figure}

On equation \ref{pde:warin3} ($d=10$, Figure \ref{fig:infr:warin3}) the network \textbf{a.} does not converge -- the standard FC networks \textbf{b.} and \textbf{c.} have lower errors on the initial conditions and greater integral errors than the Merged and LSTM networks, which show similar performance.

Overall, our new architectures seem to be more resilient to a non-linearity increase.

\begin{table}[ht]
\begin{adjustwidth}{-1.5cm}{-1.5cm}
\centering
\begin{tabular}{|p{4cm}|p{4cm}|p{4cm}|}
\hline
Network & \ref{pde:warin2} ($d=10$) & \ref{pde:warin3} ($d=10$) \\
\hline
\textbf{a.} FC DBSDE & \fillred & \fillred \\
\hline
\textbf{b.} FC ELU & \multicolumn{2}{c|}{The final loss increases exponentially with $r$ \fillorange} \\
\hline
\textbf{c.} FC Residual & \multicolumn{2}{c|}{The final loss increases exponentially with $r$ \fillorange} \\
\hline
\textbf{d.} FC Merged & \fillgreen & \fillgreen \\
\hline
\textbf{e.} FC Merged Shortcut & \fillgreen & \fillgreen \\
\hline
\textbf{f.} FC Merged Residual & \fillgreen & \fillgreen \\
\hline
\textbf{g.} LSTM & \fillred & \fillgreen \\
\hline
\textbf{h.} Augmented LSTM & Higher error for $r\geq 2.0$ & \fillgreen \\
\hline
\textbf{i.} Hybrid LSTM & Higher error for $r\geq 2.0$ & \fillgreen \\
\hline
\textbf{j.} Residual LSTM & Higher error for $r\geq 2.0$ & \fillgreen \\
\hline
\end{tabular}
\end{adjustwidth}
\caption{Influence of the \textbf{non-linearity} $r$ on the results. The cells are filled in red if the algorithm did not converge, in green if it yielded the best stable results.}
\end{table}

\clearpage
\subsubsection{Influence of the maturity}

In order to assess the influence of the maturity,  we keep a constant time step $\delta_t = 0.1$, i.e. we take $N = 100 \times T$.

\begin{figure}[ht]
\begin{adjustwidth}{-1.5cm}{-1.5cm}
\centering
\textbf{1.}\includegraphics[height=4cm, trim=0cm 0cm 5cm 0cm, clip]{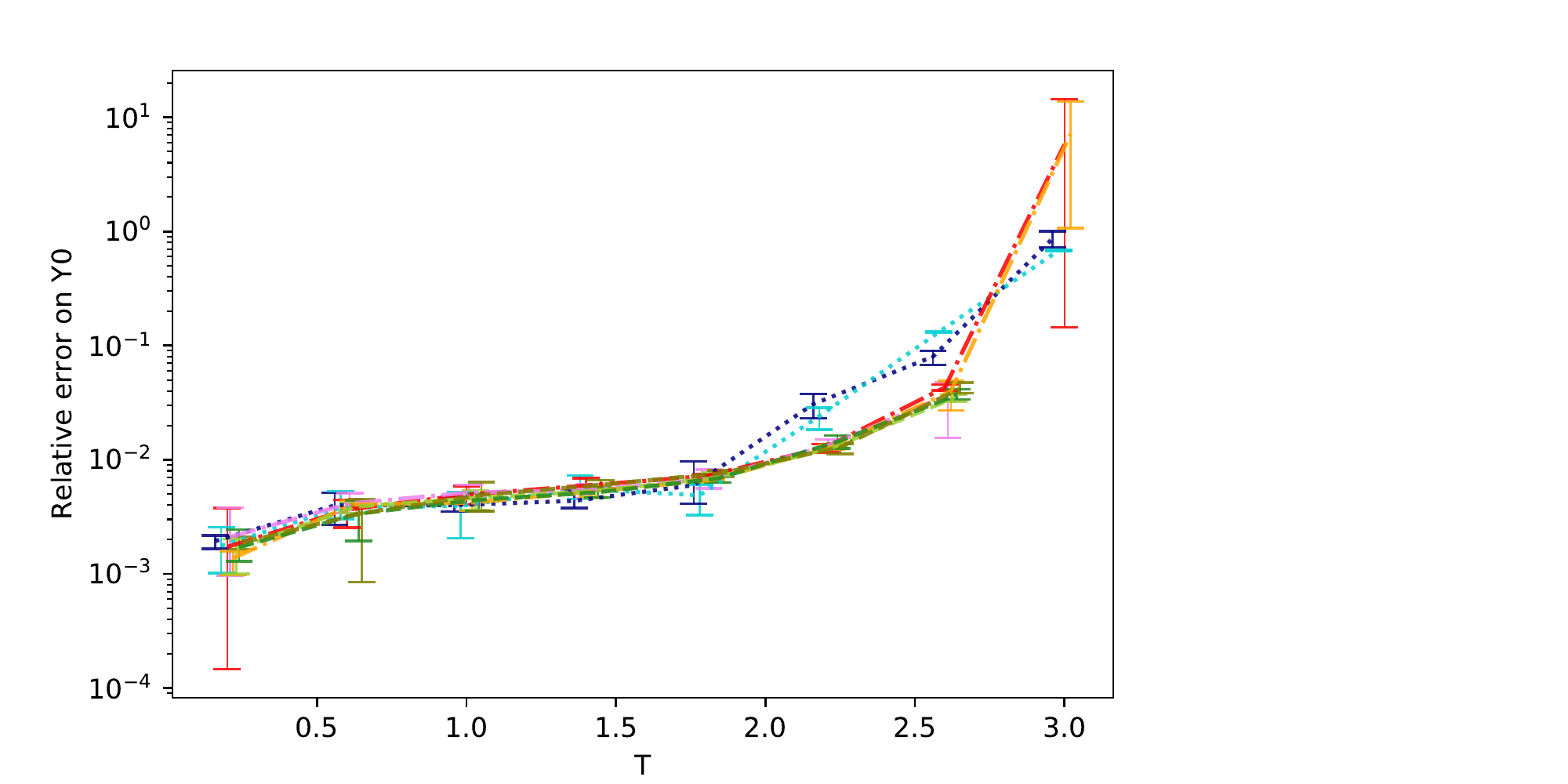}
\textbf{2.}\includegraphics[height=4cm]{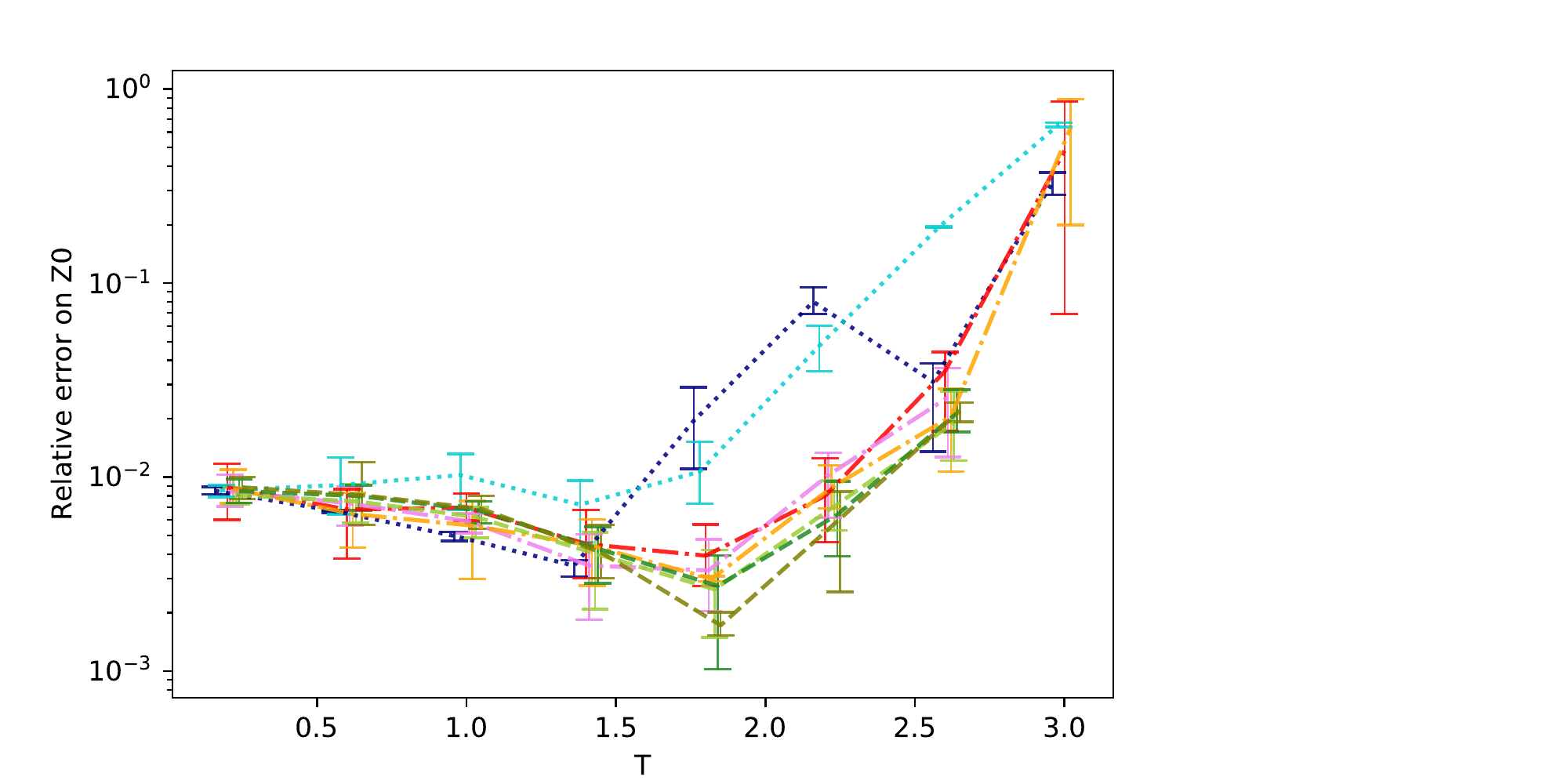} \\
\textbf{3.}\includegraphics[height=4cm, trim=0cm 0cm 5cm 0cm, clip]{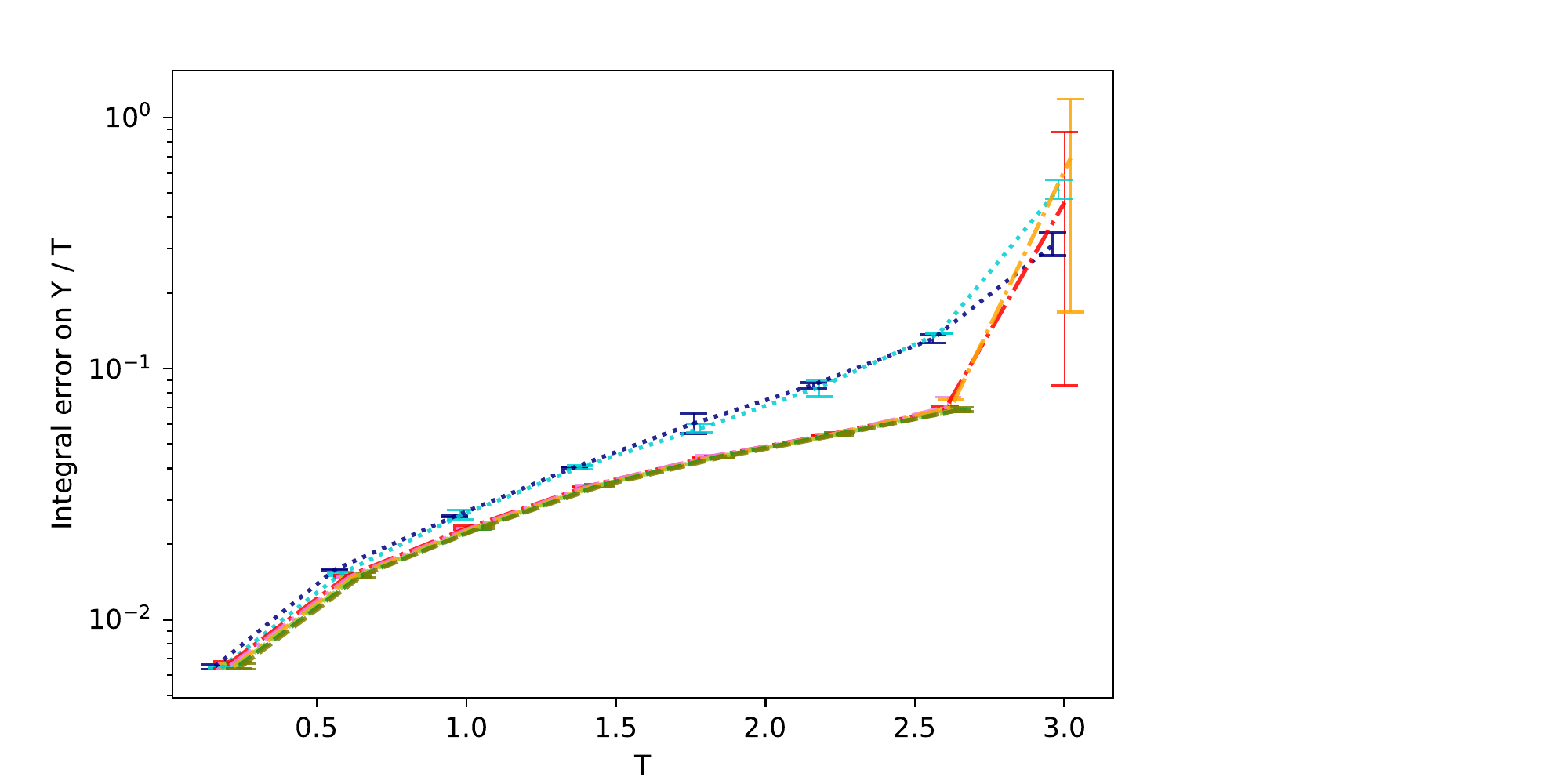}
\textbf{4.}\includegraphics[height=4cm]{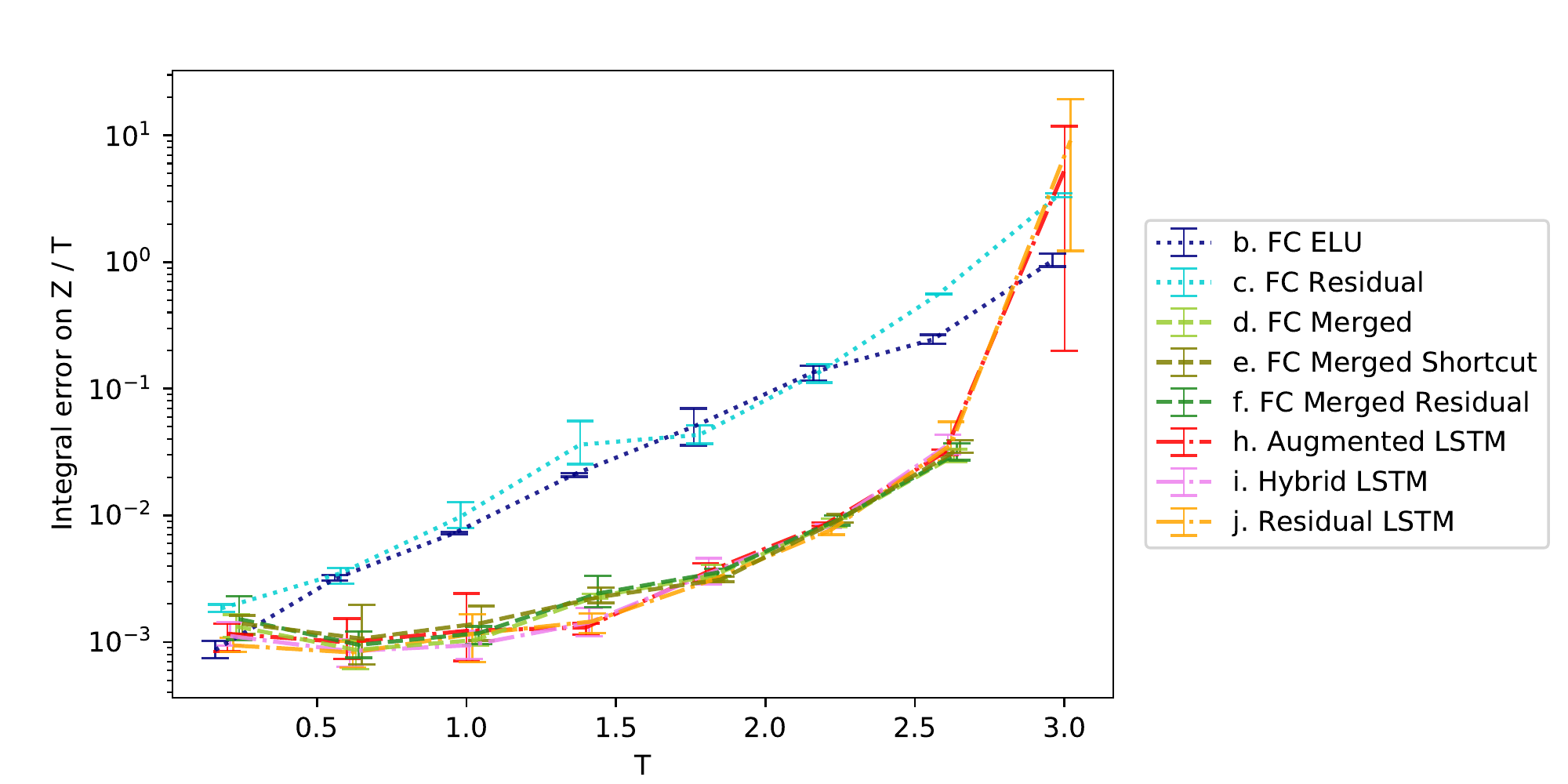} \\
\textbf{5.}\includegraphics[height=4cm]{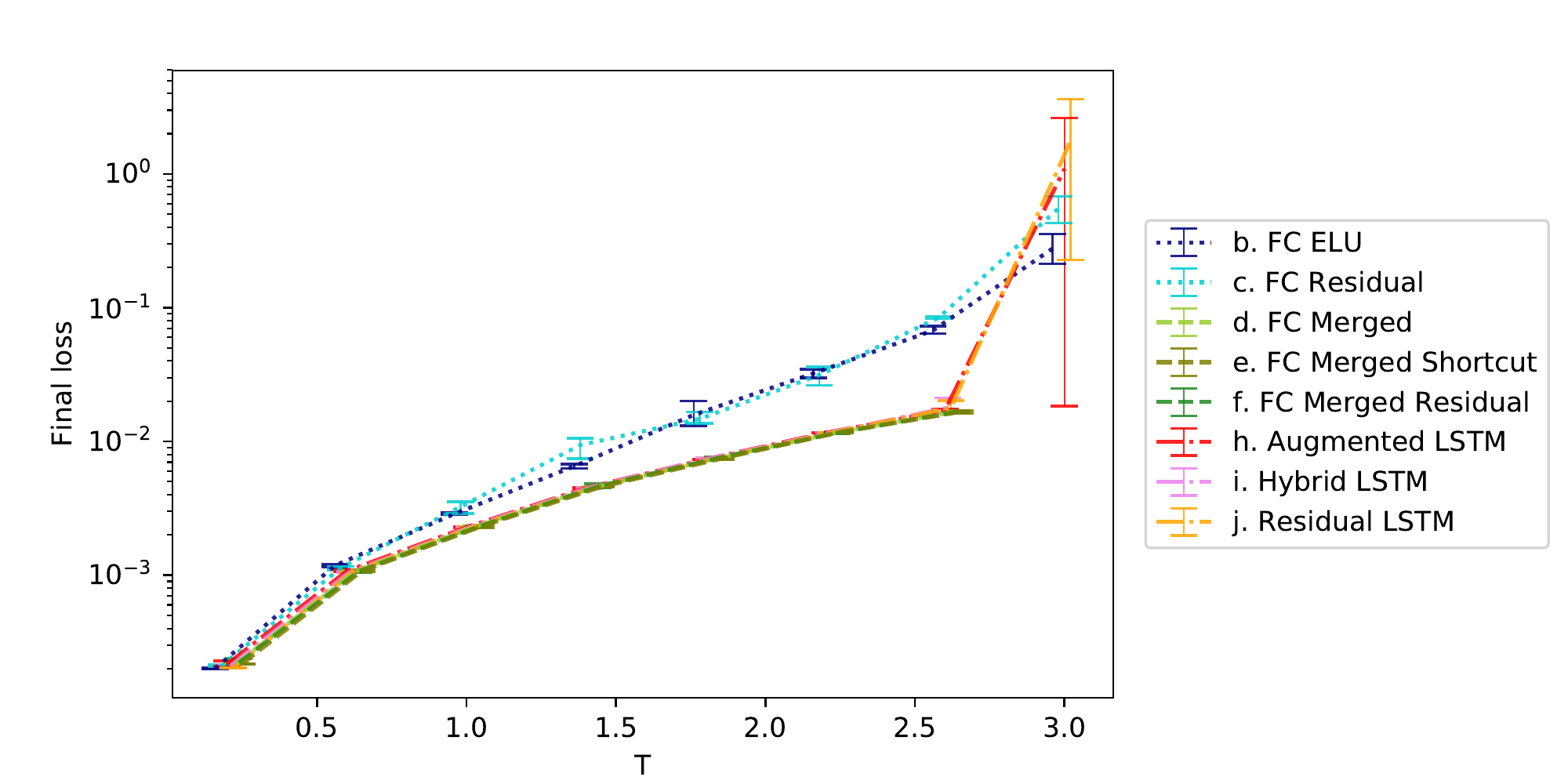} \\[0.5cm]
\textbf{6.}\includegraphics[height=4cm, trim=0cm 0cm 5cm 0cm, clip]{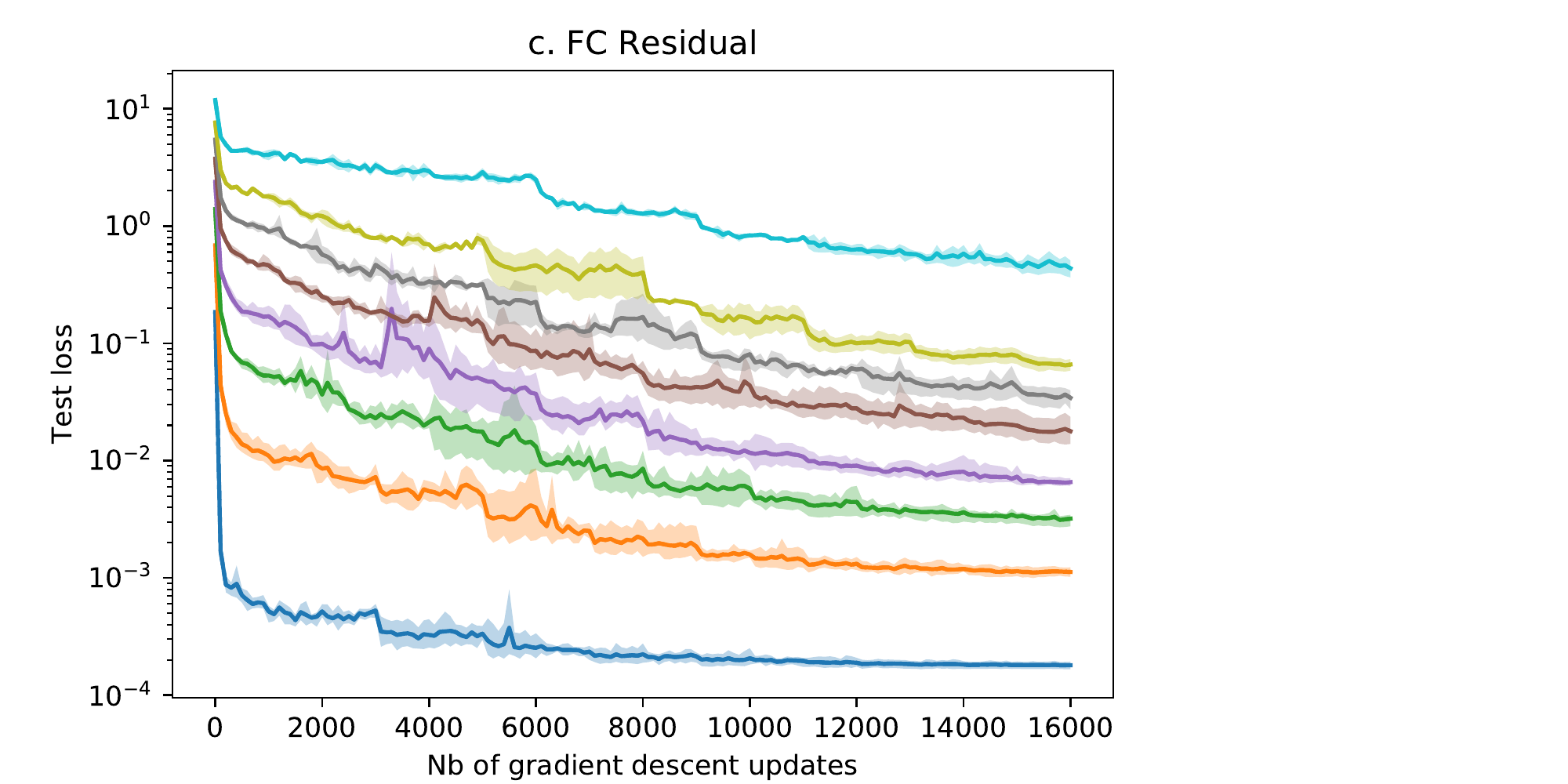}
\textbf{7.}\includegraphics[height=4cm]{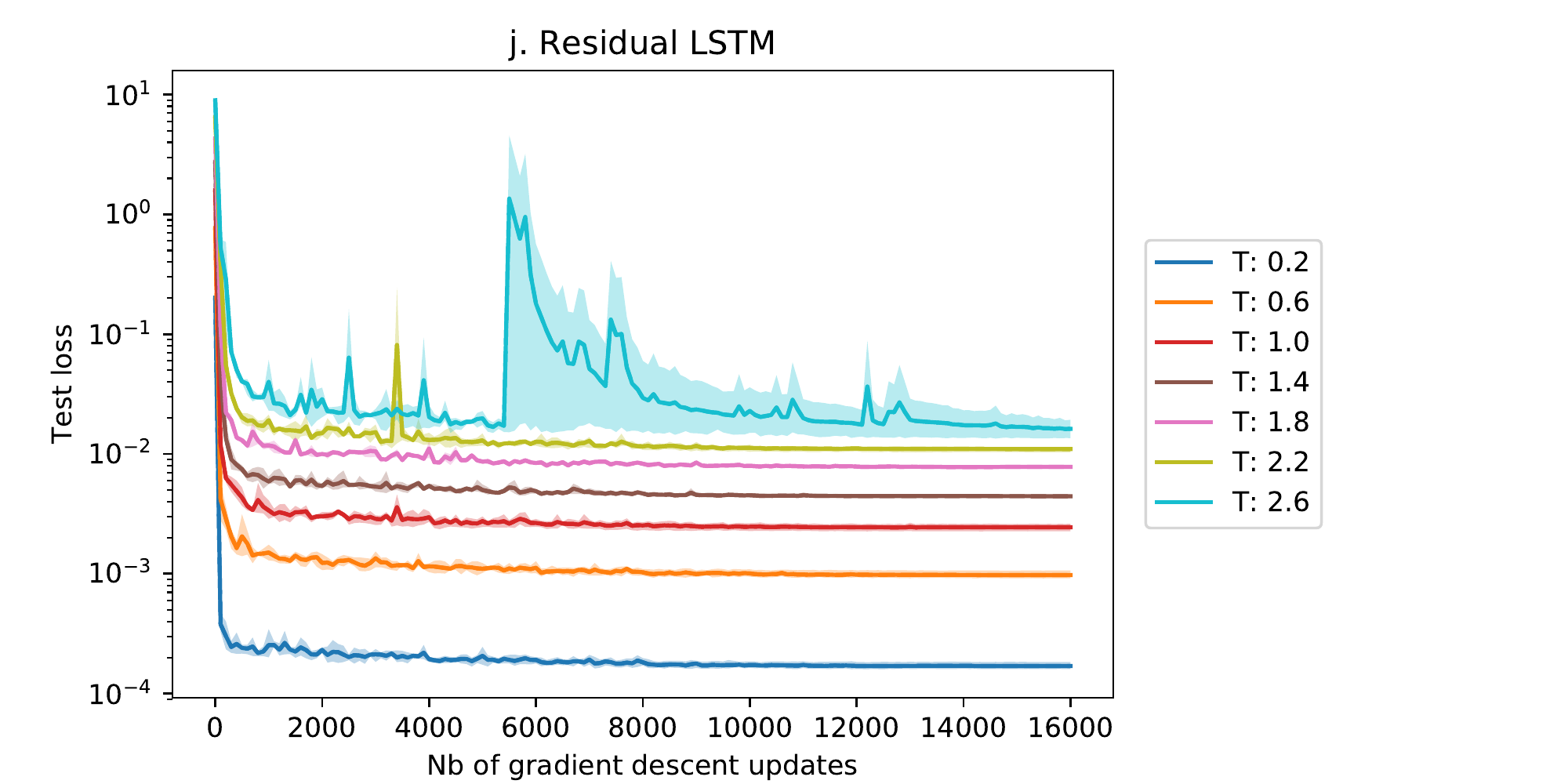}
\end{adjustwidth}
\caption{Influence of the \textbf{maturity} $T$ using equation \ref{pde:warin2} ($d=10$, $N=100$) on \textbf{1.} the relative error on $Y_0$ \eqref{eq:relerrY0}, \textbf{2.} the relative error on $Z_0$ \eqref{eq:relerrZ0}, \textbf{3.} the integral error on $Y$ \eqref{eq:interrY}, \textbf{4.} the integral error on $Z$ \eqref{eq:interrZ}, \textbf{5.} the final test loss. We represent convergence losses for networks \textbf{c.} and \textbf{j.} in \textbf{6.} and \textbf{7.}. The mean and the $5\%$ and $95\%$ quantiles are computed on $5$ independent runs and represented with the lines and error bars. Networks \textbf{a.} and \textbf{g.} do not converge on this example (their losses do not decrease during training and remain significantly higher than the other networks) and are not represented here. Above $T=2.5$ ($N=250$), only networks \textbf{e.} and \textbf{h.} are stable using the standard learning parameters.}
\label{fig:infT:warin2}
\end{figure}

\begin{figure}[ht]
\begin{adjustwidth}{-1.5cm}{-1.5cm}
\centering
\textbf{1.}\includegraphics[height=4cm, trim=0cm 0cm 5cm 0cm, clip]{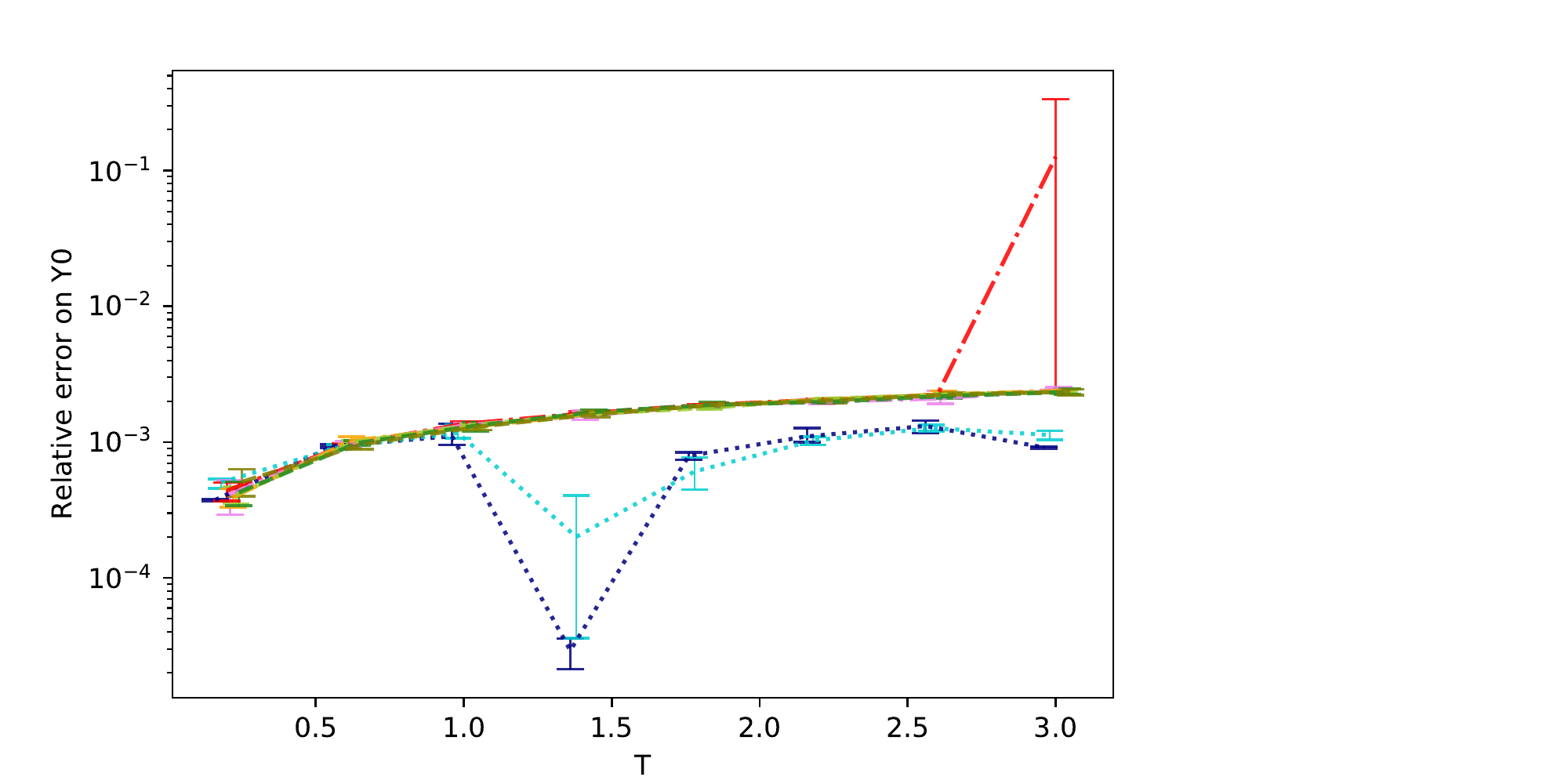}
\textbf{2.}\includegraphics[height=4cm]{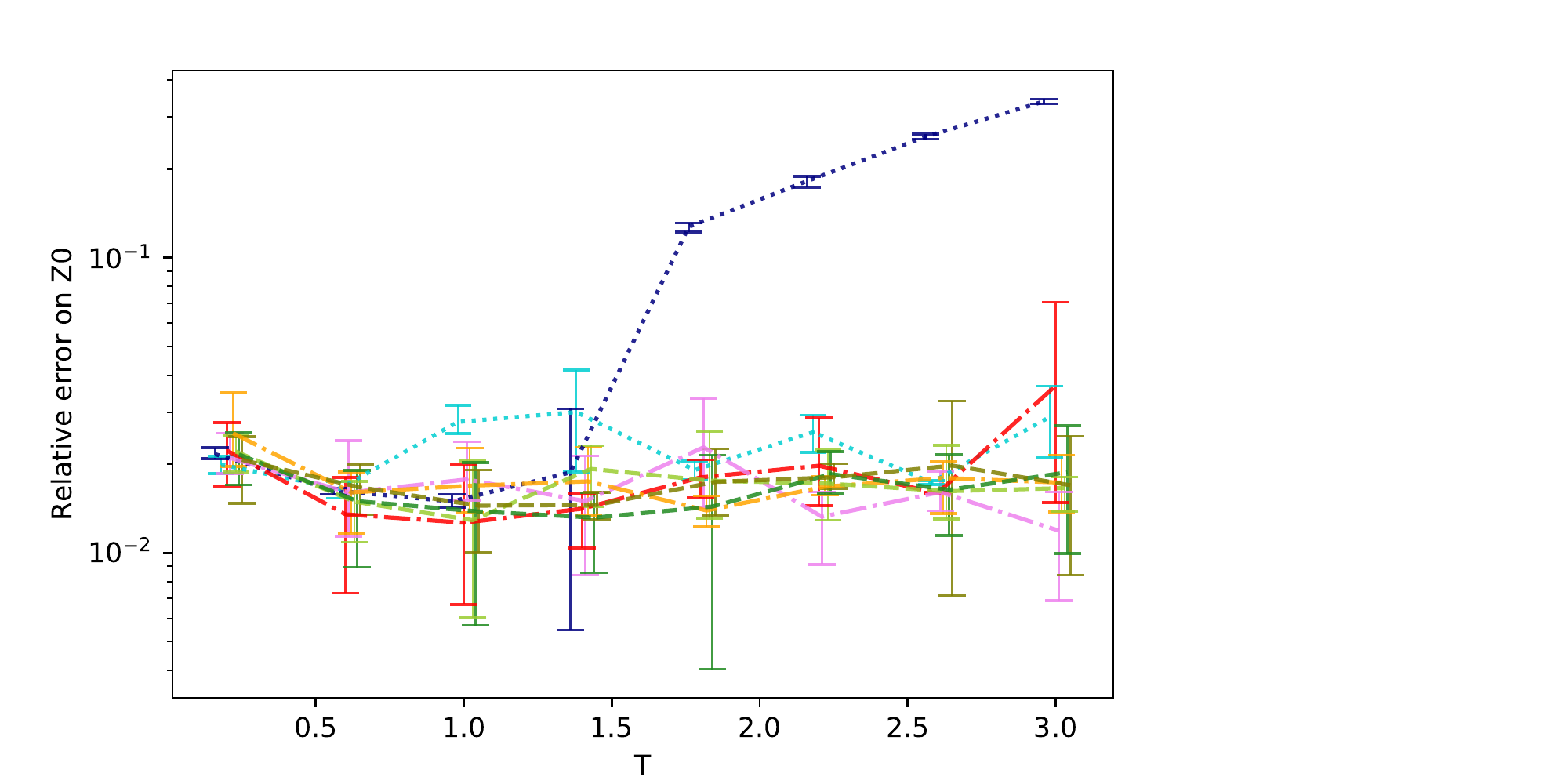} \\
\textbf{3.}\includegraphics[height=4cm, trim=0cm 0cm 5cm 0cm, clip]{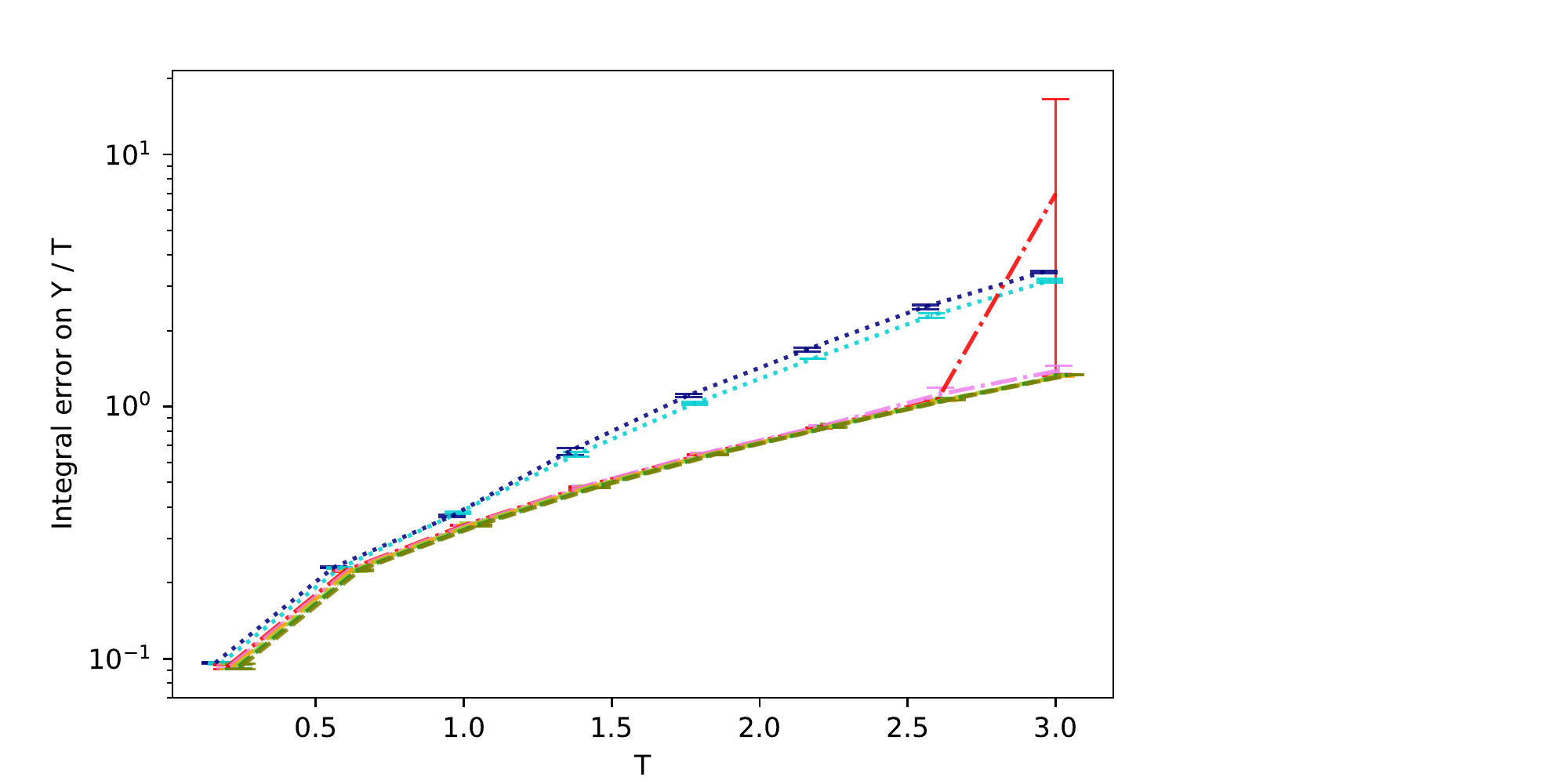}
\textbf{4.}\includegraphics[height=4cm]{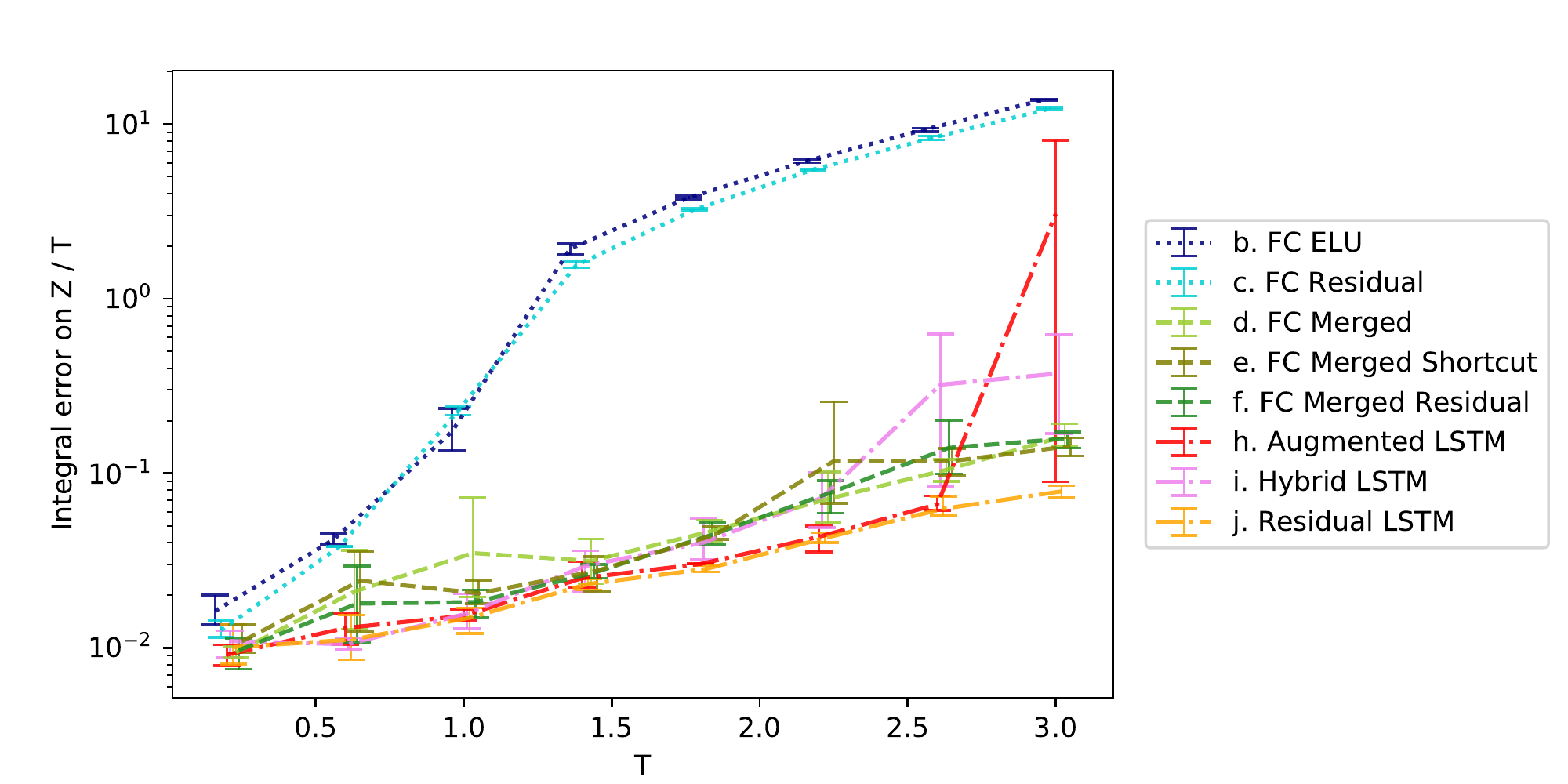} \\
\textbf{5.}\includegraphics[height=4cm]{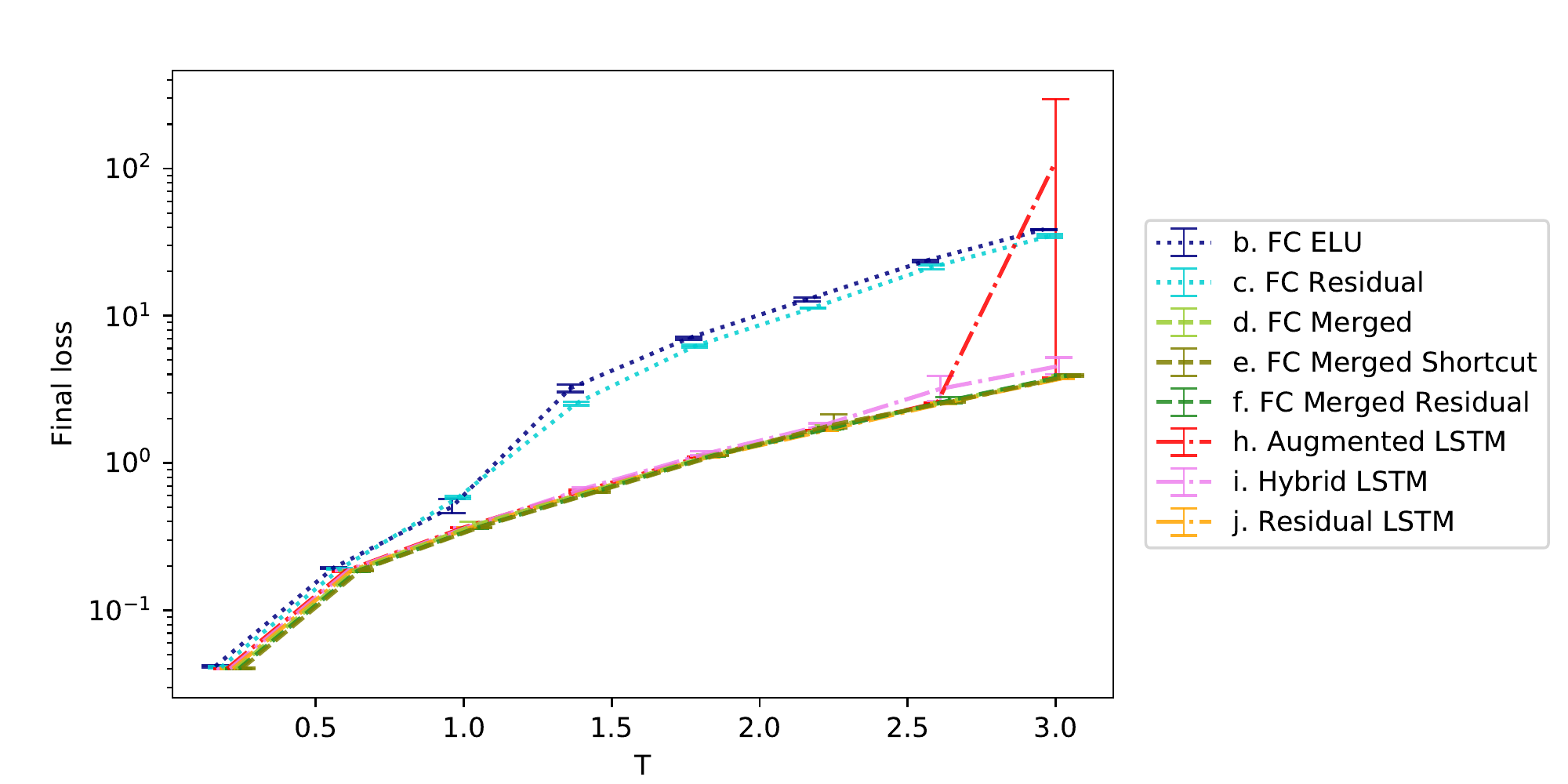} \\[0.5cm]
\textbf{6.}\includegraphics[height=4cm, trim=0cm 0cm 5cm 0cm, clip]{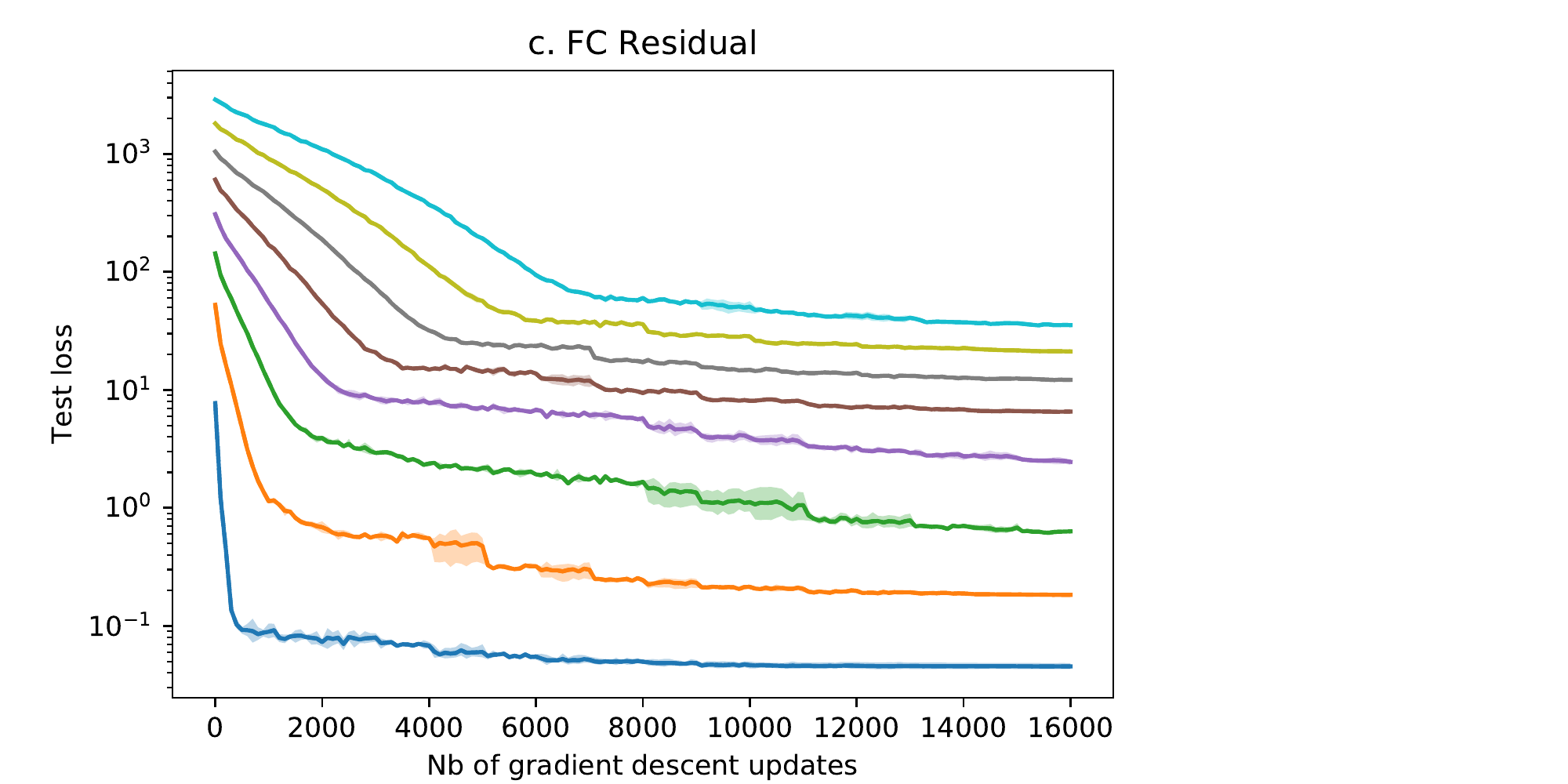}
\textbf{7.}\includegraphics[height=4cm]{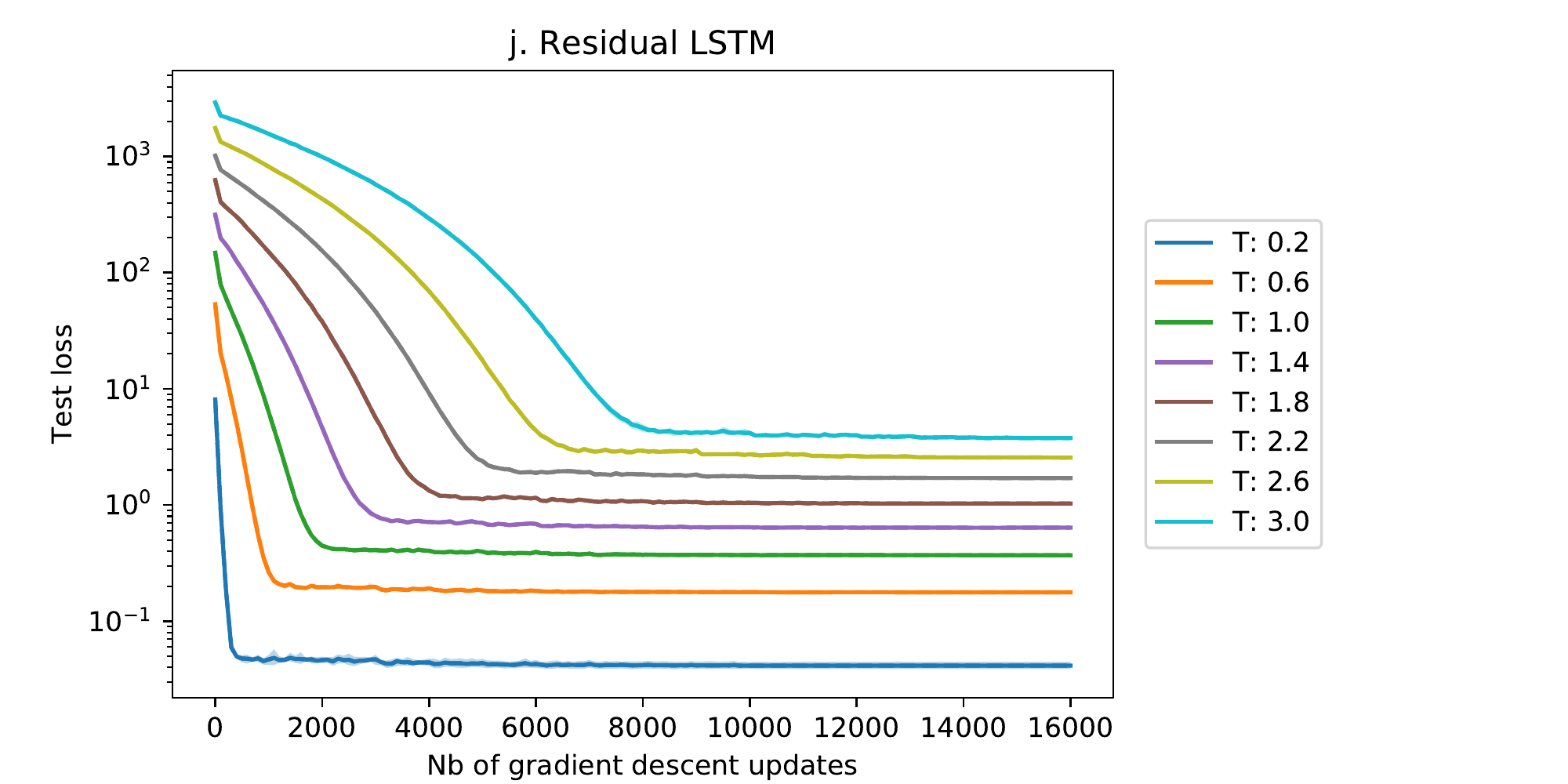}
\end{adjustwidth}
\caption{Influence of the \textbf{maturity} $T$ using equation \ref{pde:warin3} ($d=10$, $N=100$) on \textbf{1.} the relative error on $Y_0$ \eqref{eq:relerrY0}, \textbf{2.} the relative error on $Z_0$ \eqref{eq:relerrZ0}, \textbf{3.} the integral error on $Y$ \eqref{eq:interrY}, \textbf{4.} the integral error on $Z$ \eqref{eq:interrZ}, \textbf{5.} the final test loss. We represent convergence losses for networks \textbf{c.} and \textbf{j.} in \textbf{6.} and \textbf{7.}. The mean and the $5\%$ and $95\%$ quantiles are computed on $5$ independent runs and represented with the lines and error bars. Networks \textbf{a.} and \textbf{g.} do not converge on this example (their losses do not decrease during training and remain significantly higher than the other networks) and are not represented here. Above $T=2.5$ ($N=250$), the network \textbf{h.} shows instabilities.}
\label{fig:infT:warin3}
\end{figure}

Conclusions are similar for equations \ref{pde:warin2} and \ref{pde:warin3} ($d=10$): the errors seem to increase linearly with $T$, as shown in Figures \ref{fig:infT:warin2} and \ref{fig:infT:warin3}. We point out that networks \textbf{a.} and \textbf{g.} fail to converge and networks \textbf{b.} and \textbf{c.} perform slightly worse than the other networks. Finally, all the networks show some instabilities for $T \geq 2.5$ on equation \ref{pde:warin2} (with both constant $N$ and constant $\delta_t$) with our standard training procedure. This phenomenon is less visible on equation \ref{pde:warin3}, except for network \textbf{h.}. These instabilities could be solved by tuning the training hyperparameters further.

\clearpage
\subsubsection{Computation times}

We investigate the computation times for our different algorithms. In this Section the term  ``convergence'' is defined as the moment they reach a test loss $5\%$ close to the lowest test loss observed during training. In all the cases, we stopped the algorithms after $16000$ gradient descent updates. Note that the computation time and the memory usage heavily depend on the hardware used and on the implementation: the values presented below could be significantly reduced, especially for LSTM-based algorithms (\textbf{g.} to \textbf{j.}, see for instance \cite{braun2018lstm} for some comparison of existing LSTM implementations in different frameworks).  The following results are obtained using a NVIDIA Tesla K80 GPU (2014) and a Intel Xeon E5-2680 v4 CPU (2016).

\begin{figure}[ht]
\begin{adjustwidth}{-1.5cm}{-1.5cm}
\centering
\textbf{1.}\includegraphics[height=4.5cm, trim=0cm 0cm 5cm 0cm, clip]{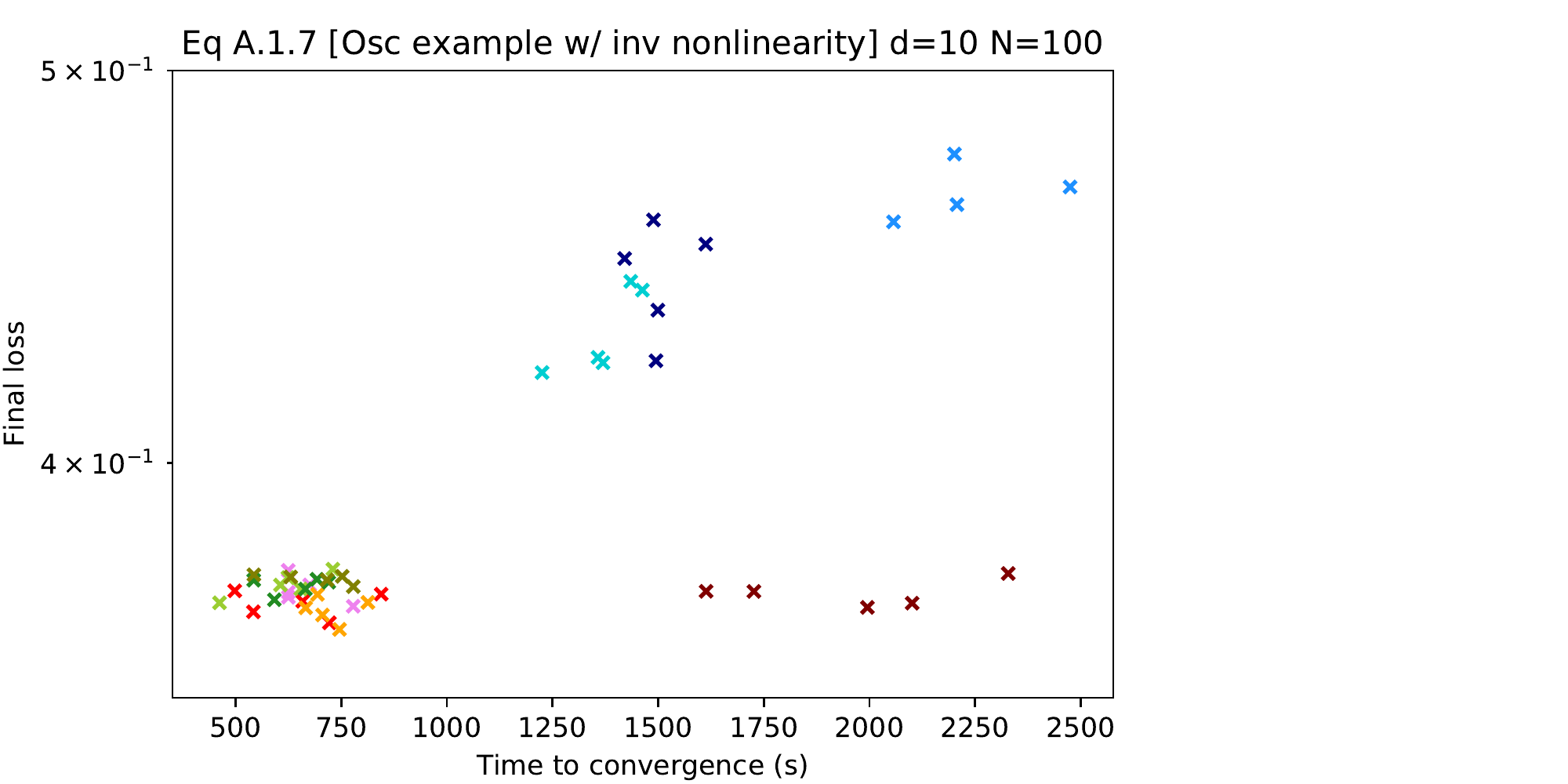}
\textbf{2.}\includegraphics[height=4.5cm]{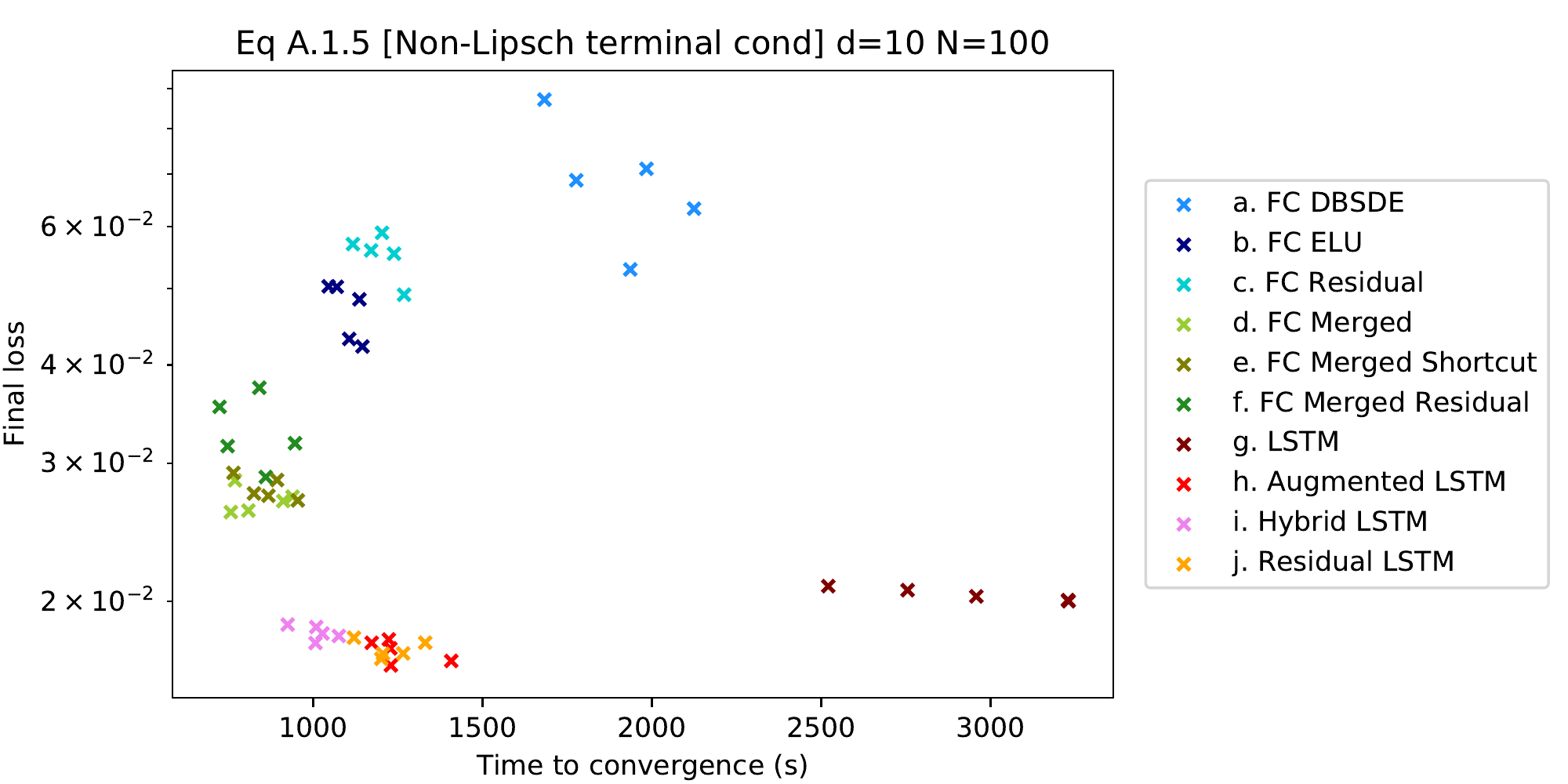} \\
\textbf{3.}\includegraphics[height=4.5cm, trim=0cm 0cm 5cm 0cm, clip]{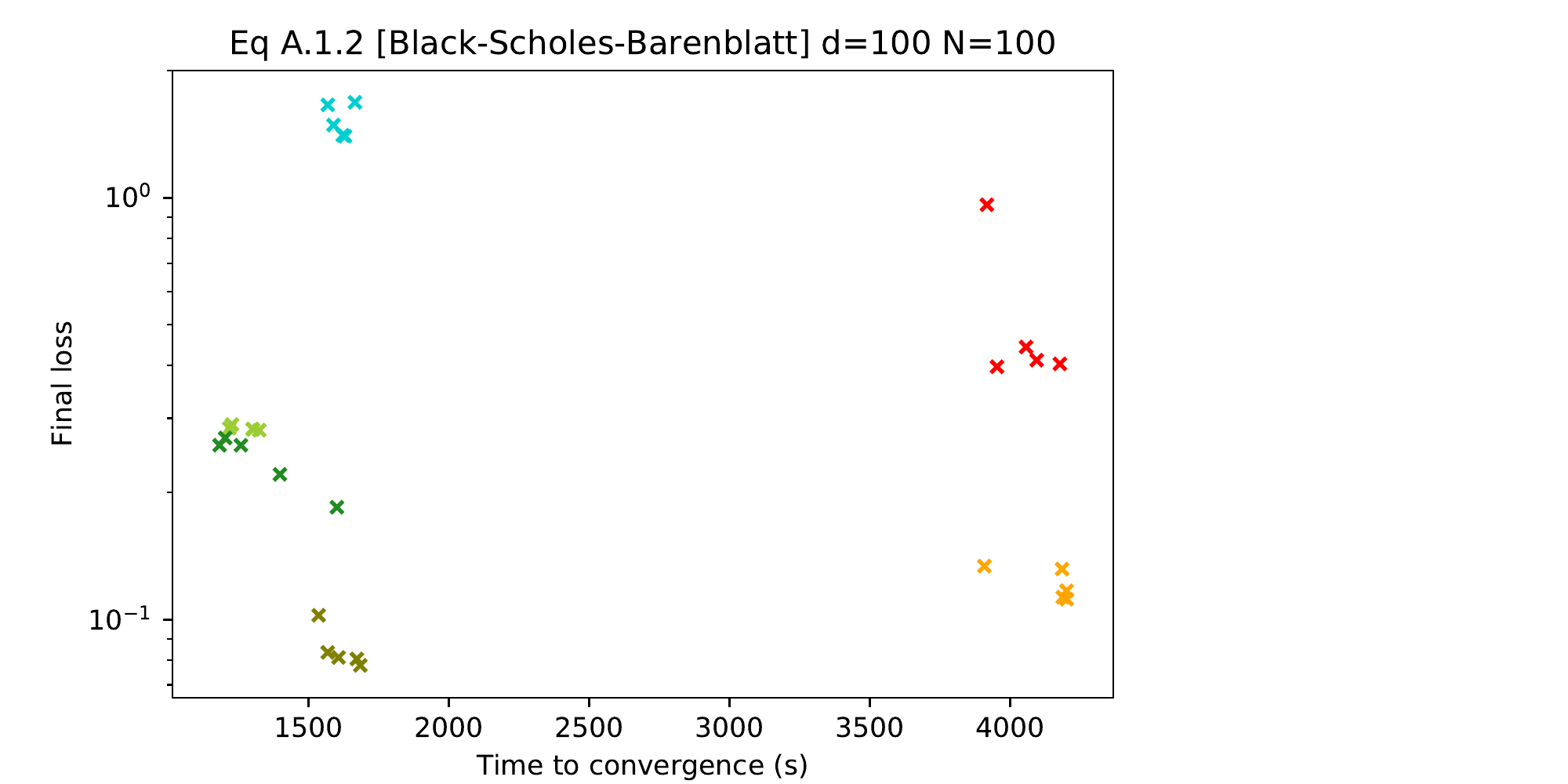}
\textbf{4.}\includegraphics[height=4.5cm]{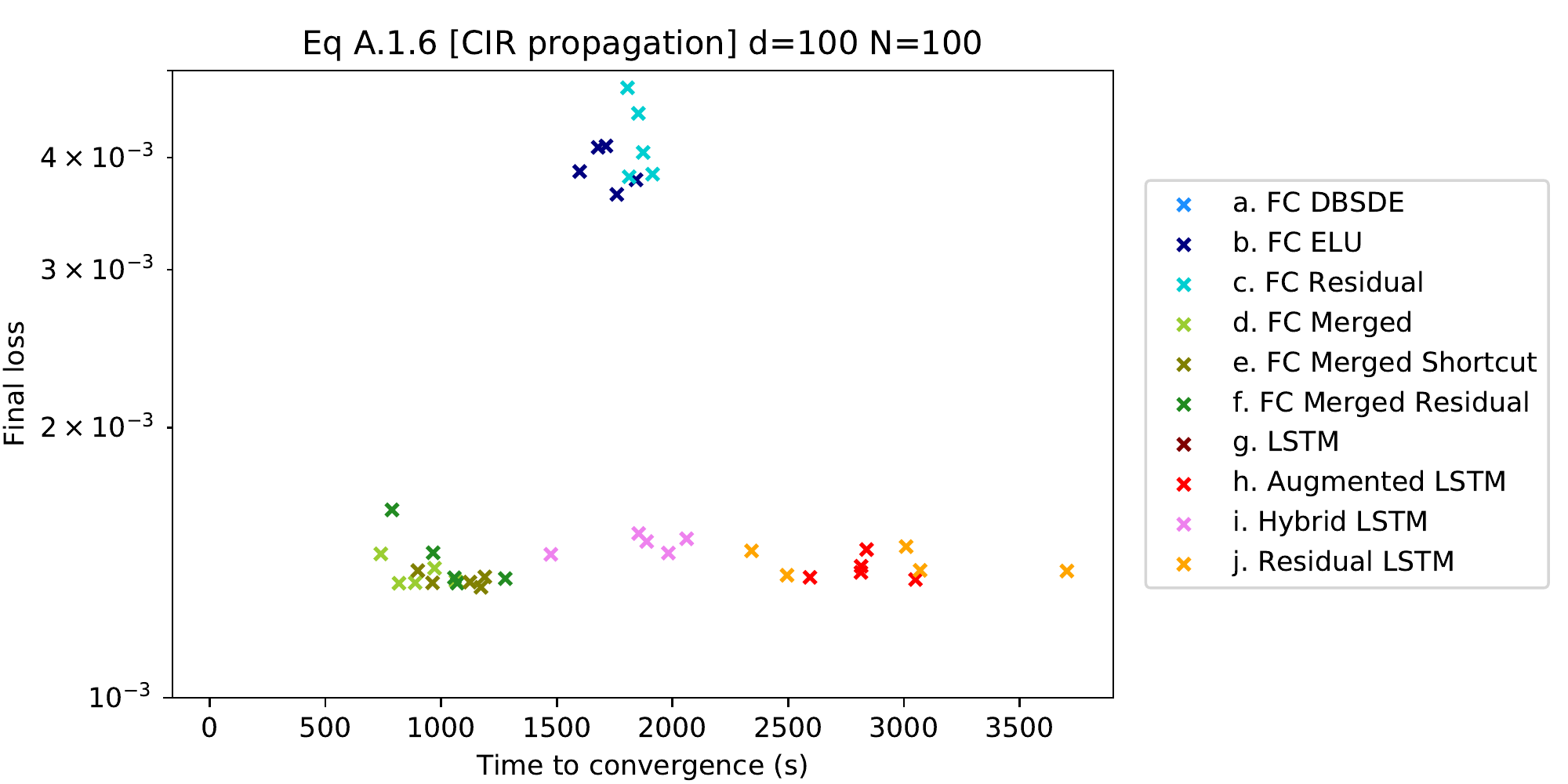}
\end{adjustwidth}
\caption{Final loss function of computation times until convergence for the different equations (with the standard parameters, $r=0.1$, $T=1.0$). We repeat the experiments $5$ times, each point represents a single run. \textbf{1.}, \textbf{2.}: the Merged and LSTM algorithms reach lower final losses in lower computation times than the other algorithms. \textbf{3.}, \textbf{4.}: the lowest losses are reached by the Merged and LSTM algorithms, with an advantage in computation times for the Merged algorithms. Note that the computation times for all algorithms remain $\simeq 1$ hour.}
\label{fig:comptimes}
\end{figure}

\begin{table}[ht]
    \centering
\begin{tabular}{|l|r|r|r|r|r|r|r|r|r|r|}
    \hline
    \textbf{1.} Time for $100$ it (s) & \textbf{a.} & \textbf{b.} & \textbf{c.} & \textbf{d.} & \textbf{e.} & \textbf{f.} & \textbf{g.} & \textbf{h.} & \textbf{i.} & \textbf{j.} \\ \hline
    \ref{pde:warin3} ($d=10$) & $16.57$ & $10.56$ & $10.35$ & $10.35$ & $10.94$ & $10.52$ & $28.19$ & $17.12$ & $15.21$ & $17.18$ \\ \hline
    \ref{pde:richou} ($d=10$) & $13.97$ & $7.25$ & $8.03$ & $6.81$ & $8.11$ & $7.6$ & $29.14$ & $12.91$ & $11.09$ & $12.99$ \\ \hline
    \ref{pde:bsbarenblatt} ($d=100$) &  &  & $10.67$ & $10.67$ & $10.87$ & $10.67$ &  & $27.66$ &  & $27.73$ \\ \hline
    \ref{pde:cir} ($d=100$) &  & $11.65$ & $12.27$ & $11.5$ & $12.62$ & $11.72$ &  & $30.19$ & $18.18$ & $30.39$ \\ \hline
    \hline
    \textbf{2.} Nb iterations to cv. & \textbf{a.} & \textbf{b.} & \textbf{c.} & \textbf{d.} & \textbf{e.} & \textbf{f.} & \textbf{g.} & \textbf{h.} & \textbf{i.} & \textbf{j.} \\ \hline
    \ref{pde:warin3} ($d=10$) & $13300$ & $14200$ & $13300$ & $6000$ & $6500$ & $6300$ & $6600$ & $3800$ & $4100$ & $4100$ \\ \hline
    \ref{pde:richou} ($d=10$) & $13800$ & $15300$ & $15000$ & $11900$ & $10700$ & $11300$ & $10100$ & $9800$ & $9100$ & $9300$ \\ \hline
    \ref{pde:bsbarenblatt} ($d=100$) &  &  & $15200$ & $12200$ & $14800$ & $11800$ &  & $14600$ &  & $15100$ \\ \hline
    \ref{pde:cir} ($d=100$) &  & $14700$ & $15000$ & $7700$ & $8900$ & $9000$ &  & $9300$ & $10400$ & $9900$ \\ \hline
\end{tabular}
    \caption{\textbf{1.} Computation times for $100$ iterations (gradient descent updates) for the different algorithms on the different equations (with the standard parameters, $r=0.1$, $T=1.0$). We computed the mean of the times achieved for $100$ gradient descent iterations for each one of our $5$ independent runs, then took the median of the $5$ values. \textbf{2.} Number of iterations (gradient descent updates) until convergence for the different algorithms on the different equations (default parameters). We took the median of the $5$ values. The cell is left black if the algorithm did not converge (had a final loss significantly higher than other algorithms).}
    \label{tab:nit_to_cv}
\end{table}

\clearpage
\subsection{Numerical results: our new algorithm} \label{subs:numres:fixedpoint}

In the following, we compare the new algorithms presented in Section \ref{sec:new_algorithms}. Then, based on our results, we compare the second version of our algorithm noted \textbf{C.} in Section \ref{sec:secondScheme} with the best neural networks developed in Section \ref{sec:other_architectures}. Integrals on $Y$ \eqref{eq:interrY}, and $Z$ \eqref{eq:interrZ} are computed with $10$ trajectories, and using a number of time steps suitable for comparison with \emph{Deep BSDE} algorithms.

We use a number of hidden layers of $h=3$ and a hidden layer size of $w=2d$ in our new algorithms. We found using higher values of $h$ and $w$ has little impact on the precision of the results and increase further computation times.

We recall the algorithm consists in approximating $u$ and $v$ by a neural network
\[
(u(\theta, t,x)  ,v(\theta,t,x) )  := \mathcal{N}^\theta(t, x)
\]
and solving a fixed point problem depending on parameters $n_\text{inner}$ (number of particles during training) and $\lambda$ (parameter of the distribution of the particles). In order to evaluate $u$ and $Du$, instead of evaluating directly the neural network $(u, v)$, we propose a post-processing step consisting in rather evaluating $(\bar{u}, \bar{v})$ using the estimator \eqref{eq:NewAlgoDis} for the first scheme, or \eqref{eq:secondScheme}, \eqref{eq:SecondSchemevbar} for the second scheme, with $n_\text{eval} \gg n_\text{inner}$ new particles. This supplementary step will be justified in Section \ref{sec:influence_n_eval}.

We investigate the influence of the parameters $n_\text{inner}$. It follows that increasing $n_\text{inner}$ over several thousands and changing $\lambda$ do not have a significant effect. Thus, as previously stated, we choose $n_\text{inner}=10000$ and $\lambda=0.5$ (if not precised otherwise) or $\lambda = 1.0$ in the following.

\subsubsection[Influence of n eval]{Influence of $n_\text{eval}$ and the post-processing step} \label{sec:influence_n_eval}

\begin{figure}[ht]
\centering
\textbf{1.}\includegraphics[height=6cm, trim=0cm 0cm 0cm 0cm, clip]{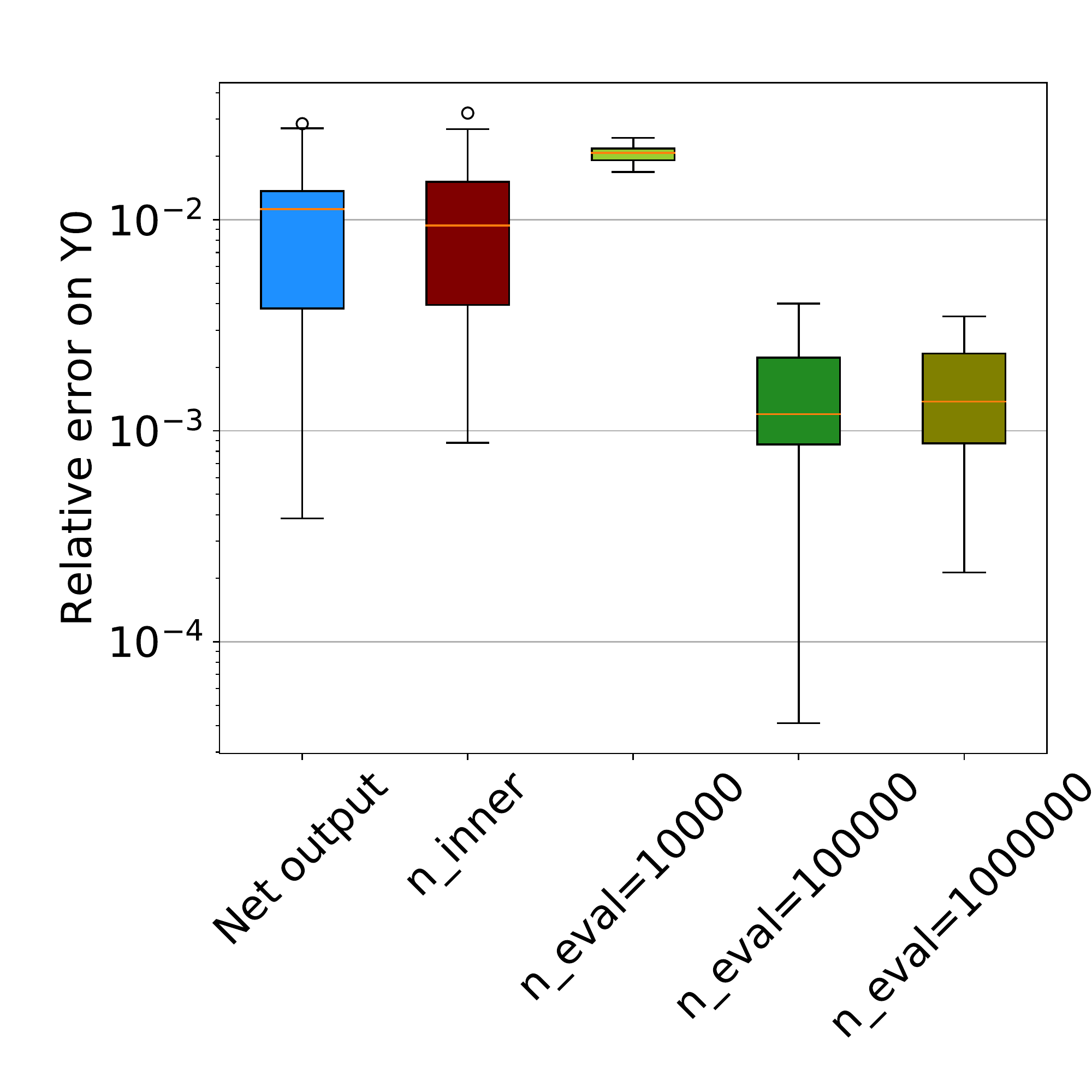}
\textbf{2.}\includegraphics[height=6cm, trim=0cm 0 0cm 0, clip]{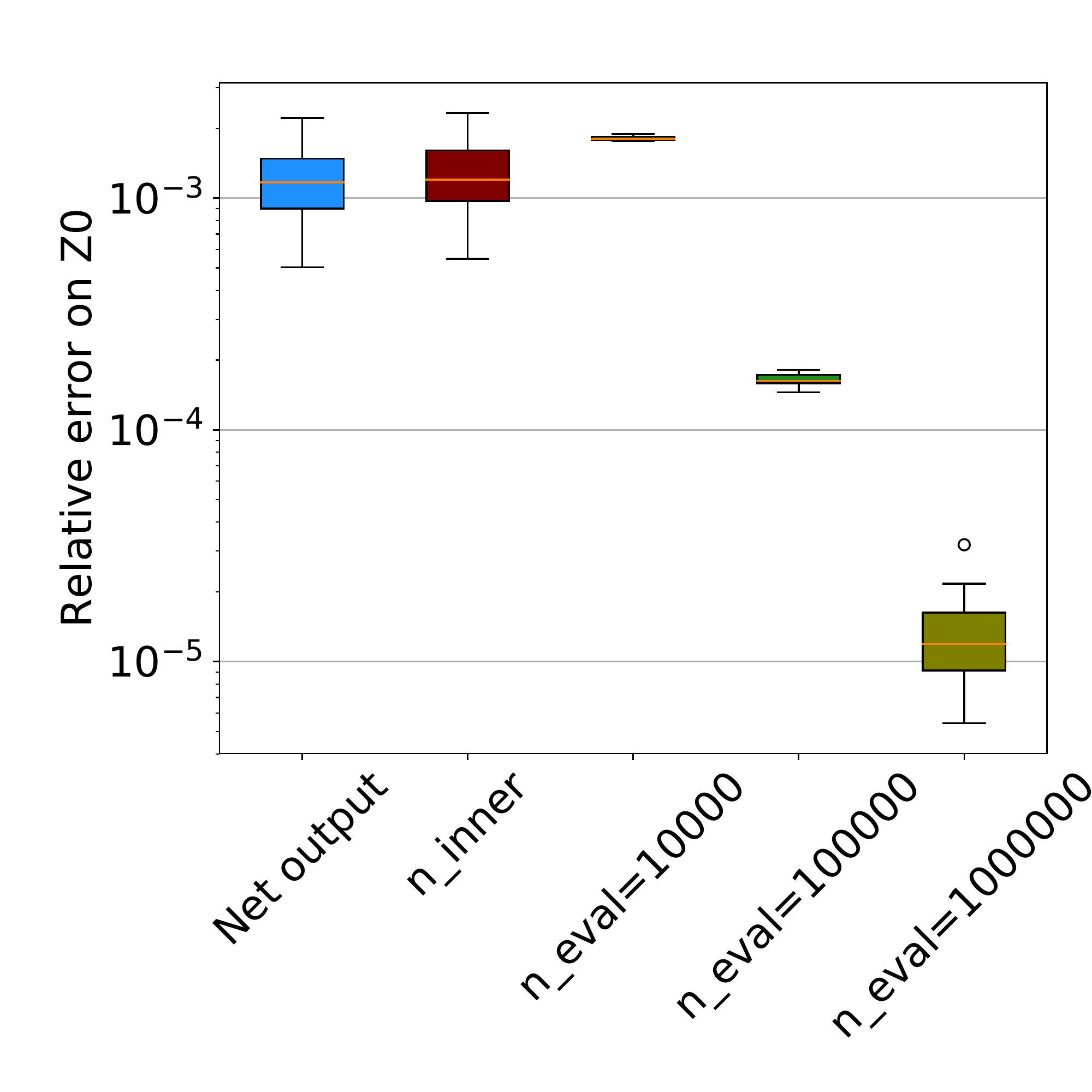} \\
\textbf{3.}\includegraphics[height=6cm, trim=0cm 0cm 0cm 0cm, clip]{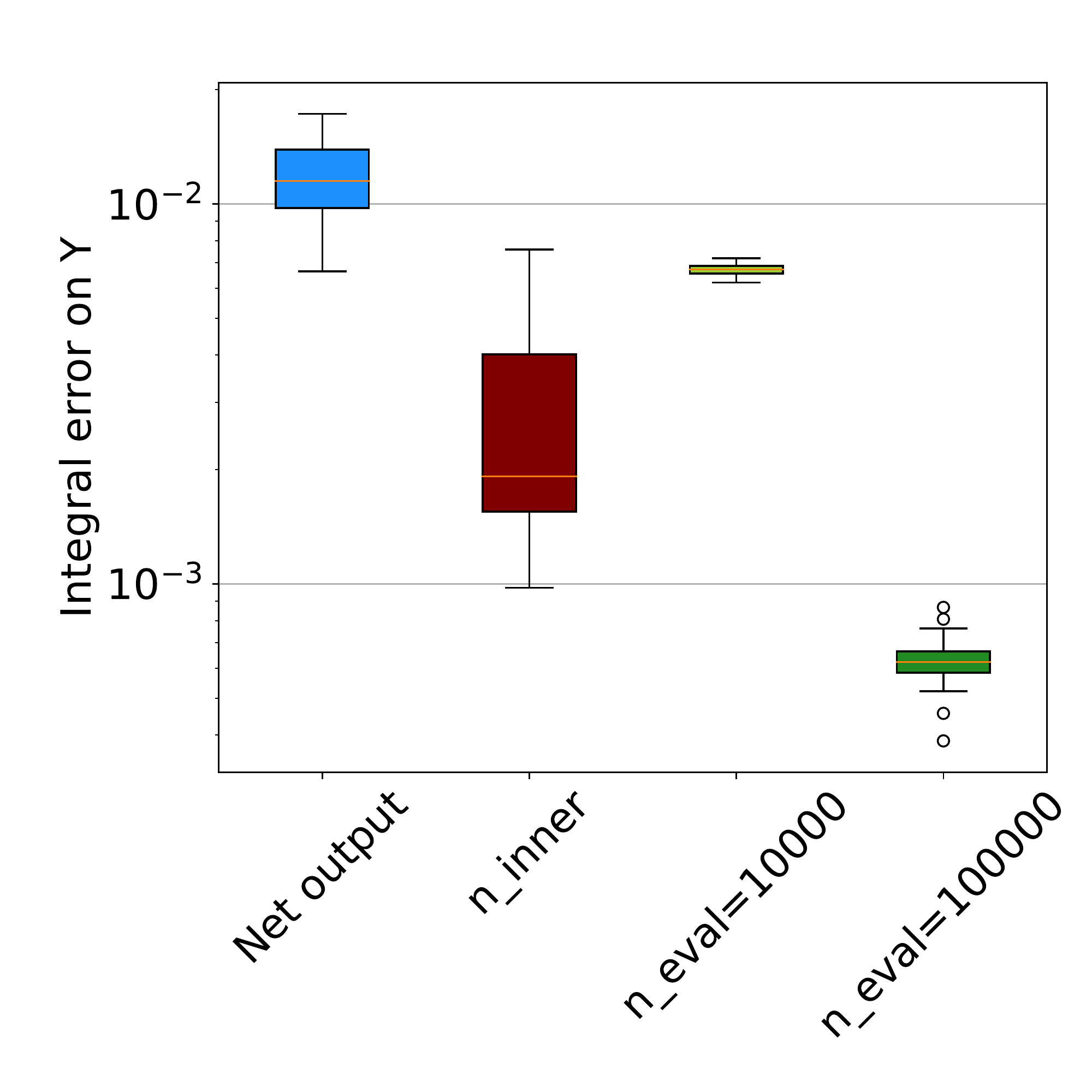}
\textbf{4.}\includegraphics[height=6cm, trim=0cm 0 0cm 0, clip]{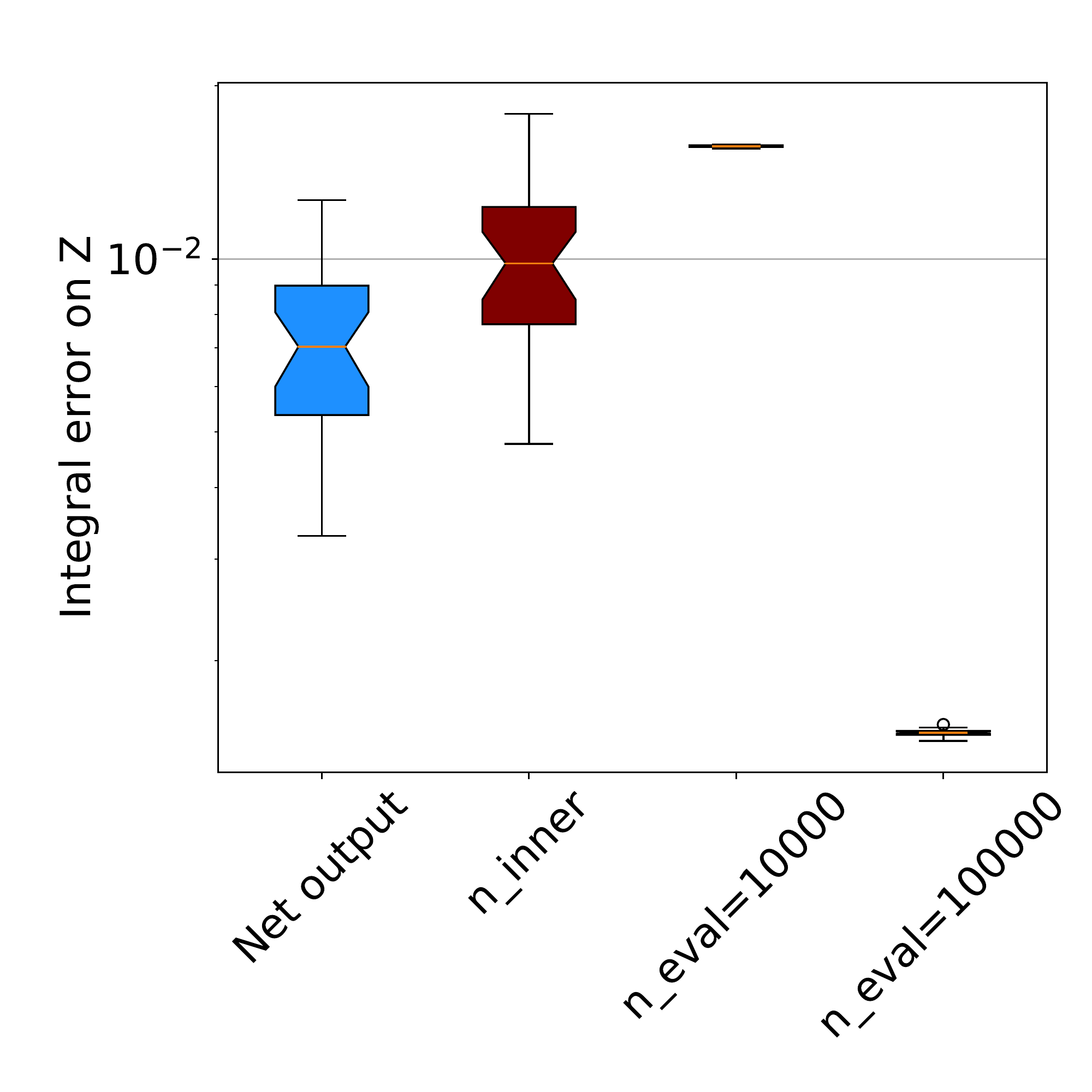}
\caption{Influence of the post-processing step on the results of algorithm \textbf{C.} on equation \ref{pde:warin2} ($d=10$, $r=0.1$). We used $n_\text{inner} = 10000$ during training. We represent the results obtained by outputting directly the network $(u, v)$ (\emph{Net output}), by outputting $(\bar{u}, \bar{v})$ using the $n_\text{inner}$ samples used during training, and by performing a post processing using different values of $n_\text{eval}$. We represent \textbf{1.} the relative error on $Y_0$ \eqref{eq:relerrY0}, \textbf{2.} the relative error on $Z_0$ \eqref{eq:relerrZ0}, \textbf{3.} the integral error on $Y$ \eqref{eq:interrY}, \textbf{4.} the integral error on $Z$ \eqref{eq:interrZ}. Integral errors were computed using $10$ trajectories. We repeat the experiment $30$ times, and we represent the median (line), the quartiles (box) and the quartiles $\pm 1.5 \times$ IQR (whiskers).
}
\label{fig:influence_n_eval}
\end{figure}

We investigate the influence of the post-processing step of our algorithm on the precision of the results. Results for the scheme \textbf{C.} are presented in Figure \ref{fig:influence_n_eval}. It follows that:
\begin{itemize}
    \item Indeed, results using directly the outputs of our neural network $(u, v)$ are similar to those obtained using the estimators
    \eqref{eq:secondScheme}, \eqref{eq:SecondSchemevbar} using the $n_\text{inner}$ samples from training, as the neural network was trained to reproduce these estimators with these $n_\text{inner}$ samples in particular.
    \item Using a post-processing step with the estimators
    \eqref{eq:secondScheme}, \eqref{eq:SecondSchemevbar} and choosing $n_\text{eval} \gg n_\text{inner}$ improve the accuracy of the solution. Large $n_\text{eval}$ increases further the accuracy.
\end{itemize}
It is feasible to use a very large $n_\text{eval}$ for pointwise evaluation (to get $Y_0$ and $Z_0$ for instance) but it is costly to use high $n_\text{eval}$ to get full trajectories. In the following, we used $n_\text{eval} = 1 000 000$ to compute the error on $Y_0$ and $Z_0$, and $n_\text{eval} = 100 000$ on $10$ trajectories to compute the integral errors.

\subsubsection{Comparison of the new architectures} \label{sec:comp_new_arch}

\begin{figure}[ht]
\begin{adjustwidth}{-1.5cm}{-1.5cm}
\centering
\textbf{1.}\includegraphics[height=4.5cm, trim=0cm 0cm 5.5cm 0cm, clip]{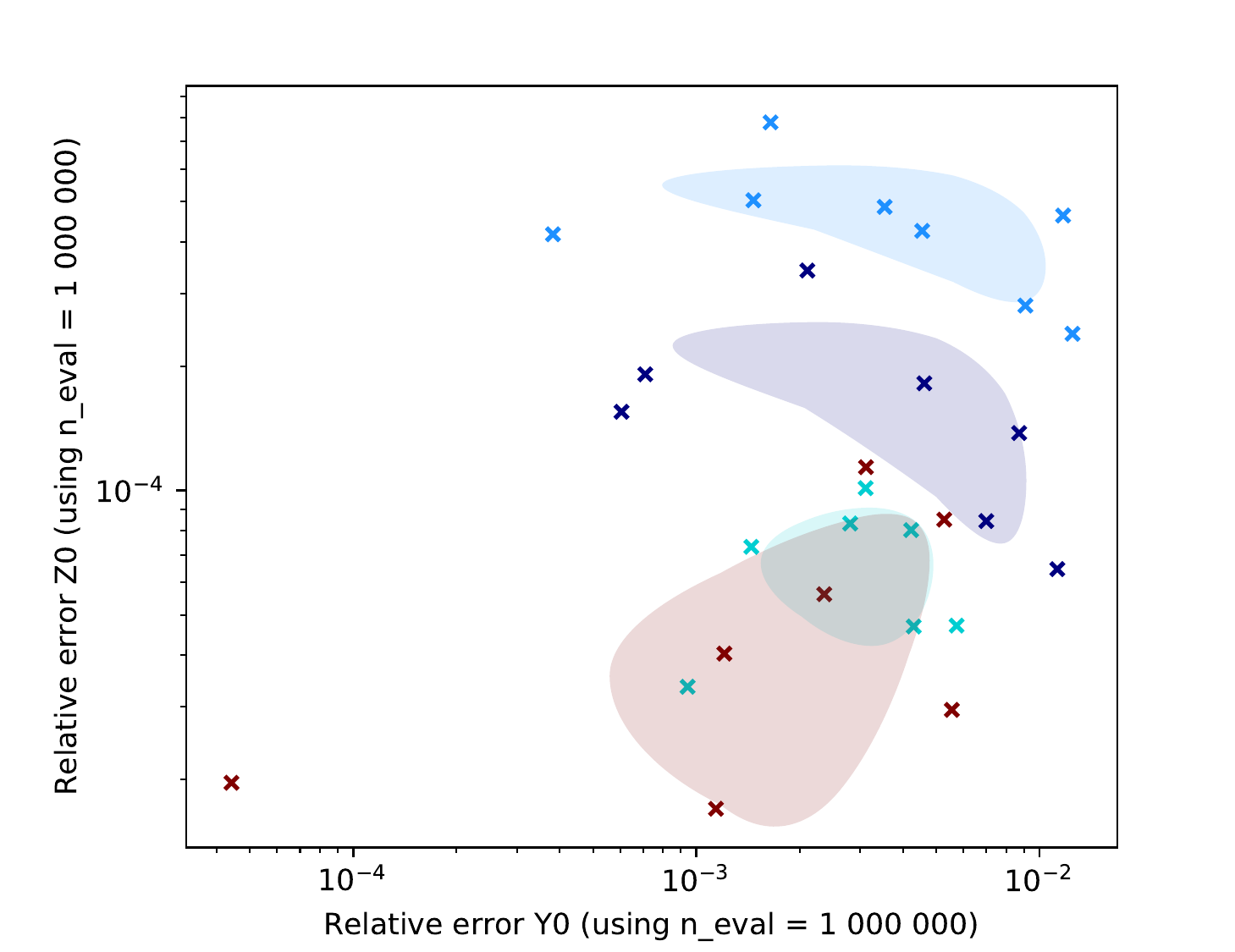}
\textbf{2.}\includegraphics[height=4.5cm]{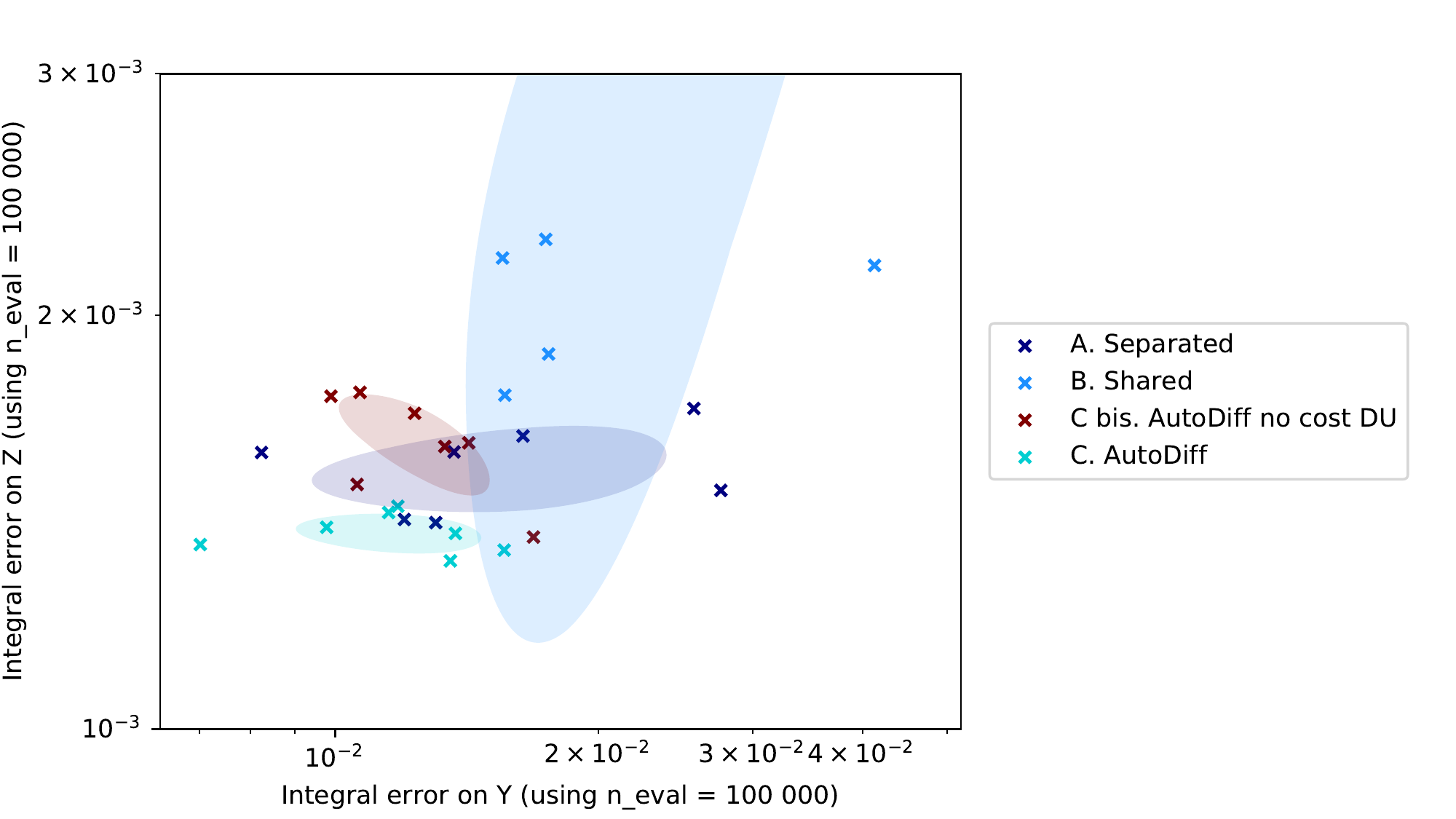} \\
\textbf{3.}\includegraphics[height=4.5cm, trim=0cm 0cm 5.5cm 0cm, clip]{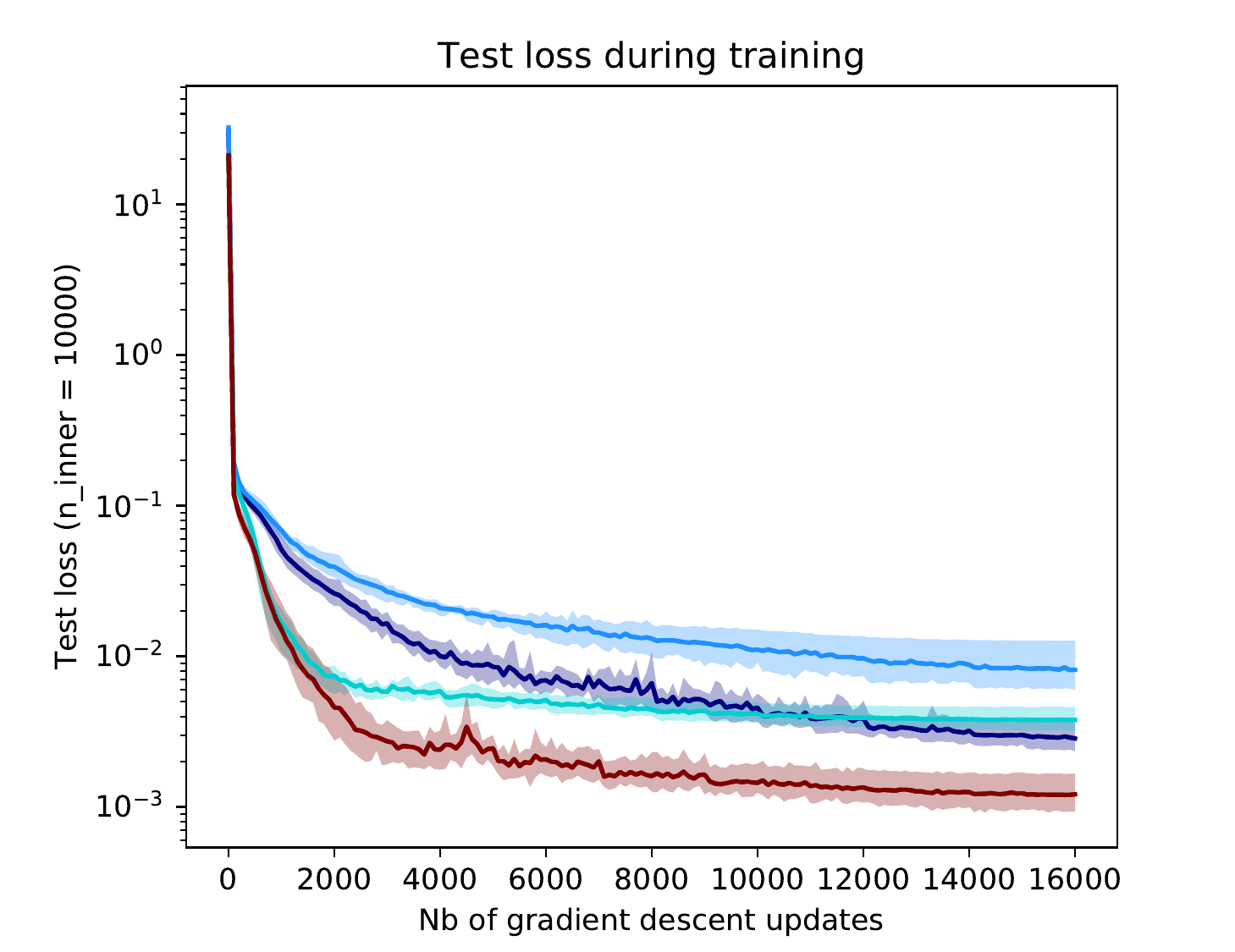}
\textbf{4.}\includegraphics[height=4.5cm]{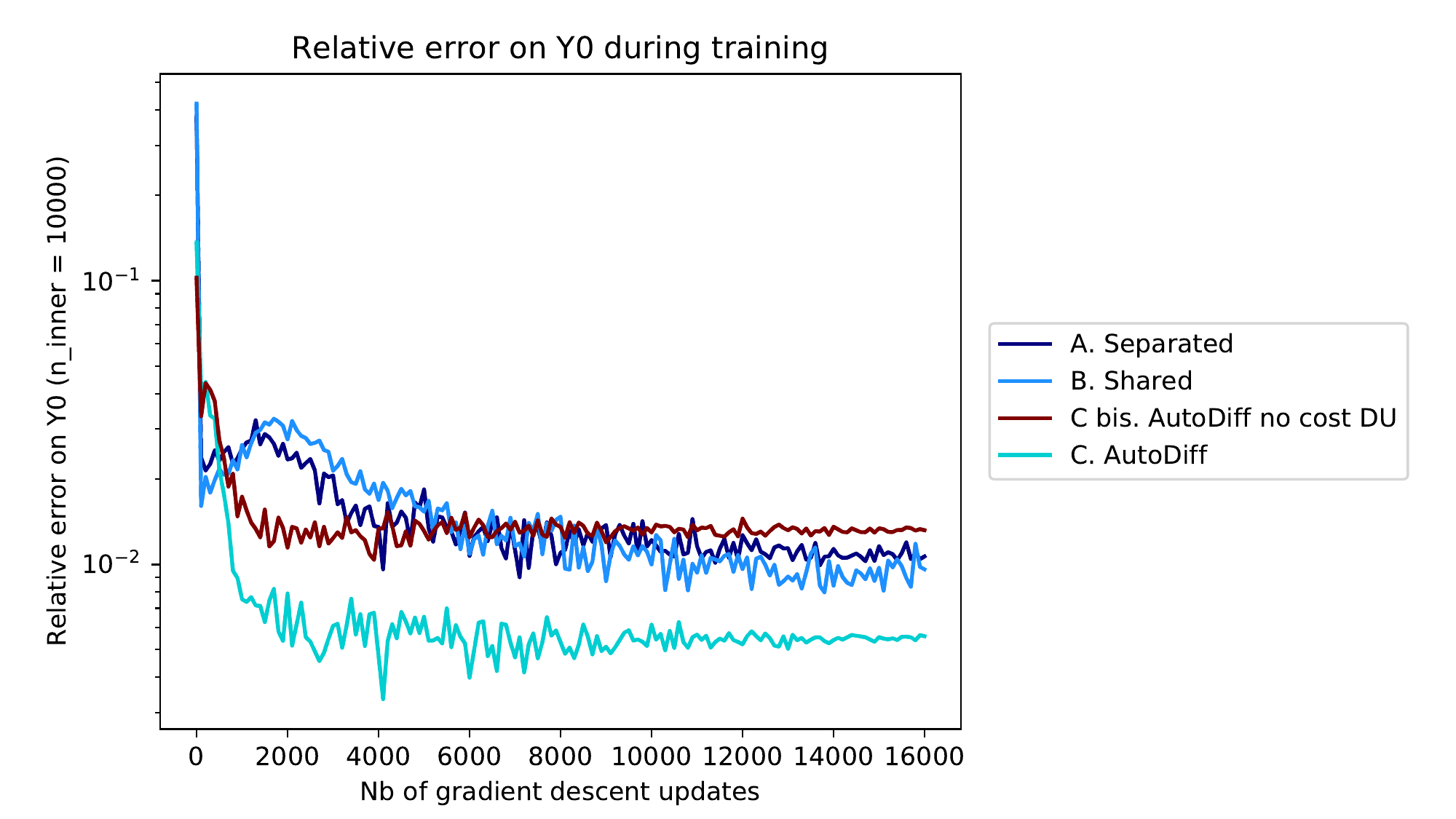} \\[0.5cm]
\begin{tabular}{|l|r|r|r|r|}
    \hline
    \ref{pde:richou} ($d=10$) [using GPU] & \textbf{A.} & \textbf{B.} & \textbf{Cbis.} & \textbf{C.} \\
    \hline
    \textbf{5.} Time for $100$ it (s) & $125.47$ & $79.03$ & $111.12$ & $119.45$ \\
    \hline 
    \textbf{6.} Time to convergence (s) & $19690$ & $11692$ & $15664$ & $15047$ \\
    \hline
\end{tabular}
\end{adjustwidth}
\caption{Comparison of algorithms \textbf{A.}, \textbf{B.}, \textbf{C.} and \textbf{C bis.} with no cost on $Du$ on equation \ref{pde:richou} ($d=10$), $n_\text{inner}=10 000$ during training. We ran each algorithm $8$ times. \textbf{1.}: relative errors on $Y_0$ \eqref{eq:relerrY0} and $Z_0$ \eqref{eq:relerrZ0} using $n_\text{eval} = 1 000 000$. \textbf{2.}: integral error on $Y$ \eqref{eq:interrY} and $Z$ \eqref{eq:interrZ} using $n_\text{eval} = 100 000$. The ellipsis $\pm$ $1 \times$ std are represented. \textbf{3.}: test loss using the $n_\text{inner} = 10000$ samples used during training, function of the number of gradient descent updates (note that \textbf{C bis.} is not comparable to the other algorithms as its definition of the ``loss'' does not include $Du$). \textbf{4.}: relative error on $Y_0$ \eqref{eq:relerrY0} using the $n_\text{inner} = 10000$ samples used during training, function of the number of gradient descent updates (quantiles are not represented for readability). \textbf{5.} and \textbf{6.}: mean computation times for $100$ gradient descent updates and time to convergence (until the test loss is $5\%$ close to the lowest loss observed during training) -- we computed the mean for each run, then took the median of the $8$ values obtained.}
\label{fig:infnetstr:richou}
\end{figure}

We compare the algorithms from our first scheme \textbf{A.}, \textbf{B.} (using a loss with a cost on $Du$ \eqref{eq:LossFirstScheme}), and from our second scheme \textbf{C.} (using a loss with a cost on $Du$ \eqref{eq:LossSecondScheme2}) and \textbf{C bis.} (\textbf{C.} using a loss with no cost on $Du$ \eqref{eq:LossSecondScheme1}). The number of parameters of each algorithm is shown in Table \ref{tab:nparam:newalgo}. Results on equation \ref{pde:richou} ($d=10$) are presented in Figure \ref{fig:infnetstr:richou}: it appears networks \textbf{A.}, \textbf{C bis.} and \textbf{C.} perform better overall and including $Du$ in the loss results in lower errors on $Y_0$ and $Z$. Computation times are comparable, with an advantage for \textbf{B.}. Setting no cost on $Du$ in \textbf{C bis.} has little influence on the results overall, but seems to slightly degrade the precision on the trajectories, especially $Z$. Thus, we chose \textbf{C.} in the following comparisons.

\begin{table}[ht]
\centering
\begin{tabular}{|p{2.5cm}|p{8.7cm}|p{1.2cm}|p{1.2cm}|}
\hline
Algorithm & Number of parameters & $d=10$ & $d=100$\\
\hline
\textbf{A.} & $2 \left[(2d)(d+1) + 2d + 2\times ((2d)^2 + 2d)\right] + 2d + 1 + 2d^2 + d$ & 2 391 & 221 901\\
\hline
\textbf{B.} & $(2d)(d+1) + 2d + 2\times ((2d)^2 + 2d) + (2d)(d+1) + d+1$ & 1 311 & 121 101\\
\hline
\textbf{C.} and \textbf{C bis.} & $(2d)(d+1) + 2d + 2\times ((2d)^2 + 2d) + 2d + 1$ & 1 101 & 101 001\\
\hline
\end{tabular}
\caption{Number of parameters for our new algorithms.} \label{tab:nparam:newalgo}
\end{table}

The memory needed by the algorithm prevents us to use it on GPU for high $n_\text{inner}$ values in high dimension and only CPU with large memory are sometimes use: it has a direct impact on the computational time as shown on Table \ref{tab:compTime2}.

Note that the computation time and the memory usage heavily depend on the hardware use and on the implementation: it is possible that the values presented below can be significantly reduced. The orders of magnitude presented show that the algorithms remain indeed tractable in high dimension. The results presented are obtained using a NVIDIA Tesla K80 GPU (2014) and a Intel Xeon E5-2680 v4 CPU (2016).

\begin{table}[h!]
    \centering
    \begin{tabular}{|l|l|c|r|r|} \hline
         Equation & Network &  Architecture &  Time for $16000$ it (s) & Time to convergence (s) \\ \hline
         & \textbf{f.} & GPU &  1 623 & 448\\
         \textbf{1.} \ref{pde:warin2} ($d=10$) & \textbf{j.} & GPU  & 2 641 & 695\\
         & \textbf{C.} & GPU & 21 260 & 9433 \\ \hline
        &\textbf{f.} & GPU &  1 706 & 1 259\\
        \textbf{2.} \ref{pde:bsbarenblatt} ($d=100$) &\textbf{j.} & GPU  & 4 436 & 4 187\\
        &\textbf{A.} & CPU & $\simeq$ 124 500 & 45 789\\
        &\textbf{B.} & CPU & $\simeq$ 78 883 & 41 820\\
        &\textbf{C bis.} & CPU & $\simeq$ 106 670 & 60 572\\ 
         &\textbf{C.} & CPU & $\simeq$ 116 000 & 51 905\\ \hline
     \end{tabular}
    \caption{\textbf{1.}: Computation times for test case \ref{pde:warin2} in dimension $d=10$. \textbf{C.} uses $n_\text{inner}=10000$. We give the time necessary for $16000$ iterations and the observed time to convergence (until the test loss first becomes $5\%$ close to the lowest loss observed during training). We took the median over $5$ independent runs for each value.
    \textbf{2.}: Computation time for test case \ref{pde:bsbarenblatt} in dimension $d=100$. \textbf{C.} uses $n_\text{inner}=4000$. We took the median over $5$ independent runs for \textbf{f.} and \textbf{j.}. For \textbf{C.}, for practical reasons, we only performed a single experiment -- we stopped the training after $10000$ iterations ($72570$ s) and we extrapolated the computation time for $16000$ iterations.} 
    \label{tab:compTime2}
\end{table}

\subsubsection{Influence of the non-linearity in the driver}

\begin{figure}[p]
\begin{adjustwidth}{-1.5cm}{-1.5cm}
\centering
\textbf{1.}\includegraphics[height=5cm, trim=0.9cm 0cm 8cm 0cm, clip]{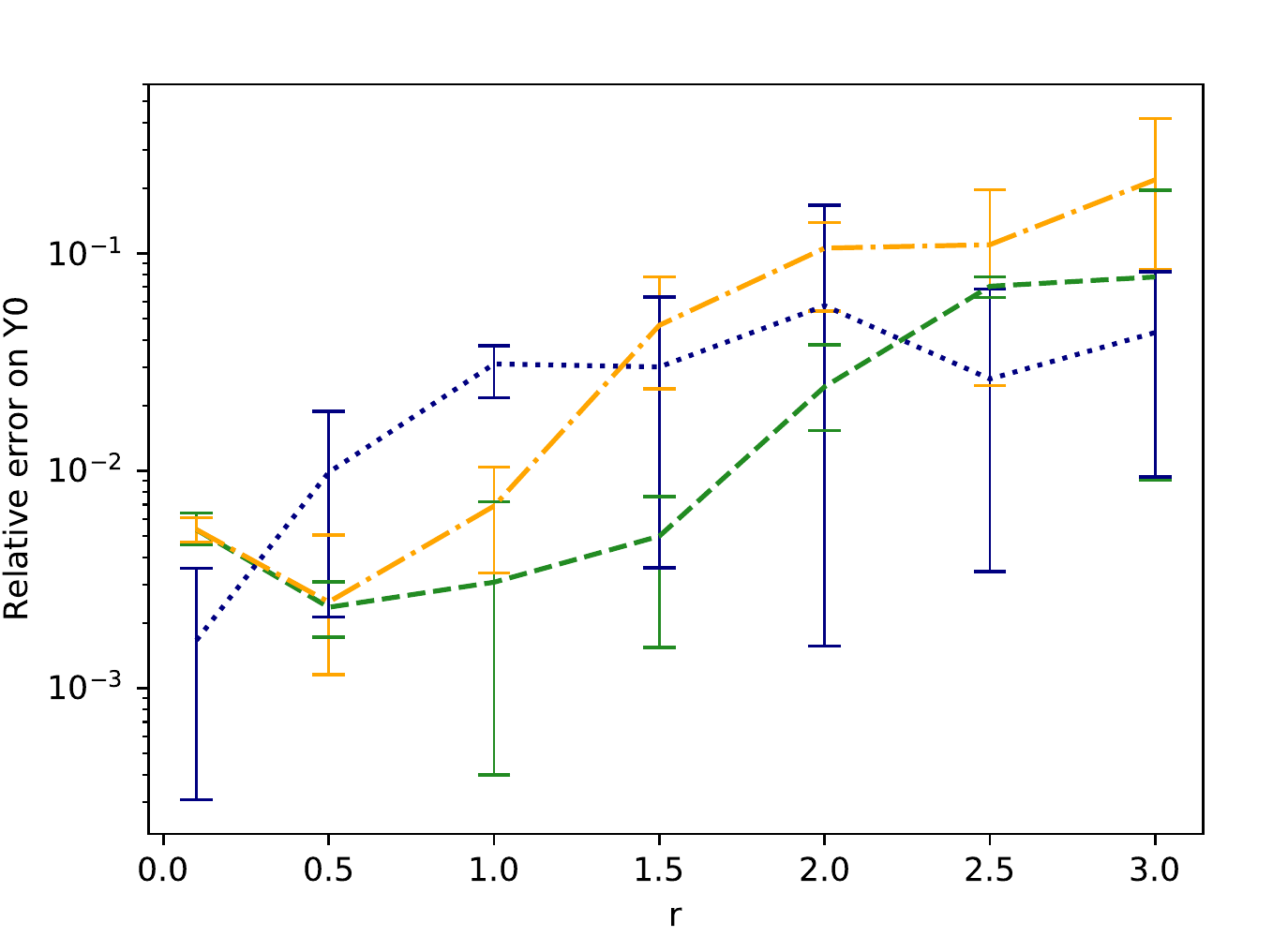}
\textbf{2.}\includegraphics[height=5cm, trim=0.9cm 0 1.5cm 0, clip]{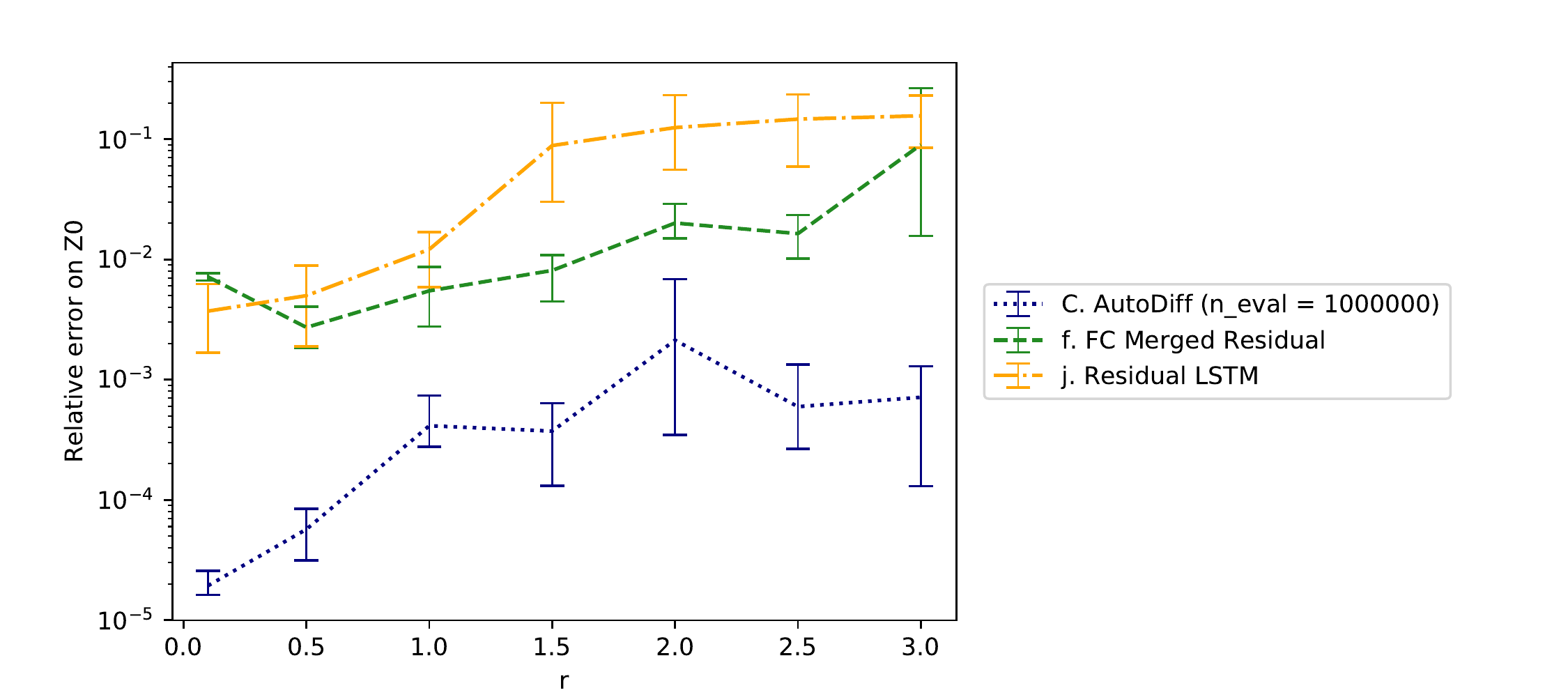} \\
\textbf{3.}\includegraphics[height=5cm, trim=0.9cm 0cm 8cm 0cm, clip]{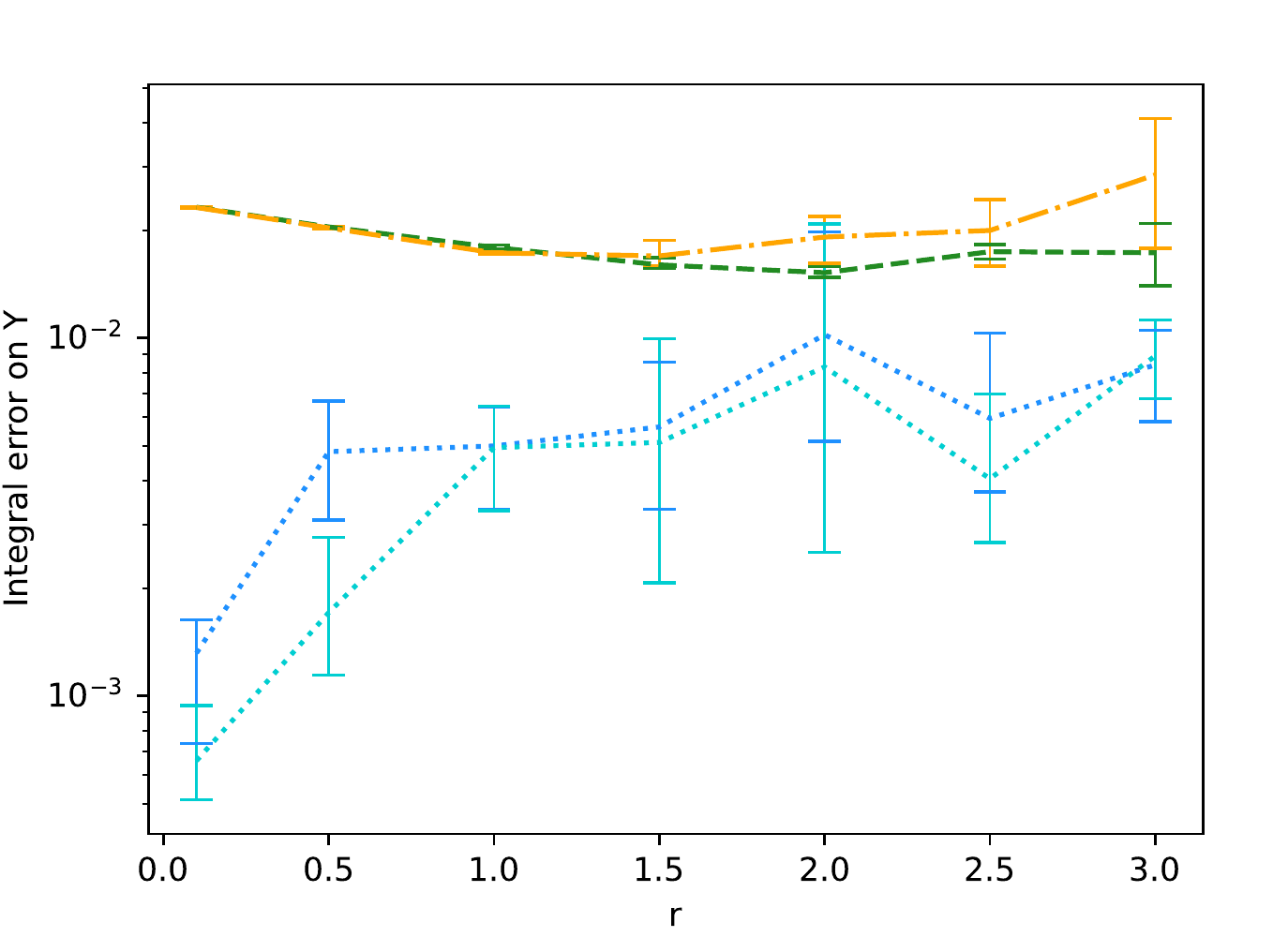}
\textbf{4.}\includegraphics[height=5cm, trim=0.9cm 0 1.5cm 0, clip]{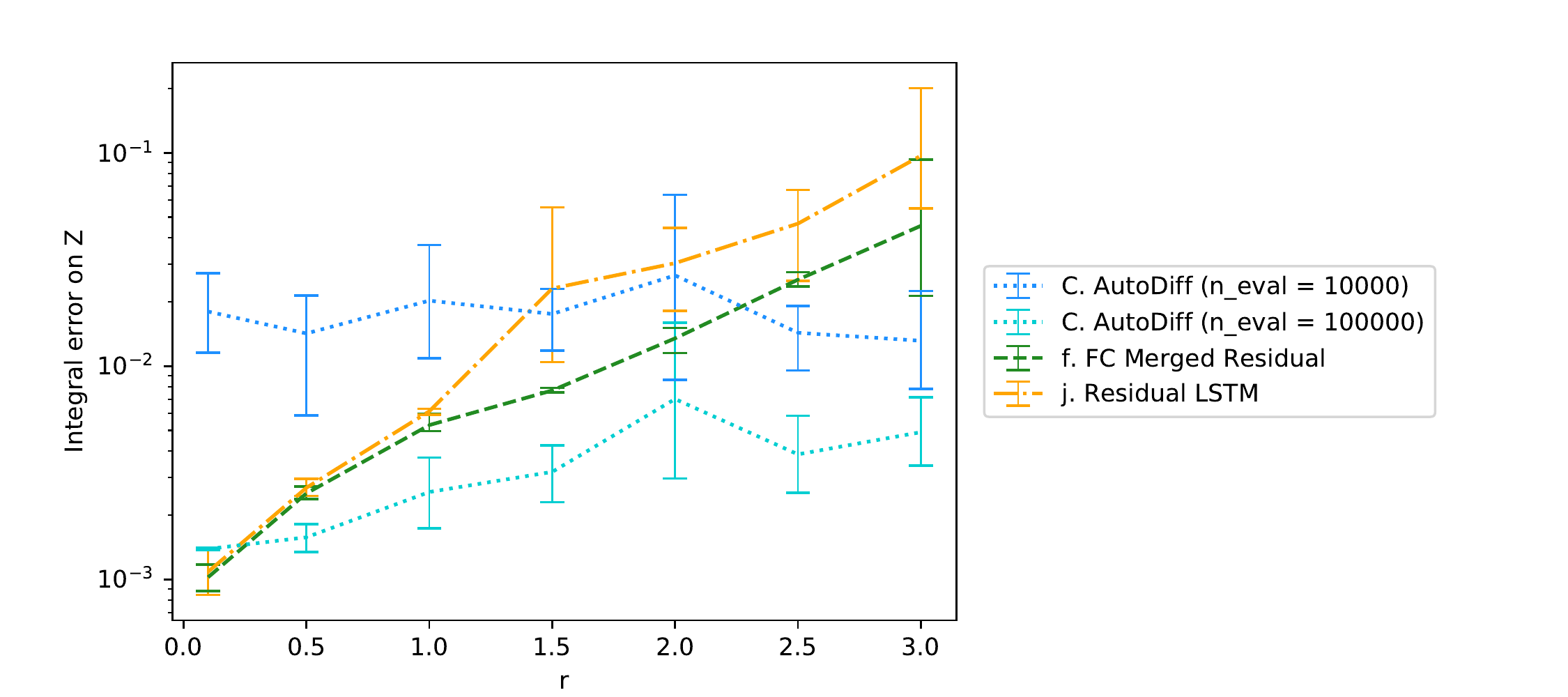} \\
\textbf{5.}\includegraphics[height=5cm,]{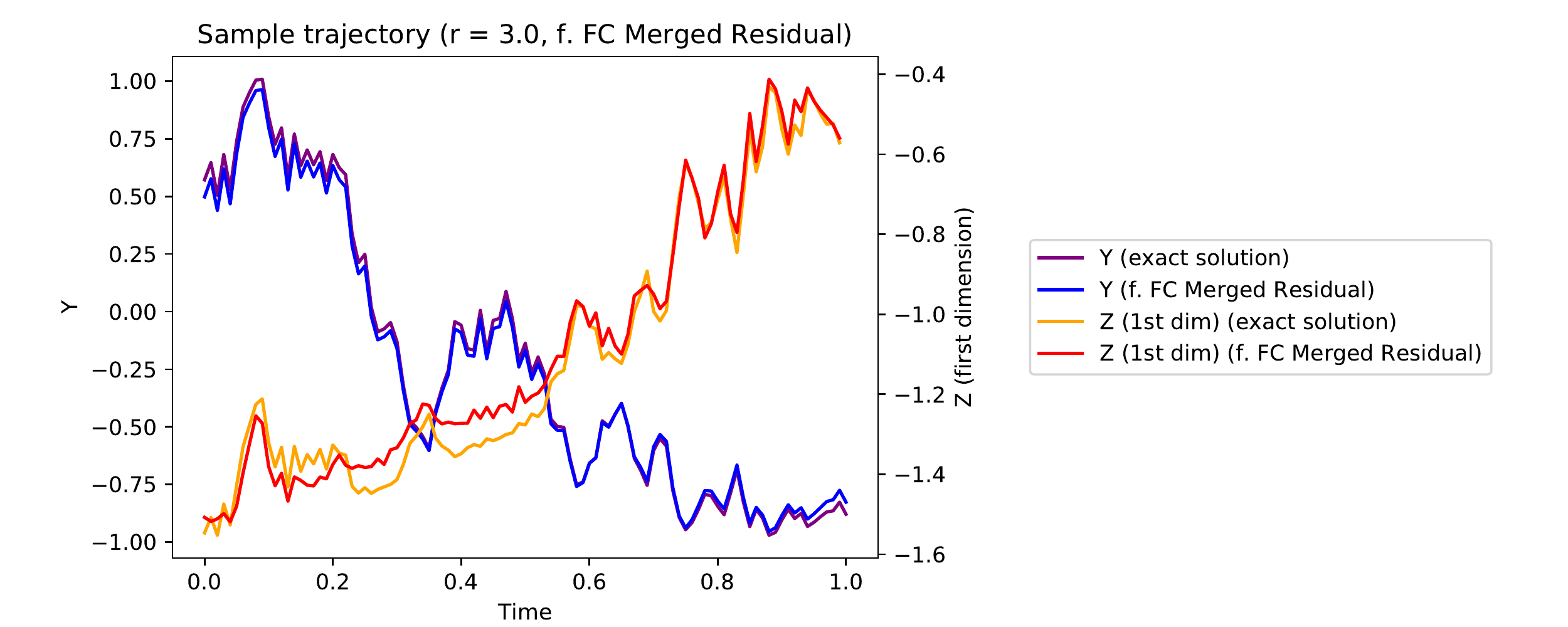} \\
\textbf{6.}\includegraphics[height=5cm,]{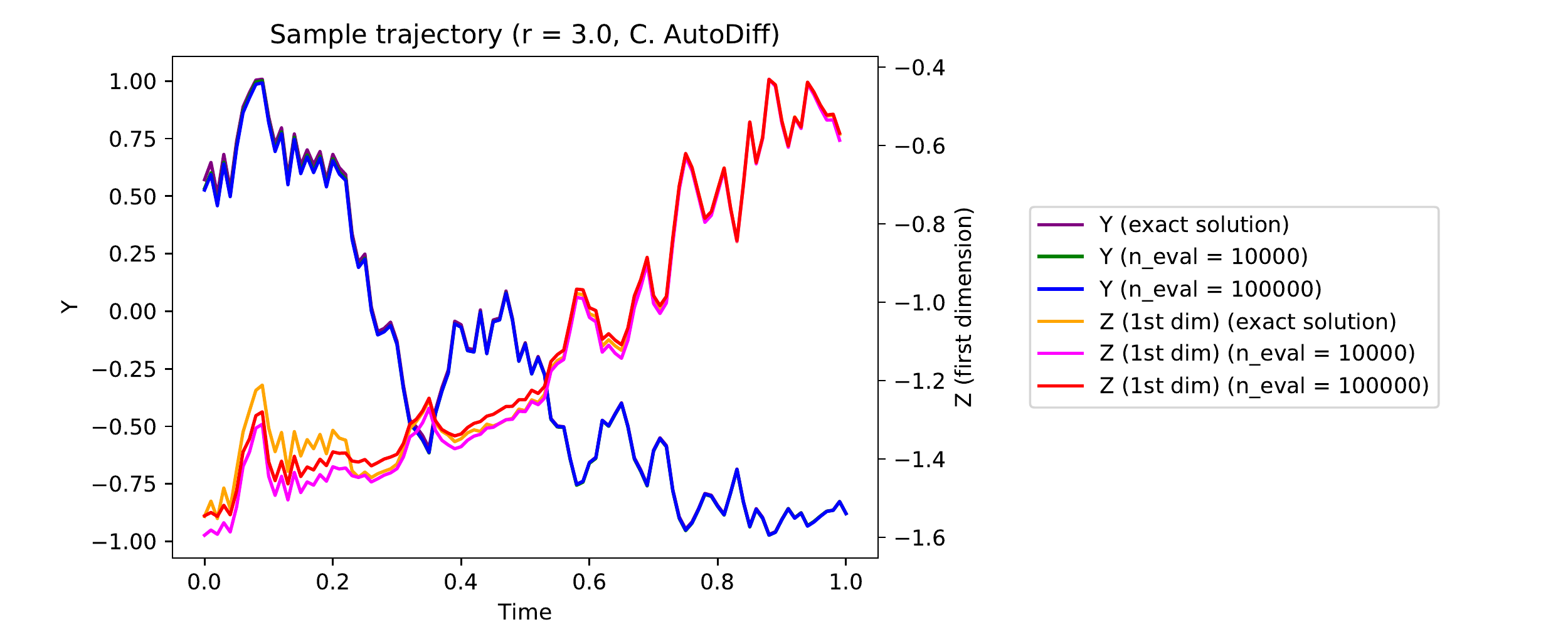}
\end{adjustwidth}
\caption{Influence of the \textbf{non-linearity} $r$ using our new algorithm \textbf{C.}  on equation \ref{pde:warin2} ($d=10$, $\lambda=0.5$) compared to networks \textbf{f.} and \textbf{j.} (our best algorithms using \emph{Deep BSDE}). We represent \textbf{1.} the relative error on $Y_0$ \eqref{eq:relerrY0}, \textbf{2.} the relative error on $Z_0$ \eqref{eq:relerrZ0}, \textbf{3.} the integral error on $Y$ \eqref{eq:interrY}, \textbf{4.} the integral error on $Z$ \eqref{eq:interrZ}. Integral errors are computed using $N=100$ time steps. The final errors increase as $r$ increases, but remains acceptable. The convergence curves (not represented) are similar for all $r$, with final losses between $10^{-3}$ and $10^{-4}$. We represent a sample trajectory using \textbf{f.} in \textbf{5.} and using \textbf{C.} in \textbf{6.} with $r=3.0$.}
\label{fig:infr:deepnesting:warin2}
\end{figure}

We investigate the influence of $r$ on equation \ref{pde:warin2} and compare the results with \emph{Deep BSDE} with networks \textbf{f.} and \textbf{j.}. The results are presented in Figure \ref{fig:infr:deepnesting:warin2}. Indeed, increasing $n_\text{eval}$ helps achieving lower errors during evaluation, but the gain decreases when $r$ increases. Overall, our new algorithm yields similar precision as \emph{Deep BSDE} on the initial condition, with better integral error.

\subsubsection{Influence of the maturity}

\begin{figure}[p]
\begin{adjustwidth}{-1.5cm}{-1.5cm}
\centering
\textbf{1.}\includegraphics[height=5cm, trim=0.9cm 0cm 8cm 0cm, clip]{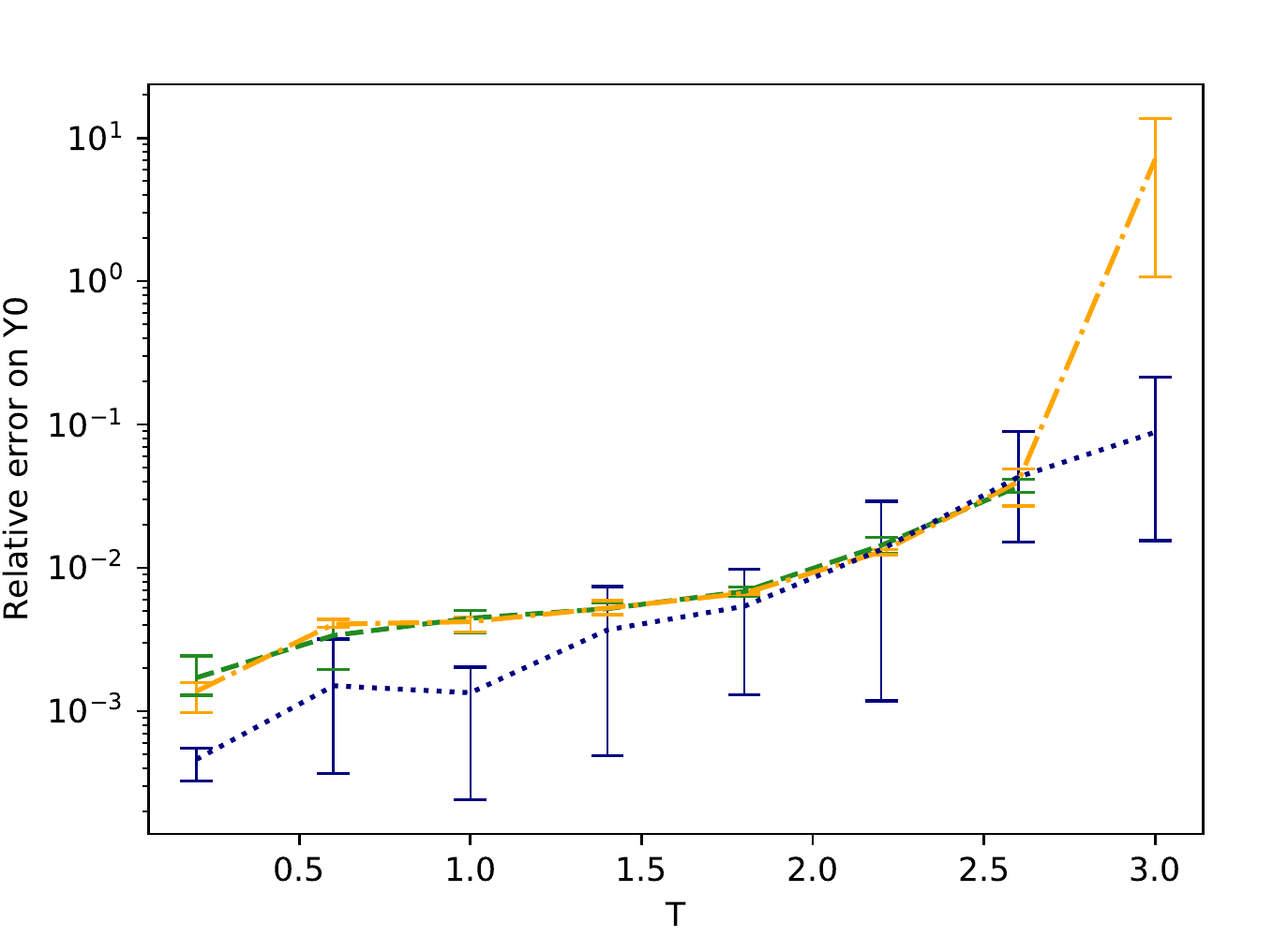}
\textbf{2.}\includegraphics[height=5cm, trim=0.9cm 0 1.5cm 0, clip]{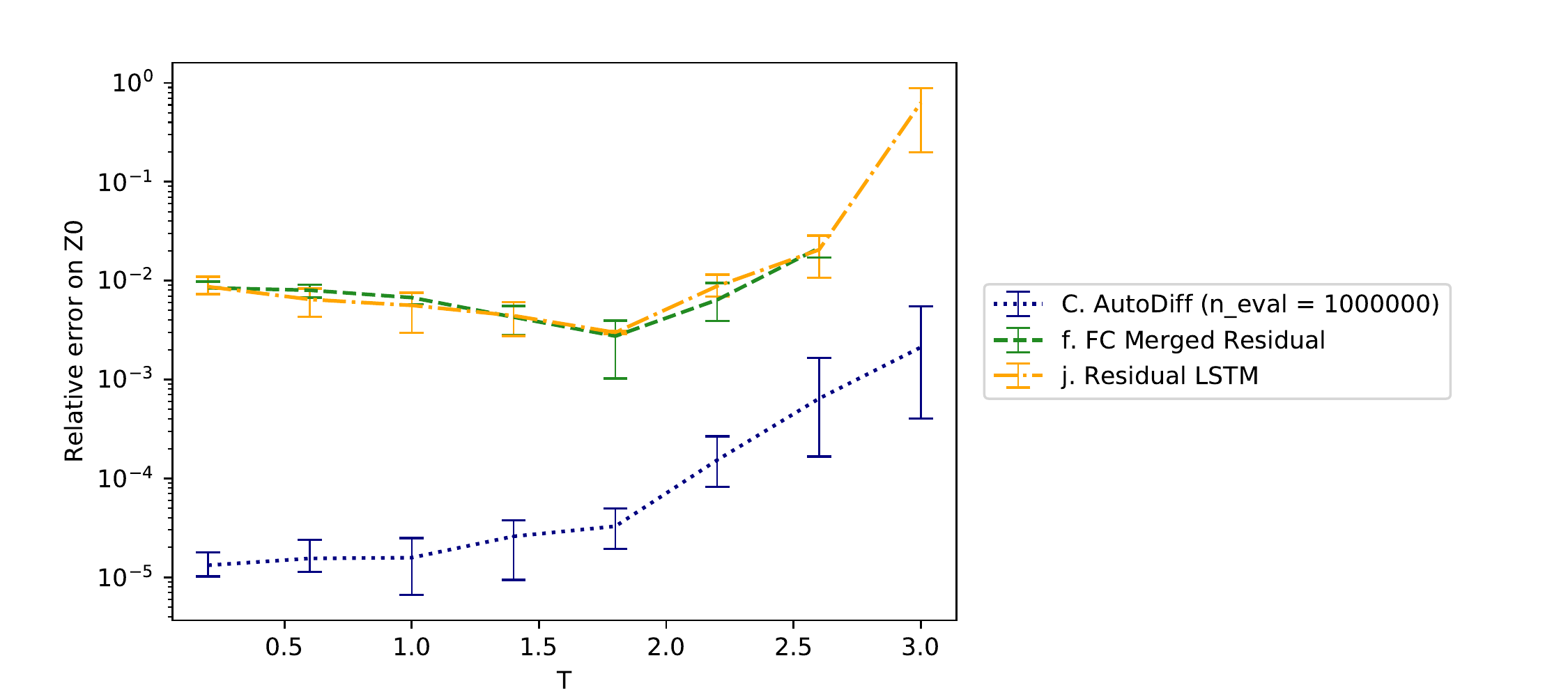} \\
\textbf{3.}\includegraphics[height=5cm, trim=0.9cm 0cm 8cm 0cm, clip]{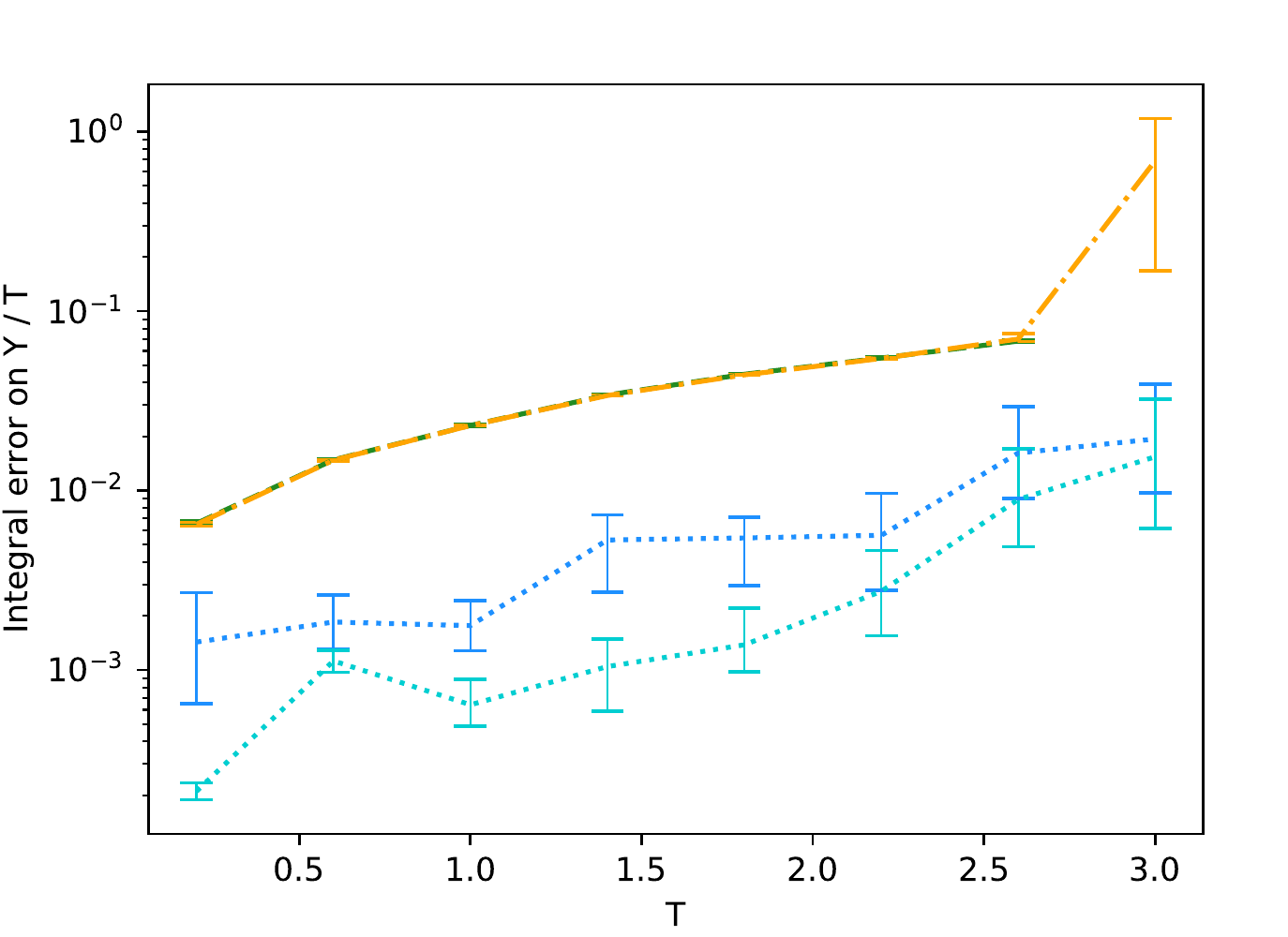}
\textbf{4.}\includegraphics[height=5cm, trim=0.9cm 0 1.5cm 0, clip]{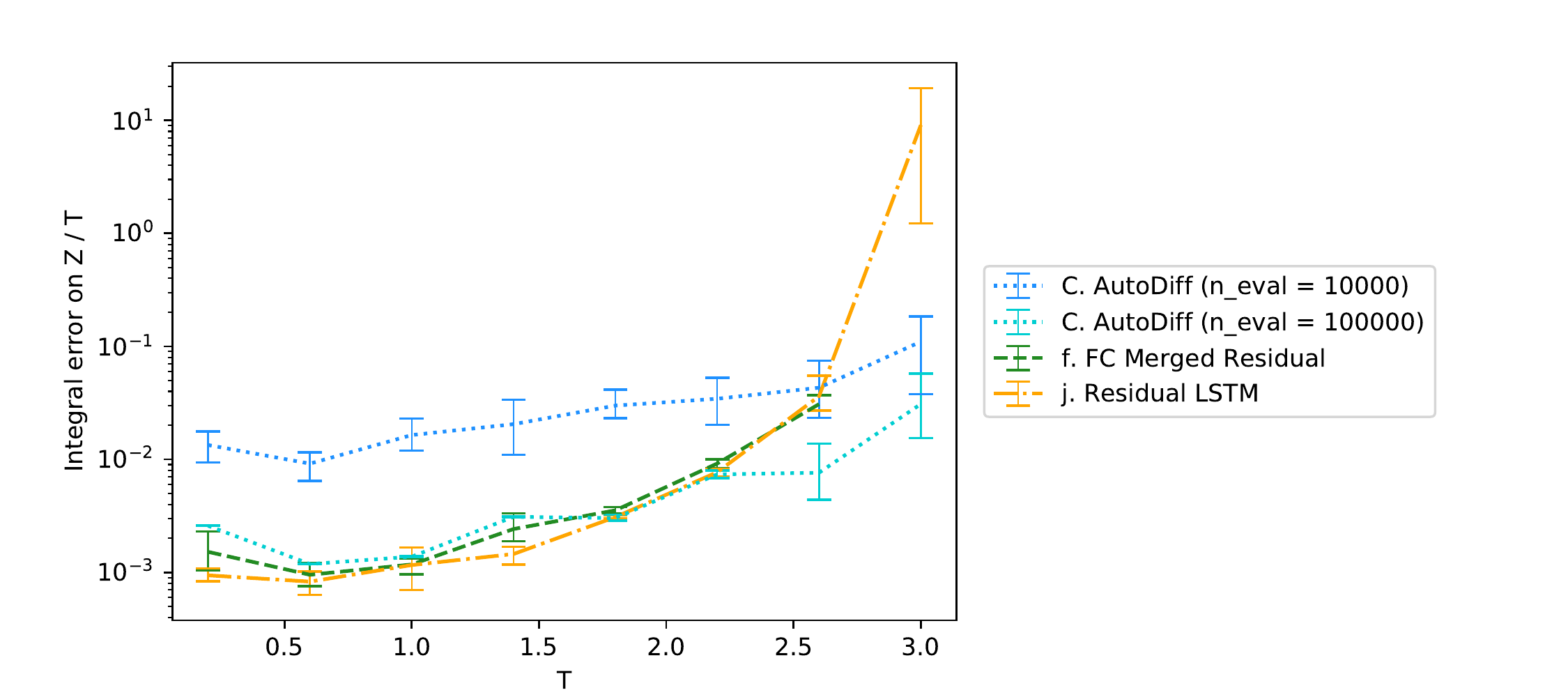} \\
\textbf{5.}\includegraphics[height=5cm,]{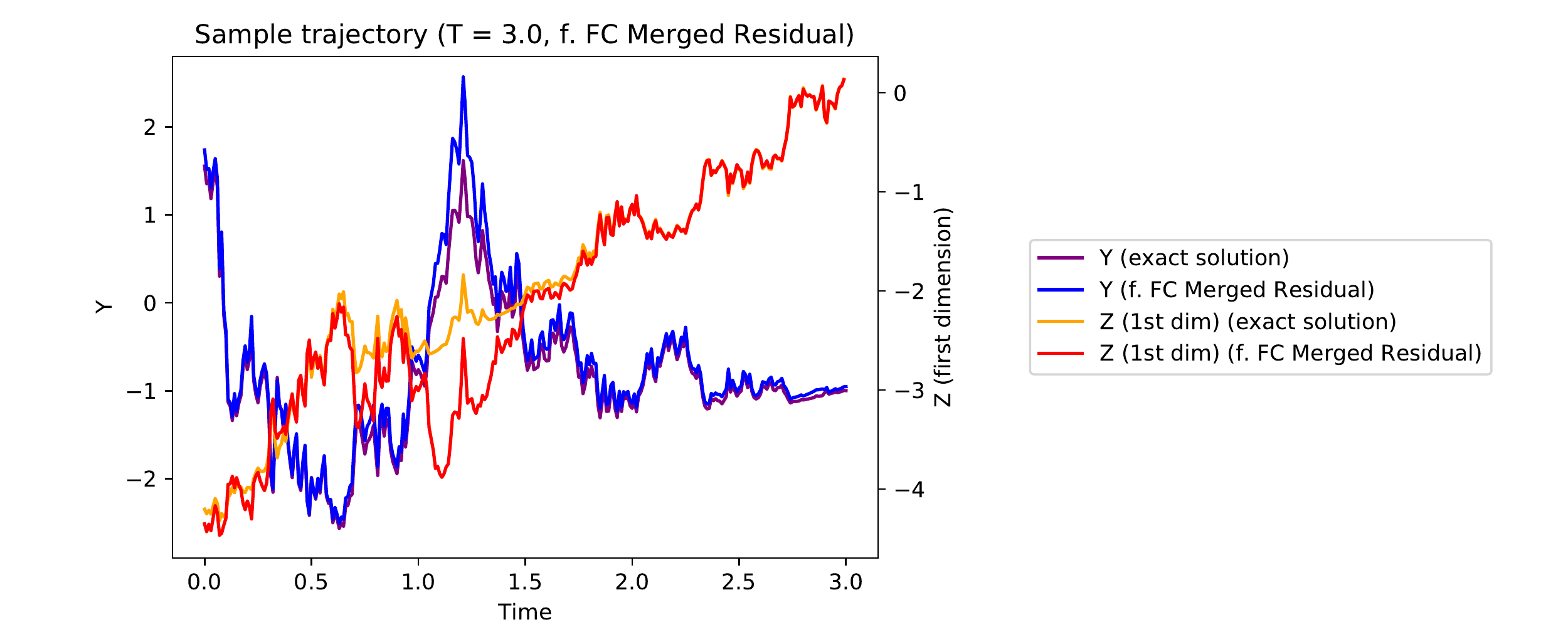} \\
\textbf{6.}\includegraphics[height=5cm,]{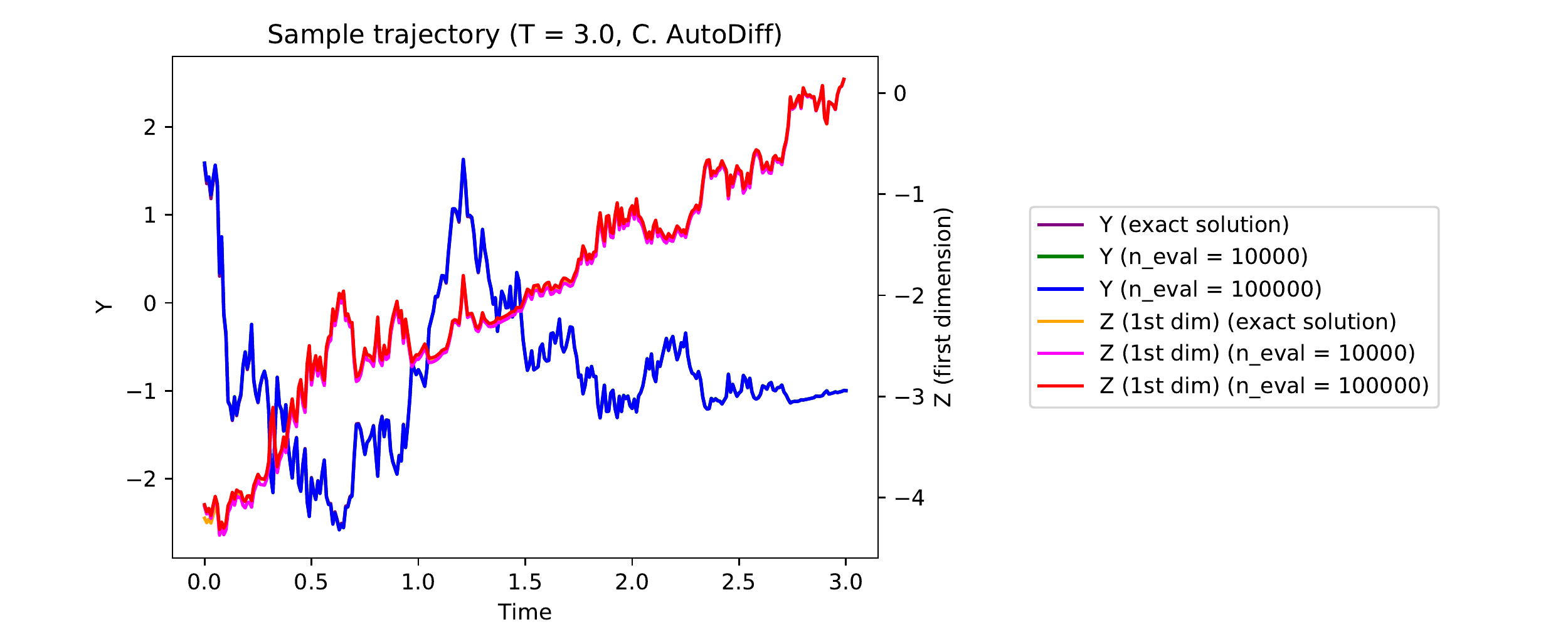}
\end{adjustwidth}
\caption{Influence of the \textbf{maturity} $T$ using our new algorithm \textbf{C.} on equation \ref{pde:warin2} ($d=10$, $\lambda=0.5$) compared to networks \textbf{f.} and \textbf{j.} (our best algorithms using \emph{Deep BSDE}). We represent \textbf{1.} the relative error on $Y_0$ \eqref{eq:relerrY0}, \textbf{2.} the relative error on $Z_0$ \eqref{eq:relerrZ0}, \textbf{3.} the integral error on $Y$ \eqref{eq:interrY}, \textbf{4.} the integral error on $Z$ \eqref{eq:interrZ}. Integral errors are computed using $N = 100 \times T$ time steps. The final errors increase as $r$ increases, but remains acceptable. The final loss (not represented) increases with $T$ from $\sim 10^{-4}$ to $\sim 10^{-3}$. We represent a sample trajectory using \textbf{f.} in \textbf{5.} and using \textbf{C.} in \textbf{6.} with $T=3.0$.}
\label{fig:infT:deepnesting:warin2}
\end{figure}

We investigate the influence of $T$ on equation \ref{pde:warin2}. We compare the results with \emph{Deep BSDE} using $N = 100 \times T$ -- for consistency, we compute the integral error for our new algorithm using the same rule for the number of time steps, even though this is not a parameter of the algorithm. The results are presented in Figure \ref{fig:infT:deepnesting:warin2}. The error on the initial condition is similar to \emph{Deep BSDE}, slightly lower on $Z_0$, and our new algorithm shows better integral errors overall. All the algorithms presented remain quite stable when $T$ increases.

\subsubsection[Other examples]{Other examples }

We run our new algorithm on some other equations \ref{pde:richou} (square root terminal condition) in dimension $d=10$ and \ref{pde:hjb} (Hamilton-Jacobi-Bellman  problem), \ref{pde:bsbarenblatt} (Black-Scholes-Barenblatt) in dimension $d=100$. Typical results are presented in Table \ref{fig:newalgo:resd100} -- these show comparable performance overall. The error on $Y_0$ is higher than the one obtained by the network \textbf{f.}. The integral errors are similar.

\begin{table}[ht]
\begin{adjustwidth}{-1.5cm}{-1.5cm}
\centering

\begin{tabular}{|P{2.5cm}|P{3cm}|P{3cm}|P{3cm}|P{3cm}|}
\hline
\ref{pde:richou} ($d=10$) & Abs error $Y_0$ & Norm error $Z_0$ & Integral error $Y$ & Integral error $Z$ \\
\hline
\textbf{f.} & $6.821155 \times 10^{-3}$ & $4.986833 \times 10^{-2}$ & $4.484242 \times 10^{-2}$ & $1.621381 \times 10^{-2}$ \\
\hline
\textbf{C.} & $1.054191 \times 10^{-2}$ & $2.332632 \times 10^{-2}$ & $2.162184 \times 10^{-2}$ & $4.105479 \times 10^{-3}$ \\
\hline
\end{tabular}
\\[0.5cm]

\begin{tabular}{|P{2.5cm}|P{3cm}|P{3cm}|P{3cm}|P{3cm}|}
\hline
\ref{pde:hjb} ($d=100$) & Abs error $Y_0$ & Norm error $Z_0$ & Integral error $Y$ & Integral error $Z$ \\
\hline
\textbf{f.} & $6.856918 \times 10^{-4}$ & $1.510367 \times 10^{-3}$  & $9.022277 \times 10^{-3}$ & $2.065553 \times 10^{-3}$ \\
\hline
\textbf{C.} & $9.320735 \times 10^{-3}$ & $9.937924 \times 10^{-4}$ & $1.322869 \times 10^{-2}$ & $1.715425 \times 10^{-3}$ \\
\hline
\end{tabular}
\\[0.5cm]

\begin{tabular}{|P{2.5cm}|P{3cm}|P{3cm}|P{3cm}|P{3cm}|}
\hline
\ref{pde:bsbarenblatt} ($d=100$) & Abs error $Y_0$ & Norm error $Z_0$ & Integral error $Y$ & Integral error $Z$ \\
\hline
\textbf{f.} & $7.350158 \times 10^{-2}$ & $2.887233 \times 10^{-1}$ & $2.009426 \times 10^{-1}$ & $3.74562 \times 10^{0}$ \\
\hline
\textbf{A.} & $5.261307 \times 10^{-1}$ & $1.180990 \times 10^{0}$ & $1.795204 \times 10^{-1}$ & $2.033016 \times 10^{0}$ \\
\hline
\textbf{B.} & $2.990570 \times 10^{-1}$ & $7.033922 \times 10^{-1}$ & $1.498886 \times 10^{-1}$ & $2.144073 \times 10^{0}$ \\
\hline
\textbf{C bis.} & $1.036530 \times 10^{-1}$ & $1.334808 \times 10^{0}$ & $2.932491 \times 10^{-1}$ & $2.841126 \times 10^{0}$ \\
\hline
\textbf{C.} & $1.506577 \times 10^{-1}$ & $8.373394 \times 10^{-1}$ & $1.501750 \times 10^{-1}$ & $2.843763 \times 10^{0}$ \\
\hline
\end{tabular}
\end{adjustwidth}
\caption{Comparison of \emph{Deep BSDE} and our new fixed point algorithm on other examples, for one typical run. We use $N=100$ in \emph{Deep BSDE} and $\lambda=1.0$ in our new algorithm. We measure the initial errors $\abs{Y_0 - Y_{0, \text{ref}}}$, $\norm{Z_0 - Z_{0, \text{ref}}}$, and the mean integral errors on $Y$ \eqref{eq:interrY} and $Z$ \eqref{eq:interrZ} using $N=100$ time steps. It should be noted that the latter integral measures are the computed using $1500$ simulations for \textbf{f.} and $10$ simulations for \textbf{C.}. For \textbf{C.}, in dimension $d=10$, we use $n_\text{inner} = 10000$ and stop the training process after $16000$ iterations. In dimension $d=100$, for \textbf{A.}, \textbf{B.},\textbf{C.} and \textbf{C bis.}, we use $n_\text{inner}=4000$ and stop the training process after $10000$ iterations. We then evaluate the errors using $n_\text{eval} = 1000000$ for the initial errors and $n_\text{eval} = 100000$ for the integral errors.}
\label{fig:newalgo:resd100}
\end{table}

\begin{figure}[ht]
\begin{adjustwidth}{-1.5cm}{-1.5cm}
\centering
\textbf{1.}\includegraphics[height=5cm, trim=0.9cm 0cm 8cm 0cm, clip]{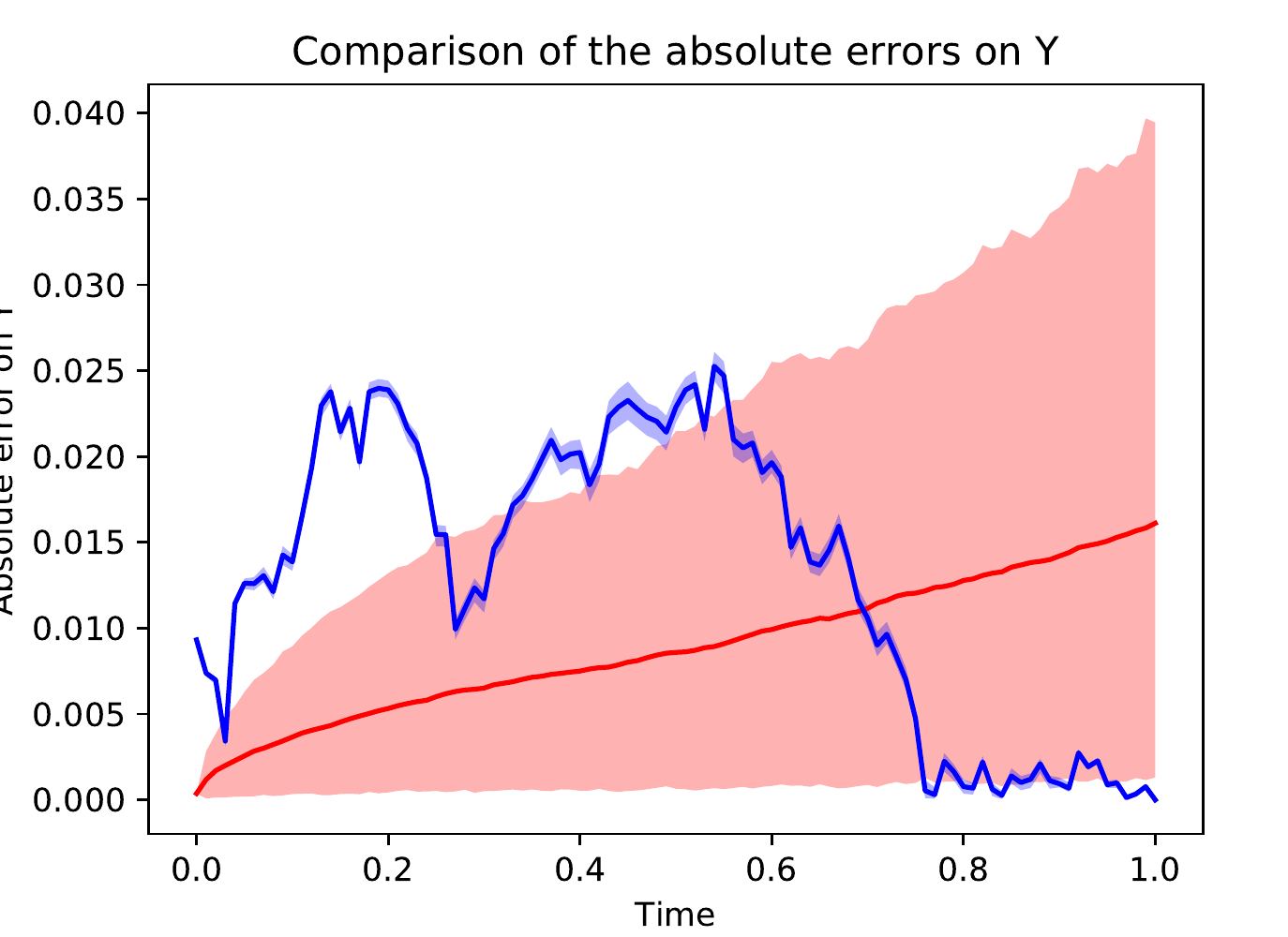}
\includegraphics[height=5cm, trim=0.9cm 0 3cm 0, clip]{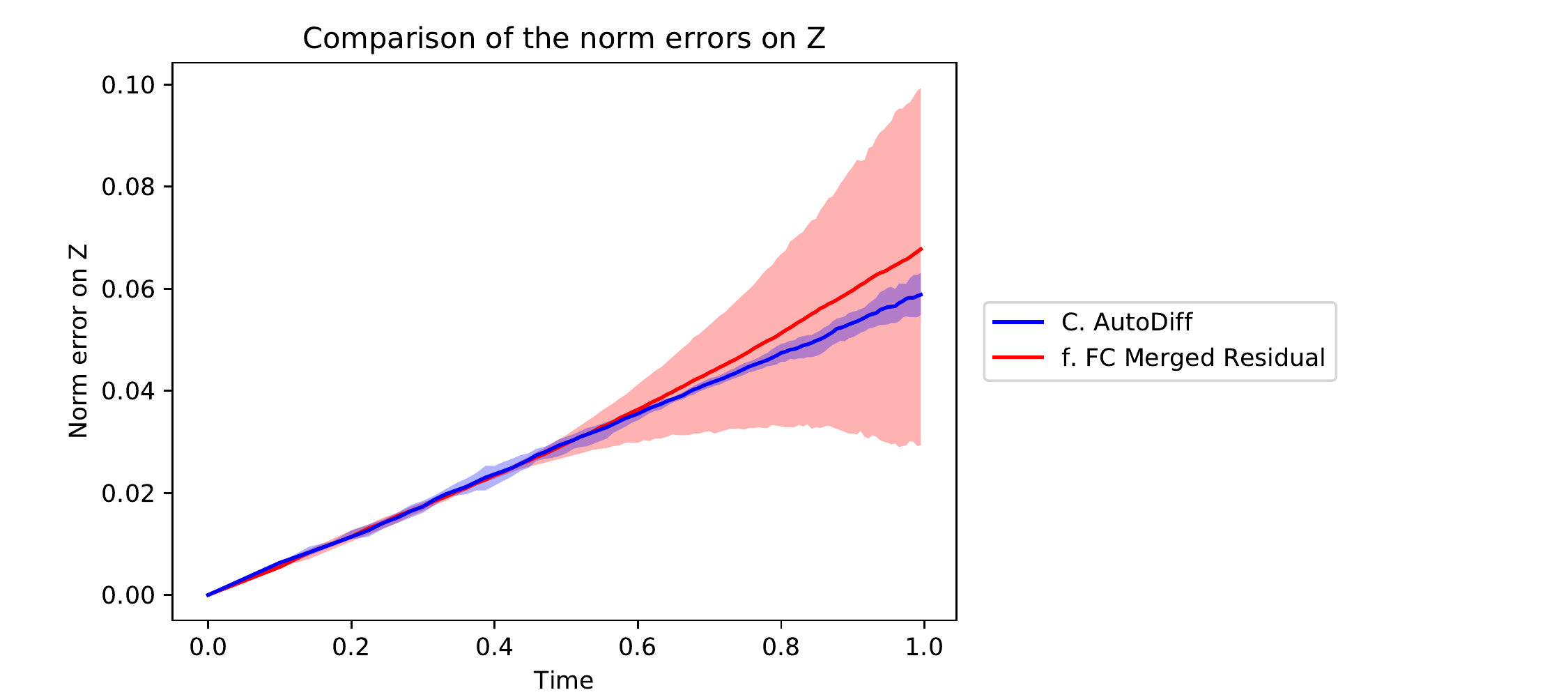} \\
\textbf{2.}\includegraphics[height=5cm, trim=0.9cm 0cm 8cm 0cm, clip]{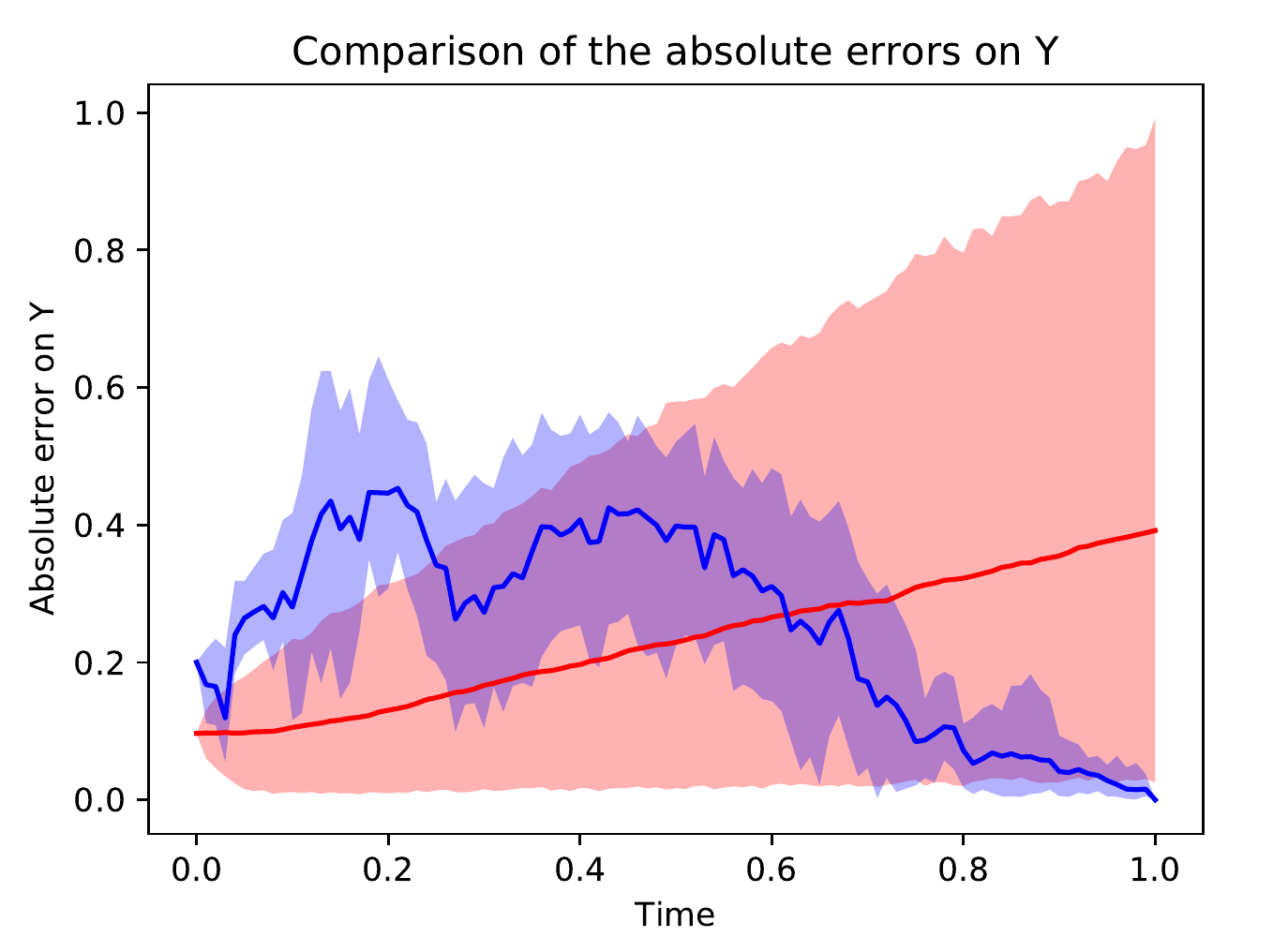}
\includegraphics[height=5cm, trim=0.9cm 0 3cm 0, clip]{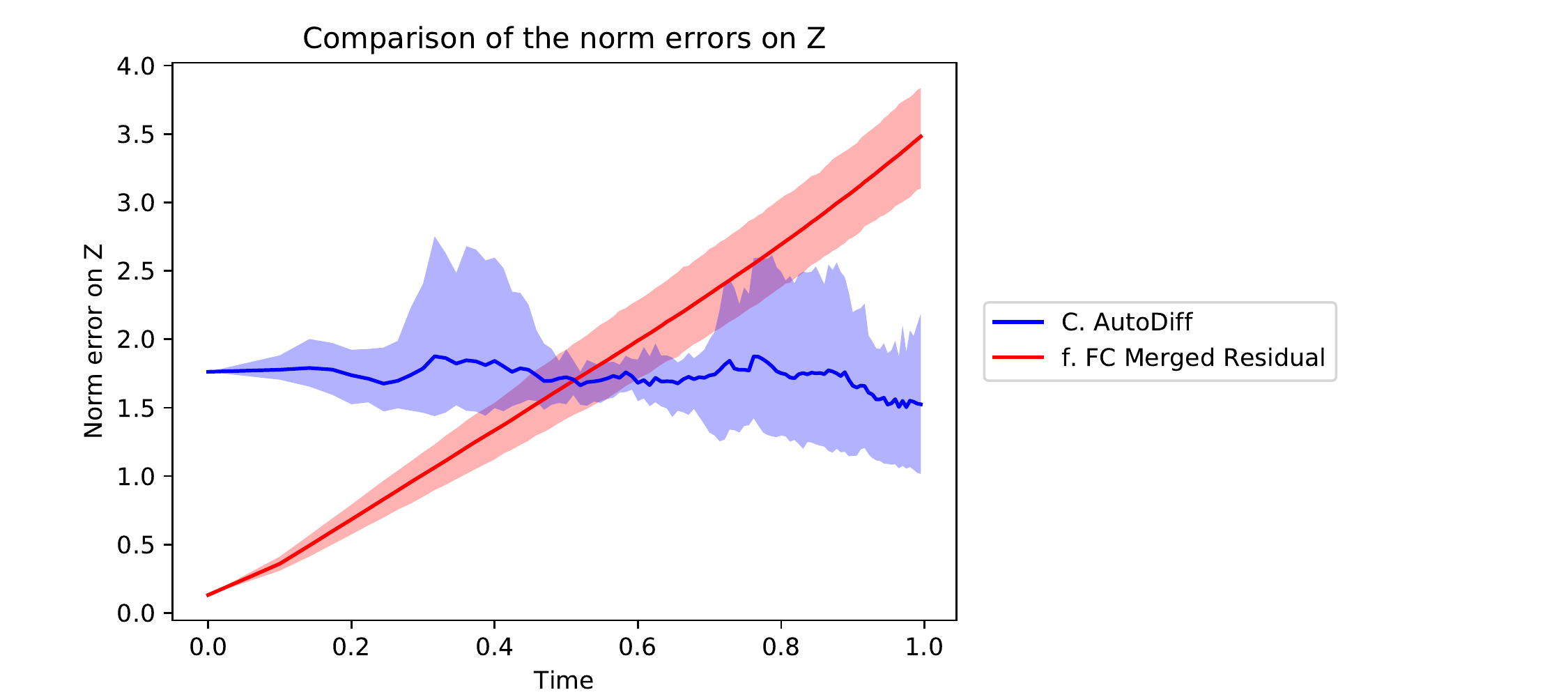}
\end{adjustwidth}
\caption{We represent the errors on the trajectories $\abs{Y-Y_\text{ref}}$ and $\norm{Z-Z_\text{ref}}$ function of the time horizon, for a typical run of our algorithms \textbf{C.} (with $\lambda=1.0$) and \textbf{f.} (with $N=100$ time steps). We represent the mean and the $5\%$ and $95\%$ quantiles on $1500$ trajectories for \textbf{f.} and $10$ trajectories for \textbf{C.}. We use equations \textbf{1.} \ref{pde:hjb} ($d=100$) and \textbf{2.} \ref{pde:bsbarenblatt} ($d=100$) with a time horizon of $T=1.0$. For \textbf{C.}, we use $n_\text{inner}=4000$ and we stop the training process after $10000$ iterations. We then compute the error measures using $n_\text{eval} = 100000$ (including the $Z_0$ and $Y_0$ use here in the error plots, thus a higher error than in Table \ref{fig:newalgo:resd100}).}
\label{fig:newalgo:resd100:errors}
\end{figure}

Finally, we compare qualitatively the shape of the errors on the trajectories using equations \ref{pde:hjb} and \ref{pde:bsbarenblatt} in dimension $d=100$. These are represented in Figure \ref{fig:newalgo:resd100:errors}. The shape of the errors are not similar, as the error increases when $t$ increases for \emph{Deep BSDE}, and it seems not to be the case for our new fixed point algorithm.

\clearpage
\section{Discussion}

In our tests, we could not find a case in which our two class of algorithms fail, i.e. a case in which the loss would seem small while the solution would be incorrect. In the worst cases, the algorithms would rather diverge or explicitly not converge. In the other cases they give results and computation times comparable (when no better) to the state-of-the-art. Thus, the loss function seems to be a reasonably robust indicator of precision, as lower losses are likely to correspond to lower error on trajectories. Most divergence cases are explained by: using a too high learning rate, initializing parameters incorrectly, encountering ``vanishing'' or ``exploding'' gradients issues when using very high numbers of time steps.

\paragraph{\emph{Deep BSDE} type algorithm.} We found that using Merged or LSTM architectures in \emph{Deep BSDE} significantly improve the precision of the results and the stability of the algorithms, while having a lower number of parameters. In particular, they enable to use more time steps to further increase the performance, and enable to use generic learning rate strategies and batch size values. In our tests, we also found that the number and width of the hidden layers could be set to typical values of $2$ layers of size $2d$. Otherwise, further tuning these hyperparameters has to be done for each combination of network and equation. 
We found that the algorithms are robust to increasing the non-linearity $r$ and the maturity $T$. As one may expect, increasing these values make the resolution more difficult, but the solutions remain acceptable.

\paragraph{Our new algorithm.} We found that our new algorithm can solve the same range of problems as \emph{Deep BSDE}, in dimensions $d=10$ and $d=100$, with tractable computation times ($\sim 1 - 10$ hours on a standard computer). This algorithm does not discretize time during training or evaluation, so that the underlying solution $u(t, x)$ can be evaluated for any $t$. A potential limitation is that this algorithm's memory usage grows with $n_\text{inner}$ and $d$, yet we found that typical values of $n_\text{inner}=4000$ or $10000$ yield good results while fitting in a standard computer's memory. Our algorithm discretizes a conditional expectation operator rather than time, which lead to different error shapes -- we found these errors to be of the same magnitude as our best algorithms based on \emph{Deep BSDE}. Finally, we found our algorithm to be slightly more robust to increasing $r$ and $T$ than our best algorithms based on \emph{Deep BSDE}, while keeping the hyperparameters constant (we did not increase $n_\text{inner}$ when increasing $T$ for instance).
As it is the case with \emph{Deep BSDE} methods, this algorithm could be generalized to second order BSDE.

\begin{figure}[ht]
\centering
\includegraphics[height=5cm, trim=0.9cm 0cm 8cm 0cm, clip]{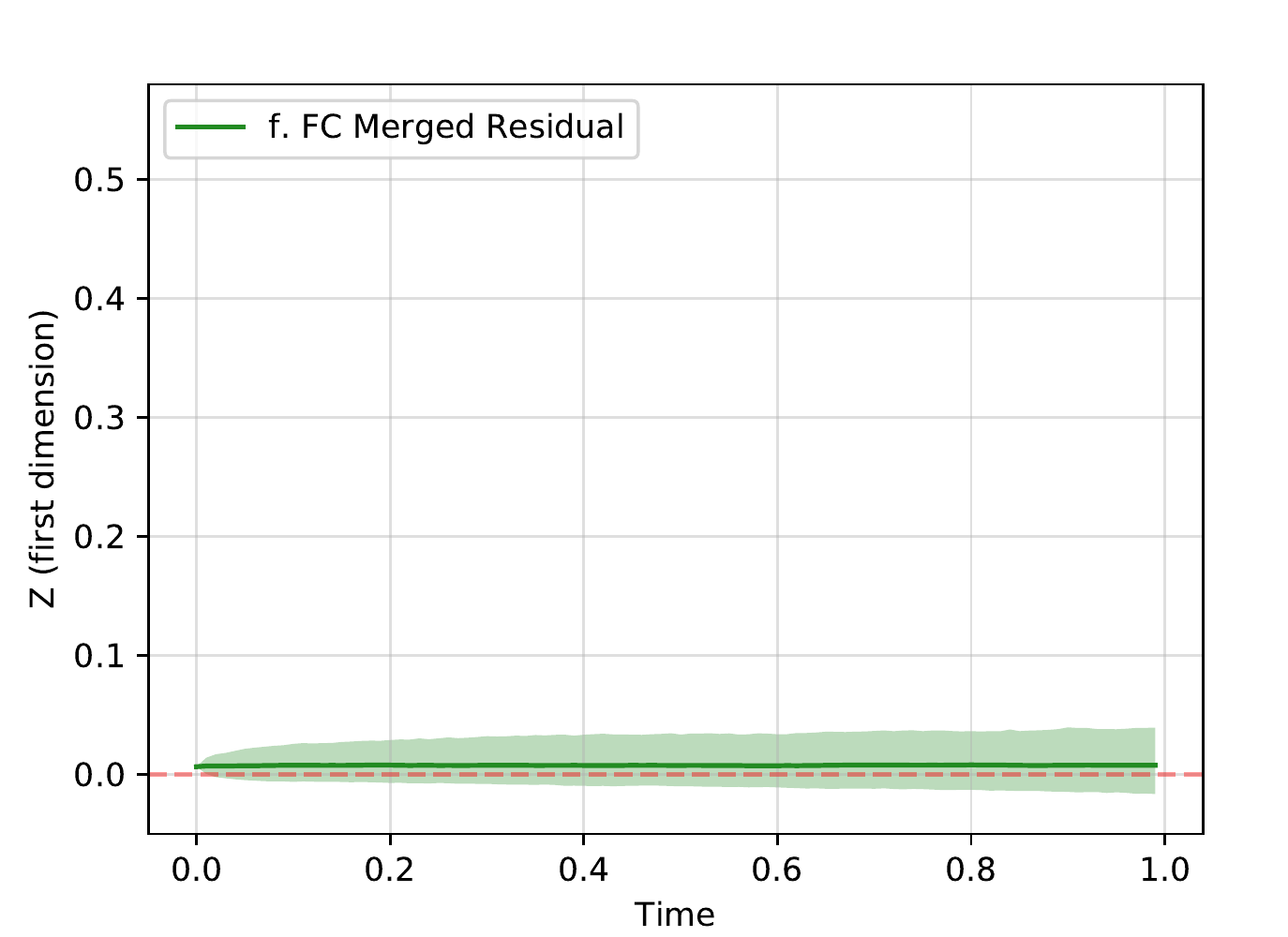}
\includegraphics[height=5cm, trim=0.9cm 0 8cm 0, clip]{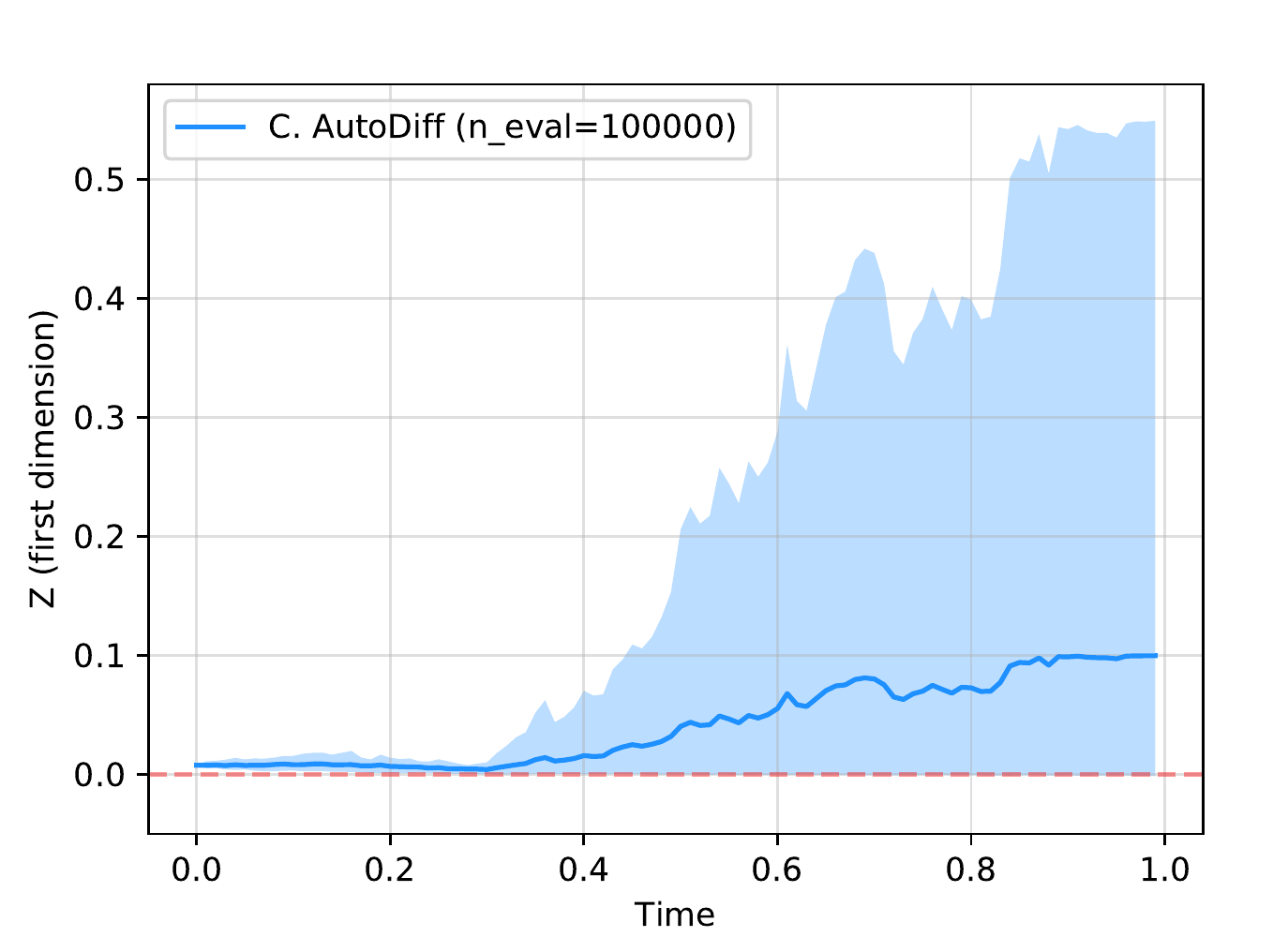}
\caption{Comparison of the distributions of the $Z$ trajectories (first dimension) on a sample run of our algorithms using \textbf{f.} and \textbf{C.} on equation \ref{pde:blackscholesdefault}. The means and $5\%$ and $95\%$ quantiles (computed using $1500$ trajectories for \textbf{f.} and $10$ trajectories for \textbf{C.} using $n_\text{eval} = 100 000$) are represented. The final losses (not comparable) are $16.84$ for \textbf{f.} and $3.87$ for \textbf{C.}. The resulting $Y_0$ are $57.11$ for \textbf{f.} and $57.37$ for \textbf{C.}.}
\label{fig:comparisonbsdefault}
\end{figure}

\paragraph{Special cases} Finally, we would like to point out that equations with ill-defined derivatives on the terminal condition remain problematic. For instance the equation \ref{pde:blackscholesdefault} has a terminal condition defined by:
\begin{align*}
g(x) &= \min_i x_i
\end{align*}
Its derivative as computed by TensorFlow is $Dg_i(x) = 1$ if $i=\textrm{argmin}(x_i)$, $0$ else. In this case, we do not have an analytical solution -- the algorithm and parameters used in \cite{han2017overcoming} lead to a slow convergence and a final loss of $\simeq 26$, showing that the algorithm could not find a way to replicate exactly the input flow. Moreover we found that the neural networks did not learn well in this case ($\kappa_{t_i}$ is constant for every value of $X_{t_i}$ for some $i > 0$). We solved the same equation with our Merged network \textbf{f.} and our new algorithm \textbf{C.}, and results presented in Figure \ref{fig:comparisonbsdefault} show that the solutions found are quite different. Our new algorithm \textbf{C.} seems more correct. Further analysis of such cases remain to be conducted.

\appendix
\section{Supplementary materials}

\subsection{Some test PDEs}

We recall our notations for the semilinear PDEs:
\begin{align*}
-\partial_t u(t, x) - \mathcal{L} u(t, x) &= f(t, x, u(t, x), \sigma^\top (t, x) \nabla u(t, x)) \\
u(T, x) &= g(x)
\end{align*}
where $\mathcal{L}u (t, x) := \frac{1}{2} \Tr \left(\sigma^\top \sigma(t, x) \nabla^2 u(t, x)\right) + \mu(t, x)^\top \nabla u(t, x)$. For each example, we thus give the corresponding $\mu$, $\sigma$, $f$ and $g$. In the implementation, we rather use a function $$\Tilde{f}(t, x, u(t, x), Du(t, x)) := f(t, x, u(t, x), \sigma^\top(t, x) Du(t, x))$$ for convenience, as this does not influence the results and allows for a more direct formulation for some PDEs.

\subsubsection{A Black-Scholes equation with default risk} \label{pde:blackscholesdefault}

From \cite{weinan2017deep,han2017overcoming}. If not otherwise stated, the parameters take values: $\overline{\mu}=0.02$, $\overline{\sigma}=0.2$, $\delta=2/3$, $R=0.02$, $\gamma_h=0.2$, $\gamma_l=0.02$, $v_h=50$, $v_l=70$. We use the initial condition $X_0 = (100, \cdots, 100)$.
\begin{align*}
\mu : (t, x) &\mapsto \overline{\mu} x \\
\sigma : (t, x) &\mapsto \overline{\sigma} \; \mathrm{diag}(\left\{x_i\right\}_{i=1..d})  \\
f : (t, x, y, z) &\mapsto -(1-\delta ) \times \mathrm{min} \left\{ \gamma_h, \; \mathrm{max}\left\{ \gamma_l, \; \frac{\gamma_h - \gamma_l}{v_h - v_l} \left( y - v_h \right) + \gamma_h \right\}\right\} y - R y\\
g : x &\mapsto \min_{i=1..d} x_i
\end{align*}
We used the closed formula for the SDE dynamic:
\begin{align*}
X_t &= X_s \exp\left[\left((\bar{\mu}-\frac{\bar{\sigma}^2}{2})\right) (t-s) + \bar{\sigma} (W_t - W_s)\right]  & \forall t > s
\end{align*}

Baseline (from \cite{han2017overcoming}):
\begin{center}
\begin{tabular}{|c|c|}
\hline
$Y_0$ & $47.300$ \\
\hline
\end{tabular}
\end{center}

\subsubsection{A Black-Scholes-Barenblatt equation} \label{pde:bsbarenblatt}

From \cite{raissi2018forward}. If not otherwise stated, the parameters take values: $\overline{\sigma}=0.4$, $r=0.05$. We use the initial condition $X_0 = (1.0, 0.5, 1.0, \cdots)$.
\begin{align*}
\mu : (t, x) &\mapsto 0 \\
\sigma : (t, x) &\mapsto \overline{\sigma} \; \mathrm{diag}(\left\{x_i\right\}_{i=1..d})  \\
f : (t, x, y, z) &\mapsto -r \left(y -  \frac{1}{\overline{\sigma}}\sum_{i=1}^{d} z_i\right)\\
g : x &\mapsto \norm{x}^2
\end{align*}

We used the closed formula for the SDE dynamic:
\begin{align*}
X_t &= X_s \exp\left[-\frac{\bar{\sigma}^2}{2} (t-s) + \bar{\sigma} (W_t - W_s)\right]  & \forall t > s
\end{align*}

Exact solution:
\begin{align*} 
u(t, x) &= \exp((r+\overline{\sigma}^2)(T-t))g(x)
\end{align*}

\subsubsection{A Hamilton-Jacobi-Bellman equation} \label{pde:hjb}

From \cite{weinan2017deep,han2017overcoming}. If not otherwise stated: $\lambda=1.0$ and $X_0 = (0, \cdots, 0)$.
\begin{align*}
\mu : (t, x) &\mapsto 0 \\
\sigma : (t, x) &\mapsto \sqrt{2} \; \mathbf{I}_d \\
f : (t, x, y, z) &\mapsto - 0.5 \; \lambda \norm{z}^2 \\
g : x &\mapsto \log \left( 0.5 \left[ 1 + \norm{x}^2 \right] \right)
\end{align*}

Monte-Carlo solution:
\begin{align*}
u(t, X_t) &= -\frac{1}{\lambda} \log \left( \mathbb{E}\left[ \; \exp \left( - \lambda g(X_t + \sqrt{2} B_{T-t}\right) \right] \right) \\
\forall j, \quad \frac{\partial u}{\partial x_j}(t, x) &=  \left(\mathbb{E}\left[ \exp \left\{ -\lambda g(X_t + \sqrt{2} B_{T-t})\right\}\right]\right)^{-1} \mathbb{E} \left[ \DPar{g}{x_j}(X_t + \sqrt{2} B_{T-t}) \exp \left\{ -\lambda g(X_t + \sqrt{2} B_{T-t})\right\} \right] \\
\text{where} \quad \DPar{g}{x_j}(x) &= \frac{2x_j}{1+\norm{x}^2}
\end{align*}

Baseline (computed using $10$ million Monte-Carlo realizations, $d=100$):
\begin{center}
\begin{tabular}{|c|c|}
\hline
$Y_0$ & $4.590119548591171217$ \\
\hline
$Z_0$ & $-0.00006060718806111$ \\
\hline
\end{tabular}
\end{center}

\subsubsection{An oscillating example with a square non-linearity} \label{pde:warin2}

From \cite{warin2018nestingsMC}. If not otherwise stated, the parameters take values: $\mu_0=0.2$, $\sigma_0=1.0$, $a=0.5$, $r=0.1$. The intended effect of the $\min$ and $\max$ in $f$ is to make $f$ Lipschitz. We used the initial condition $X_0=(1.0, 0.5, 1.0, \cdots)$.
\begin{align*}
\mu : (t, x) &\mapsto \mu_0 / d \\
\sigma : (t, x) &\mapsto \sigma_0 / \sqrt{d} \; \mathbf{I}_d \\
f : (t, x, y, z) &\mapsto \phi(t, x) + r \left(\max\left[-\exp(2a(T-t)), \min\left\{\frac{1}{\sigma_0 \sqrt{d}} y \sum_{i=1}^{d} z_i , \exp(2a(T-t)) \right\} \right] \right)^2 \\
\text{where} \quad \phi : (t, x) &\mapsto \cos\left(\sum_{i=1}^{d} x_i \right) \left( a + \frac{\sigma_0^2}{2} \right) \exp(a(T-t)) + \sin\left( \sum_{i=1}^{d} x_i \right) \mu_0 \exp(a(T-t)) \\
&\quad - r \left( \cos\left( \sum_{i=1}^{d} x_i \right) \sin\left( \sum_{i=1}^{d} x_i \right)\exp(2a(T-t))\right)^2 \\
g : x &\mapsto \cos \left( \sum_{i=1}^d x_i \right)
\end{align*}

Exact solution: 
\begin{align*} u(t, x) &= \cos\left(\sum_{i=1}^d x_i \right) \exp(a(T-t)) \\
\forall j, \quad \frac{\partial u}{\partial x_j}(t, x) &= -\sin\left(\sum_{i=1}^d x_i \right) \exp(a(T-t))
\end{align*}

\subsubsection{A non-Lipschitz terminal condition} \label{pde:richou}

From \cite{richou2010phd}. If not otherwise stated: $\alpha=0.5$, $X_0 = (0, \cdots, 0)$.
\begin{align*}
\mu : (t, x) &\mapsto 0 \\
\sigma : (t, x) &\mapsto \mathbf{I}_d \\
f : (t, x, y, z) &\mapsto - 0.5 \norm{z}^2 \\
g : x &\mapsto \sum_{i=1}^d \left(\max\left\{0, \min\left[ 1, x_i\right] \right\}\right)^\alpha
\end{align*}

Monte-Carlo solution:
\begin{align*}
u(t, X_t) &= \log\left( \mathbb{E}\left[ \; \exp \left(g(X_t + B_{T-t})\right) \right] \right) \\
\forall j, \quad \frac{\partial u}{\partial x_j}(t, x) &=  \left(\mathbb{E}\left[ \exp \left\{ g(X_t + \sqrt{2} B_{T-t})\right\}\right]\right)^{-1} \mathbb{E} \left[ g'(X_t + \sqrt{2} B_{T-t}) \exp \left\{ g(X_t + \sqrt{2} B_{T-t})\right\} \right] \\
\text{where} \quad g'(x) &= \left\{ \begin{array}{ll} 0 &\text{if $x\leq 0$ or $x\geq 1$} \\ \alpha x^{\alpha-1} &\text{else}\end{array} \right.
\end{align*}

Baseline (computed using $10$ million Monte-Carlo realizations, $d=10$):
\begin{center}
\begin{tabular}{|c|c|}
\hline
$Y_0$ & $4.658493663928657$ \\
\hline
$Z_0$ & $0.3795303954478772$ \\
\hline
\end{tabular}
\end{center}

\subsubsection{An oscillating example with Cox-Ingersoll-Ross propagation} \label{pde:cir}

From \cite{warin2018monte}. If not otherwise stated: $a=0.1$, $\alpha=0.2$, $T=1.0$, $\hat{k}=0.1$, $\hat{m}=0.3$, $\hat{\sigma}=0.2$. We used the initial condition $X_0 = (0.3, \cdots, 0.3)$. Note that we have $2 \hat{k} \hat{m} > \hat{\sigma}^2$ so that $X$ remains positive.
\begin{align*}
\mu : (t, x) &\mapsto \hat{k}\left( \hat{m} - x \right) \\
\sigma : (t, x) &\mapsto \hat{\sigma} \mathrm{diag} \left\{ \sqrt{x} \right\}
\end{align*}
\begin{align*} f : (t, x, y, z) &\mapsto \phi(t, x) + a  y \left(\sum_{i=1}^d z_i \right)\\
\text{where} \quad \phi : (t, x) &\mapsto \cos\left(\sum_{i=1}^{d} x_i \right) \left( -\alpha + \frac{\hat{\sigma}^2}{2} \right) \exp(-\alpha(T-t)) + \sin\left( \sum_{i=1}^{d} x_i \right) \exp(-\alpha(T-t)) \sum_{i=1}^{d} \hat{k}\left( \hat{m} - x_i\right) \\
&+ a \cos\left( \sum_{i=1}^{d} x_i \right) \sin \left(\sum_{i=1}^{d} x_i\right) \exp(-2\alpha(T-t)) \sum_{i=1}^d \hat{\sigma} \sqrt{x_i}
\end{align*}
\begin{align*}
g : x &\mapsto \cos \left( \sum_{i=1}^d x_i \right)
\end{align*}

Exact solution: 
\begin{align*} u(t, x) &= \cos\left(\sum_{i=1}^d x_i \right) \exp(-\alpha(T-t)) \\
\forall j, \quad \frac{\partial u}{\partial x_j}(t, x) &= -\sin\left(\sum_{i=1}^d x_i \right) \exp(-\alpha(T-t))
\end{align*}

\subsubsection{An oscillating example with inverse non-linearity} \label{pde:warin3}

If not stated otherwise, we took parameters $\mu_0 = 0.2$, $\sigma_0 = 1.0$, $a=0.5$, $r=0.1$ and the initial condition $X_0=(1.0, 0.5, 1.0, \cdots)$.

\begin{align*}
\mu : (t, x) &\mapsto \mu_0 /d \mathbbm{1}_d\\
\sigma : (t, x) &\mapsto \sigma_0 \; \mathbf{I}_d \\
f : (t, x, y, z) &\mapsto \phi(t, x) + r \frac{d y}{\sum_{i=1}^d z_i} \\
g : x &\mapsto 2\sum_{i=1}^d x_i + \cos\left(\sum_{i=1}^d x_i\right)\\
\text{where} \quad \phi : (t, x) &\mapsto 2a \sum_{i=1}^d x_i \exp(a(T-t)) + \cos\left(\sum_{i=1}^{d} x_i \right) \left( a + \frac{d \sigma_0^2}{2} \right) \exp(a(T-t)) \\
&\quad - \mu_0 \left[2- \sin\left( \sum_{i=1}^{d} x_i \right) \right]  \exp(a(T-t)) \\
&\quad - r \frac{2 \sum_{i=1}^d x_i + \cos\left(\sum_{i=1}^{d} x_i \right)}{\sigma_0 \left[2 - \sin\left(\sum_{i=1}^{d} x_i \right)\right]} 
\end{align*}

Exact solution: 
\begin{align*} u(t, x) &= \left[2\sum_{i=1}^d x_i + \cos\left(\sum_{i=1}^d x_i \right)\right] \exp(a(T-t)) \\
\forall j, \quad \frac{\partial u}{\partial x_j}(t, x) &= \left[2-\sin\left(\sum_{i=1}^d x_i \right)\right] \exp(a(T-t))
\end{align*}

\subsection{Demonstration of Proposition \ref{prop:minloss}} 
\label{sec:demo1}
In the sequel, $C$ will only depend on $u$ $f$, $g$ and may vary from one line to another.

First, note that due to the Lipschitz property in Assumption \ref{ass::lipF}, the boundedness of $f$ and $g$ and the regularity of $u$ due to Assumption \ref{ass::uReg}, the solution $u$ of \eqref{eq:PDEInit} and $Du$ satisfy a Feynman-Kac relation (see an adaptation of Proposition 1.7 in \cite{touzi2012optimal}):
\begin{align}
 u(t,x)= & \frac{1}{2}\E_{t,x} \big[ \phi\big(t, t+\tau ,X_{t+\tau}^{t,x}, u(t+\tau ,X_{t+\tau}^{t,x}), Du(t+\tau ,X_{t+\tau}^{t,x}) \big) +  \nonumber \\
 & \phi\big(t, t+\tau ,\widehat{X}_{t+\tau}^{t,x}, u(t+\tau ,\widehat{X}_{t+\tau}^{t,x}), Du(t+\tau ,\widehat{X}_{t+\tau}^{t,x}) \big) \big] \nonumber \\
 Du(t,x)= &\E_{t,x} \Big[ \sigma^{-\top} \frac{W_{(t+ (\tau  \wedge \Delta t )) \wedge T}-W_{t}}{\tau \wedge (T-t) \wedge \Delta t} \frac{1}{2} \big( \phi(t,t+\tau, X_{t+\tau}^{t,x}, u(t+\tau, X_{t+\tau}^{t,x}), Du(t+\tau, X_{t+\tau}^{t,x})) -  \nonumber \\
 & \quad  \phi(t,t+\tau, \widehat{X}_{t+\tau}^{t,x}, u(t+\tau, \widehat{X}_{t+\tau}^{t,x}), Du(t+\tau, \widehat{X}_{t+\tau}^{t,x})) \big) \big].
 \label{eq:udu}
 \end{align}
First picking $(t,y) \in [0,T] \times \Omega_\epsilon^{0,x}$, we have using equation \eqref{eq:udu}:
\begin{align*}
    B(\theta,t, y) = & | \bar u^{\epsilon}(\theta, t, y) - u(\theta ,t,y)|^2 \\
    \le &
    2 | u(t,y)-u(\theta ,t,y)|^2 + 
     2 \left( \E_{t,y}[ 1_{X^{t,y}_{t+\tau} \in \Omega^{0,x}_\epsilon}   \frac{1}{2}\left( \frac{f(t,X^{t,y}_{t+\tau},u(\theta , t+\tau, X^{t,y}_{t+\tau}), v(\theta,t+\tau, X^{t,y}_{t+\tau}))}{ \rho(\tau)} - \right. \right.\\
     & \frac{f(t,X^{t,y}_{t+\tau},u( t+\tau, X^{t,y}_{t+\tau}), Du(t+\tau, X^{t,y}_{t+\tau}))}{ \rho(\tau)} + \frac{f(t,\hat X^{t,y}_{t+\tau},u(\theta , t+\tau, \hat X^{t,y}_{t+\tau}), v(\theta,t+\tau, \hat X^{t,y}_{t+\tau}))}{ \rho(\tau)} -\\
     & \left . \frac{f(t,\hat X^{t,y}_{t+\tau},u( t+\tau, \hat X ^{t,y}_{t+\tau}), Du(t+\tau, \hat X^{t,y}_{t+\tau}))}{ \rho(\tau)} \right) ] + \\
     &  \frac{1}{2}\E_{t,y} \big[ 1_{X^{t,y}_{t+\tau} \notin \Omega^{0,x}_\epsilon} (\phi\big(t, t+\tau ,X_{t+\tau}^{t,y}, u(t+\tau ,X_{t+\tau}^{t,y}), Du(t+\tau ,X_{t+\tau}^{t,y}) \big) +\\
     & \left. \phi\big(t, t+\tau ,\widehat{X}_{t+\tau}^{t,y}, u(t+\tau ,\widehat{X}_{t+\tau}^{t,y}), Du(t+\tau ,\widehat{X}_{t+\tau}^{t,y}) \big) )\big] \right)^2 
\end{align*}
Using Jensen equality, the boundedness of $g$ and $f$, the fact that $\rho$ is bounded by below and Assumption \ref{ass::lipF}, we get:
\begin{align*}
    B(\theta,t,y) \le & 2 | u(t,y)-u(\theta ,t,y)|^2 +  C (K^2 \E[ 
    1_{X^{t,y}_{t+\tau} \in \Omega^{0,x}_\epsilon}  \left( 
     (u(t+\tau,X^{t,y}_{t+\tau})-u(\theta ,t+\tau,X^{t,y}_{t+\tau}))^2 + \right. \\
     & \left.
     || Du(t+\tau,X^{t,y}_{t+\tau}) -v(\theta ,t+\tau,X^{t,y}_{t+\tau})||^2_2
     \right)] + \E[ 1_{X^{t,y}_{t+\tau} \notin \Omega^{0,x}_{\epsilon}}])
\end{align*}
  Then using the results in \cite{hornik1991approximation}, \cite{cybenko1989approximation}, we know that we are able to find  $ (u(\theta^{*},t,x), v(\theta^*,t,x )) \in \kappa$ such that:
 \begin{align}
  \sup_{(\hat t, y) \in [t,T]\times \Omega_\epsilon^{0,x}} ( | u(\hat t, y) - u(\theta^*, \hat t, y)| +
  || Du(\hat t, y)- v(\theta^{*},t,y)||_2) \le \epsilon
  \label{eq:nnr}
\end{align}
Then we have that 
\begin{align}
B(\theta^{*},t,y) \le & C (\epsilon + \E_{t,y}[ 1_{X^{t,y}_{t+\tau} \notin \Omega^{0,x}_{\epsilon}}]) \nonumber \\
 \le & C (\epsilon + \E_{t,y}[ 1_{\exists s  \in [t,T] / X^{t,y}_{s} \notin \Omega^{0,x}_{\epsilon}}])  
\label{eq:B}
\end{align}
Similarly  introducing $\hat W^{t}_\tau = \sigma^{-\top} \frac{W_{(t+ (\tau  \wedge \Delta t )) \wedge T}-W_{t}}{\tau \wedge (T-t) \wedge \Delta t}$,
\begin{align}
C(\theta^{*},t,y) = & || \bar v^\epsilon( \theta^{*} ,t, y) - v(\theta^{*} ,t,y)||^2_2 \nonumber\\
    \le &
    2 || Du(t,y)-v(\theta^{*} ,t,y)||^2_2 + 
     2 \left( \frac{1}{2} \E_{t,y}[ 1_{X^{t,y}_{t+\tau} \in \Omega^{0,x}_\epsilon}  \hat W^{t}_\tau \left( \frac{f(t,X^{t,y}_{t+\tau},u(\theta^{*} , t+\tau, X^{t,y}_{t+\tau}), v(\theta^{*},t+\tau, X^{t,y}_{t+\tau}))}{ \rho(\tau)} - \right. \right.\nonumber \\
     & \frac{f(t,X^{t,y}_{t+\tau},u( t+\tau, X^{t,y}_{t+\tau}), Du(t+\tau, X^{t,y}_{t+\tau}))}{ \rho(\tau)} - \frac{f(t,\hat X^{t,y}_{t+\tau},u(\theta^{*} , t+\tau, \hat X^{t,y}_{t+\tau}), v(\theta^{*},t+\tau, \hat X^{t,y}_{t+\tau}))}{ \rho(\tau)} + \nonumber \\
     & \left . \frac{f(t,\hat X^{t,y}_{t+\tau},u( t+\tau, \hat X ^{t,y}_{t+\tau}), Du(t+\tau, \hat X^{t,y}_{t+\tau}))}{ \rho(\tau)} \right) ] + \nonumber \\
     &  \frac{1}{2}\E_{t,y} \big[ 1_{X^{t,y}_{t+\tau} \notin \Omega^{0,x}_\epsilon}  \hat W^{t}_\tau (\phi\big(t, t+\tau ,X_{t+\tau}^{t,y}, u(t+\tau ,X_{t+\tau}^{t,y}), Du(t+\tau ,X_{t+\tau}^{t,y}) \big) - \nonumber \\
     & \left. \phi\big(t, t+\tau ,\widehat{X}_{t+\tau}^{t,y}, u(t+\tau ,\widehat{X}_{t+\tau}^{t,y}), Du(t+\tau ,\widehat{X}_{t+\tau}^{t,y}) \big) )\big] \right)^2 \nonumber \\
     \le & 2 || Du(t,y)-v(\theta^{*} ,t,y)||^2 +   C  \E_{t,y}[ (\hat W^{t}_\tau)^\top \hat W^{t}_\tau  (\frac{||f||_\infty^2 }{\rho(\tau)^2} + \frac{|g(X^{t,y}_{t+\tau}) - g( \hat X^{t,y}_{t+\tau})|^2}{\bar F(T)^2}) ]  \E_{t,y}[ 1_{X^{t,y}_{t+\tau} \notin \Omega^{0,x}_{\epsilon}} ]+ \nonumber \\
     & C K \E_{t,y}[ 1_{X^{t,y}_{t+\tau} \in \Omega^{0,x}_\epsilon} \frac{(\hat W^{t}_\tau)^\top \hat W^{t}_\tau}{\rho(\tau)^2}  \left( u(t+\tau,X^{t,y}_{t+\tau}) - u(\theta^{*},t+\tau,X^{t,y}_{t+\tau}) \right)^2  ] 
      + \nonumber \\
      & C K \E_{t,y}[1_{X^{t,y}_{t+\tau} \in \Omega^{0,x}_\epsilon}  \frac{(\hat W^{t}_\tau)^\top \hat W^{t}_\tau}{\rho(\tau)^2} || Du(t+\tau,X^{t,y}_{t+\tau}) - v(\theta^{*},t+\tau,X^{t,y}_{t+\tau}) ||^2_2  ],
\label{eq:C}
\end{align}
where we have used Jensen and Cauchy Schwarz.\\
Introducing $G\in \R^d$ composed of centered unitary independent Gaussian random variables
$$\E_{t,x}[ \frac{(\hat W^{t}_\tau)^\top \hat W^{t}_\tau }{\rho(\tau)^2}]
= \E[\frac{1}{\tau \rho(\tau)^2}] \E[G^\top \sigma^{-1} \sigma^{-\top} G] < \infty$$
when $u<1$ in equation \eqref{eq:CCF}.\\
Similarly using the fact that $g$ is Lipschitz,
\begin{align*}
  \E_{t,y}[ (\hat W^{t}_\tau)^\top \hat W^{t}_\tau  (\frac{||f||_\infty^2 }{\rho(\tau)^2} + \frac{|g(X^{t,y}_{t+\tau}) - g( \hat X^{t,y}_{t+\tau})|^2}{\bar F(T)^2}) ] < C  < \infty   
\end{align*}
Then using equation \eqref{eq:nnr} in equation \eqref{eq:C}, we get
\begin{align}
    C(\theta^{*}, t,y) \le C  ( \epsilon +  \E_{t,y}[ 1_{\exists s  \in [t,T] / X^{t,y}_{s} \notin \Omega^{0,x}_{\epsilon}}])\label{eq:CFin}
\end{align}
Injecting \eqref{eq:B} and \eqref{eq:CFin} in equation \eqref{eq:LossFirstSchemeLoc} and using the definition of $\Omega^{0,x}_{\epsilon}$ complete the proof.

\subsection{Demonstration of Proposition \ref{prop:convergence}}

Let $\hat K$ be a compact. It is always possible to find $n_0$ such that for $n> n_0$, $ \hat K \subset \Omega^{0,x}_{\epsilon_n}$.

Let $u$ be a function from $\R \times \R^d$ to $\R$ and $v$ function from $\R \times \R^d$ to $\R^d$. Let
\begin{align*}
    \psi(t,t+\tau,x,u) =&   1_{X^{t,x}_{t+\tau} \in \Omega^{0,x}_{\epsilon_n}} \frac{1}{2} \big[ \phi\big(t, t+\tau ,X_{t+\tau}^{t,x}, u(t+\tau ,X_{t+\tau}^{t,x}) \big) +   \phi\big(t, t+\tau ,\widehat{X}_{t+\tau}^{t,x}, u(t+\tau ,\widehat{X}_{t+\tau}^{t,x}) \big) \big].
\end{align*}

We note that for $\zeta$ an uniform random variable in $[0,T]$
\begin{align*}
D_n &:= \mathbb{E} \left[  \int_0^T   1_{X^{0,x}_{t} \in \Omega^{0,x}_{\epsilon_n}} \left( \E_{t,X^{0,x}_t} \left[ \psi(t,t+\tau,X^{0,x}_t,u(\theta_n, .,.),v(\theta_n, .,.)) \right]  - u(\theta_n,t ,X^{0,x}_{t}) \right)^2 \dif t \right]  \\
&=
T \mathbb{E} \left[ 1_{X^{0,x}_{\zeta} \in \Omega^{0,x}_\epsilon}  \left( \E_{\zeta,X^{0,x}_\zeta} \left[  \psi(\zeta,\zeta+\tau,X^{0,x}_\zeta,u(\theta_n, .,.),v(\theta_n, .,.)) \right]  - u(\theta_n, \zeta ,X^{0,x}_{\zeta}) \right)^2 \dif t \right]\\
&= T \ell(\theta)
\end{align*}
Using the equations \eqref{eq:udu}, the expression of $D_n$, the Lipschitz property of $f$, Jensen inequality, the boundedness of $f$, $g$, and the fact that $\rho$ is  bounded by below by $\hat \rho(T)$:
\begin{flalign*}
  F_n= &  \E\left[  \int_0^T  1_{X^{0,x}_{t} \in K}  (u( t,X^{0,x}_t) - u( \theta_n, t,X^{0,x}_t)) ^2  \dif t \right]\\ 
  \le & \E\left[   \int_0^T  1_{X^{0,x}_{t} \in \Omega^{0,x}_{\epsilon_n}}  (u( t,X^{0,x}_t) - u( \theta_n, t,X^{0,x}_t)) ^2 \right] \dif t \\
 \le &    3 \E\left[    \int_0^T  1_{X^{0,x}_{t} \in \Omega^{0,x}_{\epsilon_n}} \left( \E_{t,X^{0,x}_t}   (\psi(t,t+\tau,X^{0,x}_{t+\tau},u) - \psi(t,t+\tau,X^{0,x}_{t+\tau}, u(\theta_n,.,.))) \right)^2 \dif t \right] + \\
 & 3 \E\left[ \int_0^T  1_{X^{0,x}_{t} \in \Omega^{0,x}_{\epsilon_n}}  \left( \E_{t,X^{0,x}_t} (  \psi(t,t+\tau,X^{0,x}_{t+\tau}, u(\theta_n,.,.))) - u( \theta_n, t,X^{0,x}_t) \right)^2 \dif t \right]  +\\
 & 3 \E\left[    \int_0^T  1_{X^{0,x}_{t} \in \Omega^{0,x}_{\epsilon_n}}  \E_{t,X^{0,x}_t} \left( 1_{X^{0,x}_{t+ \tau} \notin \Omega^{0,x}_{\epsilon_n}} \frac{1}{4} \big[ \phi\big(t, t+\tau ,X_{t+\tau}^{t,x}, u(t+\tau ,X_{t+\tau}^{t,x}) \big) + \right. \right.\\
 &\left. \left. \phi\big(t, t+\tau ,\widehat{X}_{t+\tau}^{t,x}, u(t+\tau ,\widehat{X}_{t+\tau}^{t,x}) \big) \big]^2 \right)  \dif t \right]\\
  \le &    3 K^2   \E \left[ \int_0^T 1_{X^{0,x}_{t} \in \Omega^{0,x}_{\epsilon_n}}   \E_{t,X^{0,x}_t} [ 1_{\tau <T -t}   1_{X^{0,x}_{t+\tau} \in \Omega^{0,x}_{\epsilon_n}} 
  \frac{(u( \theta_n, t+\tau,X^{0,x}_{t+\tau}) - u(t+\tau,X^{0,x}_{t+\tau}))^2}{\rho(\tau)^2 }] \dif t  \right] +
  \\ 
  &    3 T \ell(\theta_n) + 
  3  ( \frac{ ||f||_\infty^2}{\hat \rho(T)^2} +  \frac{ ||g||_\infty^2}{\bar F(T)^2}) \E[    \int_0^T  1_{X^{0,x}_{t} \in \Omega^{0,x}_{\epsilon_n}}  \E_{t,X^{0,x}_t} \left( 1_{X^{0,x}_{t+ \tau} \notin \Omega^{0,x}_{\epsilon_n}}  \right)  \dif t] \\
  \le & 3 K^2   \E \left[  \int_0^T  1_{X^{0,x}_{t} \in \Omega^{0,x}_{\epsilon_n}}   \E_{t,X^{0,x}_t} [  \int_t^T ds \left( 1_{X^{0,x}_{s} \in \Omega^{0,x}_{\epsilon_n}} 
  \frac{(u( \theta_n, s,X^{0,x}_{s}) - u(s,X^{0,x}_{s}))^2}{\rho(s-t) } \right) ds ] \dif t  \right] + 3 T \ell(\theta_n)  + C \epsilon \\
  \le &  3 \frac{K^2 T}{\hat \rho(T)}  \E \left[  \int_0^T  1_{X^{0,x}_{t} \in \Omega^{0,x}_{\epsilon_n}}  
  (u( \theta_n, t,X^{0,x}_{t}) - u(t,X^{0,x}_{t}))^2 \dif t  \right] + 3T \ell(\theta_n)  + C \epsilon 
 \end{flalign*}
 Then, if $K$ is small enough
 \begin{align*}
 F_n \le  \hat \rho(T) \frac{T \ell(\theta_n)  + C \epsilon }{ \hat \rho(T) - 3 K^2 T},
 \end{align*}
 which completes the proof.

\printbibliography

\end{document}